\documentclass[a4paper]{report}
\usepackage[nottoc,notlot,notlof]{tocbibind}
\pdfoutput=1
\usepackage[sc]{mathpazo}
\usepackage[T1]{fontenc}
\usepackage{textcomp}

\usepackage[utf8]{inputenc}
\usepackage{natbib}

\usepackage[linktocpage=true,backref=page]{hyperref}
\hypersetup{
  colorlinks   = true,    
  urlcolor     = blue,    
  linkcolor    = red,    
  citecolor    = red      
}
\usepackage{amsthm}
\usepackage{amsmath,amssymb, amstext}
\usepackage{mathabx}
\usepackage{bm}
\usepackage[font={small,it}]{caption,subcaption}
\usepackage{graphicx}
\usepackage{tikz}
\usetikzlibrary{arrows,fit,matrix,backgrounds,decorations.pathreplacing,calc}
\usetikzlibrary{positioning}
\usepackage{bm} 

\newbox\mybox
\usepackage{enumitem}

\newlist{thmlist}{enumerate}{1}
\setlist[thmlist]{label=(\roman{thmlisti}),noitemsep}

\usepackage{thmtools}

\declaretheorem[
    name=Definition]{mydef}

\declaretheorem[
    name=Theorem]{thm}

\usepackage[capitalize]{cleveref}
\crefname{paragraph}{paragraph}{paragraphs}
\Crefname{paragraph}{Paragraph}{Paragraphs}

\makeatletter

\Crefname{thm}{Theorem}{Theorems}
\Crefname{lem}{Lemma}{Lemmas}
\Crefname{mydef}{Definition}{Definitions}
\Crefname{listthm}{Theorem}{Theorems}
\Crefname{listlem}{Lemma}{Lemmas}
\Crefname{listdef}{Definition}{Definitions}

\addtotheorempostheadhook[thm]{\crefalias{thmlisti}{listthm}}
\addtotheorempostheadhook[lem]{\crefalias{thmlisti}{listlem}}
\addtotheorempostheadhook[mydef]{\crefalias{thmlisti}{listdef}}

\creflabelformat{listthm}{#2\thethm.#1#3}
\creflabelformat{listlem}{#2\thethm.#1#3}
\creflabelformat{listdef}{#2\thethm.#1#3}

\tikzset{%
    myarrow/.style={%
        thick,->,>=stealth',shorten >=2pt
    }
}

\DeclareMathOperator*{\argmin}{arg\,min}
\bmdefine\bcolon{:}

\DeclareMathOperator{\mi}{mi}
\DeclareMathOperator{\nmi}{nmi}
\newcommand{\ci}{\iota}

\DeclareMathOperator{\pa}{pa}
\newcommand{\dis}{\mathfrak{D}}
\newcommand{\dmi}{\Delta\!\mi}

\newcommand{\bs}{\setminus}
\newcommand{\Nplus}{\mathbb{N}^+}

\newcommand{\X}{\mathcal{X}}
\newcommand{\Y}{\mathcal{Y}}

\newcommand{\A}{\mathcal{A}}
\newcommand{\B}{\mathcal{B}}

\newcommand{\D}{\mathcal{D}}
\newcommand{\E}{\mathcal{E}}
\newcommand{\M}{\mathcal{M}}
\newcommand{\T}{\mathcal{T}}

\renewcommand{\S}{\mathcal{S}}

\newcommand{\Xt}{\{X_t\}_{t \in T}}

\newcommand{\Xv}{\{X_i\}_{i \in V}}
\newcommand{\Xw}{\{X_i\}_{i \in W}}
\newcommand{\Xat}{\{X_{A_t}\}_{t \in T}}
\newcommand{\Yt}{\{Y_t\}_{t \in T}}

\newcommand{\At}{\{A_t\}_{t \in T}}

\newcommand{\Et}{\{E_t\}_{t \in T}}
\newcommand{\Mt}{\{M_t\}_{t \in T}}
\newcommand{\St}{\{S_t\}_{t \in T}}

\newcommand{\pet}{{\preceq t}}

\newcommand{\xpet}{x_{\preceq t}}

\newcommand{\id}{\mathfrak{i}}

\newcommand{\Latt}{\mathfrak{L}}
\newcommand{\lunit}{\textbf{1}}
\newcommand{\lzero}{\textbf{0}}
\newcommand{\lpre}{\vartriangleleft}
\newcommand{\lsuc}{\vartriangleright}
\newcommand{\lpreeq}{\trianglelefteq}

\newcommand{\lmin}{\blacktriangleleft}

\newcommand{\inv}{\updownarrow}
\newcommand{\shift}{\rightarrow}
\newcommand{\Ent}{\mathfrak{E}}
\newcommand{\stpset}{\mathfrak{C}}

\newcommand{\perstp}{\mathfrak{S}}

\newcommand{\ext}{w}
\newcommand{\pal}{{PA}}

\newcommand{\MCconst}{{MC^{=}}}

\newcommand{\MCnoise}{{MC^\epsilon}}
\newcommand{\symg}{\Sigma}
\newcommand{\gr}{\mathfrak}
\newcommand{\Pset}{\mathcal{P}}

\newcommand{\patno}{N_{pat}}
\newcommand{\slino}{N_{sli}}

\newcommand{\corres}{\overset{\scriptscriptstyle\wedge}{=}}

\newcommand{\stp}[1]{\{X_i=x_i\}_{i \in #1}}
\newcommand{\gustp}[2]{\{X_i = x_i\}^{#2}_{i \in #1}}
\newcommand{\gdstp}[2]{\{X_i = x_{#2(i)}\}_{i \in #1}}

\newcommand{\barstp}[1]{\{X_i=\bar{x}_i\}_{i \in #1}}
\newcommand{\bargustp}[2]{\{X_i = \bar{x}_i\}^{#2}_{i \in #1}}
\newcommand{\bargdstp}[2]{\{X_i = \bar{x}_{#2(i)}\}_{i \in #1}}

\DeclareMathOperator{\HS}{H}
\DeclareUnicodeCharacter{202F}{ }
\graphicspath{ {images/} }

\author{Martin Andreas Biehl}
\title{}
\title{
	{Formal approaches to a definition of agents}\\
	{\small submitted to the University of Hertfordshire in partial fulfilment of the requirements of the degree of 
PhD}\\
	\vspace{.9cm}
}
\date{\today}

\begin{document}
\pagenumbering{gobble}
\maketitle
 
\tableofcontents

\abstract{This thesis is a contribution to the formalisation of the notion of an agent within the class of finite multivariate Markov chains. In accordance with the literature agents are are seen as entities that act, perceive, and are goal-directed. We present a new measure that can be used to identify entities (called \textit{$\ci$-entities}). The intuition behind this is that entities are spatiotemporal patterns for which every part makes every other part more probable. The measure, complete local integration (CLI), is formally investigated within the more general setting of Bayesian networks. It is based on the specific local integration (SLI) which is measured with respect to a partition. CLI is the minimum value of SLI over all partitions. Upper bounds are constructively proven and a possible lower bound is proposed. We also prove a theorem that shows that completely locally integrated spatiotemporal patterns occur as blocks in specific partitions of the global trajectory. Conversely we can identify partitions of global trajectories for which every block is completely locally integrated. These global partitions are the finest partitions that achieve a SLI less or equal to their own SLI. We also establish the transformation behaviour of SLI under permutations of the nodes in the Bayesian network. 

We then go on to present three conditions on \textit{general} definitions of entities. These are most prominently not fulfilled by sets of random variables i.e.\ the perception-action loop, which is often used to model agents, is too restrictive a setting. We instead propose that any general entity definition should in effect specify a subset of the set of all spatiotemporal patterns of a given multivariate Markov chain. Any such definition will then define what we call an entity set. The set of all completely locally integrated spatiotemporal patterns is one example of such a set. Importantly the perception-action loop also naturally induces such an entity set. We then propose formal definitions of actions and perceptions for arbitrary entity sets. We show that these are generalisations of notions defined for the perception-action loop by plugging the \textit{entity-set} of the perception-action loop into our definitions. We also clearly state the properties that general entity-sets have but the perception-action loop entity set does not. This elucidates in what way we are generalising the perception-action loop. 

Finally we look at some very simple examples of bivariate Markov chains. We present the disintegration hierarchy, explain it via symmetries, and calculate the $\ci$-entities. Then we apply our definitions of perception and action to these $\ci$-entities.}


\clearpage
\thispagestyle{plain}
\par\vspace*{.35\textheight}{\centering To my late father and my mother\par}


\chapter*{Acknowledgements}

I thank: 

Professor Daniel Polani; first and foremost for giving me the chance to pursue this particular line of research; second for great discussions, supervision, criticism, collaboration, and measured encouragement. I also want to thank my second supervisor Professor Chrystopher Nehaniv for support whenever I needed it. 

The friends and colleagues at the University of Hertfordshire: Andres Burgos, Christoph Salge, Cornelius Glackin, Nicola Catenacci-Volpi, Lukas Everding, Sander van Dijk, Dari Trendafilov, Martin Greaves, Marcus Scheunemann, Frank Foerster, Antoine Hiolle, Joan Saez.

Professors Takashi Ikegami and Nathaniel Virgo for inviting me as a doctoral fellow of the Japan Society for the Promotion of Science and as a long term visitor to the Earth-Life-Science Institute's Origins Network (EON) respectively. The external interest in my research that I experienced during these two stays in Tokyo were great motivation to continue the path that lead to much of this thesis. 

The colleagues and friends at the Earth-Life-Science Institute and Ikegami Lab. Especially: Nicolas Guttenberg, Julien Hubert, Stuart Bartlett,
Lana Sinapayen, Olaf Witkowski, Kanjin Yoneda, and Yoichi Mototake. 


Professor Florentin W\"org\"otter for providing a path back into science when it was starting to look unlikely. I also thank him for generously supporting my decision to leave his group when it became clear that our short term goals in research were not aligned closely enough.

The colleagues and friends in G\"ottingen for discussions and good past times: Frank Hesse, Jan Braun, Christian Tetzlaff, Harm Surkamer, Xiaofeng Xiong, Michael Fauth, Mohammad Aein, Alexey Abramov, Minja Tamosiunaite, Sakyashinga Dasgupta, Tomas Kulvicius, Christoph Kolodziejski, Dennis Goldschmidt, Ahmed Tarek, Christopher Battle, Niko Deuschle, Bernhard Althaner, Clemens Buss, Lukas Geyrhofer, and David Hofmann.

Professors Auke Ijspeert and Karl Svozil for continuing to support my scientific development long after I had been their student. They both also had a profound influence on my thinking.

Basement flat 346 for friendship and hospitality.

Stefanie and Urs Schrade for a great place to write my thesis.

Cornelia and Jan Loewengut as well as Ulla Biehl and Michael Breuner for supporting my education until late into my thirties. 

All the people whose friendship I can rely on even after long phases without any feedback from my side. They are in my heart (even if some of them apparently think that it is grey). 


\chapter{Introduction}
\pagenumbering{arabic}

On the most general level this thesis is a contribution to existing research that tries to reconcile a physicalist worldview with the notion of \textit{agents}. The physicalist worldview holds that the laws of physics determine (whether in any way stochastically or not) everything that happens in the universe. The notion of an agent relies fundamentally on the agent's capacity to act. This, however, means that the agent can \textit{make} something happen and that there are not only things happening \textit{to} it \citep{sep-action,mcgregor_more_2016}. It seems that if the agent can make something happen then there must be something that is not determined by the laws of physics. Conversely, if the laws of physics determine everything that happens then the agent did not make it happen. So either the laws of physics do not determine everything or agents do not exist. Let us assume the laws of physics determine everything. We do not know the actual laws of physics but let us also assume that the laws of physics are the same everywhere (in every inertial frame of reference). Then wherever there is a human (the primary example of an agent) and wherever there is no human the laws of physics are the same. These laws of physics do not care about what is happening, they just make it happen. The question remains whether there are agents. Even if humans (or animals, bacteria, plants) cannot really make things happen, our intuition tells us that there is a difference between the volumes of space that contain humans and the volumes of space that do not. The ones that do not contain humans (or other agents) usually are vastly less dangerous for example.

The question is still open what the difference is or even how a difference in danger between volumes of (physically identical) space can arise. In other words it is still an open question \citep{mcgregor_more_2016} what it is that makes some volumes of space (and their time evolution) \textit{agents}. Specifically human characteristics are not the focus of this research, simple living organisms are sufficient from this point of view.

We want to ascertain that we do not fall prey to our own imagination and give an account of agents that only \textit{seems} compatible with laws of physics. Therefore we choose a completely formal setting. This means we choose a well defined class of ``universes'' that have laws which are equal basically everywhere. We do not choose the leading theories of physics. Our target here is the seeming incompatibility between lawfulness and agent containment of a system/universe. There is no need to use complicated systems if we are not sure that the simple ones are not sufficient. We also do not want to assume a priori the existence of notions from physics; most prominently the notion of energy, which gives rise to the notion of work. If it turns out that we need such a concept for agents to exist within a lawful universe then even better. In summary we are firmly in the field of artificial life with its intention to study ``life as it could be'' \citep{langton_artificial_1989}. More precisely, we study agents as they could be.

%
The literature \citep{barandiaran_defining_2009} tells us that agents are entities that act, perceive, and pursue goals. Accordingly, we should try to define each of these notions for a universe governed by (basically everywhere equal) laws of physics. This thesis, building on previous research, proposes formal definitions for \textit{entities}, \textit{action}, and \textit{perception} in such systems. A definition of what it means to pursue goals for entities (that may perceive and act) is not part of this thesis.

As the setting for the formal definitions we choose (possibly driven) multivariate Markov chains. As they include cellular automata, these are suited to model universes with basically everywhere equal laws of physics \citep{toffoli_cellular_1984}. They also include the famous game of life cellular automaton. This is the setting for one of the most complete attempts at a formal definition of agents to date by \citet{beer_cognitive_2014}. Driven multivariate Markov chains are important because they contain computer implementations of (also continuous) reaction-diffusion systems that exhibit life-like phenomena \citep{virgo_thermodynamics_2011,froese_motility_2014,bartlett_emergence_2015,bartlett_precarious_2016}.

Within this setting this thesis splits up into two parts. The first part is the introduction and formal investigation of a newly conceived measure of integration, complete local integration. The second then contains four smaller contributions, a proposition of three requirements for entity definitions, a proposal and motivation of using complete local integration as a definition of entities, a formal definition of actions for arbitrary entities, and a formal definition of perceptions for arbitrary entities. 

Apart from \citet{barandiaran_defining_2009} which contains a review of agent definitions we ignore in this thesis all work on agents that is not formal. This means we will not discuss the historical background and philosophical considerations that enable us to even try and formalise agents. We highlight, however, that the formal approaches we are building on are almost all in turn strongly influenced by the work of \citet{maturana_autopoiesis_1980} 
.
This is true for \citet{barandiaran_defining_2009} themselves but equally so for \citet{bertschinger_information_2006,bertschinger_autonomy_2008,beer_characterizing_2014,beer_cognitive_2014}. Our work can be seen to a certain extend as a synthesis of these publications.

Recently \citet{beer_characterizing_2014,beer_cognitive_2014} has thoroughly investigated the application of the ideas of \cite{maturana_autopoiesis_1980} to the glider in the game of life. In \citet{beer_characterizing_2014} he informally introduces criteria for the organisational closure of spatiotemporally extended structures in the game of life (the block, blinker, and glider). From the organisational closure he derives the boundary of the structures and thereby arrives at a definition of entities. It seems to us that this approach can be further formalised. However, as \citeauthor{beer_characterizing_2014} notes for a formalisation that also accounts for edge cases where closure is temporarily lost, or transformation into other closure regimes further decisions have to be taken. In the end probably for this reason no formal definition of which structures in general constitute entities is given. In \citet{beer_cognitive_2014} the glider is then investigated in more detail with respect to its cognitive domain. The cognitive domain contains a notion of perception that we generalise in this thesis and an implicit notion of action which is also similar to the notion of action we propose here. Ignoring goal-directedness this work represents a significant step to a formal definition of agents as it is well defined in a cellular automaton with the same dynamical/physical law at every cell. 

A fully formal definition of entities is still missing, however, and the account of perception is also specific to deterministic systems. Furthermore the account of perception (or cognitive domain) may seem peculiar and unrelated to concepts outside the theory of \citet{maturana_autopoiesis_1980}. 

If we look for formal concepts that can be used as entity definitions outside of \citeauthor{beer_characterizing_2014}'s work we find that most candidate notions also have problems discerning entities in edge cases. Such edge cases occur where multiple entities collide, appear or disappear. A common reason for this is that many notions that discern ``important'' structures/patterns from ``less important'' structures or more generally just some structures from other structures do so by evaluating structures only spatially. This means they discern among different structures that exist at a single time-step $t$. Then at the next time-step $t+1$ they again discern between structures at that time-step. The question then remains how to identify which of the structures at $t$ match up with which structures at $t+1$ to form spatiotemporal structures. Often this is unambiguous but when similar structures (like multiple gliders) collide or even overlap then this approach usually fails. An example of such notions are the spatiotemporal filters \citep{shalizi_automatic_2006,lizier2008local,flecker_partial_2011} developed for cellular automata. These can highlight gliders, but if two gliders collide they make no claim about the identity of a possibly ensuing glider. Note that these structures were also not conceived for the purpose of detecting entities. The same problem occurs however for the Markov blanket entity underlying the ``living organism'' of \citet{friston_life_2013} at least in its current formulation. 

An obvious solution to this problem is to directly evaluate spatiotemporal structures for their identity. Then no matching up of time-slices is needed anymore. The existing work identifying such spatiotemporal structures is limited. \citet{balduzzi_detecting_2011} detects spatiotemporal coarse-grainings in cellular automata (and multivariate Markov chains in general). This approach may be an alternative to our proposal. It has never been used even for small systems however. Our proposal seems much simpler to express but computationally both are unfeasible for large systems without significant approximations. Another similar work also resulting in a spatiotemporal coarse-graining is the work by \citet{hoel_quantifying_2013} which identifies causally efficient macrostates. These are however random variables themselves and not spatiotemporal structures like gliders as we will argue. The latter work can be combined with \citet{oizumi_phenomenology_2014}, according to the authors, to get spatiotemporal structures more similar to gliders. In this case after the spatiotemporal coarse-graining of \citet{hoel_quantifying_2013} the approach of \citet{oizumi_phenomenology_2014} is used to detect spatial structures on top. This looses some flexibility compared to our approach since the coarse-graining does not allow arbitrary combinations of fine-grained spatiotemporal structures anymore. It also does not treat the spatial and temporal dimension on equal footing which is a desirable theoretical property considering the success of relativity theory. These spatiotemporal coarse-grainings have not been specifically proposed as definitions of entities and and do not come with definitions of perception and action. The work by \citet{oizumi_phenomenology_2014} \citep[and predecessors][]{tononi_information_2001,tononi_measuring_2003,tononi_information_2004,balduzzi_integrated_2008} are somewhat related to agents since they try to formally define consciousness. However, for the same reason they identify a single entity (the main complex) in the systems they are used for. These systems are also conceptualised to be applied to neural networks i.e.\ the ``inside'' of agents and not to universes to detect the agents. However, investigating the relations of our measure to these will be interesting work for the future.

In summary, currently there is no formally defined and accepted way of identifying entities in multivariate Markov chains. In this thesis we contribute a new measure for this purpose. this measure is called complete local integration (CLI) and we denote the resulting entities as $\ci$-entities. Identifying entities is important for our general research project because it allows to unambiguously and consistently attribute sequences of actions and perceptions over the course of time. This seems to be needed in order to reveal any goal-directed behaviour. This in turn is a defining feature of agents. 

The underlying idea of complete local integration is quite simple. We require that every part of an entity makes all other parts of it more probable. Intuitively this can be related to the fact that partial living organisms are extremely rare or at least much rarer than whole living organisms. Not all entities are agents, however, since for example soap-bubbles also have this property\footnote{The author thanks Eric Smith for pointing out this example.}. 

In \cref{ch:stp} we analyse the notion of complete local integration formally. This is done in the general setting of Bayesian networks. These are a generalisation of multivariate Markov chains to cases where notions of time and space are irrelevant or not so simple. This is done since SLI and CLI may be of interest in different contexts as well. First we define the more basic notion of specific local integration with respect to a particular partition. For SLI we constructively prove upper bounds and construct an example of a pattern with strongly negative SLI. These results are of general technical interest and also provide examples. Then we introduce CLI which is the minimum value of SLI with respect to any partition.  We then introduce the disintegration hierarchy and the refinement-free disintegration hierarchy. These constructions help reveal the structure of the completely locally integrated patterns and underlie the main formal contribution of this thesis, which is the disintegration theorem (\cref{thm:disintegration}). The disintegration theorem connects the SLI of an entire trajectory (time-evolution) with respect to a partition with the CLI of the blocks of that partition. More precisely for a given trajectory the blocks of the finest partitions among those leading to a particular value of SLI consists only of completely locally integrated blocks. Conversely each completely locally integrated pattern is a block in such a finest partition among those leading to a particular value of SLI. This connection is new. This theorem may lead to further theoretical results and suggests an additional interpretation of completely integrated patterns as independently encoded parts in a code adapted to the specific trajectory (see \cref{sec:interpret}). We then go on and investigate the symmetry properties of SLI. We establish its transformation under permutations of the nodes in the Bayesian network in the SLI symmetry theorem and its corollary (\cref{thm:symmpart,thm:symmpartcor}). This can be used to explain the structure of the disintegration hierarchy as we will see in \cref{sec:exstp} where we present simple examples. Symmetry properties are also expected to be important for further formal analysis of SLI/CLI. For convenient reference we also show how symmetries spread in multivariate Markov chains, our main application here.

We then come to the second part of this thesis. We have already stated that the notion of perception (part of the cognitive domain) in \citet{beer_cognitive_2014} may seem idiosyncratic. However, it turns out to be closely related to the notion of perception that is formalised in the perception-action loop. The perception-action loop is a model of agent-environment interaction that goes back at least to \citet{uexkull_theoretische_1920}. Renewed interest possibly started with \citet{beer_dynamical_1995} and the dynamical systems view of cognition. Later it was formally captured as a Bayesian network by \cite{klyubin_organization_2004} and has been used extensively since then for information theoretic investigations into the interaction of agents and environments \citep{klyubin_empowerment_2005,bertschinger_information_2006,bertschinger_autonomy_2008,salge2014changing,ay_information-driven_2012,zahedi_quantifying_2013}. 

It is therefore safe to say that the perception-action loop is a powerful tool to investigate such interactions. However, it makes some assumptions that make it unsuitable as a tool for investigating entities. The reason for this is that it models agents as random variables/processes.

We argue in \cref{sec:entinmvmc} that a formal notion of entities in multivariate Markov chains should satisfy three criteria. These are \textit{compositionality}, \textit{degree of freedom traversal}, and \textit{counterfactual variation}. It becomes clear in the course of this argument that subsets of the set of random variables in the multivariate Markov chain are not suitable for agent definitions. This includes in particular the perception-action loop since there the agent is just a sequence of random variables.

The three criteria are derived by using what we call the non-preclusion argument. Definitions of entities must allow every phenomenon that is known to be exhibited by any agent (since all agents are entities). For example, if we know that there is a green agent somewhere then an entity definition which says all entities are blue must be wrong. So greenness must not be \textit{precluded} by the entity definition. We argue that, because the glider and other life-like structures in known simulations \citep{virgo_thermodynamics_2011,froese_motility_2014,bartlett_emergence_2015,schmickl_how_2016} exhibit compositionality, degree of freedom traversal, and counterfactuality both \textit{in value} and \textit{in extent}, entity definitions must not preclude these phenomena. Roughly, compositionality means that it must be possible that entities have spatial and temporal extension. Degree of freedom traversal (in the game of life for example) means that over time the cells that the entity occupies can change. Counterfactual variation means that entities can be different from one trajectory or time-evolution to another depending on the initial condition for example. Counterfactual variation in value means that there are entities in both trajectories and they occupy exactly the same cells but the occupied cells have different values (e.g.\ some black ones are white). Counterfactual variation in extent means that the entity in the first trajectory and possibly an entity in the second trajectory occupy different cells. If there is no entity in the second trajectory this is a special kind of counterfactual variation in extent. Apart from ruling out the definition of entities as sets of random variables these three phenomena can help guide future entity definitions. We also believe the non-preclusion argument can be extended to further phenomena such as growth or replication.

The three requirements then convince us that subset of random variables are unsuitable for a general entity definition. We then propose to define entities in general as subsets of the set of spatiotemporal patterns. These are formally defined in \cref{ch:fb} but are basically just subset of the cells with fixed values. Importantly the fixed cells are not limited to one time-step but can spread across arbitrary times. We then call any chosen subset of all spatiotemporal patterns an entity-set. How to arrive at the entity-set is a matter of choice. We propose to use the completely locally integrated spatiotemporal patterns, the $\ci$-entities but our definitions of entity action and entity perception are for arbitrary entity-sets. 

These definitions of entity actions and entity perceptions combine ideas from \citet{bertschinger_autonomy_2008} and \citet{beer_cognitive_2014}. 
Let us first come back to the initial problem since we are about to define actions in a lawful system. In a multivariate Markov chain only the transition matrix makes things happen and since all entities are within the chain they cannot possibly make anything happen. Furthermore for each entity in the entity set we are given the full spatiotemporal extension of the entity at once. There is \textit{no choice} for these entities they are completely determined for their entire lifetime. The trick we use to define actions in such a system is to rely on counterfactual entities. That is we use entities that are indistinguishable for the environment. We then say that an entity performs an action at time $t$ if it has a \textit{co-action entity} that cannot be distinguished from the original one by \textit{any} observer in the system. This is ensured if there is a single environment at $t$ that can occur together with both entities. Since the environment is identical nothing in it and therefore no observer can know what the next configuration or time-slice of the entity is. At least if the next time-slices of the co-action entities are actually different. This is another requirement we make of co-action entities. Since the two co-action entities can differ in value or extent at the next time-step we also can differentiate between value and extent actions. 

We show that this definition of entity action implies the notion of non-heteronomy due to \citet{bertschinger_autonomy_2008} in the special case where the entities are the perception-action loop entities. It is possible to show this formally because the perception-loop can be seen as consisting of a special case of an entity set. This entity set is not composite in space, not degree of freedom traversing, and only counterfactual in value but it is still an entity set (it is also exhaustive which means there is an entity in every trajectory). Due to the generality of the entity set we can therefore treat the perception-action loop as a test case for our definitions. This will also be useful in future research since we can rely on the existing body of work in the perception action-loop and generalise it. The entity set can then serve as a bridge between the perception-action loop and a more general theory of agents in multivariate Markov chains (or even more general Bayesian networks in the future).

We then come to our definition of perception for entity-sets. The basic idea behind entity perception is to capture all influences from the environment on an entity. For the perception-action loop there is a well defined procedure for doing this and we will show this in \cref{sec:paloopformal}. There we show that we can capture the influence from the environment by a partition of the environment states into states that have the same influence on the agent process. This kind of construction is known in the literature and has been used for example by \citet{balduzzi_detecting_2011} in basically the same way. This is also related to the older notion of causal states \citep{shalizi_causal_2001}. In this thesis we generalise this construction for entity sets. It turns out that the result is also a generalisation of the cognitive domain (more precisely the macroperturbations) defined in \citet{beer_cognitive_2014}\footnote{We do not prove this. But we are quite sure.}. So how does this generalisation work? This is formally more involved then we originally envisioned. Again we are forced to deal with the fact that the entities are already defined for their entire lifetimes. So we actually cannot ``test'' influences on them. Again we rely on other, similar entities to formally capture perception. For a given entity, we take the set of entities that has identical pasts up to some time $t$. We also make sure that those entities still all exist at $t+1$. These entities are the co-perception entities. We then classify the environments that can occur with at least one of these entities. These are the co-perception environments. Since the multivariate Markov chains can be stochastic we cannot identify which environment leads to which future  \citep[as in][]{beer_cognitive_2014} we have to do this probabilistically. For this we have to define a probability distribution over the futures of the co-perception entities. In the perception-action loop setting this is straightforward since the futures of the co-perception entities are just the possible values of a random variable\footnote{Due to the special case of the entity-set in the perception-action loop, which consists of all possible combinations of all agent random variables.}. For arbitrary entity sets, the co-perception entities can have undesirable properties. One such property is that they may not exhaustive. This means that the sum of their probabilities does not sum to one as is needed for probability distributions. This can be dealt with in a standard way if the the co-perception entities are mutually exclusive. However, for arbitrary entity sets this is not the case. We then have two options. 

\begin{itemize}
  \item Either we take a subset of the co-perception entities that is mutually exclusive and define the probability distribution over this set. Due to the arbitrary choice of the subset however this leads to a non-unique perceptions; another choice of a subset produce another set of perceptions.
\item Or we find that the entity set is non-interpenetrating. In that case we can use the whole set of co-perception entities for the definition of the probability distribution. This ensures a uniquely defined set of perceptions. Non-interpenetration is the assumption that two (different) entities with identical pasts cannot occur together. Note that this is not something akin to cell division. Cell division corresponds to a single entity that just becomes two separate \textit{spatial} patterns. Two non-identical entities with the same past that occur together would be more akin to two aligned light beams projected onto a wall unaligning. 
\end{itemize}
In both cases we then arrive at a probability distribution which allows us to classify the co-perception environments. However, this probability distribution is over the entire futures of the co-perception entities. Say there are only two co-perception entities. These may not only be identical up to time $t$ they may be identical up to some arbitrary time $t+r$ in the future. It then seems wrong to interpret the classification of the environments based on the difference in the far future between the two co-perception entities as a perception at time $t$. To solve this problem we introduce the branching partition. This partitions the co-perception entities according to their next configuration or time-slice. Co-perception entities with equal next configurations are considered as equivalent and part of the same future \textit{branch}. We can then easily derive the probability distribution over these branches by summing over the probabilities in the branch. We call this probability distribution the \textit{branch-morph}. 

We then go on to show explicitly that in the special case of the perception-action loop the branch-morph specialises to the standard construction we used to define perceptions in the perception-action loop. We therefore successfully generalise this construction to the case of arbitrary entity sets. In particular these entity sets can be non-exhaustive, degree of freedom traversing, and counterfactual in extent (not only in value). We establish that the branch morph is uniquely defined if the entity-set is non-interpenetrating. 
This is significant since non-interpenetration then seems like a possible axiom for entity-sets. The branch-morph itself and possibly similar constructions can be used to carry over information theoretic notions from the perception-action loop to entity-sets. This may lead to a definition of goal-directedness. We note here already that the entity set of $\ci$-entities does not satisfy non-interpenetration. 

On the technical side we also show how the assumption that all co-perception environments must occur with at least one of the co-perception entities translates to a seemingly weaker requirement in case of the perception-action loop and related cases. This shows conversely that this requirement on the co-perception environments is not stronger than the assumptions made in the perception-action loop case. This is important because we want to generalise the perception-action loop without making extra assumptions.

We then come to the final chapter which presents two extremely simple bi-variate Markov chains that have three time-steps. We calculate and visualise the disintegration hierarchy for the first and explain its structure using the SLI symmetry theorems. We also calculate the $\ci$-entities for both chains. We then verify that $\ci$-entities indeed satisfy the three criteria of compositionality, degree of freedom traversal, and counterfactual variation (in value and extent). We also find some counter intuitive examples of $\ci$-entities.

Then we apply our definitions of action and perception to the calculated $\ci$-entities. We find actions in value and extent. One of the actions we find however seems to question the motivation of our construction this will need further investigation. We also find that $\ci$-entities are generally interpenetrating. We nonetheless construct two branch-morphs based on two different mutually exclusive subsets of the co-perception entities and find that they differ slightly. This is expected.

Finally we discuss the results and give some outlook for future work.

\section{Original contributions}
In summary the original contributions are:

\subsection*{\Cref{ch:stp}} 
\begin{itemize}
  \item Definition of specific local integration (SLI).
  \item Constructive proof of upper bound of SLI.
  \item Construction of negative SLI example. 
\item Definition of complete local integration (CLI).
  \item Definition of disintegration hierarchy and refinement-free disintegration hierarchy.
\item Proof of the disintegration theorem.
\item Proof of the SLI symmetry theorems.
\end{itemize}

\subsection*{\Cref{ch:agents}} 
\begin{itemize}
  \item An argument (via compositionality, degree of freedom traversal, and counterfactual variation) for a spatiotemporal pattern-based definition of entities. 
\item The abstraction of entity-sets which enables the formal connection to perception-action loop.
\item A tentative\footnote{For some context on what we mean by ``tentative'' see \cref{ch:agents}.} formal definition of entities as completely locally integrated spatiotemporal patterns.
  \item A tentative formal definition of action for arbitrary entity-sets.
  \item A classification of actions into value actions and extent actions.
  \item A tentative formal definition of perception for arbitrary entity-sets. 
  \item An exposition of the role of non-interpenetration of entity-sets in perception. Namely, it makes perception naturally unique.
  \item The formal exposition of the connection of the action definition to non-heteronomy of \citet{bertschinger_autonomy_2008} in the perception-action loop.
  \item The formal exposition of the way the perception definition specialises to the perception-action loop.
  \item A construction of a conditional probability distribution (the branch-morph, including branching partition) over the futures of entities which allows the definition of perception.
  \item Proof that the condition on co-perception environments is not stronger than the assumptions about environment states inherent in the perception-action loop.
\end{itemize}

\subsection*{\Cref{sec:exstp}} 
\begin{itemize}
  \item Computation and presentation of disintegration and refinement-free disintegration hierarchies for two simple systems.
  \item Explanation of the occurrence of multiple disconnected components in the partially ordered disintegration levels via the SLI symmetry theorems.
  \item Computation and presentation of the completely locally integrated spatiotemporal patterns of two simple systems.
  \item Examples of $\ci$-entities that exhibit the three phenomena compositionality, degree of freedom traversal, and counterfactual variation that we argued for in \cref{sec:entinmvmc}.
  \item Examples of entity actions of $\ci$-entities. 
  \item Example of interpenetrating $\ci$-entities showing that they do not necessarily obey non-interpenetration.
  \item Example of an entity perception and a branch-morph using a proxy for a co-perception partition.
  \item Example of an entity action and entity perception of the same $\ci$-entity at the same time-step.
  \item Discussion of the results on $\ci$-entities as entity sets in the example systems.
\end{itemize}

\chapter{Related work}
Here we discuss closely related work in the literature. First we point to the formal origin of the new measures of specific local integration and complete local integration. Then we discuss work that is related to our notion of entities. In cases where the entities are part of conceptions of agents we also discuss perception and action.  We have tried to write this chapter without relying too much on our own formalism for accessibility. It might also serve as a further introduction into the field which is why we have left it in front of the technical part of the thesis. Nonetheless, after reading this thesis some arguments will be easier to understand. 

\section{Formally related work}
\label{sec:formallyrelated}

\subsection{Specific local integration and complete local integration}
In \cref{ch:stp} we define specific local integration (SLI). This is a local measure in the sense of the measures of local information dynamics proposed by \citet{lizier_local_2012}. We use the same method of localization presented there only on a different original measure namely multi-information \citep{mcgill_multivariate_1954,tononi_measure_1994,amari_information_2001}. The method of localising information-theoretic notions like mutual-information and transfer entropy was developed to measure information of specific realisations $x,y$ of random variables $X,Y$. This is in contrast to the original measures which are averages of the local versions. We argue in \cref{sec:cfvar} that entities should be trajectory dependent. This is equivalent to saying that entities are composed of specific realisations of random variables. Therefore we follow \citeauthor{lizier_local_2012} in using a localised measure.

In contrast to the work by \citeauthor{lizier_local_2012} we are not trying to reveal information storage, transfer and processing to characterise computation within a dynamical system but instead we are trying to find entities and agents within such a system. While the dynamics of information are certainly relevant for agents in dynamical systems we focus here directly on the identification of spatially and temporally composite structures. The measures discussed by \citeauthor{lizier_local_2012} are not designed for the purpose of identifying spatially composite structures (see also \cref{sec:stpfiltering}).


The measure of complete local integration (CLI), which we define in \cref{sec:cli} builds on the notion of SLI. The measure of SLI is defined with respect to a particular partition of a set of random variables. In order to get CLI we evaluate SLI with respect to every possible partition of the set of random variables. We then take the minimum of all the values found in this way to be the value of CLI. A spatiotemporal pattern that has positive CLI value is then defined as an \textit{$\ci$-entity}. The procedure of passing through all partitions has been used for measures similar (and originally equal \citep{tononi_measure_1994}) to multi-information in \citet{tononi_information_2001,tononi_information_2004,balduzzi_integrated_2008,balduzzi_detecting_2011,oizumi_phenomenology_2014}. We have adopted it from these publications. Apart from \citet{balduzzi_detecting_2011} these publications are part of the integrated information theory approach which we will discuss further below.

It is worth mentioning that our choice of taking the minimum value of SLI found when evaluating all possible partitions of a set of random variables is not without alternatives. Another approach would be to take the (possibly weighted) average of all these values. This has been proposed by \citet{ay_information_2015} for the non-local multi-information.

In summary the measures of SLI and CLI which we introduce in this thesis are a combination of the idea of localisation of information theoretic measures by \citeauthor{lizier_local_2012} and the origins of integrated information theory by \citeauthor{tononi_information_2001}.

\section{Work related to our notion of entities}

\label{sec:entrelated}
\subsection{Spatiotemporal filtering and entities}
\label{sec:stpfiltering}
A basic notion in this thesis is that of an entity. At the most basic level the intuition behind this notion is that some spatiotemporal patterns are more important than others. This is also the problem of spatiotemporal filtering. We here discuss work that is similar on this most basic level and then indicate how $\ci$-entities essentially differ due to the problem of identity over time. 

Defining (and usually finding) more important spatiotemporal patterns or structures (also called coherent structures) has a long history in the theory of cellular automata and distributed dynamical systems. As \citet{shalizi_automatic_2006} have argued most of the earlier definitions and methods \citep{wolfram_computation_1984,grassberger_chaos_1984,hanson_attractor_1992,pivato_defect_2007} require previous knowledge about the patterns being looked for. They are therefor not suitable for a general definition of what entities are. More recent definitions based on information theory \citep{shalizi_automatic_2006,lizier2008local,flecker_partial_2011} do not have this limitation anymore. As argued above our method of identifying $\ci$-entities is also based on information theoretic notions similar to those used by \citet{lizier2008local}. Like the information based definitions in the literature it also requires no knowledge about the system or the patterns that are supposedly interesting. The main difference of our approach is again that it directly results in spatiotemporal patterns and does not go via an intermediate step of evaluating a measure / criterion time-step by time-step. This has certain advantages for our particular purpose. 

Applying any one of the definitions (or associated methods) proposed by \citep{shalizi_automatic_2006,lizier2008local,flecker_partial_2011} to the time-evolution (what we call a trajectory) of a cellular automaton assigns each cell (or group of cells) $j$  at each time $t$ a value (usually a real number, but can be discrete as for local statistical complexity in \citet{shalizi_automatic_2006}) that measures an important property of the current state of $(j,t)$ \footnote{In the case of local statistic complexity, the value is the causal state not only of the state at $(j,t)$ but of the state of the entire past light-cone. This makes no difference to the following argument however as the result is still just a (discrete) value at $(j,t)$.}. The result is then a ``filtered'' time evolution of the cells (or groups of cells) in the cellular automaton where each cell $j$ at time $t$ now takes its value of the measured property. These filtered time evolutions then highlight the important spatiotemporally extended structures like gliders and domains. However these methods make no claim about the identity of the revealed patterns. This means that there is no criterion given that tells us which cells and their values at time $t_1$ and which cells and their values at time $t_2$ are part of the same entity or object. For isolated gliders this may not seem like a problem but whenever gliders collide it is not clear whether they both loose their identity and become a new thing (or no thing) or whether one of them survived the collision and maybe just changed direction. These questions are not addressed by these publications since the problem of identity over time (or identity of entities in general) is not the focus of these publications. The goal of these publications is to quantify and identify emergent computation and coherent structures and not resolving the identity of entities that may be agents. In order to assign sequences of action and perceptions to entities (or structures) we have to be able to identify them over time. Our approach assigns a measure of integration (CLI) directly to groups of cells that are not only spatially but also temporally extended. We then select the spatiotemporal patterns that have a value above zero as the $\ci$-entities in a given time evolution. If gliders are such $\ci$-entities our approach could make clear whether and which gliders survive collisions.  

Note that it could be possible introduce criteria for identity over time via the measured values of the above publications. An example criterion would be to define a threshold and say that all cells whose measured values are above this threshold belong to one entity. However this would often lead to all highlighted structures to be identified as one entity and it is not directly obvious how to define a more detailed entity criterion.

With respect to the criteria for entities we propose in this thesis we find the following 
\begin{description}
  \item[Compositionality] Both spatial (e.g.\ in \citet{shalizi_automatic_2006}) and temporal compositionality can occur. 
\item[Degree of freedom traversal] Degree of freedom traversal can occur. The highlighted spatiotemporal patterns cross from one degree of freedom at one time to another degree of freedom at the next. Just like the gliders they capture.
\item[Counterfactual variation] Counterfactual variation can occur. The highlighted spatiotemporal patterns depend on the particular time-evolution (trajectory) of the system.
\item[Identity] Only spatial identity is defined e.g.\ in \citet{shalizi_automatic_2006}. Identity over time is not addressed. 
\item[Perception, action, goal-directedness]There is no intention to define these.
\end{description}

\subsection{Emergent coarse-graining}
The approach most closely related to our own approach and an important inspiration for our work is that of \citet{balduzzi_detecting_2011}. It proposes a method for coarse-graining the time evolutions (trajectories) of multivariate Markov chains. Using a cellular automaton as an example, the value $x_{j,t}$ of a cell $j$ at each time $t$ is represented by a random variable $X_{j,t}$. This is a common practice in information theoretic/stochastic conceptions of such systems \citep[e.g.][]{shalizi_automatic_2006,lizier_local_2012}, which we follow as well. Then, for a given time evolution ($x_{out}$ for \citeauthor{balduzzi_detecting_2011} and $x_V$ in our formalism), spatiotemporally extended groups of the random variables are combined to form units $\bm{U}_i$ of the coarse-graining $\mathcal{K}$. The coarse graining $\mathcal{K}$ is formed not only of units but also of ground $\bm{G}$ and channel $\bm{C}$. The ground can be related to driving variables (cf.\ the driven multivariate Markov chain \cref{def:drivenchain}) in our case whereas the channel has no analogue in our approach. Ignoring the channel, the coarse-graining is equivalent to a partition of the time evolution like those we investigate in \cref{ch:stp}. This means that the units resulting from the coarse-graining method can (by design) be spatiotemporally extended and could correspond to spatiotemporally extended entities that require no additional concept of identity over time. One difference is that the units are also random variables with an associated state space (the coarse-grained alphabet). In our case the $\ci$-entities have a fixed state for all random variables they occupy. They are not random variables themselves. We note that the approach of \citeauthor{balduzzi_detecting_2011} is then peculiar in the sense that it generates spatiotemporally extended and located coarse-grained random variables that depend on the particular time evolution of a system. This means that our argument against using sets of random variables as agents / entities (see \cref{sec:relatedstochpro,sec:entinmvmc}) does not apply to this approach.
 
The coarse-graining $\mathcal{E}_X(x_{out})$ that best describes the particular time evolution $x_{out}$ for a given system $X$ is also chosen in a way that exhibits some similarities with our approach. First, only \textit{emergent} coarse-grainings are considered. Emergent coarse-grainings satisfy two properties which are too involved to state concisely but which essentially ensure the following:
\begin{enumerate}
  \item Emergent coarse-grainings are special among the coarse-grainings with equal cardinality. This is makes them similar to refinement-free partitions at a particular disintegration level (see \cref{def:rfreedishier}).
\item Every unit in these coarse-grainings satisfies a particular condition with respect to its refinements. More precisely, it has more ``excess information'' than its refinements with respect to the units it is connected to. This is similar to the blocks of the refinement-free partitions at a disintegration level. These are locally integrated with respect to each of their refinements i.e.\ they have a positive CLI value.
\end{enumerate}
The emergent coarse-grainings are then conceptually somewhat related to the partitions in the refinement-free disintegration hierarchy. One difference is that the units are obtained by looking at how they are connected to other units. In our case we focus only on the internal connection of $\ci$-entities. It would therefore be surprising if the two approaches were measuring the same thing. At the same time it should be noted that excess information as defined by \citeauthor{balduzzi_detecting_2011} is a partially localised\footnote{Partially localised refers to measure where the averages over some of the random variables in an information theoretic measure are omitted but others are still taken \citep[see][]{lizier_local_2012}.} information theoretic measure that considers all possible partitions of the \textit{inputs} of a set of random variables. It is therefore closely related to CLI, which we use. The difference is that CLI partitions the random variables in a group/block/unit directly and not the input variables.

The best coarse-graining $\mathcal{E}_X(x_{out})$ is the one that maximises the ``excess information'' among all emergent coarse-grainings. A similar requirement could be made in our case by selecting the partition at the lowest level of the refinement-free disintegration hierarchy. We make no such final selection in this thesis but plan to investigate this further in the future. 

In summary the approach of \citeauthor{balduzzi_detecting_2011} has many parallels to our notion of entities (agent properties like actions, perception, or goal-directedness are not treated) and it would be interesting to investigate how the two approaches are related in detail. This will be future work. 

With respect to the criteria for entities we propose in this thesis we find the following 
\begin{description}
  \item[Compositionality] Both spatial and temporal compositionality can occur. The units are spatiotemporally defined.
\item[Degree of freedom traversal] Degree of freedom traversal can occur. Units can cross arbitrarily from one degree of freedom at one time to another degree of freedom at the next.
\item[Counterfactual variation] Counterfactual variation can occur. The units depend on the particular time-evolution (trajectory) of the system.
\item[Identity] Both spatial and temporal identity are defined in a unified way.
\item[Perception, action, goal-directedness] There is no intention to define these.
\end{description}

\subsection{Integrated information theory}

Integrated information theory \citet{tononi_information_2001,tononi_measuring_2003,tononi_information_2004,balduzzi_integrated_2008,oizumi_phenomenology_2014} is an attempt to develop a measure of consciousness of physical configurations. Similar to our setting it is defined for the setting of multivariate Markov chains. While the main focus of this theory is consciousness it becomes conceptually related to our work if it is slightly reinterpreted. One of its main goals is to quantify the unity of conscious experiences. In \citet{tononi_information_2004} the authors also mention that informationally integrated sets form ``entities'' (also called complexes) that have ``ports-in'' and ``ports-out'' to connect to parts that are not within the entity. This is very similar to the program of this thesis which is to establish a formal definition of acting and perceiving entities. As far as we know there is no formal definition of these ports-in and ports-out and what constitutes perceptions and actions of them.
In its modern formulation \citep{oizumi_phenomenology_2014} IIT measures the IIT-integration of all spatial patterns $x_{A_t}:=(x_{j,t})_{(j,t) \in A_t}$ with $A_t \subset V_t$ \footnote{We write $V_t$ for all random variables in the system at time $t$.} at some time-step $t$. For this all possible partitions of the parent and child nodes in the multivariate Markov chain are evaluated and the minimal value is used to define the IIT-integration of the spatial pattern. This leads to the most integrated patterns which are called complexes. Like in the case of spatiotemporal filtering, no criterion is given as to what patterns at $t_1$ and what patterns at $t_2$ belong to the same spatiotemporally extended pattern (or complex). The problem of identity over time is then not solved in this publication. However, the authors refer to \citet{hoel_quantifying_2013} when mentioning that the spatial patterns should be evaluated over optimal ``grains''. In \citet{hoel_quantifying_2013} a method is presented which coarse-grains multivariate Markov chains spatiotemporally. This means that multiple random variables at multiple times are grouped together to form new coarser random variables. Unlike in \citet{balduzzi_detecting_2011} these coarse-grainings are not dependent on the particular time-evolution of the chain. They do, however, create also temporally extended structures (random variables) and can therefore be seen to solve the problem of identity over time. If IIT is now used on these coarse-grained random variables we again find the IIT-integrated spatial patterns which are now also temporally extended since the coarse-grained variables are themselves temporally extended on the underlying (not coarse-grained) level. This would lead us to a notion of entity where each entity is a coarse-grained ``spatial'' pattern that is based on temporally extended underlying patterns. Each such entity / coarse pattern would then correspond to a set of underlying spatiotemporal patterns.  

With respect to the criteria for entities we propose in this thesis we find the following 
\begin{description}
  \item[Compositionality] Both spatial and temporal compositionality can occur.
\item[Degree of freedom traversal] Degree of freedom traversal can occur. The method by \citet{hoel_quantifying_2013} can create coarse-grained random variables lumping together variables at different times and that belong to different degrees of freedom.
\item[Counterfactual variation] A restricted kind of counterfactual variation can occur. IIT evaluates spatial patterns which are values of random variables and therefore change from one time evolution to another. However, we cannot have both degree of freedom traversal and counterfactual variation of the degree of freedom traversal. This means we cannot have full counterfactual variation in extent. More precisely, assume we have two binary degrees of freedom and look at two time steps. Then we have the random variables $\{X_{1,t_1},X_{2,t_1},X_{1,t_2},X_{2,t_2}\}$. Say the coarse-graining selects the two variables $X_{1,t_1},X_{2,t_2}$ at different times to form a coarse-grained variable $Y$ then the underlying spatiotemporal patterns exhibit degree of freedom traversal (they switch from the first to the second degree of freedom). These can be identified as IIT integrated if $Y$ is integrated by itself (this is possible). However, now that $Y$ is fixed there can be no entity that does not traverse the degrees of freedom e.g.\ one occupying only $X_{1,t_1}$ and $X_{1,t_2}$ since these are not together part of a coarse-grained variable and if they are joined via IIT then they must always include all of $Y$ since $X_{1,t_1}$ is part of $Y$. This means that the coarse-graining restricts the possible counterfactual variation in extent.
\item[Identity] Spatial identity is realised by IIT. The coarse graining realises both spatial and temporal identity. The two kinds of identity are therefore not treated in the same way.
\item[Perception, action, goal-directedness] There are no formal definitions for these. Parts of the investigated network are sometimes defined as sensor and actuator variables \citep{albantakis_evolution_2014} but in that case the whole network is the ``brain'' of a animat and not a general universe or biosphere like system.
\end{description}

%

\subsection{Kolmogorov complexity of patterns}
Recently \citet{zenil_two-dimensional_2015} have proposed a method of evaluating spatiotemporal patterns directly (instead of concatenating spatial patterns) by approximating the Kolmogorov complexity. They evaluate 2D patterns (one time and one space dimension) according to the (algorithmic) probability that they are generated by a 2D Turing machine. 


The algorithmic probability of one of the patterns is the number of 2D Turing machines that generate the pattern divided by all halting 2D Turing machines. The (Kolmogorov) complexity is then estimated as the self-information (negative logarithm) of this probability. This results in a very general measure for the complexity of patterns. For the purpose of this thesis this approach is too general. We want to explicitly evaluate patterns according to the dynamical laws that generate them i.e.\ we want to find the spatiotemporal patterns that can be agents within particular multivariate Markov chains. From our point of view some patterns that are agents in one multivariate Markov chain could well be an arbitrary pattern in another chain. If the patterns look the same however the approach of \citeauthor{zenil_two-dimensional_2015} will ascribe the same value to them independent of the underlying dynamics of the system.
%
It is therefore not applicable to our problem. 

With respect to the criteria for entities we propose in this thesis we find the following 
\begin{description}
  \item[Compositionality] Both spatial and temporal compositionality can occur.
\item[Degree of freedom traversal] Degree of freedom traversal can in principle be evaluated. In the present version however only rectangular patterns are treated this is means there are no degree of freedom traversals.
\item[Counterfactual variation] Counterfactual variation can occur. All occurring patterns in a trajectory can be evaluated and these differ in general from trajectory to trajectory. The problem is that all patterns have the same value across all systems/multivariate Markov chains.
\item[Identity] Both spatial and temporal identity are defined in a unified way.
\item[Perception, action, goal-directedness] There is no intention to define these. 
\end{description}

\section{Work related to our definition of agents (and entities)}
\subsection{Interacting stochastic processes as agents or entities}
\label{sec:relatedstochpro}
In the literature it is common to model agents as stochastic (including deterministic) processes interacting with an environment. 
In its most general formulation this view assumes that at each time $t$ there is a random variable $M_t$ that represents the agent (or its ``memory'') and a random variable $E_t$ that represents the environment. Interactions can then be modelled via conditional probabilities (see \cref{sec:paloopformal}). This is also a discretised version of interacting dynamical systems as proposed by \citet{beer_dynamical_1995} to model agent and environment. Furthermore, this model includes as an important subclass the Markov decision problems and partially observable Markov decision problems \citet{tishby_information_2011}. Note that in cases of the Markov decision problems the agent memory $M_t$ is often not explicitly modelled but is implicitly assumed to be a part of the system. In the perception-action loop setting various features of agents have been formally investigated. Examples include learning (e.g.\ reinforcement learning) \citep{sutton_reinforcement_1998}, empowerment \citep{klyubin_empowerment_2005,anthony_impoverished_2009}, informational closure \citep{bertschinger_information_2006}, autonomy \citep{bertschinger_autonomy_2008,seth_measuring_2010}, digested information \citep{salge_digested_2011}, self-organisation \citep{ay_information-driven_2012}, thermodynamics of prediction \citep{still_thermodynamics_2012}, morphological computation \citep{zahedi_higher_2010,zahedi_quantifying_2013}, and individuality \citep{krakauer_information_2014}.

In this thesis we deliberately do not assume that there is a random variable $M_t$ at each time $t$ which corresponds to an agent. Neither do we assume that there is an environment random variable at each time $t$. We take a multivariate Markov chain whose state at each point in time $t$ is represented by a (finite) set of random variables $\{X_{j,t}\}_{j \in J}$. Whether there exists an agent (or even an entity) at that time is left open. Furthermore, even if there exists an agent at time $t$ it may only exist at time $t$ in one particular time-evolution or trajectory of the system. In another trajectory there might again be no agent at time $t$ or there might be one occupying a different subset of the random variables than in the first case. These situations are not modelled by the perception-action loop framework. They have been ignored or modelled away in order to focus on different aspects of agents. The success of this approach justifies this choice. Since we are interested in a fundamental and general definition of agents in multivariate Markov chains we cannot follow this choice. We will argue this in more detail in \cref{sec:entinmvmc}. 

Since our definitions also accommodate systems where there is an agent and environment at every time step we will also connect our approach to the perception-action loop after we have defined actions and perception in \cref{sec:palooprelation}. In the future we hope that our work contributes to the extension of the work cited above to the more general setting treated here. We take some steps in this direction by generalising perception and action but more detailed investigations are needed to see whether these notions are sufficient.

Among the work cited above we will make use of and are also generally inspired by the fundamental work on the \textit{autonomy} of agents in perception-action loops by \citet{bertschinger_autonomy_2008}. This work has more recently been extended in \citep{krakauer_information_2014} where it is proposed as the basis of a method to detect the random variables that represent a biological individual at some time $t$. Note that also in this newer work, unlike in our case, the individual/agent is assumed to be represented by a set of random variables (not their values) and it is assumed that it is represented by the \textit{same} set of random variables in every time-evolution. Nonetheless, the underlying ideas of \citeauthor{bertschinger_autonomy_2008} autonomy namely non-heteronomy and self-determination both reappear in our conception of agents. The role of self-determination which refers to the influence of the agent's state at one time on its state at a subsequent time is played by the requirement of integration of the $\ci$-entities. We only consider patterns as candidates for agents if their parts are interrelated according to complete local integration. The role of non-heteronomy, which requires that the environment state does not determine the agent's next state is played by our notion of entity action. We relate this notion of action to \citeauthor{bertschinger_autonomy_2008}'s measure of non-heteronomy in \cref{sec:palooprelation}.


We also note here that within the formalism of reinforcement learning and in response to the definition of universal intelligence by \citet{legg_universal_2007} \citet{orseau_space-time_2012} have argued against the assumption that the agent's random variable $M_t$ (which in this case is seen as the memory/tape of a Turing machine) is guaranteed to exist. The idea there is that in a more realistic setting the environment can also overwrite the agent's memory. They conclude that in the most realistic case there only ever is one memory that the agent's data is embedded in. These arguments for a single system and a blurred boundary between agent and environment then lead to similar conclusions as our arguments for spatiotemporal patterns as entities (that can be agents) in \cref{ch:agents}. 

Speculating at the end \Citeauthor{orseau_space-time_2012} propose (also in the setting of cellular automata) to define a utility function which is $1$ as long as some chosen ``heart'' pattern exists and $0$ otherwise. The agent is then not further specified but supposed to protect the heart pattern against destructive influence and accordingly regarded the longer it succeeds. The only choice possible is that of the initial condition. We agree that the only choice is the initial condition but a prior choice of a pattern that must be maintained does not seem in accordance with our viewpoint here. Here the kinds of patterns that constitute agents depend on the dynamics of the system / multivariate Markov chain. It is possible that in some systems having some form of ``heart'' pattern (a better analogy might be a ``gene'' pattern) turns out to be just what agents need to persist. This, however, would be a consequence of the dynamics of the chain again and the gene pattern would be the gene pattern under those dynamics and not one that can be chosen externally. The only way we see to make sense of using a ``heart'' pattern is in a setting where finding the dynamics of the system that preserve it for the maximum amount of time is the goal. This however seems trivial to achieve with dynamics that leave every cell fixed. So for an definition of agents in our setting this approach does not seem to work.

\Citeauthor{orseau_space-time_2012} then also pose it as open questions in the ``one memory'' setting what part of the system the agent is and what an agent is (where the boundaries between agent and environment are). 
%
This thesis is also an attempt to contribute to the answers to these questions.


With respect to the criteria for entities we propose in this thesis we find the following 
\begin{description}
  \item[Compositionality] Both spatial and temporal compositionality can occur. The random variable $M_t$ can be composed out of multiple random variables and it has multiple time-steps.
\item[Degree of freedom traversal] Degree of freedom traversal is possible. The random variable $M_t$ can be defined to correspond to a different set of random variables at each time \citet[e.g.][]{krakauer_information_2014}. 
\item[Counterfactual variation] Counterfactual variation cannot occur. If the entity is a set of random variables then it is always the same set and only the values change.
\item[Identity] Usually in the perception-action loop both spatial and temporal identity are given without any justification. \Citet{krakauer_information_2014} have dropped the spatial assumption and search for the right spatial composition of the individual. Both, spatial and temporal identity could be defined via the coarse-graining method by \cite{hoel_quantifying_2013}. However, no claims have been made that these coarse-grained variables have anything to do with agents.
\item[Perception, action] Perception and action are implicitly defined as the interactions between the agent and the environment.
\item[Goal-directedness] Goal-directedness is ongoing research. \citet{bertschinger_autonomy_2008} note that their notion of non-trivial informational closure indicates that the agent has some information about the environment or even models it. This may be related to-goal directedness. Another route is to take cues from inverse reinforcement learning \citep{ng_algorithms_2000} or work on inferring intentions \citep[e.g.][]{pantelis_inferring_2014}. 
\end{description}

\subsection{Autopoiesis and cognition in the game of life}
\subsubsection{Autopoiesis and entities}
In \citet{beer_characterizing_2014} the author constructs an account of spatiotemporal patterns in the game of life cellular automaton based on the ideas of Maturana and Varela \citep{varela_principles_1979,maturana_autopoiesis_1980}. This can be seen as a definition of entities. Moreover it defines entities as spatiotemporal patterns and therefore the set of all such entities may constitute an entity-set in the sense of our \cref{def:entset}. This would make it a direct alternative to our own notion of $\ci$-entities. 

The construction of the entities proceeds roughly as follows.
First the maps from the Moore neighbourhood to the next state of a cell are classified into five classes of \textit{local processes}. Then these are used to reveal the dynamical  structure in the transitions from one time-slice of a spatiotemporal pattern to the next. The used example patterns are the famous block, blinker, and glider and are considered including their temporal extension. Using both the processes and the spatial patterns/values/components (the black and white values of cells are called components) networks characterising the organisation of the spatiotemporally extended patterns are constructed. These can then be investigated for their organisational closure. This is defined to occurs if the same process component relations as before reoccur at a later time. Boundaries of the spatiotemporal patterns are identified by determining the conditions necessary for the reoccurence of the organisation.  

\Citeauthor{beer_characterizing_2014} mentions that the current version of this method of identifying entities has its limitations. If the closure is perturbed or delayed and then recovered the entity still looses its identity according to this definition. Two possible alternatives are also suggested. The first is to define the \textit{potential for closure} as enough for the ascription of identity. This is questioned as well since a sequence of perturbations can take the entity further and further away from its ``defining'' organisation and make it hard to still speak of a defining organisation at all. The second alternative is to define that the persistence of any organisational closure indicates identity. It is suggested that this would allow blinkers to transform to gliders. 

We note that our definition of $\ci$-entities does not need similar choices to be made since it is not based on the reocurrence of any organisation. As mentioned before, it takes entire spatiotemporal patterns and evaluates their integration. It is then possible that later time-slices of $\ci$-entities have no organisational similarity to earlier ones. This is most similar to the latter proposal where blinkers can transform to gliders. However, in our case not even the blinker or the glider would necessarily need to exhibit a reoccurring organisation explicitly.
%
%

It still seems to us that any of the choices proposed by \citeauthor{beer_characterizing_2014} may be used to construct an automatic way to identify autopoietic patterns as a kind of special patterns or entities. By automatic we mean that no knowledge of the structures we are looking for is necessary. Ignoring computational issues again it may be possible to search through all spatiotemporal patterns and look for closures. Once we find a closure we could try to reconstruct the associated boundaries and thereby obtain complete entities. It is not stated in the paper whether this is possible in principle. If we assume it is then the resulting set of entities is a set of spatiotemporal patterns and therefore an entity-set according to our \cref{def:entset}. This means our own definitions of action and perception could be applied to these autopoietic entities. This is not surprising since \citet{beer_cognitive_2014} defines a closely related notion of perception himself. This will be discussed next.

\subsubsection{Cognitive domain and perception}
%
%
%
%

In \citet{beer_cognitive_2014} the author constructs \citep[again following][]{varela_principles_1979,maturana_autopoiesis_1980} the cognitive domain of the glider in the game of life. Our concept of perception can be seen as a generalisation not only of perception in the perception-action loop but also as a generalisation of the cognitive domain in this publication.

To get the cognitive domain \citet{beer_cognitive_2014} employs a series of concepts that have analogues in our definition of perception. The glider is defined as an autopoietic entity in the sense of \citet{beer_characterizing_2014}. This is also a spatiotemporally extended entity $x_A$ in accordance with our definition.
First he defines the \textit{microperturbations} $\mathcal{P}$ of the glider. These are the possible states of the boundary around the glider. The set of microperturbations can also be restricted to the nondestructive perturbations. This means those were the glider does not die at the next time-step. The role of the nondestructive microperturbations that preserve the glider identity is played in our case by the co-perception environments $\perstp(x_A,t)$ of entity $x_A$ at time $t$. 
The set of microperturbations are then classified according to the induced next state of the glider (including its death state if destructive perturbations are allowed\footnote{We do not use the death state since we don't allow destructive environments.}). This results in a set of equivalence classes called the \textit{macroperturbations}. In our case these equivalence classes are the \textit{perceptions} of the entity $x_A$ which are the blocks of the \textit{co-perception partition} $\pi^\perstp(x_A,t)$. In \citet{beer_cognitive_2014} the cognitive domain $\mathcal{C}$ is the collection of all macroperturbations of all possible glider states. In our formalism the cognitive domain $\mathcal{C}(x_A)$ of an entity $x_A$ would be the set of all perceptions that occur along the the time-slices of a given entity $x_A$:
\begin{equation}
  \mathcal{C}(x_A):=\{b \in \pi^\perstp(x_A,t): A_t, A_{t+1}\neq \emptyset\}
\end{equation}  
 where the condition $A_t, A_{t+1}\neq \emptyset$ just picks the times where the entity exists at $t$ and $t+1$. If it doesn't exist at $t+1$ then it cannot perceive anything about the environment at $t$.

%
%
%

The cognitive domain in \citet{beer_cognitive_2014} is defined for the autopoietic entities \citet{beer_characterizing_2014}. Via its macroperturbations it contains a notion of perceptions which is suitable for systems/entity sets that do not contain agents in every trajectory. Recall that this was not the case for perception in the perception action loop. In this thesis we present a generalisation of \citet{beer_cognitive_2014}'s macroperturbations to arbitrary entity sets in arbitrary (possibly stochastic) multivariate Markov chains. This reveals the requirement of \textit{non-interpenetration} for entity sets, which allows uniquely defined perception and, accordingly, uniquely defined cognitive domains. Finally we connect the general notion of perception to the perception-action loop setting. This means we also expose a connection between \citet{beer_cognitive_2014} and the perception-action loop.

Our work on perception can therefore be regarded as an extension of the cognitive domain notion proposed in \citet{beer_cognitive_2014}.

\subsubsection{Summary}
With respect to the criteria for entities we propose in this thesis as well as action, perception and goal-directedness we find the following:
\begin{description}
  \item[Compositionality] Both spatial and temporal compositionality can occur (see the glider).
\item[Degree of freedom traversal] Degree of freedom traversal is possible (see the glider).
\item[Counterfactual variation] Counterfactual variation can occur (see different gliders in different time-evolutions / trajectories).
\item[Identity] Temporal identity is defined via the closure condition. Spatial identity is then derived from there. There are some choices left to make. So the notion is not yet unique.
\item[Perception] Perception is defined via the macroperturbations. Our notion is a generalisation to arbitrary entities and stochastic settings.
\item[Action] There is no explicit definition of action in this work. However, it is mentioned that the sequence of the entity's time-slices i.e.\ its ``behavioural trajectory'' would be interpreted as actions by an observer. This is compatible with our notion even if we make an additional explicit requirement. \citeauthor{beer_characterizing_2014} does not require that there must be different possible next time-slices given the \textit{same} environment. Without this requirement the connection of actions to autonomy that we obtain in this thesis is lost. We note that the glider according to our definition can perform an action. 
\item[Goal-directedness] There is no notion of goal-directedness defined. 
\end{description}

\subsection{Life as we know it}
\Citet{friston_life_2013} argues that life is an emergent property of some dynamical systems and that the emergent living organisms are characterised by Markov blankets. Since living organisms are the primary examples of agents we can focus on implicit properties of agents. The Markov blankets define the entities in this publication. This works in the following way. We are given a particle like system \footnote{Particles are referred to as subsystems in the original, we deviate from this terminology here. The notation is the original however.} with (each two dimensional) position $p^{(i)}$ and velocity $p^{'(i)}$ for each particle $i$. The particles also have inner degrees of freedom $\tilde{q}^{(i)}$ but they play no role in the entity definition (they do in persistence etc.). The particles positions and velocities obey some equations of motion that involve the inner degrees of freedom $\tilde{q}$ but apart from this model particles with some friction term in a potential well. The particles only ever interact if they are closer to each other than some threshold (which happens to be equal to $1$). From the position of these particles a time dependent adjacency matrix $A(t)$ is derived. The matrix entry $A_{ij}(t)$ is set to $1$ if the particles $i$ and $j$ were closer than the threshold within a time-window (of length $256$ seconds) preceding $t$. The matrix $A(t)$ is then used to find the Markov blanket. This is done by constructing the Markov blanket matrix $B(t):=A(t)+A(t)^T+A(t)^TA(t)$ where $C^T$ denotes the transpose of $C$. At each point in time $t$ the eigenvectors of the Markov blanket matrix $B(t)$ are then calculated. The eigenvector with the largest eigenvalue then contains positive real numbers and indicates in how far the according particle is part of the most interconnected cluster. The particles with the largest $k=8$ values were then picked to be the internal states i.e.\ the inside of the entity. So setting $k=8$ is arbitrary. If we now construct the vector $\chi= (\chi_1,...,\chi_{128})$ such that $\chi_i =1$ if $i$ is one of the internal particles (and $\chi_i =0$ otherwise) the matrix product $A(t) \chi$ will indicate the children, $A(t)^T \chi$ the parents, and $A(t)^T A(t) \chi$ the parents of the children of the internal particles according to the adjacency matrix $A(t)$.

Now let us define:
\begin{itemize}
  \item $\Lambda(t) \subset \{1,...,128\}$ as the set of internal particles at time $t$,
  \item $B(t)\subset \{1,...,128\}$ as the set of children of the internal particles, the particles of the ``active states''\footnote{In the original these are denoted by $A$ but this would be confusing here.},
  \item $S(t)\subset \{1,...,128\}$ as the set of parents of the internal particles, the particles of the ``sensory states'',
\item $\Psi(t)\subset \{1,...,128\}$ as the rest, the particles of the ``external states''.\footnote{It is not clear to us which set the parents of the children $A(t)^T A(t) \chi$ are supposed to belong to. It is probably either the action states or the sensory states. We ignore them as where they belong to does not affect the reasoning here.} 
\end{itemize}

Let us consider position and velocity $\tilde{p}^{(i)}=(p^{(i)},p^{'(i)})$ and internal degrees of freedom $\tilde{q}^{(i)}$ of each particle $i$ together as one variable $x_i:= (\tilde{p}^{(i)},\tilde{q}^{(i)})$ and let us define for each time $t$ the random variable $X_{i,t}$ to represent the value of $x_i$ at time $t$. Then an entity is defined at each time $t$ by:
\begin{itemize}
  \item the internal states $x_{\Lambda(t),t}:= (x_{i,t})_{i \in \Lambda(t)}$,
  \item the active states $x_{B(t),t}:= (x_{i,t})_{i \in B(t)}$,
  \item the sensory states $x_{S(t),t}:= (x_{i,t})_{i \in S(t)}$. 
\end{itemize}
Together these form a spatiotemporal pattern $(x_{\Lambda(t),t},x_{B(t),t},x_{S(t),t})$ in accordance with our definition. 

This results in temporally changing entities that can be different from time-evolution to time-evolution. Note that, since $k=8$ particles are always chosen as the internal states there is always exactly one entity at each time $t$. If no particles interacted with these $k=8$ particles in the current time window then there are no active or sensory states. So an entity is an agent if such interactions happen. 

Importantly, the problem of identity over time is not addressed. A single Markov blanket is calculated at each time-step this of course leads to a uniquely defined next Markov blanket but why the two share some identity is not argued. In cases where the interconnections between one set of $k=8$ particles weakens while those of another set of external particles strengthens there can be a discontinuous jump of all internal states at some point. This would still be seen as the time-evolution of the one Markov blanket in the system. Even if multiple Markov blankets were defined there is no proposed method of discerning what would happen in edge cases where Markov blankets collide for example.

In summary this means that the entities of this approach can \textit{vary counterfactually} (they can occupy different random variables from one trajectory to another \cref{def:cfvar}) can traverse degrees of freedom but do not solve the problem of identity over time. They are therefore entities that are more general than the ones in the perception-action loop but less general than some spatiotemporal pattern based entities like the autopoietic entities of \citet{beer_characterizing_2014} or our $\ci$-entities.


So much for the entities of this approach. We now discuss perception and action. The condition that interactions have to occur between internal states and other states for there to perception is also contained in our own notion of perception. In our definition whenever there is an influence by the environment on an entity the entity perceives something. This is also the case here. If there is influence by some non-internal states on the internal states these states are defined to be sensory states. This happens since the adjacency matrix will indicate this interaction. Then the set of sensory states becomes non-empty. However, the notion of perception in \citet{friston_life_2013} includes more than just the existence of sensory states. If the active states are also non-empty then it is argued that the internal states will ``appear to solve Bayesian inference about the external states''. This is the notion of perception in this publication. It is a more ambitious and higher-level notion of perception than the merely influenced based notion we propose. We note that our notion of perception is not necessarily carried by a set of sensory states (we would say sensory random variables) i.e.\ in our case there are no random variables that explicitly represent the sensor values/perceptions. Perceptions are a classification of the environment which may or may not be explicitly represented in the system by random variables.

The action states of \citeauthor{friston_life_2013} also have no direct analogue in our concept of action. Our concept of action is deliberately weak and does not even require an influence on the environment. It only requires that the environment does not determine the next state of the entity (non-heteronomy). This condition is also met by \citeauthor{friston_life_2013}'s concept of action states as he states that the ``flow of action states does not depend on external states''. So if we consider the action states as part of the entity (which is also done by \citeauthor{friston_life_2013}) our notion of actions is exhibited by such entities. Conversely, our entities can exhibit actions without having action-states such that our notion of action is weaker than \citeauthor{friston_life_2013}'s.

In contrast to this thesis \citeauthor{friston_life_2013} also argues that the Markov blanket entities if they have sensory and action states exhibit a kind of goal-directedness. He states that ``action will appear to conserve the structural and dynamical integrity'' of the Markov blanket. If this is true then we have entities, perception, action, and even goal-directedness (the conservation of integrity) and therefore a complete agent definition.  

Evaluating these claims is still an open research question.
 
With respect to the criteria for entities we propose in this thesis as well as action, perception and goal-directedness we find the following:
\begin{description}
  \item[Compositionality] Both spatial and temporal compositionality can occur for the Markov blanket entities.
\item[Degree of freedom traversal] Degree of freedom traversal can occur (and does so in the paper). The Markov blanket contains different particles at different times. 
\item[Counterfactual variation] Counterfactual variation can occur but is restricted. The Markov blankets change from trajectory to trajectory. In the current version, there is always exactly one such entity however. For agents we can say that there is no agent if the internal states of the $k=8$ internal particles don't interact with any other particles. More than one agent is not defined currently. 
\item[Identity] Only spatial identity is treated by the Markov blanket. The identity over time is not addressed.
\item[Perception] Perception is defined as Bayesian inference about the external states. According to the author it occurs whenever there is a Markov blanket with active states and sensory states. 
\item[Action] Action is defined via the active states which flow independently of the external states.
\item[Goal-directedness] Goal-directedness is emergent since the actions ``appear to conserve the structural and dynamical integrity'' of the Markov blanket.
\end{description}

\chapter{Formal background}
\label{ch:fb}
\label{formal1}
In this section the formal background that will be used throughout the thesis is presented. It is assumed the reader is at least vaguely familiar with 
\begin{itemize}
\item 
elementary probability theory of discrete random variables, and
\item Bayesian networks.
\end{itemize}
We will nonetheless present many basic definitions for quick reference and to clarify our notation. For the most part we tried to stick to standard notation of probabilities. However the section on symmetries of spatiotemporal patterns (\cref{sec:symm}) requires a more elaborate, if more basic, notation than usual. This will be introduced in \cref{sec:rvandpat} and related to more standard notation.   

In \cref{sec:polapa} we first recall the definitions of partially ordered sets (\textit{posets}) and special posets called \textit{lattices}. Then we look at \textit{partitions} and the \textit{partition lattices} they form. Partitions will be used in multiple ways in this thesis. They are a basic concept in our definitions of specific local interaction in \cref{sec:slicli}, of $\ci$-entities in \cref{sec:entdef} and of entity perception in \cref{sec:perceptions}. The partition lattice is also needed for the definition of $\ci$-entities and forms the underlying structure of the disintegration theorem (\cref{thm:disintegration}) which is one of the main contributions of this thesis.

In \cref{sec:rvandpat} we introduce our notation for random variables and probability distributions. We also introduce the notion of \textit{patterns} which is equivalent to the notion of \textit{spatiotemporal patterns}\footnote{Spatiotemporal patterns are just patterns in systems where notions of time and space are defined.}. This simple notion is fundamental to this thesis. Under the name of cylinder sets it is a well known notion that usually does not play a dominant role. In this thesis it is the basis for the formal as well as the conceptual part. 

In \cref{sec:bn} we define \textit{Bayesian networks} which are a generalisation of \textit{multivariate Markov chains}. Most of the theorems (including the disintegration theorem and the SLI symmetry theorem) in \cref{sec:slicli} hold for Bayesian networks in general and not only for multivariate Markov chains. Multivariate Markov chains are the systems used in the conceptual part of this thesis. There they represent universes or geospheres that may contain agents. A special kind of multivariate Markov chains are the driven multivariate Markov chains which are relevant for some applications that involve external influences on a system (e.g.\ heat baths). In \cref{sec:slicli} we prove \cref{thm:drivensymm}, which can be used to relate the symmetries of such a driven multivariate Markov chain to the transformation of specific local integration. \Cref{sec:bn} also presents the definition of the \textit{perception-action loop} and the \textit{extended perception-action loop}. We will frequently refer to the perception-action loop in the conceptual part since it is a common way to model agents formally. Furthermore, the method of extracting of perceptions from the perception-action loop to get to the extended version is the starting point for our own definition of entity perception in \cref{sec:perceptions}. The proof we present there that the extracted perceptions (or sensor-values) and actions capture all interactions between agent and environment process also supports the interpretation of perception as the total of all influences of the environment on the agent.

This chapter contains no original contributions and only provides the vocabulary and notation for the following. We are not aware of a reference for the proof of \cref{thm:paloopext} concerning the extraction of perceptions and actions from the perception-action loop. However, we are sure that it is well known among researchers in the field.

%
%
%

\section{Posets, lattices, and partitions}
\label{sec:polapa}

Here we introduce the terminology of partially ordered sets (short: posets) and state some facts without proofs. We also present the definition of a lattice as a special poset. For a more thorough treatment as well as proofs we refer to \citet{gratzer_lattice_2011}. 

In this thesis, we will mostly use the partial order of ``refinement'' to relate ``set partitions'' to each other. These will be defined in \cref{sec:partitions}. All set partitions of a set will turn out to form a lattice. However, we will sometimes look at a subsets of all set partitions and the posets they form. For this reason we start with the more general notion of posets.

\subsection{Partially ordered sets and lattices}

\begin{mydef}[Partial orders and posets]
\label{def:partialorder}
  A \textit{partial order} $\preceq$ on a set $A$ is a binary relation that is 
  \begin{thmlist}
  \item reflexive: $\forall a\in A$:$a\preceq a$, 
  \item antisymmetric: $\forall a,b \in A$: if $a\preceq b$ and $b\preceq a$ then $a=b$, and 
  \item transitive: $\forall a,b,c \in A$: if $a \preceq b$ and $b\preceq c$ then $a \preceq c$.
  \end{thmlist}
  $A$ together with the partial order $\preceq$ forms a \textit{partially ordered set} or shorter a \textit{poset}.
\end{mydef}
Remarks:
\begin{itemize}
  \item Technically, a poset is a tuple $\langle A, \preceq \rangle$ where the partial order is explicitly specified. We will usually call $A$ a poset and imply the partial order if it is clear from context.
  \item A partial order is partial in the sense that there may be elements $a,b \in A$ such that neither $a \preceq b$ nor $b \preceq a$. In the total order defined next this possibility is excluded. 
\end{itemize}

\begin{mydef}
  A total order on a set $A$ is a partial order on $A$ such that for all $a,b \in A$ either $a \preceq b$ or $b \preceq a$.
\end{mydef}
$A$ together with a total order $\preceq$ forms a \textit{totally ordered set}.
\begin{itemize}
  \item The natural numbers together with the usual less or equal relation $\leq$ are a totally ordered set.
\end{itemize}

\begin{mydef}[Minimal and maximal elements]
\label{def:minandmaxelements}
  A minimal element of a poset $A$ with partial order $\preceq$ is an element $a \in A$ such that if for any $b \in A$ we have $b \preceq a$ then $b =a$. Conversely, a maximal element is an element $c \in A$ such that if for any $b \in A$ we have $c \preceq b$ then $b =c$.
\end{mydef}
Remarks:
\begin{itemize}
\item A minimal element then has no lesser element within the poset and a maximal element has no greater element within it.
\item There can be multiple minimal and maximal elements in a poset.
\item Minimal and maximal elements should not be confused with least and greatest elements which we define next.
\end{itemize}



\begin{mydef}
\label{def:lunit} \label{def:lzero} Given a poset $A$ an element $a \in A$ is called a \textit{least element} if for all $b \in A$, $a \preceq b$. We then denote $a$ by $\lzero$. An element $c \in A$ is called a \textit{greatest element} if for all $b \in A$, $b \preceq c$. We then denote $c$ by $\lunit$. 
\end{mydef}
Remarks:
\begin{itemize}
\item The least element is lesser than all elements in the poset and the greatest element is greater than all elements.
  \item If they exist, least and greatest elements are unique.
  \item A least (greatest) element is always a minimal (maximal) element but not vice versa.
\end{itemize}

\begin{mydef}[Boundedness]
  A poset $A$ is bounded if it has both a least and a greatest element.
\end{mydef}

\begin{mydef}[Covering relation]
\label{def:cover}
  Given a two elements $a,b \in A$ of a poset $A$ with $a \neq b$ we say \textit{$b$ covers $a$} and write $a \preceq: b$ if there is no $c \in A$ with $a \neq c \neq b$ such that $a \preceq c \preceq b$. 
\end{mydef}
Remarks:
\begin{itemize}
  \item If $b$ covers $a$ then $b$ is greater than $a$ and there is no element in between the two.
  \item If $A$ is finite then knowing all covering relations determines the partial order of $A$ completely \citep[p.6]{gratzer_lattice_2011}.
  \item The join, which is defined below, can be seen as the generalisation of the case where one element covers another to the case where one element ``covers'' a whole set of elements.
  \item The covering relation is helpful for the construction of Hasse diagrams of posets. These provide an informative visual impression of partial orders and are introduced next.
\end{itemize}

\begin{mydef}[Hasse diagram]
\label{def:hassediagram}
  A \textit{Hasse diagram} is a visualisation of a poset. Given a poset $A$ the Hasse diagram represents the elements of $A$ by dots. The dots representing the elements are arranged in such away that if  $a,b \in A$, $a \neq b$, and $a \preceq b$ then the dot representing $a$ is drawn below the dot representing $b$. An edge is drawn between two elements $a,b \in A$ if $a \preceq: b$ i.e.\ if $b$ covers $a$. If edges cross in the diagram this does not mean that there is an element of $A$ where they cross and edges never pass through a dot representing an element.
\end{mydef}
Remarks:
\begin{itemize}
  \item No edge is drawn between two elements $a,b \in A$ if $a \preceq b$ but not $a \preceq: b$. 
  \item Only drawing edges for the covering relation does not imply a loss of information about the poset since the covering relation determines the partial order completely (see remark to \cref{def:cover}).
  \item For some example Hasse diagrams see \cref{fig:hassediagrams}
\end{itemize}

\begin{figure}[ht!]
\input{./images/hassediagrams.tex}
\caption{%
        Hasse diagrams of three different posets. \subref{fig:two} Two element poset with $\lzero \preceq \lunit$. This poset is bounded because it has a zero $\lzero$ and unit $\lunit$. \subref{fig:three} A bounded poset with five elements. Note that neither $a \preceq b$ nor $b \preceq a$, so this is a poset but not a totally ordered set. The partial order of this set is $\{\lzero \preceq a,\lzero \preceq b,\lzero \preceq c,\lzero \preceq \lunit,a\preceq \lzero,b\preceq \lzero,c\preceq \lzero\}$. The covering relation is $\{\lzero \preceq: a,\lzero \preceq: b,\lzero \preceq: c,a\preceq: \lzero,b\preceq: \lzero,c\preceq: \lzero\}$. Note that the covering relation does not contain a relation between $\lzero$ and $\lunit$ but the relation $\lzero \preceq \lunit$ (no ``$:$'') is a consequence of the covering relation (see remarks to \cref{def:cover,def:hassediagram}). \subref{fig:nounitnozero} A poset without a zero or a unit. Note that $d,e$ are minimal elements and $a,f$ are maximal elements.
     }%
   \label{fig:hassediagrams}
\end{figure}

\begin{mydef}[Join and meet]
\label{def:joinandmeet}
  Given a subset $B \subseteq A$ of elements of a poset $A$ a least upper bound, supremum, or \textit{join $\bigvee B$}, if it exists, is an element $c \in A$ such that for all $b \in B$ we have $b \preceq c$ and if there exists $a \in A$ such that we also have for all $b \in B$ that $b \preceq a$ then $c \preceq a$. Conversely, a greatest lower bound, infimum, or \textit{meet $\bigwedge B$}, if it exists, is an element $c \in A$ such that for all $b \in B$ we have $c \preceq b$ and if there exists $a \in A$ such that we also have for all $b \in B$ that $a \preceq b$ then $a \preceq c$.
\end{mydef}
Remarks:
\begin{itemize}
  \item $\bigvee B$ and $\bigwedge B$ are unique if they exist.
  \item For a pair of elements $b_1,b_2 \in A$ we also write $b_1 \vee b_2$ for the join and $b_1 \wedge b_2$ for the meet.
  \item Join and meet are both associative,
\begin{align}
 \pi_1 \wedge (\pi_2 \wedge \pi_3) &= \pi_1 \wedge \pi_2 \wedge \pi_3,\\
 \pi_1 \vee (\pi_2 \vee \pi_3) &= \pi_1 \vee \pi_2 \vee \pi_3,
\end{align}
commutative,
\begin{align}
 \pi_1 \wedge \pi_2  &= \pi_2 \wedge \pi_1,\\
 \pi_1 \vee \pi_2  &= \pi_2 \vee \pi_1,
\end{align}
and idempotent,
\begin{align}
 \pi \wedge \pi  &= \pi,\\
 \pi \vee \pi  &= \pi.
\end{align}
\end{itemize}

\begin{mydef}[Lattice]
\label{def:lattice}
A poset $A$ is a \textit{lattice} if for every pair of elements $a,b \in A$ both join and meet exist, i.e.\ $a \vee b \in A$ and $a \wedge b \in A$.   
\end{mydef}
Remarks:
\begin{itemize}
  \item If join and meet exist for every pair they also exist for every finite subset $B \subseteq A$ \citep[p.9]{gratzer_lattice_2011}. In this thesis we only encounter finite sets so we have join and meet for every subset, which in general is an additional property of lattices called completeness.
  \item Every non-empty finite lattice $A$ is a bounded poset with $\lzero = \bigwedge A$ and $\lunit = \bigvee A$.
  \item The posets in \cref{fig:hassediagrams}\subref{fig:two} and \cref{fig:hassediagrams}\subref{fig:three} are lattices.
\end{itemize}

\begin{mydef}[Atoms and dual atoms]
  Given a bounded poset $A$ an atom is an element $a \in A$ that covers the zero element, i.e.\ $\lzero \preceq: a$. A dual atom is an element $b \in A$ that is covered by the unit element, i.e.\ $b \preceq: \lunit$.
\end{mydef}

%
%

\subsection{Partitions and the partition lattice}
\label{sec:partitions}
This section recalls the definitions of 
\begin{itemize}
 \item set partitions,
 \item refinement and coarsening of set partitions,
 \item join and meet operation between partitions,
 \item the partition lattice.
\end{itemize}

The following definitions are due to \citet[p.359]{gratzer_lattice_2011}.
\begin{mydef}
A \emph{(set) partition} $\pi$ of a set $\X$ is a set of non-empty subsets (called \emph{blocks}) of $\X$ satisfying 
\begin{enumerate}
 \item for all $x_1,x_2 \in \pi$, if $x_1 \neq x_2$, then $x_1 \cap x_2 = \emptyset$,
 \item $\bigcup_{x \in \pi} = \X$.
\end{enumerate}
We write $\Latt(\X)$ for the set of all partitions of $\X$. 
\end{mydef}
Remark:
\begin{itemize}
  \item In words, a partition of a set is a set of disjoint non-empty subsets whose union is the whole set.
\end{itemize}

\begin{mydef}
If two elements $x_1, x_2 \in \X$ belong to the same block of a partition $\pi$ of $\X$ write $x_1 \equiv_\pi x_2$. Also write $x_1/\pi$ for the block $\{x_2 \in \X: x_2 \equiv_\pi x_1\}$. 
\end{mydef}

\begin{mydef}[Refinement and coarsening]
\label{def:refinement}
 We define the binary relation
 $\lpreeq$ between partitions $\pi, \rho \in \Latt(\X)$ as:
\begin{equation}
 \pi \lpreeq \rho \text{ if } x_1 \equiv_{\pi} x_2 \text{ implies } x_1 \equiv_{\rho} x_2.
\end{equation}
In this case $\pi$ is called a \emph{refinement} of $\rho$ and $\rho$ is called a \emph{coarsening} of $\pi$.
\end{mydef}
Remark:
\begin{itemize}
\item More intuitively, $\pi$ is a refinement of $\rho$ if all blocks of $\pi$ can be obtained by further partitioning the blocks of $\rho$. Conversely $\rho$ is a coarsening of $\pi$ if all blocks in $\rho$ are unions of blocks in $\pi$. 
\item Refinement and coarsening will be used frequently throughout this thesis.
\end{itemize}

\begin{figure}[ht!]
\begin{center}
  \includegraphics[width=.6\textwidth]{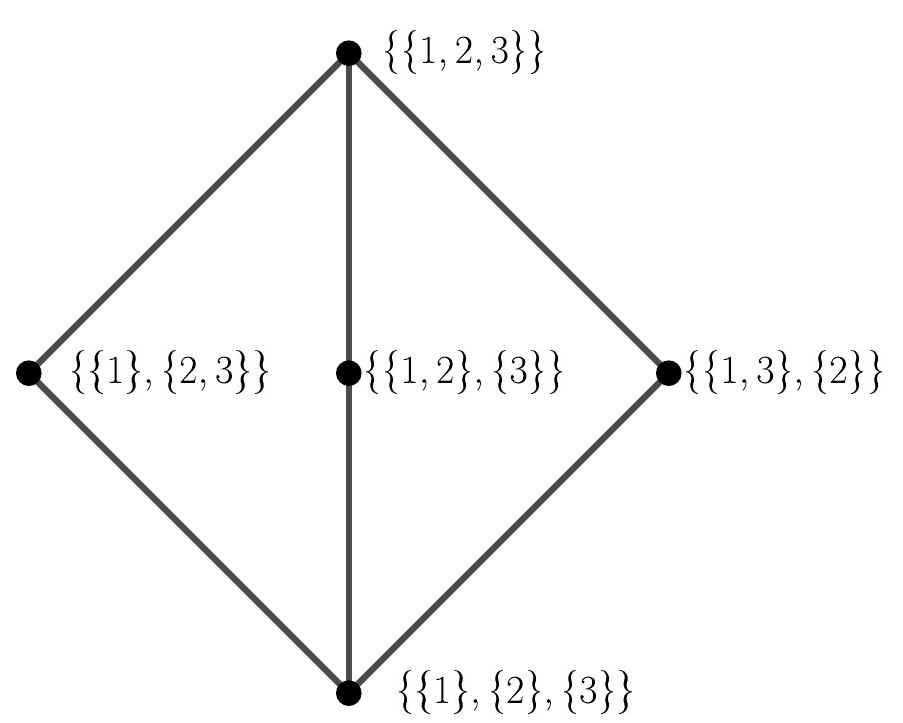}
\end{center}
\caption{%
        Hasse diagrams of the partition lattice of the three element set.
     }%
   \label{fig:threepart}
\end{figure}

\begin{figure}[ht!]
\begin{center}
  \includegraphics[width=\textwidth]{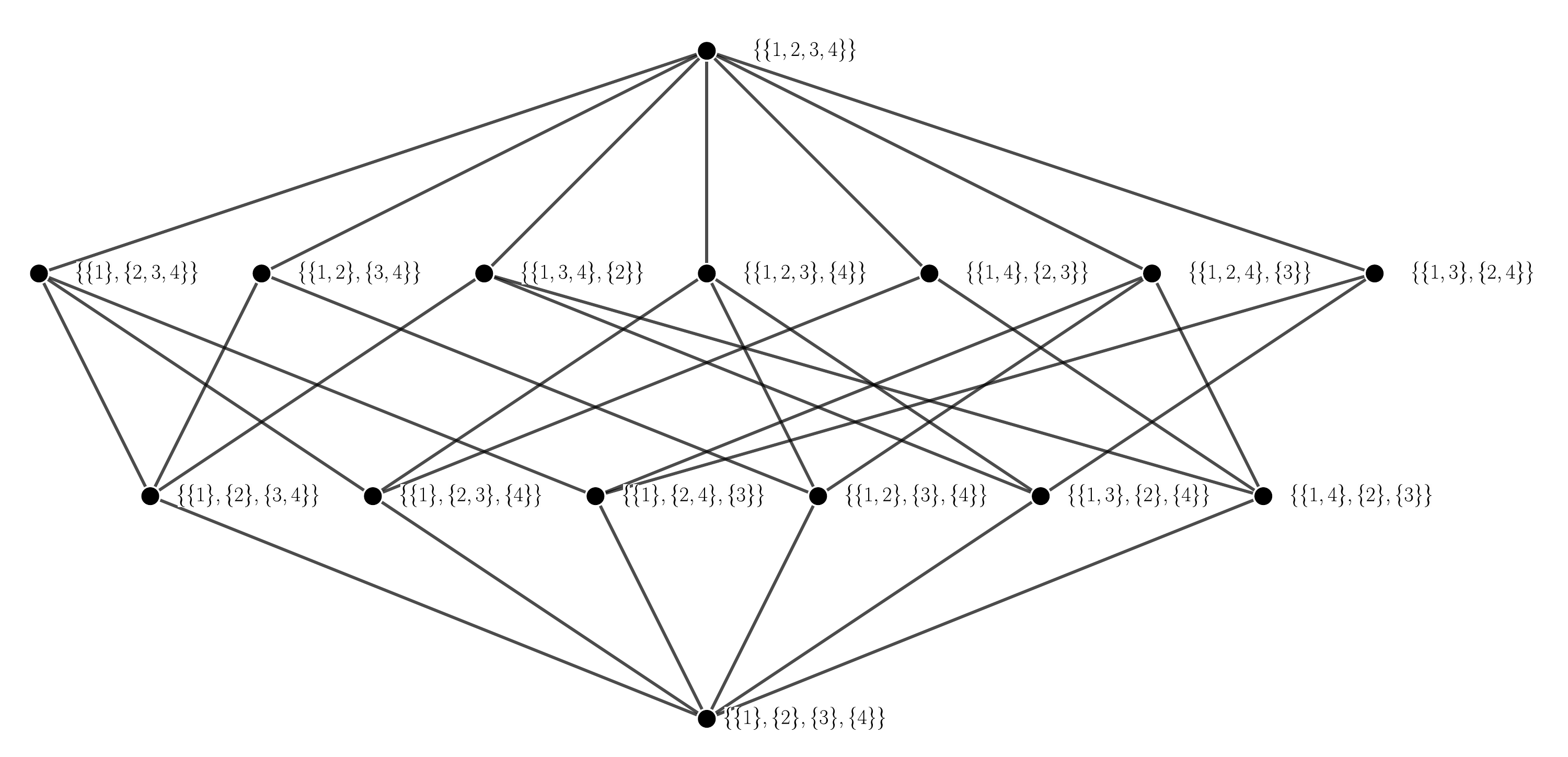}
\end{center}
\caption{%
        Hasse diagrams of the partition lattice of the four element set.
     }%
   \label{fig:fourpart}
\end{figure}

\begin{thm}[Partition lattice]
\label{def:partitionlattice}
  Let $\X$ be a finite set and $\Latt(\X)$ its set of all partitions.
  \begin{thmlist}
    \item Refinement $\lpreeq$ is a partial order of $\Latt(\X)$.
    \item $\Latt(X)$ together with refinement $\lpreeq$ is a (bounded) lattice. When we write $\Latt(\X)$ in the following we consider it as a lattice with refinement as its partial order.
    \item The zero of $\Latt(\X)$ is the partition consisting only of singleton sets as blocks, $\lzero = \{\{x\}\}_{x \in \X}$.
    \item The unit of $\Latt(\X)$ is the partition consisting of a single block containing all elements, $\lunit = \{\X\}$.
    \item The atoms of $\Latt(\X)$ are the partitions consisting of a single block containing two elements and apart from this block only of blocks that are singletons. 
    \item The dual atoms of $\Latt(\X)$ are the partitions that consist of two blocks.
    \item A partition $\pi \in \Latt(\X)$ covers another partition $\xi \in \Latt(X)$, i.e.\ $\xi \lpreeq: \pi$ iff $\pi$ is the result of replacing two blocks of $\xi$ by their union.
  \end{thmlist}
\end{thm}
\begin{proof}
  See \citet[p.360]{gratzer_lattice_2011}.
\end{proof}
Remarks:
\begin{itemize}
\item 
The join $\pi_1 \vee \pi_2$ of two partitions $\pi_1,\pi_2 \in \Latt(\X)$ is the unique coarsening of both $\pi_1$ and $\pi_2$ which can be obtained by taking the union of the fewest blocks in either of the two partitions. The meet is the partition composed of the blocks obtained by taking the set intersections of all blocks of $\pi_1$ and $\pi_2$
. 
\item The Hasse diagrams of the partition lattices of the three and four element set are shown in \cref{fig:threepart,fig:fourpart}.
\end{itemize}

\begin{thm}
\label{thm:bellno}
  The number of partitions of a set $\X$ of cardinality $|\X|=n$ is given by the Bell numbers:
  \begin{equation}
    |\Latt(\X)| =\B_n:= \sum_{k=0}^n \S(n,k) = \sum_{k=0}^n \frac{1}{k!}\sum_{i=0}^k (-1)^i \binom{k}{i} (k-i)^n.
  \end{equation} 
Where $\S(n,k)$ are the Sterling numbers of the second kind which are the number of partitions of a set with $n$ elements into $k$ blocks.
  \end{thm}
\begin{proof}
  This is well known.
\end{proof}

\section{Sets of random variables and patterns}

\label{sec:rvandpat}

\subsection{Overview}
This section introduces the notion of patterns. In the form of spatiotemporal patterns (which are just patterns in systems where space and time have clear interpretations) this notion forms an important concept in this thesis. The main idea is to capture things/structures that can occur within \textit{single} trajectories of multivariate dynamical systems and stochastic processes. In other words, we would like to capture structures that very from one trajectory of a system to another instead of remaining fixed throughout all trajectories. At the same time these structures should only concern part of a trajectory and not the whole. A reader familiar with probability theory should have no big trouble with the following definitions.
\subsection{Patterns}

\begin{mydef}
  Let $\{X_i\}_{i \in V}$ be set of random variables with index set $V$ and state spaces $\{\X_i\}_{i \in V}$ respectively. Then for $A \subseteq V$ define:
 \begin{thmlist}
  \item $X_A := (X_i)_{i \in A}$ as the random variable composed of the random variables indexed by $A$,
  \item $\X_A:=\prod_{i \in A} \X_i$ as the state space of $X_A$,
  \item $x_A := (x_i)_{i \in A} \in \X_A$ as a value of $X_A$,
 \end{thmlist}
\end{mydef}

\begin{mydef}[Elementary pattern]
   Let $\{X_i\}_{i \in V}$ be set of random variables with index set $V$ and state spaces $\{\X_i\}_{i \in V}$ respectively. An \textit{elementary pattern in $\{X_i\}_{i \in V}$} is an assignment $$X_j = y$$ of a value $y \in \X_j$ to a single one of the random variables $X_j \in \{X_i\}_{i \in V}$. 
   
   We often choose to denote the value $y \in \X_j$ we assign to a random variable $X_j$ by $x_j$ so that it indicates the index $j$ of the random variable $X_j$ it is assigned to. This often allows us to just write $x_j$ for an elementary pattern without danger of ambiguity. 
\end{mydef}
Remark:
\begin{itemize}
  \item Note that it will later sometimes be necessary to move the values assigned to one random variable to another random variable. This can result in elementary patterns of the form $X_j = x_k$ where the index $k$ of the assigned value is not identical to the index $j$ of the random variable that it is assigned to. In such cases we will always use the unambiguous full notation $X_j = x_k$ and cannot use just $x_k$ to denote the elementary pattern.
 \end{itemize}

\begin{mydef}[Pattern]
\label{def:pattern}
  Let $\{X_i\}_{i \in V}$ be set of random variables with index set $V$ and state spaces $\{\X_i\}_{i \in V}$ respectively. Then a \textit{pattern in $\{X_i\}_{i \in V}$} is an assignment $$\{X_j = x_j\}_{j \in A}$$ of values $\{x_j\}_{j \in A}$ with $x_j \in \X_j$ for all $j \in A \subseteq V$ to a subset $\{X_j\}_{j \in A}$ of the set of random variables. 
%
%
  
  An alternative form of specifying a pattern is via $X_A = x_A$ in which case the identical ordering of the tuples $X_A = (X_j)_{j \in A}$ and $(x_j)_{j \in A}$ ensure that the value $x_i$ is assigned to the random variable $X_i$. 
  
  Just like for elementary patterns, if the index set $A$ of a value $x_A$ specifies the index set of the joint random variable that it is assigned to (and in the right order) we use the short notation $x_A$ for the pattern.  
\end{mydef}
Remark:
\begin{itemize}
  \item Note that by definition a random variable $X_i$ can only occur once and be assigned a single value in a pattern. This is due to the requirement that we can index the elementary patterns that occur in a pattern by a subset $A \subseteq V$. In other words a pattern is just a set of elementary patterns where each random variable occurs on the left hand side in at most one elementary pattern.
  \item Formally the only difference between the two ways of expressing a pattern is that $\{X_i = x_i\}_{i \in A}$ is a set of elementary patterns and $X_A = x_A \Leftrightarrow (X_i = x_i)_{i \in A}$ is a tuple of elementary patterns. A set is unordered while a tuple is ordered. The expressed assignments are the same.
   \item  This notion of patterns is similar to ``patterns'' as defined in \citet{ceccherinisilberstein_cellular_2009} and to ``cylinders'' as defined in \citet{busic_probabilistic_2010}. However the notions there are explicitly limited to single time-slices. Our notion of patterns purposely extends to spatiotemporal patterns. These are patterns in specific sets of random variables endowed with notions of time and space like multivariate Markov chains (see \cref{def:mvmarkov}).
  \item This notion of patterns is also similar to the notion of propositions. However, propositions are usually more general as they allow any logical combination of the elementary patterns as defined here. For example ${X_i = x\; OR \;X_j = y}$ is a proposition but not a pattern, since patterns are only defined as lists of elementary patterns not as logical combinations thereof. A set of elementary patterns might be seen as a logical conjunction (AND) of elementary patterns, but a disjunction (OR) is not allowed by the definition.  

\end{itemize}

\begin{mydef}[Trajectory]
   Let $\{X_i\}_{i \in V}$ be set of random variables with index set $V$ and state spaces $\{\X_i\}_{i \in V}$ respectively. A \textit{trajectory} is a pattern $\{X_j = x_j\}_{j \in V}$ that assigns a value to every random variable in $\{X_i\}_{i \in V}$. We also write $x_V$ for trajectories
\end{mydef}
Remark:
\begin{itemize}
  \item The use of the term ``trajectory'' might be somewhat surprising at this stage as the set of random variables doesn't have a structure which justifies its use yet. However, the sets of random variables in this thesis will usually correspond to Markov chains which makes trajectory an intuitive choice. 
  \item Trajectories are just particular patterns. However they are special because they determine the whole set of random variables. Since we will need to refer to them frequently it is convenient to give them a particular name. 
\end{itemize}

\begin{mydef}
\label{def:stpoccurrence}
 Given a set of random variables $\Xv$ we say \textit{pattern $x_A$ occurs in trajectory $\bar{x}_V \in \X_V$} if $\bar{x}_A = x_A$. For formal reasons we also define that all empty patterns are equal i.e.\ $\bar{x}_\emptyset =x_\emptyset$.
\end{mydef}
Remark:
\begin{itemize}
\item 
The definition of $\bar{x}_\emptyset =x_\emptyset$ implies that $x_\emptyset$ occurs in every trajectory.
\end{itemize}
\begin{mydef}
\label{def:stptraset}
  Let $\{X_i\}_{i \in V}$ be set of random variables with index set $V$ and state spaces $\{\X_i\}_{i \in V}$ respectively. Given a pattern $x_A$ let $\T(x_A)$ denote the \textit{set of trajectories of $\Xv$ in which $x_A$ occurs} i.e.\
  \begin{equation}
    \T(x_A):= \{\bar{x}_V \in \X_V : \bar{x}_A = x_A\}.
  \end{equation} 
\end{mydef}
Remark:
\begin{itemize}
  \item Note that $\T(x_\emptyset) = \X_V$ because of \cref{def:stpoccurrence}.
  \item Each pattern thus defines a set of trajectories in which it occurs. However, the converse is not true. As we will see below (\cref{thm:invisiblesubset}) there are subsets of trajectories of a set of random variables that are not captured by any of the patterns in the network. The set of trajectories defined by a pattern is therefore of secondary importance. Primarily patterns are structures that can occur within trajectories.
\end{itemize}

\begin{thm}
  Given a set of random variables $\{X_i\}_{i \in V}$ with index set $V$ and state spaces $\{\X_i\}_{i \in V}$ respectively, the set of all patterns in $\Xv$ is isomorphic to $\bigcup_{A \subseteq V} \X_A$.
\end{thm}
\begin{proof}
  Each element $x_B=(x_i)_{i \in B}$ in $\bigcup_{A \subseteq V} \X_A$ defines a pattern $\{X_i = x_i\}_{i \in B}$. And each pattern $\{X_i = x_i\}_{i \in B}$ defines an element $x_B=(x_i)_{i \in B} \in \X_B \subset \bigcup_{A\subseteq V} \X_A$. 
\end{proof}
Remark:
\begin{itemize}
  \item Note $x_\emptyset \in \X_\emptyset \subset \bigcup_{A\subseteq V} \X_A$ is included here since $\emptyset \subset V$. The set $\X_\emptyset$ only contains a single element which is $x_\emptyset$. We could have defined this in an extra definition but it is of no fundamental consequence and so we just note it here.
\end{itemize}

\subsection{Patterns and invisible subsets of trajectories}
\begin{mydef}[Anti-pattern]
\label{def:antistp}
  Given a pattern $x_O$ define its set of anti-patterns $\neg(x_O)$ that have values different from those of $x_O$ on all variables in $O$:
  \begin{equation}
   \neg(x_O):= \{\bar{x}_O \in \X_O : \forall i \in O, \bar{x}_i \neq x_i\}.
  \end{equation} 
\end{mydef}
Remark:\begin{itemize}
        \item It is important to note that for an element of $\neg(x_O)$ to occur it is not sufficient that $x_O$ does not occur. Only if \textit{every} random variable $X_i$ with $i \in O$ differs from the value $x_i$ specified by $x_O$ does an element of $\neg(x_O)$ necessarily occur. This is why we call $\neg(x_O)$ the anti-pattern of $x_O$.
        \item Anti-patterns are useful in the construction of examples of patterns. A generalisation of this anti-patterns will be presented in \cref{def:antistppart}.
        \end{itemize}

\begin{thm}
\label{thm:invisiblesubset}
  Given a set of random variables $\Xv$ where $|V| \geq 2$ there are subsets of trajectories $\D \subset \X_V$ such that there is no pattern $x_A \in \bigcup_{C \subseteq V} \X_C$ with $\D = \T(x_A)$.
\end{thm}
\begin{proof}
   We construct one such subset $\D$ for an arbitrary set of random variables $\Xv$. Take an arbitrary pattern $x_A$ with $|A| \geq 2$ and choose another pattern $\bar{x}_A$ from $\neq(x_A)$. Then let
     \begin{equation}
    \D:= \{\T(x_A) \cup \T(\bar{x}_A)\}.
  \end{equation} 
  To see that there is no pattern $\tilde{x}_B \in \bigcup_{C \subseteq V} \X_C$ with $\D = \T(\tilde{x}_B)$ note that we can write
  \begin{align}
    \tilde{x}_C = (\tilde{x}_{C \bs A},\tilde{x}_{C \cap A}).
  \end{align}
  If $C \cap A \neq \emptyset$ we must have either $\tilde{x}_{C\cap A} = x_A$ or $\tilde{x}_{C \cap A} \neq x_A$. First, let $\tilde{x}_{C\cap A} = x_A$ but then $\T(\bar{x}_A) \nsubseteq \T(\tilde{x}_C)$ so $\D \nsubseteq \T(\tilde{x}_C)$. Next choose $\tilde{x}_{C \cap A} \neq x_A$ but then $\T(x_A) \nsubseteq \T(\tilde{x}_C)$ so also $\D \nsubseteq \T(\tilde{x}_C)$. So we must have $C \cap A = \emptyset$. 
  
  Now we show that if $C \cap A = \emptyset$ there are trajectories in $\T(\tilde{x}_C)$ that are not in 
  $\D$. Consider the following trajectory: $\hat{x}_V:=(\tilde{x}_C,x_{A_1},\bar{x}_{A_2},\check{x}_D)$ where $A_1 \cup A_2 = A$, $A_1 \cap A_2 =\emptyset$, $D = V \bs (C \cup A)$, and $\check{x}_D \in \X_D$ is arbitrary. We can split up $A$ into $A_1$ and $A_2$ like this because $|A| \geq 2$ by assumption. Now $\hat{x}_V \in \T(\tilde{x}_C)$ but $\hat{x}_V \neq \D$ because $\hat{x}_A=(x_{A_1},\bar{x}_{A_2}) \neq x_A$ and $\hat{x}_A \neq \bar{x}_A$ due to our initial choice of $\bar{x}_A \in \neg(x_A)$.
\end{proof}
Remark:
\begin{itemize}
\item 
We explicitly construct a simple example set $\D$ for $V = \{1,2\}$ and $\Xv=\{X_1,X_2\}$ the set of random variables. Let $\X_1 = \X_2=\{0,1\}$. Then $\X_V = \{(0,0),(0,1),(1,0),(1,1)\}$. Now let $A=V=\{1,2\}$, choose pattern $x_A=(0,0)$ and pattern $\bar{x}_A\in \neg(x_A) = (1,1)$ from its set of anti-patterns. Then let
  \begin{equation}
    \D:= \{\T(x_A) \cup \T(\bar{x}_A)\} = \{(0,0),(1,1)\}.
  \end{equation} 
  In this case we can easily list the set of all patterns $\bigcup_{C \subseteq V} \X_C$: 
  \begin{equation}   
    \begin{array}{|c|c|c|}
\hline
C \subseteq V & x_C & \T(x_C) \\
\hline
\emptyset & x_\emptyset & \X_V \\
\{1\} & (0) & \{(0,0),(0,1)\} \\
      & (1) & \{(1,0),(1,1)\} \\
\{2\} & (0) & \{(0,0),(1,0)\} \\
      & (1) & \{(0,1),(1,1)\} \\     
\{1,2\} & (0,0) & \{(0,0)\} \\      
& (0,1) & \{(0,1)\} \\
& (1,0) & \{(1,0)\} \\
& (1,1) & \{(1,1)\} \\
\hline
\end{array}
\end{equation}
and verify that $\D$ is not among them. This suggests the first part of the proof above i.e.\ that $C \cap A =\emptyset$ or else $\D \nsubseteq \T(\tilde{x}_C)$. If there were a further random variable $X_3$ then any pattern $x_3$ would contain the trajectory $(x_1,\bar{x}_2,x_3)=(0,1,x_3)$ which is not in $\D$ and corresponds to $\hat{x}_V$ of the proof.
\end{itemize}

\begin{mydef}[Visible and invisible subsets]
  Given a set of random variables $\Xv$ a \textit{visible subset of $\Xv$} is a subset of trajectories $\D \subseteq \X_V$ such that there is an pattern $x_A$ in $\Xv$ with $\D = \T(x_A)$. Subsets $\D \subseteq \X_V$ that are not visible are called \textit{invisible subsets of $\Xv$}. 
\end{mydef}

Remark:
\begin{itemize}
\item Visible subsets are completely defined by a pattern whereas this is impossible for invisible subsets. For a given trajectory the question whether a given pattern occurs within this trajectory is well defined via \cref{def:stpoccurrence}. Intuitively this also makes sense since we can just look at the trajectory of a cellular automaton for example to check whether a pattern has occurred. However, for invisible subsets there is no defining pattern, and we cannot inspect a given single trajectory and look for a pattern within it. We can check whether a given single trajectory is an element of the invisible set but this is not the same thing. While this difference might seem to be an inessential subtlety, in this thesis we take this difference seriously.
  \item The invisible subsets of a set of random variables do not occur within a trajectory in the same way that the patterns do. Let $\D$ be an invisible set. To check whether it occurs in a trajectory $\bar{x}_V$ by the method defined in \cref{def:stpoccurrence} we cannot directly check whether $\bar{x}_V = \D$ because one is a vector of numbers $(\bar{x}_i)_{i \in V}$ and the other a set of trajectories $\{x_V\}_{x_V \in \D}$. We can check whether $\T(\bar{x}_V)=\D$ (which will always fail by the definition of invisible sets) but that is not our definition of occurrence. A definition of occurrence that would allow checking all subsets of $\X_V$ would be to require only that $\bar{x}_V \in \D$. Considering our application to Markov chains and dynamical systems, this definition however 
  cannot capture the intuition behind a pattern occurring \textit{in} a single trajectory. This is due to the fact that subsets of trajectories can be constructed from patterns and their anti-patterns together as in the proof of \cref{thm:invisiblesubset}. We would therefore end up in a situation where a subset could ``occur in a trajectory'' even though it consists of contradictory patterns. Our definitions are specifically designed to talk about occurrences of things (here represented by patterns) within trajectories and not, as the other definition would offer, to talk about trajectories that are contained in subsets. The whole idea behind patterns is to end up being able to talk about things within single trajectories. Maybe it will turn out at some point that patterns are not the right structure for this purpose but in order to evaluate their suitability we are forced to make this distinction between patterns and subsets via the distinction of visible and invisible subsets.
\end{itemize}

\subsection{Probabilities of patterns}

\begin{mydef}
\label{def:probabilities}
  Let $\{X_i\}_{i \in V}$ be set of random variables with state spaces $\{\X_i\}_{i \in V}$ respectively and let $A,B \subseteq V$. Also, for all $i \in A \cup B$ let $x_i \in \X_i$. Then:
   \begin{thmlist}
    \item The \textit{joint probability} that the pattern $\{X_i = x_i\}_{i \in A}$ occurs is denoted by:
  \begin{align}
    \Pr\left(\{X_i = x_i\}_{i \in A}\right).
  \end{align}
  It satisfies the usual conditions:
  \begin{enumerate}
  \item
  \begin{equation}
    \Pr\left(\{X_i = x_i\}_{i \in A}\right) \in [0,1]
  \end{equation}
  \item
    \begin{equation}
    \prod_{j \in A} \sum_{x_j \in \X_j} \Pr\left(\{X_i = x_i\}_{i \in A}\right) = 1.
  \end{equation}
\end{enumerate}
  \item The \textit{conditional probability} that the pattern $\{X_i = x_i\}_{i \in A}$ occurs given that the pattern $\{X_j = x_j\}_{j \in B}$ occurs is denoted:
\begin{align}
  \Pr\left(\{X_i = x_i\}_{i \in A}\mid \{X_j = x_j\}_{j \in B}\right).
\end{align}
and defined by:
\begin{equation}
  \Pr\left(\{X_i = x_i\}_{i \in A}\mid \{X_j = x_j\}_{j \in B}\right):= \frac{\Pr\left(\{X_i = x_i\}_{i \in A \cup B}\right) }{\Pr\left(\{X_j = x_j\}_{j \in B}\right)}.
\end{equation} 
\end{thmlist}
\end{mydef}

\begin{mydef}[Probability distribution]
  Let $\{X_i\}_{i \in V}$ be set of random variables with index set $V$ and state spaces $\{\X_i\}_{i \in V}$ respectively and let $A \subseteq V$. We define the probability distribution of $X_A$ as the function $p_A:\X_A \rightarrow [0,1]$ with
  \begin{thmlist}
    \item \begin{equation}
            p_A(x_A) := \Pr(\{X_i = x_i\}_{i \in A}) = \Pr(X_A = x_A) 
          \end{equation} 
  \end{thmlist}
\end{mydef}
Remark:
\begin{itemize}
  \item The more technically precise term for the probability distribution is a ``probability mass function''.
  \item The probability distribution takes the arguments in the order specified by $A$ i.e.\ the $i$-th argument is interpreted as the value of the random variable $X_i$.  
\end{itemize}

\begin{mydef}
   Let $\{X_i\}_{i \in V}$ be set of random variables with index set $V$ and state spaces $\{\X_i\}_{i \in V}$ respectively and let $A,B \subseteq V$ and $A \cap B =\emptyset$. Then we also define:
   \begin{equation}
     p_{A,B}(x_A,x_B):= \Pr(X_A = x_A,X_B=x_B) = \Pr(\{X_i = x_i\}_{i \in A},\{X_i = x_i\}_{i \in B})
   \end{equation} 
   For convenience we often just write $p_{A \cup B}(x_A,x_B)$ instead of $p_{A,B}(x_A,x_B)$ e.g.\ if $A \cup B =V$ we write $p_V(x_A,x_B)$ for $p_{A,B}(x_A,x_B)$ this causes no confusion again if the index sets $A$ and $B$ unambiguously indicate the random variables $X_A$ and $X_B$ which they are assigned to. 
\end{mydef}
Remark:
\begin{itemize}
  \item Note that, technically,
   \begin{equation}
     p_{A,B}(x_B,x_A)= \Pr(X_A = x_B,X_B=x_A) \neq p_{A,B}(x_A,x_B).
   \end{equation} 
   In all such cases (where $A$ does not index both the values and the random variables they are assigned to) we will refrain from writing $p_{A \cup B}(x_B,x_A)$. Such cases will arise when we look at symmetries of patterns and their probabilities in \cref{sec:symm}. This is the main reason for introducing the more cumbersome full notation of patterns $\stp{A}$ and their probabilities $\Pr(\stp{A})$ in addition to the short notation $x_A$ and the probability distribution $p_A$.
\end{itemize}


\begin{mydef}[Conditional probability distribution]
   Let $\{X_i\}_{i \in V}$ be set of random variables with index set $V$ and state spaces $\{\X_i\}_{i \in V}$ respectively and let $A,B \subseteq V$. Then we define:
   \begin{equation}
     p_{A|B}(x_A|x_B):= \Pr(X_A = x_A|X_B=x_B) = \Pr(\{X_i = x_i\}_{i \in A}\mid \{X_i = x_i\}_{i \in B})
   \end{equation} 
   If it is clear from the index set of the values we condition on which random variables they are assigned to, we often just write $p_A(x_A|x_B)$ instead of $p_{A|B}(x_A|x_B)$. 
\end{mydef}

\begin{mydef}[Morph of a pattern]
\label{def:morph}
  Let $\{X_i\}_{i \in V}$ be set of random variables with index set $V$ and state spaces $\{\X_i\}_{i \in V}$ respectively and let $x_A$ be a pattern in $\Xv$. Then we define the \textit{morph} denoted by $p_{V \bs A}(X_{V \bs A}|x_A)$ of $x_A$ as the probability distribution $p_{V\bs A | A}( . | x_A):\X_{V\bs A} \rightarrow [0,1]$.
\end{mydef}
Remark:
\begin{itemize}
  \item The morph is the probability distribution over the rest of the set of random variables given a pattern $x_A$.
  \item This terminology is inspired by \citet{shalizi_causal_2001}. In the case where $A$ indicates all past variables in some stochastic process and $V\bs A$ indicates all future variables, the definition here coincides with the original. 
\end{itemize}


\begin{thm}[Marginalisation]
  Let $\{X_i\}_{i \in V}$ be set of random variables with index set $V$ and state spaces $\{\X_i\}_{i \in V}$ respectively and let $A,B \subseteq V$ and $A \cap B =\emptyset$. Then we have:
  \begin{equation}
    p_A(x_A) = \sum_{\bar{x}_{V\bs A}} p_{A,V \bs A}(x_A,\bar{x}_{V \bs A}).
  \end{equation} 
\end{thm}
\begin{proof}
 We do not give a proof here. It follows from the axioms of probability. For a proof we would need to invoke these axioms which is beyond the scope of this thesis.
\end{proof}

\begin{thm}[Chain rule of probability]
  Let $V$ be a set of indices for a set of random variables $\{X_i\}_{i \in V}$ with state spaces $\X_i$. Also for all $i \in V$ let $x_i \in \X_i$. Then for any (re-)labelling $i_1,i_2,...,i_{|V|}$ of the index set $V$ we have:
  \begin{equation}
  \label{eq:factorization}
     \Pr\left(\{X_i = x_i\}_{i \in V}\right)=\prod_{j=1}^{|V|} \Pr(\{X_{i_j}=x_{i_j}\} \mid \{X_{i_k}=x_{i_k}\}_{k \in \{j+1,...,|V|\}}).
  \end{equation} 
\end{thm}
\begin{proof}
  Follows directly from the definition of the conditional probability, see \cref{def:probabilities}. Just replace the conditional probabilities by their defining fractions and reduce.
\end{proof}

 The Kronecker-delta is used in this thesis to represent deterministic conditional distributions.
\begin{mydef}[Delta]
Let $X$ be a random variable with state space $\X$ then for $x \in \X$ and a subset $C \subset \X$ define
 \begin{equation}
  \delta_{x}(C) := \begin{cases}
                         1 & \text{if } x \in C, \\
                         0 & \text{else.}
                        \end{cases}
 \end{equation} 
We will abuse this notation if $C$ is a singleton set $C=\{\bar{x}\}$ by writing
 \begin{align}
  \delta_{x}(\bar{x}) :&= \begin{cases}
                         1 & \text{if } x \in \{\bar{x}\}, \\
                         0 & \text{else.}
                        \end{cases} \\
                   &= \begin{cases}
                         1 & \text{if } x = \bar{x}, \\
                         0 & \text{else.}
                        \end{cases} 
 \end{align} 
 The second line is a more common definition of the Kronecker-delta.
\end{mydef}

Remark:
\begin{itemize}
\item 

Let $X,Y$ be two random variables with state spaces $\X,\Y$ and $f:\X \rightarrow \Y$ a function such that
\begin{align}
 p(y|x) = \delta_{f(x)}(y),
\end{align}
then 
\begin{align}
 p(y) &= \sum_x p_Y(y|x)p_X(x) \\
 &= \sum_x \delta_{f(x)}(y) p_X(x) \\
 &= \sum_x \delta_{x}(f^{-1}(y)) p_X(x) \\ 
 &= \sum_{x \in f^{-1}(y)} p_X(x) \\
 &= p_X(f^{-1}(y)).
\end{align}
\end{itemize}

\section{Bayesian networks}
\label{sec:bn}
In this section we introduce Bayesian networks and the special cases of it that we will use in this thesis. Our main formal original contributions, the disintegration theorem \cref{thm:disintegration}, and the sli symmetry theorem \cref{thm:symmpart} in later sections hold for Bayesian networks in general. In the conceptual part of this thesis (\cref{ch:agents}) we use multivariate Markov chains (\cref{sec:multimc}) which are a special kind of Bayesian network as systems that can contain agents. A famous example of a deterministic multivariate Markov chain which we also use in \cref{sec:entinmvmc} is the game of life cellular automaton. Driven (multivariate) Markov chains (\cref{sec:driven}) are multivariate Markov chains where the focus is on a subset of the degrees of freedom. Such systems are often used in practice and we therefore include them in our formal considerations. In the conceptual part they lay a lesser role. Finally, in \cref{sec:paloopformal} we present the definition of the perception-action loop. This is also a multivariate Bayesian network. The perception-action loop plays a role in this thesis as a reference system that is used to formally represent agents in the literature. 
Furthermore we present a method to extract perceptions (and actions) from the perception-action loop that capture all influences from the environment on the agent (the actions capture all influences from the agent on the environment). We formally prove that this is the case and in the conceptual part (\cref{sec:perceptions}) we generalize this method of extracting perceptions/influences to ``spatiotemporal pattern-based entities''. 


We therefore present these as special cases of Bayesian networks. For a more thorough treatment of Bayesian networks we refer to \citet{pearl_causality_2000}.

\subsection{Bayesian networks and mechanisms}

\begin{mydef}
 A directed acyclic graph $G=(V,E)$ with nodes $V$ and edges $E$ is \emph{factorization compatible} with the joint probabilities the probabilities of a probability distribution $p_V:\X_V\rightarrow [0,1]$ iff the latter can be factorized in the way suggested by $G$ which means:
\begin{align}
\label{eq:graphfactorization}
 p_V(x_V)=\prod_{i \in V} p(x_i|x_{\pa(i)}).
\end{align}
Where $\pa(i)$ denotes the parents of node $i$ according to $G$.
\end{mydef}
Remark:
\begin{itemize}
 \item In general there are multiple directed acyclic graphs that are factorization compatible with the same probability distribution. If we choose any total order for the nodes in $V$ and define a graph by $\pa(i) = \{j \in V: j < i\}$ then \cref{eq:graphfactorization} becomes \cref{eq:factorization} which always holds. This means every probability distribution is compatible with all graphs that can be constructed in this way.
\end{itemize}

\begin{mydef}[Bayesian network]
\label{def:bnet}
A \emph{Bayesian network} is a finite set of random variables $\Xv$ and a directed acyclic graph $G=(V,E)$ with nodes indexed by $V$ such that the joint probability distribution $p_V:\X_V \rightarrow [0,1]$ of $\{X_i\}_{i \in V}$ is factorization compatible with $G$. We also refer to the graph set of random variables $\{X_i\}_{i \in V}$ as a Bayesian network implying the graph $G$.
\end{mydef}

Remark:
\begin{itemize}
 \item Since $\Xv$ is finite and $G$ is acyclic there is a set $V_0$ of nodes without parents.
 \item
We will see specific kinds of Bayesian networks with restricted the graph structures in \cref{sec:unimc,sec:multimc,sec:driven}.
\end{itemize}

%

\begin{mydef}[Mechanism]
 Given a Bayesian network $\{X_i\}_{i \in V}$ with index set $V$ for each node with parents i.e. for each node $i \in V \bs V_0$ (with $V_0$ the set of nodes without parents) the \textit{mechanism of node $i$} or also called the \textit{mechanism of random variable $X_i$} is the conditional probability (also called a transition kernel) $p_i: \X_{\pa(i)} \times \X_i \rightarrow [0,1]$ mapping $(x_{\pa(i)},x_i) \mapsto p_i(x_i|x_{\pa(i)})$. For each $x_{\pa(i)}$ the mechanism defines a probability distribution $p_i(.|x_{\pa(i)}):\X_i \rightarrow [0,1]$ satisfying (like any other probability distribution)
 \begin{equation}
  \sum_{x_i \in \X_i} p_i(x_i|x_{\pa(i)}) = 1. 
 \end{equation} 
\end{mydef}

Remark:
\begin{itemize}
\item We could define the set of all mechanisms to formally also include the mechanisms of the nodes without parents $V_0$. However in practice it makes sense to separate the nodes without parents as those that we choose an initial probability distribution over (similar to a boundary condition) which is then turned into a probability distribution $p_V$ over the entire Bayesian network $\Xv$ via \cref{eq:graphfactorization}. Note that in \cref{eq:graphfactorization} the nodes in $V_0$ are not explicit as they are just factors $p_i(x_i|x_{\pa(i)})$ with $\pa(i) = \emptyset$.
\item 
To construct a Bayesian network, take graph $G=(V,E)$ and equip each node $i \in (V \bs V_0)$ with a mechanism $p_i: \X_{\pa(i)} \times \X_i \rightarrow [0,1]$ and for each node $i \in V_0$ choose a probability distribution $p_i:\X_i \rightarrow [0,1]$. The joint probability distribution is then calculated by the according version of \cref{eq:graphfactorization}:
\begin{align}
\label{eq:v0factorization}
 p_V(x_V)=\prod_{i \in V \bs V_0} p_i(x_i|x_{\pa(i)}) \prod_{j \in V_0} p_j(x_j).
\end{align}
\end{itemize}

\subsection{Deterministic Bayesian networks}

\begin{mydef}[Deterministic mechanism]
\label{def:detmech}
 A mechanism $p_i: \X_{\pa(i)} \times \X_i \rightarrow [0,1]$ is \textit{deterministic} if there is a function $f_i:\X_{\pa(i)} \rightarrow  \X_i$ such that 
 \begin{equation}
  p_i(x_i|x_{\pa(i)})= \delta_{f_i(x_{\pa(i)})}(x_i)=\begin{cases} 1 &\text{   if }  x_i = f_i(x_{\pa(i)}), \\
				     0 &\text{   else.}
                       \end{cases}
 \end{equation} 
\end{mydef}

\begin{mydef}[Deterministic Bayesian network]
\label{def:detbn}
 A Bayesian network $\Xv$ is deterministic if all its mechanisms are deterministic.
\end{mydef}

\begin{thm}
\label{thm:functiontotrajectory}
 Given a deterministic Bayesian network $\Xv$ there exists a function $f_{V \bs V_0}:\X_{V_0} \rightarrow \X_{V \bs V_0}$ which given a value $x_{V_0}$ of the random variables without parents $X_{V_0}$ returns the value $x_{V \bs V_0}$ fixing the values of all remaining random variables in the network. 
\end{thm}
\begin{proof}
 According to \cref{eq:graphfactorization}, the definition of conditional probabilities, and using the determinisitic mechanisms we have:
 \begin{align}
  p_{V \bs V_0}(x_{V \bs V_0} | x_{V_0}) &= \prod_{i \in V \bs V_0} p_i(x_i|x_{\pa(i)}) \\
  &= \prod_{i \in V \bs V_0} \delta_{f_i(x_{\pa(i)})}(x_i).
 \end{align} 
 For every $x_{V_0}$ the product on the right hand side is a probability distribution and therefore is always greater or equal to zero and maximally one. Also for every $x_{V_0}$ the sum of the probabilities over all $x_{V \bs V_0} \in X_{V \bs V_0}$ is equal to one. As a product of zeros and/or ones the right hand side on the second line can only either be zero or one. This means for every $x_{V_0}$ there must be a unique $x_{V \bs V_0}$ such that the right hand side is equal to one. Define this as the value of the function $f_{V \bs V_0}(x_{V_0})$.
 \end{proof}


\begin{thm}[Pattern probability in a deterministic Bayesian network]
 \label{thm:detprob}
 Given a deterministic Bayesian network (\cref{def:detbn}) and uniform initial distribution $p_{V_0}:\X_{V_0}\rightarrow [0,1]$ the probability of the occurrence of an pattern $x_A$ is:
 \begin{equation}
  \label{eq:detprob}
  p_A(x_A) = \frac{N(x_A)}{|\X_{V_0}|}
 \end{equation} 
 where $N(x_A)$ is the number of trajectories $\bar{x}_V$ in which $x_A$ occurs.  
\end{thm}
\begin{proof}
Recall that in a deterministic Bayesian network we have a function $f_{V \bs V_0}:\X_{V_0}\rightarrow \X_{V \bs V_0}$ (see \cref{thm:functiontotrajectory}) which maps a given value of $x_{V_0}$ to the value of the rest of the network $x_{V \bs V_0}$. We calculate $p_A(x_A)$ for an arbitrary subset $A \subset V$.
To make this more readable let $A \cap V_0 = A_0$, $A \bs V_0 = A_r$, $B:=V\bs A$, $B \cap V_0 =B_0$, and $B \bs V_0 = B_r$. Then
\begin{align}
 p_A(x_A) &= \sum_{\bar{x}_B} p_V(x_A,\bar{x}_B) \\
 &= \sum_{\bar{x}_{B_0},\bar{x}_{B_r}} p_V(x_{A_r},\bar{x}_{B_r}|x_{A_0},\bar{x}_{B_0}) p_{V_0}(x_{A_0},\bar{x}_{B_0}) \\
 &= \sum_{\bar{x}_{B_0},\bar{x}_{B_r}} \delta_{f_{V \bs V_0}(x_{A_0},\bar{x}_{B_0})}(x_{A_r},\bar{x}_{B_r}) p_{V_0}(x_{A_0},\bar{x}_{B_0}) \\
 &= \sum_{\bar{x}_{B_r}} \sum_{\{\bar{x}_{B_0} : (x_{A_0},\bar{x}_{B_0}) \in f_{V \bs V_0}^{-1}(x_{A_r},\bar{x}_{B_r})\}} p_{V_0}(x_{A_0},\bar{x}_{B_0}) \\
 &= \frac{1}{|\X_{V_0}|} \sum_{\bar{x}_{B_r}} |\{\bar{x}_{B_0} \in \X_{B_0} : (x_{A_0},\bar{x}_{B_0}) \in f_{V \bs V_0}^{-1}(x_{A_r},\bar{x}_{B_r})\}| \\
 &= \frac{1}{|\X_{V_0}|} N(x_A) 
\end{align}
In the second to last line we used the uniformity of the initial distribution $p_{V_0}$. The second sum in the second to last line counts all initial conditions that are compatible with $x_{A_0}$ and lead to the occurrence of $x_{A_r}$ together with some $\bar{x}_{B_r}$. The first one then sums over all such $\bar{x}_{B_r}$ to get all initial conditions that are compatible with $x_{A_0}$ and lead to the occurrence of $x_{A_r}$. Together these are all initial conditions compatible with $x_A$. In a deterministic system the number of initial conditions that lead to the occurrence of an pattern $x_A$ is equal to the number of trajectories $N(x_A)$ since every different initial condition will produce a single, unique trajectory.
\end{proof}

Remark:
\begin{itemize}
 \item Due to the finiteness of the network, deterministic mechanisms, and chosen uniform initial distribution the minimum possible non-zero probability for an pattern $x_A$ is $1 / |\X_{V_0}|$. This happens for any pattern that only occurs in a single trajectory. Furthermore the probability of any pattern is a multiple of $1/|\X_{V_0}|$.
\end{itemize}

\subsection{Univariate Markov chain}
\label{sec:unimc}
\begin{figure}[ht]
\begin{center}
  \begin{tikzpicture}
    [myarrow,node distance=2cm]

    \node[] (1) [] {$X_1$};
    \node[] (2) [right of=1] {$X_{2}$};
    \node[] (3) [left of=1] {$X_{0}$};
    \node[] (5) [right of=2] {};

    \path[myarrow]
      (1) edge node {} (2)
      (3) edge node {} (1)
      (2) edge[-,dotted] node {} (5)


      ;
  \end{tikzpicture}
  \caption{First time steps of the Bayesian network representing a univariate Markov chain $\Xt$.}
  \label{fig:markovchain}
\end{center}
\end{figure}
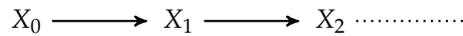

Markov chains are commonly used to model processes that have no memory of past states.
The Bayesian network\footnote{We abuse the terminology here by also referring to the directed acyclic graph $G$ associated to a Bayesian network as a ``Bayesian network''. Context always resolves this ambiguity however.} in \cref{fig:markovchain} shows three initial time steps of a univariate Markov chain. This is a discrete time stochastic process. This means that the index set $V$ is isomorphic to a contiguous subset of the integers. We assume that $V$ is also finite such that we can also assume (without further loss of generality) $V:=T=\{0,...,n-1\}$ with $n\in \Nplus$. Then the defining feature is that for all $t \in T$ the random variable with index $t+1$ only depends on the random variable indexed by $t$. We also assume that the state spaces $\X_t$ of the random variables $X_t$ are all equal. From the graph in \cref{fig:markovchain} we can read:
\begin{align}
\label{eq:markovjoint}
 p_T(x_T)&=\prod_{t\in T} p_t(x_t|x_{\pa(t)}),\\
 &=\prod_{t=1}^{n-1} p_t(x_t|x_{t-1}) p_0(x_0),
\end{align}
where $p_0$ is the initial distribution.
From \cref{eq:markovjoint} we can see the usual Markov chain condition:
\begin{equation}
\label{eq:markovcond}
 p_{t+1}(x_{t+1}|\xpet) = p_{t+1}(x_{t+1}|x_t),
\end{equation} 
where $\xpet=(x_i)_{i \leq t}$ is the entire history up to and including $t$. Starting from a Bayesian network we can then define the Markov chain as follows.
\begin{mydef}
\label{def:unimc}
 Let $T=\{0,1,...,n-1\}$, $n \in \Nplus$. 
 Then a \textit{univariate Markov chain} is a Bayesian network with random variables $\Xt$ and graph $G=(T,E)$ such that 
\begin{itemize}
 \item $\forall t_1,t_2 \in T: \X_{t_1} = \X_{t_2}$,
 \item $\pa(0)=\emptyset$,
 \item $\forall t\in T\bs \{0\}: \pa(t)=\{t-1\}$. 
\end{itemize}
\end{mydef}

We can also write the right hand side of \cref{eq:markovcond} as a Matrix called a Markov matrix.
\begin{mydef}
 Given a Markov chain $\{X_i\}_{i \in V}$ the Markov matrix $P_t$ at time step $t$ is the matrix with entries:
 \begin{equation}
  (P_t)_{x_t,x_{t-1}}:= p_t(x_t|x_{t-1}).
 \end{equation} 
\end{mydef}
Correspondingly we can write the probability distribution over any random variable $X_t$ as a column vector $p_t=(p_t(x_t))_{x_t \in \X_t}$. Then we can propagate thes probability distributions forward in time using the Markov matrix:
\begin{equation}
 p_{t+1} =  P_{t+1} p_t.
\end{equation} 

An especially simple case of Markov chains are time-homogenous Markov chains.
\begin{mydef}
A time-homogenous Markov chain is a Markov chain with index set $T$ such that for all $t_1,t_2 \in V$
\begin{equation}
 P_{t_1} = P_{t_2}.
\end{equation} 

\end{mydef}
So the dynamics of time-homogenous Markov chains do not change over time.

\begin{figure}
\begin{center}
  \begin{center}
\begin{tikzpicture}[transform shape,->,>=stealth,shorten >=2pt,auto,thick]
    \pgfmathtruncatemacro{\cols}{4}
    \pgfmathtruncatemacro{\firstcol}{0}
    \pgfmathtruncatemacro{\rows}{5}
    \pgfmathtruncatemacro{\fadingcols}{1}
    \pgfmathsetlengthmacro{\vdist}{1cm}
    \pgfmathsetlengthmacro{\hdist}{.2\textwidth}

    \pgfmathtruncatemacro{\secondcol}{\firstcol+1}
    \pgfmathtruncatemacro{\fullcols}{\cols-\fadingcols}
        
    \node (N-1-\firstcol) [] {$X_{1,0}$};

    \foreach \y in {\secondcol,...,\fullcols}{%
       \pgfmathtruncatemacro{\yonleft}{\y-1}
       \node (N-1-\y) [node distance=\hdist,right of=N-1-\yonleft] {$X_{1,\y}$};
    }
  \foreach \x in {2,...,\rows}{%
     \pgfmathtruncatemacro{\xabove}{\x-1}
     \foreach \y in {\firstcol,...,\fullcols}{%
       \node (N-\x-\y) [below of=N-\xabove-\y] {$X_{\x,\y}$};
     }
  } 
  
  \node (space) [node distance=1cm, left of=N-2-\firstcol, rotate=270] {degrees of freedom (DOFs) $\rightarrow$};

  \node (time) [node distance=1cm, below of=N-\rows-\firstcol, xshift=-.5cm] {time $\rightarrow$};

  \pgfmathtruncatemacro{\rowsminusone}{\rows-1}
  \foreach \x in {2,...,\rowsminusone}{%
    \foreach \y in {\secondcol,...,\fullcols}{%
	\pgfmathtruncatemacro{\yonleft}{\y-1}
	\pgfmathtruncatemacro{\xabove}{(Mod(\x-1-1,\rows)+1)}
	\pgfmathtruncatemacro{\xbelow}{Mod(\x-1+1,\rows)+1}
        \draw (N-\xabove-\yonleft) edge (N-\x-\y);
        \draw (N-\x-\yonleft) edge (N-\x-\y);
        \draw (N-\xbelow-\yonleft) edge (N-\x-\y);
    }
  }
  \pgfmathtruncatemacro{\x}{1}
  \foreach \y in {\secondcol,...,\fullcols}{%
	\pgfmathtruncatemacro{\yonleft}{\y-1}
	\pgfmathtruncatemacro{\xbelow}{2}
        \draw (N-\x-\yonleft) edge (N-\x-\y);
        \draw (N-\xbelow-\yonleft) edge (N-\x-\y);
  }
  \pgfmathtruncatemacro{\x}{\rows}
  \foreach \y in {\secondcol,...,\fullcols}{%
    \pgfmathtruncatemacro{\yonleft}{\y-1}
    \pgfmathtruncatemacro{\xabove}{4}
    \draw (N-\xabove-\yonleft) edge (N-\x-\y);
    \draw (N-\x-\yonleft) edge (N-\x-\y);
  }

    \foreach \y in {\cols}{%
       \pgfmathtruncatemacro{\yonleft}{\y-1}
       \node (N-1-\y) [node distance=\hdist,right of=N-1-\yonleft] {};
    }
  \foreach \x in {2,...,\rows}{%
     \pgfmathtruncatemacro{\xabove}{\x-1}
     \foreach \y in {\cols}{%
       \node (N-\x-\y) [below of=N-\xabove-\y] {};
     }
  }

  \pgfmathtruncatemacro{\rowsminusone}{\rows-1}
  \foreach \x in {2,...,\rowsminusone}{%
    \foreach \y in {\cols}{%
	\pgfmathtruncatemacro{\yonleft}{\y-1}
	\pgfmathtruncatemacro{\xabove}{(Mod(\x-1-1,\rows)+1)}
	\pgfmathtruncatemacro{\xbelow}{Mod(\x-1+1,\rows)+1}
        \draw (N-\xabove-\yonleft) edge[dotted,-] (N-\x-\y);
        \draw (N-\x-\yonleft) edge[dotted,-] (N-\x-\y);
        \draw (N-\xbelow-\yonleft) edge[dotted,-] (N-\x-\y);
    }
  }
  \pgfmathtruncatemacro{\x}{1}
  \foreach \y in {\cols}{%
	\pgfmathtruncatemacro{\yonleft}{\y-1}
	\pgfmathtruncatemacro{\xbelow}{2}
        \draw (N-\x-\yonleft) edge[dotted,-] (N-\x-\y);
        \draw (N-\xbelow-\yonleft) edge[dotted,-] (N-\x-\y);
  }
  \pgfmathtruncatemacro{\x}{\rows}
  \foreach \y in {\cols}{%
    \pgfmathtruncatemacro{\yonleft}{\y-1}
    \pgfmathtruncatemacro{\xabove}{4}
    \draw (N-\xabove-\yonleft) edge[dotted,-] (N-\x-\y);
    \draw (N-\x-\yonleft) edge[dotted,-] (N-\x-\y);
  }
  



\end{tikzpicture}
\end{center}
  \caption{First time steps of the Bayesian network representing a multivariate Markov chain $\Xt$. The shown edges are just an example, any two nodes within the same or subsequent columns can be connected.}
  \label{fig:mv}
\end{center}
\end{figure}
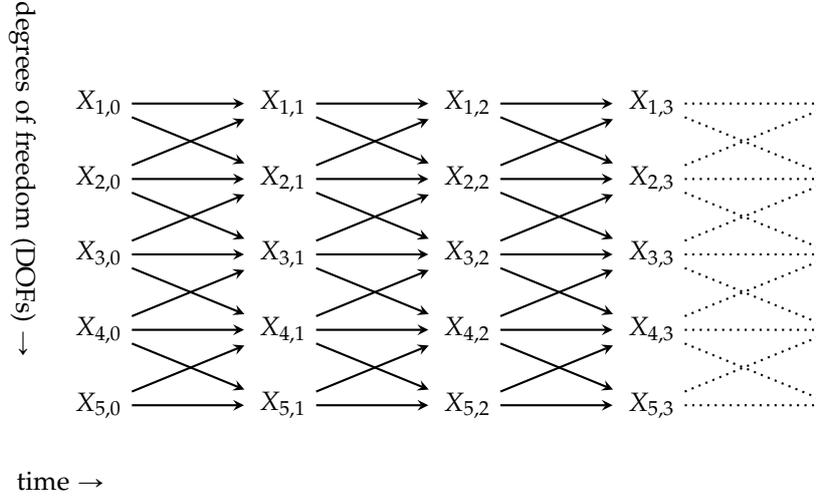

\subsection{Multivariate Markov chain}
\label{sec:multimc}
A multivariate Markov chain is just a Markov chain where the random variable $X_t$ at each point in time is replaced by a whole set of random variables. We also define their time-slices here which will be used throughout this thesis. We then show that time-slices obey the Markov property (this is not surprising and only included for technical reference). Finally we define the notions of space- and time-homogeneity, and instantaneous interactions.
\begin{mydef}[Multivariate Markov chain]
\label{def:mvmarkov}
 Let $T = \{0,...,n\}$ with $n \in \Nplus$, $J = \{1,...,d\}$ with $d \in \Nplus$ and $V=\{(j,t) \in J \times T\}$. 
Then a \textit{multivariate Markov chain} is a Bayesian network with random variables $\Xv$ and graph $G=(V,E)$ such that 
\begin{itemize}
 \item $\pa(j,0)\subseteq \{(k,0) \in V:k \in J\bs j\}$,
 \item $\pa(j,t)\subseteq \{(k,t) \in V:k \in J\bs j\} \cup \{(k,t-1):k \in J\}$.  
\end{itemize}
We call $j$ the \textit{spatial index} and $t$ the \textit{temporal index}. 
\end{mydef}
Remark:

\begin{itemize}
\item 
In essence we get a set of random variables $\{X_{j,t}\}_{j \in J}$ at each point in time $t$ which only depends either on variables at the same point in time or the previous point in time $t-1$ (see \cref{fig:mv} for the graph structure). This ensures that the joint random variable $(X_{j,t})_{j \in J}$ composed of all random variables at time $t$ only depends on the joint random variable $(X_{j,t-1})_{j \in J}$ composed of all random variables at time $t-1$ which ensure the Markov property of the joint random vairables.  Since these joint random variables occur repeatedly throughout this thesis we introduce a specific terminology and notation for them.
\item Dependencies among random variables $\{X_{j,t}\}_{j \in J}$ at the same point in time $t$ are explicitly allowed in our definition of multivariate Markov chains as they do not break the Markov property as we will see in \cref{thm:timeslicemarkov}. Note also that the graph of the entire Bayesian network of the multivariate Markov chain is still directed and acyclic so that we have no cycles among random variables at the same point in time either.
\end{itemize}

\begin{mydef}[Time-slices]
Let $V= J \times T$ be an index set composed of a spatial index $J$ and a temporal index $T$ then:
\begin{thmlist}
\item The \textit{time-slice $V_t$ of $V$ at time $t$} is the set of indices 
\begin{equation}
 V_t:=\{(j,t) \in V:j \in J\}.
\end{equation} 
\item Similarly, for any subset $A \subseteq V$ the \textit{time-slice $A_t$ of $A$ at time $t$} is the set of indices
\begin{equation}
 A_t:=\{(j,t) \in A:j \in J\}.
\end{equation} 
\item Given a multivariate Markov chain $\{X_j\}_{j \in V}$ with index set $V$ and a subset $A \subset V$ of indices the \textit{time-slice $X_{A_t}$ of $X_A$ at time $t$} is the joint random variable indicated by the time-slice $A_t$ of $A$
\begin{equation}
 X_{A_t}:=(X_j)_{j \in A_t}.
\end{equation} 
\end{thmlist}
\end{mydef}

\begin{thm}
\label{thm:timeslicemarkov}
 The time-slices $X_{V_t}$ of multivariate Markov chains $\Xv$ of \cref{def:mvmarkov} satisfy a the Markov property and therefore form a Markov chain. Formally, for all $t \in T$:
 \begin{equation}
p_{V_{t+1}}(x_{V_{t+1}}|x_{V_{\pet}})=p_{V_{t+1}}(x_{V_{t+1}}|x_{V_t}).                                                                                        \end{equation} 
\end{thm}
\begin{proof}
 According to \cref{def:mvmarkov} the parents of each node are composed of two subsets 
 \begin{align}
  \pa(j,t+1)&= \{(k,t+1) \in V:k \in J\bs j\} \cup \{(k,t):k \in J\}\\
  &=(\pa(j,t+1) \cap V_{t+1}) \cup (\pa(j,t+1) \cap V_t).
 \end{align}
This means that using $i=(j,t+1)$ we can write
\begin{align}
  p_{V_{t+1}}(x_{V_{t+1}}|x_{V_{\pet}}) &=\prod_{i \in V_{t+1}} p_i(x_i|x_{\pa(i)}) \\
 &= \prod_{i \in V_{t+1}} p_i(x_i|x_{(\pa(i) \cap V_{t+1}) \cup (\pa(i) \cap V_t)}) \\
 &= p_{V_{t+1}}(x_{V_{t+1}}|x_{(\bigcup_{i \in V_{t+1}} \pa(i))\bs V_{t+1}}) \\
 &= p_{V_{t+1}}(x_{V_{t+1}}|x_{V_t}). 
 \end{align}
\end{proof}

\begin{mydef}
 Given a multivariate Markov chain $\Xv$ the Markov matrix $P_t$ at time step $t$ is the matrix with entries:
 \begin{equation}
  (P_t)_{x_{V_t},x_{V_{t-1}}}:= p_{V_t}(x_{V_t}|x_{V_{t-1}}).
 \end{equation} 
\end{mydef}

\begin{mydef}[Space- and time-homogeneity]
\label{def:homogeneity}
 A multivariate Markov chain with random variables $\Xv$ and index set $V=J \times T$ is 
 \begin{thmlist}
  \item \textit{time-homogenous} if for all $t_1,t_2 \in T$
   \begin{enumerate}
    \item $\X_{V_{t_1}}=\X_{V_{t_2}}$,
    \item  for all $x_{V_{t_1}},x_{V_{t_2}} \in \X_{V_{t_1}}$ we have
  \begin{equation}
   p_{V_{t_1+1}}(x_{V_{t_1+1}}|x_{V_{t_1}})=p_{V_{t_2+1}}(x_{V_{t_2+1}}|x_{V_{t_2}}),
  \end{equation} 
  or in terms of the Markov matrices:
  \begin{equation}
    P_{t_1}=P_{t_2};
  \end{equation} 
   \end{enumerate}

  \item \textit{space-homogenous} if for all $j_1, j_2 \in J$ and all $t \in T$
   \begin{enumerate}
    \item $\X_{j_1,t}=\X_{j_2,t}$, 
    \item $\X_{\pa(j_1,t)}=\X_{\pa(j_2,t)}$,
    \item for all $x \in \X_{j_1,t}$, and $y \in \X_{\pa(j_1,t)}$ we have
  \begin{equation}
  \label{eq:mechanismequality}
   p_{j_1,t}(x|y) = p_{j_2,t}(x|y).
   \end{equation} 
   \end{enumerate}
 \end{thmlist}
\end{mydef}
Remark:
\begin{itemize}
\item 
So space-homogeneity means that all mechanisms associated to the random variables within a given time-slice are the same. 
\end{itemize}

\begin{mydef}
An \textit{instantaneous interaction} is an edge $(X_{j,t},X_{k,t}) \in E$ between random variables within the same time-slice. 
\end{mydef}
Remarks:
\begin{itemize}
 \item An example of a multivariate Markov chain without instantaneous interactions 
 is the perception-action loop of \cref{sec:paloopformal}.  
 \item Examples of space- and time-homogenous, deterministic, multivariate Markov chains without instantaneous interactions include the elementary cellular automata \citep{wolfram_statistical_1983} as well as the Game of Life cellular automaton \citep{conway_game_1970}. 
\end{itemize}

%

\subsection{Driven processes}
\label{sec:driven}
\begin{figure}
\begin{center}
  \begin{tikzpicture}
    [myarrow,node distance=2cm]

    \node[] (1) [] {$X_0$};
    \node[] (2) [right of=1] {$X_{1}$};
    \node[] (3) [right of=2] {$X_{2}$};
    \node[] (4) [right of=3] {};
    \node[] (5) [below of=1] {$Y_0$};
    \node[] (6) [below of=2] {$Y_1$};
    \node[] (7) [below of=3] {$Y_2$};
    \node[] (8) [below of=4] {};


    \path[myarrow]
      (1) edge node {} (2)
      (2) edge node {} (3)
      (3) edge[-,dotted] node {} (4)
      (5) edge node {} (1)
      (6) edge node {} (2)
      (7) edge node {} (3)

      (5) edge node {} (6)
      (6) edge node {} (7)
      (7) edge[-,dotted] node {} (8)

      (1) edge node {} (6)
      (2) edge node {} (7)
      (3) edge[-,dotted] node {} (8)
      



      ;
  \end{tikzpicture}
  \caption{First time steps of the Bayesian network representing a process $\Yt$ driving a process $\Xt$.}
  \label{fig:drivenchain}
\end{center}
\end{figure}
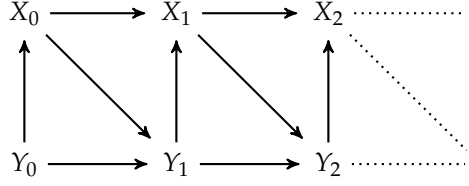
A driven process can model systems under the influence of changing external factors or control parameters. An example for such a system is the geosphere which is driven by influence from cosmic and solar radiation. At the same time earth also emits radiation into the cosmos.
Acccordingly, in general we allow interaction in both directions with the driving process $\Yt$. To define the driven process $\Xt$ we define a Bayesian network with two interacting processes and rename one of them $\Yt$ in order to simplify discussions. We assume that both processes have constant state spaces. For a visualization of the graph of the according Bayesian network see \cref{fig:drivenchain}.
\begin{mydef}[Driven process]
\label{def:drivenchain}
Let $T = \{0,...,n\}$ with $n \in \Nplus$ and $V=\{(j,t) \in \{1,2\} \times T\}$. 
Consider the Bayesian network with:
\begin{itemize}
\item $\forall (j,t_1),(j,t_2) \in V: \X_{j,t_1} = \X_{j,t_2}$
 \item $\pa(1,0):=\emptyset$,
 \item $\pa(2,0):=\{(1,0)\}$,
 \item $\forall t \in T\bs \{0\} : \pa(1,t))=\{(1,t-1),(2,t-1)\}$,
 \item $\forall t \in T\bs \{0\} : \pa(2,t))=\{(1,t),(2,t-1)\}$.  
\end{itemize}
Rename:
\begin{itemize}
 \item $\{X_{1,t}\}_{t \in T}=:\{Y_t\}_{t \in T}$,
 \item $\{X_{2,t}\}_{t \in T}=:\{X_t\}_{t \in T}$.
\end{itemize}
Then $\{Y_t\}_{t \in T}$ is called the \textit{driving process} and $\{X_t\}_{t \in T}$ the \textit{driven process}.
\end{mydef}
Remark:
\begin{itemize}
\item 
We note that ${(X_t,Y_t)}_{t \in T}$ is a bivariate Markov chain. 
\item We choose the driving process to interact instantaneously with the driven process as a convention. The main aspect of a driven process is that it highlights the possibility to pay particular attention to the driven process's dynamics and ignore those of the driving process.
\end{itemize}

\begin{mydef}[Driven multivariate Markov chain]
\label{def:mvdrivenchain}
Let $T = \{0,...,n\}$ with $n \in \Nplus$, $J= A \cup B$ with $A \cap B = \emptyset$, and $V=\{(j,t) \in J \times T\}$. 
Consider the Bayesian network with:
\begin{itemize}
\item $\forall  (j,t_1),(j,t_2) \in V: \X_{j,t_1} = \X_{j,t_2}$
 \item for $j \in B$, $\pa(j,0):=\emptyset$,
 \item for $j \in A$, $\pa(j,0)\subset \{(k,0): k \in B\}$,
 \item $\forall t \in T\bs \{0\}, j \in B : \pa(j,t))\subset \{(k,t-1): k \in A \cup B \}$,
 \item $\forall t \in T\bs \{0\}, j \in A : \pa(j,t))\subset \{(k,t-1): k \in A\} \cup \{(k,t) : k \in B \}$.  
\end{itemize}
Then $\{X_{B_t}\}_{t \in T}$ is called the \textit{driving chain} and $\Xat$ the \textit{(multivariate) driven Markov chain}.
\end{mydef}
Remark:
\begin{itemize}
\item 
We note that ${(X_{V_t})}_{t \in T}$ is a multivariate Markov chain.
\item If $B =\emptyset$ then $\Xv=\Xat$ is a multivariate Markov chain.
\item A trajectory of the driven Markov chain is a pattern $x_{A,T} \in \X_{A,T}$ where $(A,T) := A \times T$.
\item The multivariate driven Markov chain models a situation where the focus is on the multivariate process $\Xat$ even if it is influenced by other (usually very simple) processes contained in $\{X_{B_t}\}_{t \in T}$.
\item From the definition we also can write:
\begin{equation}
  p_{V_{t+1}}(x_{V_{t+1}}|x_{V_t})=p_{A_{t+1}}(x_{A_{t+1}}|x_{B_{t+1}},x_{A_t}) p_{B_{t+1}}(x_{B_{t+1}}|x_{A_t},x_{B_t}).
\end{equation} 
\item See \cref{fig:drivenmultimc} for an example of a driven multivariate Markov chain.
\end{itemize}

\begin{figure}
\begin{center}
  \begin{center}
\begin{tikzpicture}[transform shape,->,>=stealth,shorten >=2pt,auto,thick]
    \pgfmathtruncatemacro{\cols}{4}
    \pgfmathtruncatemacro{\firstcol}{0}
    \pgfmathtruncatemacro{\rows}{3}
    \pgfmathtruncatemacro{\drivingrows}{1}
    \pgfmathtruncatemacro{\fadingcols}{1}
    \pgfmathsetlengthmacro{\vdist}{1cm}
    \pgfmathsetlengthmacro{\hdist}{.2\textwidth}
    \pgfmathsetlengthmacro{\drivedist}{.1\textwidth}

    \pgfmathtruncatemacro{\secondcol}{\firstcol+1}
    \pgfmathtruncatemacro{\fullcols}{\cols-\fadingcols}
    \pgfmathtruncatemacro{\firstdrivingrow}{\rows+1}
    \pgfmathtruncatemacro{\totalrows}{\rows+\drivingrows}
    
    \node (N-1-\firstcol) [] {$X_{1,0}$};

    \foreach \y in {\secondcol,...,\fullcols}{%
       \pgfmathtruncatemacro{\yonleft}{\y-1}
       \node (N-1-\y) [node distance=\hdist,right of=N-1-\yonleft] {$X_{1,\y}$};
    }

  \foreach \x in {2,...,\rows}{%
     \pgfmathtruncatemacro{\xabove}{\x-1}
     \foreach \y in {\firstcol,...,\fullcols}{%
       \node (N-\x-\y) [below of=N-\xabove-\y] {$X_{\x,\y}$};
     }
  } 

    \pgfmathtruncatemacro{\firstdriverow}{\rows+1}
    \node (N-\firstdriverow-\firstcol) [node distance=\drivedist, below of=N-\rows-\firstcol] {$Y_{0}$};

    \foreach \y in {\secondcol,...,\fullcols}{%
       \pgfmathtruncatemacro{\yonleft}{\y-1}
       \node (N-\firstdriverow-\y) [node distance=\hdist,right of=N-\firstdriverow-\yonleft] {$Y_{\y}$};
    }
  

  \node (space) [node distance=1cm, left of=N-2-\firstcol, rotate=270] {degrees of freedom (DOFs) $\rightarrow$};

  \node (time) [node distance=1cm, below of=N-\totalrows-\firstcol, xshift=-.5cm] {time $\rightarrow$};

  \pgfmathtruncatemacro{\rowsminusone}{\rows-1}
  \foreach \x in {2,...,\rowsminusone}{%
    \foreach \y in {\secondcol,...,\fullcols}{%
	\pgfmathtruncatemacro{\yonleft}{\y-1}
	\pgfmathtruncatemacro{\xabove}{(Mod(\x-1-1,\rows)+1)}
	\pgfmathtruncatemacro{\xbelow}{Mod(\x-1+1,\rows)+1}
        \draw (N-\xabove-\yonleft) edge (N-\x-\y);
        \draw (N-\x-\yonleft) edge (N-\x-\y);
        \draw (N-\xbelow-\yonleft) edge (N-\x-\y);
    }
  }
  \pgfmathtruncatemacro{\x}{1}
  \foreach \y in {\secondcol,...,\fullcols}{%
	\pgfmathtruncatemacro{\yonleft}{\y-1}
	\pgfmathtruncatemacro{\xbelow}{2}
        \draw (N-\x-\yonleft) edge (N-\x-\y);
        \draw (N-\xbelow-\yonleft) edge (N-\x-\y);
  }
  \pgfmathtruncatemacro{\x}{\rows}
  \foreach \y in {\secondcol,...,\fullcols}{%
    \pgfmathtruncatemacro{\yonleft}{\y-1}
    \pgfmathtruncatemacro{\xabove}{4}
    \draw (N-\xabove-\yonleft) edge (N-\x-\y);
    \draw (N-\x-\yonleft) edge (N-\x-\y);
  }
  
  \pgfmathtruncatemacro{\rowsminusone}{\rows-1}
  \foreach \y in {\secondcol,...,\fullcols}{%
	\pgfmathtruncatemacro{\yonleft}{\y-1} 
	\pgfmathtruncatemacro{\yonright}{\y+1}
	\draw (N-\firstdrivingrow-\yonleft) edge (N-\firstdrivingrow-\y);
    \foreach \x in {1,...,\rows}{%
%
        \draw (N-\x-\yonleft) edge (N-\firstdrivingrow-\y);
        \draw (N-\firstdrivingrow-\yonleft) edge[bend right=45] (N-\x-\yonleft);
        
    }
  }

    \foreach \y in {\cols}{%
       \pgfmathtruncatemacro{\yonleft}{\y-1}
       \node (N-1-\y) [node distance=\hdist,right of=N-1-\yonleft] {};
    }
  \foreach \x in {2,...,\rows}{%
     \pgfmathtruncatemacro{\xabove}{\x-1}
     \foreach \y in {\cols}{%
       \node (N-\x-\y) [below of=N-\xabove-\y] {};
     }
  } 
  \node (N-\firstdrivingrow-\cols) [node distance=\drivedist, below of=N-\rows-\cols] {};
  
    	\pgfmathtruncatemacro{\yonleft}{\cols-1} 
  	\pgfmathtruncatemacro{\y}{\cols} 
	\draw (N-\firstdrivingrow-\yonleft) edge[dotted,-] (N-\firstdrivingrow-\y);
    \foreach \x in {1,...,\rows}{%
%
        \draw (N-\x-\yonleft) edge[dotted,-] (N-\firstdrivingrow-\y);
        \draw (N-\firstdrivingrow-\yonleft) edge[bend right=45] (N-\x-\yonleft);
        
    }

  \pgfmathtruncatemacro{\rowsminusone}{\rows-1}
  \foreach \x in {2,...,\rowsminusone}{%
    \foreach \y in {\cols}{%
	\pgfmathtruncatemacro{\yonleft}{\y-1}
	\pgfmathtruncatemacro{\xabove}{(Mod(\x-1-1,\rows)+1)}
	\pgfmathtruncatemacro{\xbelow}{Mod(\x-1+1,\rows)+1}
        \draw (N-\xabove-\yonleft) edge[dotted,-] (N-\x-\y);
        \draw (N-\x-\yonleft) edge[dotted,-] (N-\x-\y);
        \draw (N-\xbelow-\yonleft) edge[dotted,-] (N-\x-\y);
    }
  }
  \pgfmathtruncatemacro{\x}{1}
  \foreach \y in {\cols}{%
	\pgfmathtruncatemacro{\yonleft}{\y-1}
	\pgfmathtruncatemacro{\xbelow}{2}
        \draw (N-\x-\yonleft) edge[dotted,-] (N-\x-\y);
        \draw (N-\xbelow-\yonleft) edge[dotted,-] (N-\x-\y);
  }
  \pgfmathtruncatemacro{\x}{\rows}
  \foreach \y in {\cols}{%
    \pgfmathtruncatemacro{\yonleft}{\y-1}
    \pgfmathtruncatemacro{\xabove}{4}
    \draw (N-\xabove-\yonleft) edge[dotted,-] (N-\x-\y);
    \draw (N-\x-\yonleft) edge[dotted,-] (N-\x-\y);
  }
  



\end{tikzpicture}
\end{center}
  \caption{First time steps of the Bayesian network representing a multivariate process $\Xt$ driven by a process $\Yt$. Note that the process $\Yt$ can also be multivariate, but this would further clutter the graph. Also note that not all edges depicted here must be present. Here, each random variable in each time-slice of the driven process is influenced by the driving process and influences it.  
  }
  \label{fig:drivenmultimc}
\end{center}
\end{figure}
%
\subsection{Perception-action loop}
\label{sec:paloopformal}
Here we formally introduce the Bayesian network of the perception-action loop. We first introduce a simple version and then show how to extract implied actions and perceptions without altering the global probability distribution over agent and environment processes. The perception-action loop has been employed to define informational closure \citep{bertschinger_information_2006}, autonomy \citep{bertschinger_autonomy_2008}, and morphological computation \citep{zahedi_quantifying_2013} of agents. In \cref{sec:palooprelation} we discuss the relaiton of our concept of agents to the agent concept that is implicit in the perception-action loop.

Conceptually perception-action loops go back at least to \citet{uexkull_theoretische_1920}. Recent formalizations of the perception-action loop (also perception-action cycle, sensorimotor loop) due to \citet{beer_dynamical_1995} using dynamical systems and \citet{klyubin_organization_2004} using Bayesian networks. Since then it has also been employed, sometimes with minor alterations, by \citet{bertschinger_information_2006,bertschinger_autonomy_2008,zahedi_higher_2010,salge_digested_2011,ay_causal_2014}.


For a rough intuition think of the  \textit{perception-action} as modelling the interactions between an ``agent'' and its ``environment'' that occur over time. Roughly speaking, at each time-step the agent influences the environment via actions and the environment influences the agent via the latter's perceptions (sensors). It is important to note that in the formal models \textit{all} interactions between agent and environment are captured by the sensor values and actions \citep{beer_dynamical_1995}.

%
%

\begin{figure}
\begin{center}
  \begin{tikzpicture}

  \matrix (m) [matrix of math nodes, nodes in empty cells,row sep=1.5cm,column sep=1.5cm]
  {
     E_0 &      E_1 &      E_2 & \vphantom{E_3} \\
     M_0 &      M_1 &      M_2 & \vphantom{M_3} \\
   };
  \path[myarrow]
    (m-1-1) edge node {} (m-2-2)
            edge node {} (m-1-2)
    (m-2-1) edge node {} (m-2-2)
            edge node {} (m-1-2)
    

    (m-1-2) edge node {} (m-2-3)
            edge node {} (m-1-3)
    (m-2-2) edge node {} (m-2-3)
            edge node {} (m-1-3)
            
    (m-1-3) edge[-,dotted] node {} (m-2-4)
            edge[-,dotted] node {} (m-1-4)
    (m-2-3) edge[-,dotted] node {} (m-2-4)
            edge[-,dotted] node {} (m-1-4)

%
    ;            
\end{tikzpicture}
  \caption{First timesteps of the Bayesian network of the perception-action loop. 
The processes represent environment $\Et$, 
and agent memory $\Mt$. 
}
  \label{fig:paloop}
\end{center}
\end{figure}
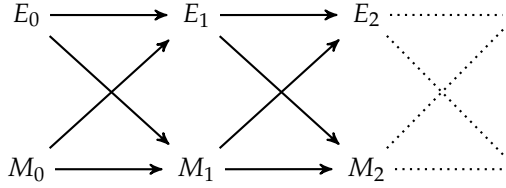

Let us first consider a simple perception-action loop consisting of agent and environment only. We assume here that there are no instantaneous interactions between agent and environment. This is particularly suitable for the situation where we relate our conception of agents to that underlying the perception-action loop in \cref{sec:palooprelation}. In the perception-action loop the agent memory is represented by a sequence of random variables $\Mt$, the environment state by $\Et$. At each time-step the agent memory $M_t$ is influenced by the last memory state $m_{t-1}$ and the last environment state $M_{t-1}$. Conversely the environment state $E_t$ is influenced by $E_{t-1}$ and $M_{t-1}$. 
See \cref{fig:paloop} for the Bayesian network.
%
%
Next, we define the perception-action loop formally.
\begin{mydef}[Perception-action loop]
\label{def:paloop}
Let $T = \{0,...,n\}$ with $n \in \Nplus$ and $V=\{(j,t) \in \{1,2\} \times T\}$. 
Consider the Bayesian network $\Xv$ with:
\begin{itemize}
 \item $\pa(1,0):=\emptyset$,
 \item $\pa(2,0):=\emptyset$,
 \item $\forall t \in T\bs \{0\} : \pa(1,t))=\{(1,t-1),(2,t-1)\}$,
 \item $\forall t \in T\bs \{0\} : \pa(2,t))=\{(1,t-1),(2,t-1)\}$,
\end{itemize}
Rename:
\begin{itemize}
 \item $\{X_{1,t}\}_{t \in T}=:\{E_t\}_{t \in T}$,
 \item $\{X_{2,t}\}_{t \in T}=:\{M_t\}_{t \in T}$.
\end{itemize}
The Bayesian network is called the \textit{perception-action loop}, $\Et$ is called the \textit{environment process}
and $\Mt$ the \textit{agent} or \textit{memory process}.
\end{mydef}
It is customary to explicitly represent actions and perceptions in the perception-action loop. In \cref{def:paloop} these are implicit. The interactions between agent and environment are not represented by random variables. As mentioned before in the tradition of perception action loops all interactions between agent and environment are considered to be captured by the actions and the perceptions \citep{beer_dynamical_1995}. 

We can construct sequences of random variables representing for each $t$ the influences of $\Mt$ on $\Et$ and vice versa such that the probability distribution $p_V$ over all random variables remains the same. This means we introduce another Bayesian network, which we will also call the perception-action loop containing two more processes, the action process $\At$ and the sensor process $\St$. The result will be the perception-action loop considered in \citet{bertschinger_information_2006,bertschinger_autonomy_2008}.

The action at a time $t$ is a block in a partition $\mu_t$ of the state space $\M_t$ of the agent at $t$. These blocks are composed out of agent states that have the same effect on the environment's transitions from $E_t$ to $E_{t+1}$. Formally, 
we define the partition $\mu_t$ via the equivalence relation $\equiv_{\mu_t}$ relating the elements of each block in $\mu_t$. 

\begin{mydef}
  Given a perception-action loop $\Xv$. For each time $t \in T$ and $m^1_t, m^2_t \in \M_t$ let 
  \begin{equation}
     m^1_t \equiv_{\mu_t} m^2_t \Leftrightarrow \forall e_{t+1} \in \E_{t+1}, e_t \in \E_t: p_{E_{t+1}}(e_{t+1} | m^1_t, e_t) = p_{E_{t+1}}(e_{t+1} | m^2_t, e_t).
  \end{equation} 
  Then:
  \begin{thmlist}
  \item The \textit{action partition $\mu_t$ }is then defined as the set of equivalence classes of the equivalence relation $\equiv_{\mu_t}$.
  \item The \textit{set of actions} is defined as $\A_t := \mu_t$ and an element $a_t \in \A_t$ (which is also a block in $\mu_t$ is called an \textit{action}.
  \item The \textit{action function $f_{A_t}:\M_t \rightarrow \A_t$} is defined by
  \begin{equation}
    f_{A_t}(m_t)=m_t/\mu_t,
  \end{equation} 
  where $m_t/\mu_t$ is the block in $\mu_t$ containing $m_t$ (which is also an action).
  \end{thmlist}
\end{mydef}
Remark:
\begin{itemize}
  \item The construction of the action partition\footnote{The author thanks Benjamin Heuer for originally pointing us to this construction.} is not new. It is also used for example in \citet{balduzzi_detecting_2011} to obtain coarser states (alphabet) of joint random variables. It is also similar to the construction of causal states \citep{shalizi_causal_2001}. Causal states are usually a partition of pasts $x_{\preceq t}$ according to equal future morphs $p(X_{t \prec}|x_{\preceq t})$. Here we use equal transition probabilities (``transition morphs'') of another process to partition the current states. 
\end{itemize}

In the same way we define perceptions or sensor values via a partition $\epsilon_t$ of $\E_t$.
\begin{mydef}
\label{def:paloopperception}
  Given a perception-action loop $\Xv$. For each time $t \in T$ and $\hat{e}_t, \bar{e}_t \in \E_t$ let 
  \begin{equation}
\label{eq:perceptionequiv}
     \hat{e}_t \equiv_{\epsilon_t} \bar{e}_t \Leftrightarrow \forall m_{t+1} \in \M_{t+1}, m_t \in \M_t : p_{M_{t+1}}(m_{t+1} |  m_t,\hat{e}_t) = p_{M_{t+1}}(m_{t+1} | m_t, \bar{e}_t).
  \end{equation} 
  Then:
  \begin{thmlist}
  \item The \textit{sensor partition $\epsilon_t$ }is then defined as the set of equivalence classes of the equivalence relation $\equiv_{\epsilon_t}$.
  \item The \textit{set of sensor values} is defined as $\S_t := \epsilon_t$ and an element $s_t \in \S_t$ (which is also a block in $\epsilon_t$ is called a \textit{perception} of a \textit{sensor value}.
  \item The \textit{sensor function $f_{S_t}:\E_t \rightarrow \S_t$} is defined by
  \begin{equation}
    f_{S_t}(e_t)=e_t/\epsilon_t,
  \end{equation} 
  where $e_t/\epsilon_t$ is the block in $\epsilon_t$ containing $e_t$ (which is also a sensor value).
  \end{thmlist}
\end{mydef}

With these definitions we can extend the Bayesian network of the perception-action loop by the action process $\At$ and the sensor process $\St$ without altering the probability distribution $p_V$ over all random variables $\Xv$ in the orginal perception-action loop of \cref{def:paloop}. 

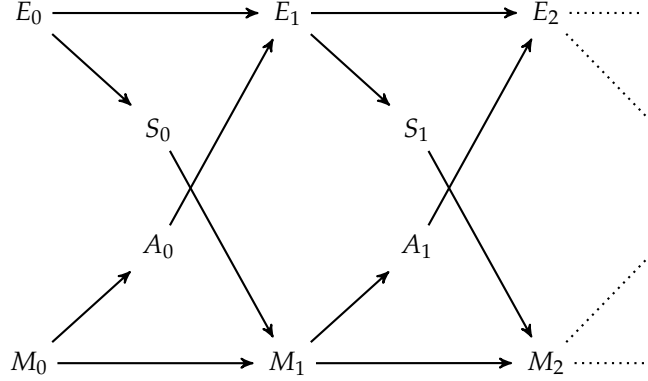
\begin{figure}
\begin{center}
  \begin{tikzpicture}

  \matrix (m) [matrix of math nodes, nodes in empty cells,row sep=1cm,column sep=1cm]
  {
     E_0 &     & E_1 &     & E_2 & \vphantom{E_3} \\
         & S_0 &     & S_1 &     & \vphantom{S_2} \\
         & A_0 &     & A_1 &     & \vphantom{A_2} \\
     M_0 &     & M_1 &     & M_2 & \vphantom{M_3} \\
   };
  \path[myarrow]
    (m-1-1) edge node {} (m-2-2)
            edge node {} (m-1-3)
    (m-2-2) edge node {} (m-4-3)
    (m-3-2) edge node {} (m-1-3)
    (m-4-1) edge node {} (m-3-2)
            edge node {} (m-4-3)

    (m-1-3) edge node {} (m-2-4)
            edge node {} (m-1-5)
    (m-2-4) edge node {} (m-4-5)
    (m-3-4) edge node {} (m-1-5)
    (m-4-3) edge node {} (m-3-4)
            edge node {} (m-4-5)
            
    (m-1-5) edge[-,dotted] node {} (m-2-6)
            edge[-,dotted] node {} (m-1-6)
    (m-4-5) edge[-,dotted] node {} (m-3-6)
            edge[-,dotted] node {} (m-4-6)
    
    ;            
\end{tikzpicture}
  \caption{First time-steps of the Bayesian network of the extended perception-action loop. The processes $\At$ and $\St$ mediate all interactions between $\Mt$ and $\Et$ without changing the probability distributions over the latter (see \cref{thm:paloopext}).}
  \label{fig:expaloop}
\end{center}
\end{figure}

First, we define the extended perception-action loop. For the Bayesian network see \cref{fig:expaloop}.
\begin{mydef}[Extended perception-action loop]
\label{def:expaloop}
Let $\Xv= \{M_t,E_t\}_{t \in T}$ be a perception-action loop. Then add to $\Xv$ the sets of random variables $\At$ called the action process and $\St$ called the sensor process such that 
\begin{itemize}
\item $\forall t \in T$
 \item$\pa(M_{t+1}):=\{M_t,S_t\}$,
 \item $\pa(A_t):=M_t$,
 \item $\pa(S_t):=E_t$,
 \item$\pa(E_{t+1}):=\{E_t,A_t\}$.
\end{itemize}
Furthermore, with $f_{A_t}$ the action functions, $f_{S_t}$ the sensor functions, $p^\ext$ denoting probability distributions of the extended perception-action loop, and $p$ denoting probability distributions of the original perception-action loop: 
\begin{itemize}
\item $p^\ext_{M_0,E_0}(m_0,e_0):=p_{M_0,E_0}(m_0,e_0)$
  \item $p^\ext_{A_t}(a_t | m_t):=\delta_{f_{A_t}(m_t)}(a_t)$,
  \item $p^\ext_{S_t}(s_t | e_t):=\delta_{f_{S_t}(e_t)}(s_t)$,
  \item $p^{\ext}_{M_{t+1}}(m_{t+1} | m_t,s_t):=p_{M_{t+1}}(m_{t+1} | m_t, e_t \in f^{-1}_{S_t}(s_t))$,
  \item $p^{\ext}_{E_{t+1}}(e_{t+1} | a_t,e_t):=p_{E_{t+1}}(e_{t+1} | m_t \in f^{-1}_{A_t}(a_t), e_t)$.
\end{itemize}
Then the resulting Bayesian network $\Xw = \{M_t,A_t,S_t,E_t\}_{t \in \Nplus}$ with probability distribution $p^\ext_W$ is called the \textit{extended perception-action loop} of $Xv$.
\end{mydef}
Remark:
\begin{itemize}
  \item Since, by definition of $f_{S_t}$, for any $t \in T$ and $s_t \in \S_t$ all $\bar{e}_t \in f^{-1}_{S_t}(s_t)$ have the same $p_{M_{t+1}}(m_{t+1} | m_t, \bar{e}_t)$ the definition of $p^{\ext}_{M_{t+1}}(m_{t+1} | m_t,s_t)$ is unambiguous. The corresponding argument holds for $f_{A_t}$ so that the definition of $p^{\ext}_{E_{t+1}}(e_{t+1} | a_t,e_t)$ is also unambiguous.
\end{itemize}

We then have the following theorem:
\begin{thm}[Invariant extension theorem]
\label{thm:paloopext}
Given a perception action loop $\Xv =\{M_t,E_t\}_{t \in T}$ and its extended perception-action loop $\Xw=\{M_t,A_t,S_t,E_t\}_{t \in \Nplus}$. Let $p_V=p_{M_T,E_T}$ be the probability distribution over the entire perception action loop $\Xv$ and let $p^{\ext}_{M_T,E_T}$ be the marginal probability distribution over the memory and environment process obtained from the probability distribution $p^{\ext}_W$ over the entire extended perception-action loop. Then
\begin{equation}
p_{M_T,E_T} = p^{\ext}_{M_T,E_T}.
\end{equation} 
\end{thm}
\begin{proof}
\begin{align}
\begin{split}   p^{\ext}_{M_T,E_T}(m_T,e_T)&=
    \sum_{a_T}\sum_{s_T} \prod_{t=1}^{n-1} p^\ext_{M_t}(m_t|m_{t-1},s_{t-1}) p^\ext_{S_{t-1}}(s_{t-1}|e_{t-1}) \\
   &\phantom{=\sum_{a_T}\sum_{s_T} \prod_{t=1}^{n-1}}p^\ext_{E_t}(e_t|a_{t-1},e_{t-1}) p^\ext_{A_{t-1}}(a_{t-1}|m_{t-1}) p^\ext_{M_0,E_0}(m_0,e_0)  
\end{split}\\
\begin{split}  &= \sum_{a_T}\sum_{s_T} \prod_{t=1}^{n-1} p^\ext_{M_t}(m_t|m_{t-1},s_{t-1}) \delta_{f_{S_{t-1}}(e_{t-1})}(s_{t-1}) \\
&\phantom{=\sum_{a_T}\sum_{s_T} \prod_{t=1}^{n-1}}p^\ext_{E_t}(e_t|a_{t-1},e_{t-1}) \delta_{f_{A_{t-1}}(m_{t-1})}(a_{t-1}) p^\ext_{M_0,E_0}(m_0,e_0) 
\end{split}\\
\begin{split}   &= \prod_{t=1}^{n-1} p^\ext_{M_t}(m_t|m_{t-1},f_{S_{t-1}}(e_{t-1})) \\
&\phantom{=\prod_{t=1}^{n-1}}p^\ext_{E_t}(e_t|f_{A_{t-1}}(m_{t-1}),e_{t-1}) p^\ext_{M_0,E_0}(m_0,e_0)  
\end{split}
\\
\begin{split}   &=\prod_{t=1}^{n-1} p_{M_t}(m_t|m_{t-1},\bar{e}_{t-1} \in f^{-1}_{S_{t-1}}\circ f_{S_{t-1}}(e_{t-1})) \\
&\phantom{=\prod_{t=1}^{n-1}} p_{E_t}(e_t|\bar{m}_{t-1} \in f^{-1}_{A_{t-1}}\circ f_{A_{t-1}}(m_{t-1}),e_{t-1}) p_{M_0,E_0}(m_0,e_0)   
\end{split}\\
   &=p_{M_T,E_T}(m_T,e_T).
\end{align}
\end{proof}
Remarks:
\begin{itemize}
  \item This proof shows that the introduction of action and sensor process in the way described by \cref{def:expaloop} only makes the interactions between agent and environment processes explicit. Action and sensor processes are not essential to the perception-action loop and do not introduce any additional dynamics. They only represent what the environment ``sees'' of the agent and vice versa. In other words the dynamics of agent and environment do not \textit{require} that the states space $\M_t \times \E_t$ of a time-slice is extended to $\M_t \times \A_t \times \S_t \times E_t$. This insight is also in line with \citet{bertschinger_information_2006} which uses the extended perception-action loop and refers to the action and sensor processes as ``channels''.
  \item This proof also shows that the sensor process (and conversely the actions) captures all influences from the environment on the agent. Else the dynamics of the extended perception-action loop could remain identical. We use this fact as a starting point for our conception of entity perception in \cref{sec:perceptions}. There we want to capture all influences of the environment on a set of ``spatiotemporal patterns'' or ``entities'' instead of on a stochastic process like $\Mt$. This will require a generalization of the perception extraction procedure in \cref{def:paloopperception}.
\end{itemize}

\chapter{Spatiotemporal patterns}
\label{sec:slicli}
\label{ch:stp}

%

This chapter constitutes the formal part of this thesis. We present here a first investigation of the properties and structure of spatiotemporal patterns\footnote{In this section we will always speak of spatiotemporal patterns in Bayesian networks instead of patterns in sets of random variables. There is no formal difference however. We only want to emphasise that we will consider spatiotemporally extended patterns whenever there are well defined notions of space and time in this thesis. Also note that every set of random variables can just be seen as a Bayesian network (possibly without any edges).} (STP) in Bayesian networks in relation to the novel measure of \textit{specific local integration} (SLI). Apart from SLI we also present the derived measure of complete local integration (CLI). The main result is the \textit{disintegration theorem} which relates the SLI of whole trajectories of Bayesian networks to the CLI of parts of these trajectories and vice versa. The connection between the two is revealed by the \textit{disintegration hierarchy} (\cref{def:dishier}) and its \textit{refinement-free} version (\cref{def:rfreedishier}). These are our own constructions. In \cref{sec:symm} we also formally define the effect of symmetry transformations on STPs in Bayesian networks and derive the behaviour of SLI under such transformations. The main result in this respect are the SLI symmetry theorems (\cref{thm:symmpart,thm:symmpartcor}). Finally, we establish under what circumstances spatial symmetries spread throughout the entire Bayesian network if the Bayesian networks are multivariate Markov chains or driven multivariate Markov chains. The according theorems are not new but provide the connection to more practical scenarios of cellular automata and reaction diffusion systems. Later, in \cref{sec:exstp} we visualise the disintegration hierarchies, completely locally integrated STPs, and use the SLI symmetry theorems to explain its structure.  

In more detail the chapter contains the following:

\begin{itemize}
\item In \cref{sec:plattbn} we give the definition of the partition lattice of Bayesian networks. This lattice is an underlying structure throughout this thesis.
\item In \cref{sec:sli} we define SLI. We constructively prove its upper bounds and construct an example of a STP with a particularly low (and negative) SLI. These constructions are mainly of technical interest and not employed conceptually. We also propose a normalised version of SLI which will not be further used but may be of interest for future research. Finally we derive some algebraic properties of differences between specific local integrations. These are given for technical reference.
\item In \cref{sec:cli} we state the definition of CLI. This is an important notion throughout this thesis and the basis for the definition of entities in multivariate Markov chains in \cref{sec:entinmvmc}.
\item In \cref{sec:disint} we define the disintegration hierarchies and prove the disintegration theorem.
\item In \cref{sec:symm} we first introduce notation and terminology to express the effect of permutations of nodes within a Bayesian network on STPs, partitions, and probabilities of spatiotemporal patterns. Then we prove the SLI symmetry theorems which specify the behaviour of SLI under permutation symmetries of the Bayesian network. These can provide insights into the structure of the disintegration hierarchies for systems with high degrees of symmetries like cellular automata. We will see how this can be done in \cref{sec:exstp}. Furthermore, symmetry properties are of general interest for future theoretical developments.
\item In \cref{sec:symmmc} we look at the Bayesian networks that are multivariate Markov chains and derive conditions under which spatial symmetries of initial distribution and Markov matrix spread the spatial symmetry over the entire Bayesian network. We also do this for driven multivariate Markov chains. The SLI symmetry theorems concern symmetries of STPs i.e.\ they depend on symmetries that are not purely spatial. The theorems in this section provides a way to obtain such extended symmetries from more simple and often well known ones. The inclusion of driven multivariate Markov chains also extends the applicability of our formal results beyond cellular automata to driven systems.  
\end{itemize}

In summary the original contributions of this chapter are:
\begin{itemize}
  \item Definition of SLI.
  \item Constructive proof of upper bound of SLI.
  \item Construction of negative SLI example. 
\item Definition of CLI.
  \item Definition of disintegration hierarchy and refinement-free disintegration hierarchy.
\item Proof of the disintegration theorem.
\item Proof of the SLI symmetry theorems.
\end{itemize}

\section{Partition lattice of Bayesian networks}
\label{sec:plattbn}
In this section we introduce the partition lattice of a Bayesian network which is the underlying structure for most of the developments in this chapter. All STPs occupy blocks of partitions in this lattice and can be further partitioned into blocks that are also blocks in the partition lattice of the Bayesian network. We also introduce the set of anti-STPs with respect to a partition which is a generalisation of the set of anti-patterns $\neg(x_A)$ (\cref{def:antistp}) of a STP $x_A$. This will be mainly used in proofs and is not important for the rest of this thesis.

\begin{mydef}[Partition lattice of a Bayesian network]
  Given a Bayesian network $\Xv$ and a subset $A \subseteq V$ we denote the \textit{partition lattice of $A$ by $\Latt(A)$}. Every partition $\pi \in \Latt(A)$ also naturally induces the following partitions:
  \begin{thmlist}
    \item partition $\pi(X_A)$ of the joint random variable $X_A$ by defining for every block $b \in \pi$ the corresponding block $X_b \in \pi(X_A)$
    \item partition $\pi(x_A)$ of every STP $x_A \in X_A$ by defining for each block $b \in \pi$ the corresponding block $x_b \in \pi(x_A)$.
  \end{thmlist}
  When it will be clear from context which instance of $\pi$ we are referring to we will denote these two partitions in the following also just by $\pi$.
 \end{mydef}   
Remark:
\begin{itemize}
\item Partition lattices were defined in \cref{def:lattice}
  \item $\Latt(V)$ is the partition lattice of all nodes in the Bayesian network.
  \item The partition lattice $\Latt(A)$ of the index set $A$ or equivalently the nodes in the Bayesian network indexed by $A$ must not be confused with the partition lattice $\Latt(\X_A)$ of the state space $\X_A$ of the joint random variable of those nodes. 
\end{itemize}


%
\begin{mydef}
\label{def:antistppart}
  Given a Bayesian network $\Xv$, an STP $x_A$ and a partition $\pi \in \Latt(A)$ of $x_A$ the set of \textit{anti-STPs of $x_A$ with respect to $\pi$} denoted by $\neg_\pi(x_A)$ is defined via 
  \begin{equation}
    \neg_\pi(x_A) := \{\bar{x}_A \in \X_A : \forall b \in \pi, \bar{x}_b \neq x_b\}.
  \end{equation} 
\end{mydef}
Remark:
\begin{itemize}
  \item Note that $\neg(x_A) = \neg_{\lzero}(x_A)$. Recall that $\lzero$ is the finest partition in a partition lattice and contains only blocks that are singletons (see \cref{def:lzero}).
\end{itemize}

\begin{thm}
  Given a Bayesian network $\Xv$, an STP $x_A$ and two partitions $\xi,\pi$ of $x_A$ with $\xi \lpre \pi$ we have:
  \begin{equation}
    \neg_\xi(x_A) \subseteq \neg_\pi(x_A).
  \end{equation} 
\end{thm}
\begin{proof}
\begin{align}
\neg_\xi(x_A) :&= \{\bar{x}_A \in \X_A: \forall b \in \xi, \bar{x}_b \neq x_b\} \\     
&=  \{\bar{x}_A  \in\X_A: \forall b \in \xi, \exists i \in b, \bar{x}_i \neq x_i\} \\
            &=  \{\bar{x}_A \in \X_A: \forall c \in \pi, \forall b \in \xi \text{ with } b \subseteq c, \exists i \in b, \bar{x}_i \neq x_i\} \\
            &\subseteq  \{\bar{x}_A \in \X_A: \forall c \in \pi, \exists i \in c, \bar{x}_i \neq x_i\} \\
            &= \{\bar{x}_A \in \X_A: \forall c \in \pi, \bar{x}_c \neq x_c\} \\
            &= \neg_\pi(x_A).
\end{align}
\end{proof}
Remark:
\begin{itemize}
  \item This theorem shows that the anti-STP $\neg_\xi(x_A)$ of a partition $\xi$ that refines a partition $\pi$ is the more restrictive anti-STP. It requires for at least one variable $\bar{x}_i$ of each block $b$ that it differs from $x_i$ whereas the anti-STPs with respect to the coarser partition $\pi \lsuc \xi$ only requires one such differing value across all the blocks of $\xi$ that make up each block of $\pi$.
\end{itemize}

\section{Specific local integration}
\label{sec:sli}
This section introduces the specific local integration (SLI). It also proves its upper bounds constructively and constructs an example of negative SLI. We state a definition of normalised SLI and algebraic properties of differences between specific local integrations. 

\subsection{General and deterministic case}

\begin{mydef}[Specific local integration (SLI)]
\label{def:mipi}
\label{def:sli}
 Given a Bayesian network $\{X\}_{i\in V}$ and a STP $x_O$ the \textit{specific local integration $\mi_\pi(x_O)$ of $x_O$ with respect to a partition $\pi$ of $O \subseteq V$} is defined as
\begin{equation}
\label{eq:mipi}
 \mi_\pi(x_O):= \log \frac{p_O(x_O)}{\prod_{b \in \pi} p_b(x_b)}.
\end{equation}  
In this thesis we use the convention that $\log \frac{0}{0} := 0$.
\end{mydef}

\begin{thm}[Deterministic specific local integration]
 Given a deterministic Bayesian network (\cref{def:detbn}) and uniform initial distribution the SLI of $x_O$ with respect to partition $\pi$ can be expressed in another way:
 Let $N(x_O)$ refer to the number of trajectories in which $x_O$ occurs. Then
 \begin{equation}
 \label{eq:detmipi}
  \mi_\pi(x_O)= (|\pi|-1) \log |\X_{V_0}| - \log \frac{\prod_{b \in \pi} N(x_b)}{N(x_O)}.
 \end{equation} 
\end{thm}
\begin{proof}
Follows by replacing the probabilities $p_O(x_O)$ and $p_b(x_b)$ in \cref{eq:mipi} with their deterministic expressions from \cref{thm:detprob}, i.e.\ $p_A(x_A)=N(X_A) / |\X_{V_0}|$. Then:
\begin{align}
\mi_\pi(x_O):&= \log \frac{p_O(x_O)}{\prod_{b \in \pi} p_b(x_b)} \\
&=\log \frac{\frac{N(x_O)}{|\X_{V_0}|}}{\prod_{b \in \pi} \frac{N(x_b)}{|\X_{V_0}|}} \\
&=\log \frac{\frac{N(x_O)}{|\X_{V_0}|}}{|\X_{V_0}|^{-|\pi|} \prod_{b \in \pi} N(x_b)} \\
&=\log \frac{|\X_{V_0}|^{|\pi|-1} N(x_O)}{\prod_{b \in \pi} N(x_b)} \\
&=(|\pi|-1) \log |\X_{V_0}| - \log \frac{\prod_{b \in \pi} N(x_b)}{N(x_O)}.
\end{align}

\end{proof}

\subsection{Upper bounds}
In this section we prove upper bounds of SLI. It is not essential for the rest of the thesis and is presented mainly for technical reference since bounds are important aspects of measures. However it is also useful to familiarise the reader with the measure of SLI since we prove the bounds constructively. We first show constructively that if we can choose the Bayesian network and the STP then SLI can be arbitrary large. This construction sets the probabilities of all blocks equal to the probability of the STP. In the subsequent theorem we show that this property in general gives the upper bound of SLI if the cardinality of the partition is fixed. This leads directly tho the upper bound if the cardinality of the partition is not fixed in the next theorem. Finally we give the expressions of the bounds in the deterministic case for convenient reference.

\begin{thm}[Construction of a STP with maximum SLI]
\label{thm:slistp}
Given a probability $q \in (0,1)$ and a positive natural number $n$ we can construct a Bayesian network $\Xv$ and an STP $x_O$ such that
\begin{equation}
 \mi_\pi(x_O) = -(n-1) \log q.
\end{equation} 
 \end{thm}
\begin{proof}
We construct a Bayesian network which realises two conditions on the probability $p_O$. From these two conditions (which can also be realised by other Bayesian networks) we can then derive the theorem.

Choose a Bayesian network $\Xv$ with binary random variables $\X_i =\{0,1\}$ for all $i \in V$. Choose all nodes in $O$ dependent only on node $j \in O$, the dependence of the nodes in $V \bs O$ is arbitrary: 
 \begin{itemize}
 \item for all $i \in O \subset V$ let $\pa(i) \cap (V \bs O) = \emptyset$, i.e.\ nodes in $O$ have no parents in the complement of $O$,
 \item for a specific $j \in O$ and all other $i \in O \bs \{j\}$ let $\pa(i) = \{j\}$, i.e.\ all nodes in $O$ apart from $j$  have $j \in O$ as a parent,
 \item for all $i \in O \bs \{j\}$ let $p_i(\bar{x}_i | b\bar{x}_j) = \delta_{\bar{x}_j}(\bar{x}_i)$, i.e.\ the state of all nodes in $O$ is always the same as the state of node $j$, 
 \item also choose $p_j(x_j) = q$ and $\sum_{\bar{x}_j \neq x_j} p_j(x_j)=1-q$.
\end{itemize}
Then it is straightforward to see that:
\begin{enumerate}
  \item $p_O(x_O)=q$,
  \item $\sum_{\bar{x}_O \in \neg(x_O)} p_O(\bar{x}_O)=1-q$.
 \end{enumerate}
Note that there are many Bayesian networks that realise the latter two conditions for some $x_O$. These latter two conditions are the only requirements for the following calculation. 

Next note that the two conditions imply that $p_O(\bar{x}_O)=0$ if neither $\bar{x}_O =x_O$ nor $\bar{x}_O \in \neg(x_O)$. Then for every partition $\pi$ of $O$ with $|\pi|=n$ and $n>1$ we have 
 \begin{align}
  \mi_\pi(x_O)&= \log \frac{p_O(x_O)}{\prod_{b \in \pi} p_b(x_b)} \\
  &= \log \frac{p_O(x_O)}{\prod_{b \in \pi} \sum_{\bar{x}_{O\bs b}} p_O(x_b,\bar{x}_{O\bs b})} \\
  &= \log \frac{p_O(x_O)}{\prod_{b \in \pi} \left( p_O(x_O) + \sum_{\bar{x}_{O\bs b}\neq x_{O\bs b}} p_O(x_b,\bar{x}_{O\bs b})\right)} \\
  &= \log \frac{p_O(x_O)}{\prod_{b \in \pi} p_O(x_O) } \\
  &= \log \frac{p_O(x_O)}{ p_O(x_O)^n} \\
  &= -(n-1) \log q. 
 \end{align}
\end{proof}
Remark:
\begin{itemize}
\item We will use this construction to reveal the general tight upper bound of $\mi_\pi(x_O)$.
\item 
The construction used here ensures that the probability $p_b(x_b)$ of each block $b \in \pi$ is equal to the probability of the STP $p_O(x_O)=q$. In other words, the parts of $x_O$ that are indicated by $\pi$ all occur if and only if the whole STP $x_O$ occurs. Note that in general $x_b$ always occurs if $x_O$ occurs but not vice versa. 
\end{itemize}

\begin{thm}[Upper bound of SLI]
\label{thm:slibound}
For any Bayesian network $\{X\}_{i\in V}$ and STP $x_O$ 
\begin{thmlist}
 \item \label{thm:slibound1} The tight upper bound of the SLI with respect to partition $\pi$ is 
 \begin{equation}
  \label{eq:slibound}
   \mi_\pi(x_O)\leq -(|\pi|-1) \log p_O(x_O).
 \end{equation} 
 \item \label{thm:slibound2} The upper bound is achieved if and only if for all $b \in \pi$ we have 
 \begin{equation}
  p_b(x_b)=p_O(x_O).
 \end{equation} 
 \item \label{thm:slibound3} The upper bound is achieved if and only if for all $b \in \pi$ we have that $x_O$ occurs if and only if $x_b$ occurs.
 \end{thmlist}
 \end{thm}
\begin{proof}

\begin{description}
 \item[ad \ref{thm:slibound1}] By \cref{def:sli} we have
\begin{equation}
\label{eq:sliboundproof1}
 \mi_\pi(x_O)= \log \frac{p_O(x_O)}{\prod_{b \in \pi} p_b(x_b)}.
\end{equation} 
Now note that for any $x_O$ and $b\subseteq O$
\begin{align}
 p_b(x_b) &= \sum_{\bar{x}_{O\bs b}} p_O(x_b,\bar{x}_{O\bs b})\\
          &= p_O(x_O) + \sum_{\bar{x}_{O\bs b} \neq x_{O\bs b}} p_O(x_b,\bar{x}_{O\bs b}) \label{eq:blockprob}\\
          &\geq p_O(x_O).\label{eq:blockprobineq}
\end{align}
Plugging this into \cref{eq:sliboundproof1} for every $p_b(x_b)$ we get
\begin{align}
 \mi_\pi(x_O)&= \log \frac{p_O(x_O)}{\prod_{b \in \pi} p_b(x_b)}\\
 &\leq \log \frac{p_O(x_O)}{p_O(x_O)^{|\pi|}}  \\
 &= -(|\pi|-1) \log p_O(x_O). 
\end{align}
This shows that $ -(|\pi|-1) \log p_O(x_O)$ is indeed an upper bound. To show that it is tight we have to show that for a given $p_O(x_O)$ and $|\pi|$ there are Bayesian networks with STPs $x_O$ such that this upper bound is achieved. The construction of such a Bayesian network and an STP $x_O$ was presented in \cref{thm:slistp}.

\item[ad \ref{thm:slibound2})] If for all $b \in \pi$ we have $p_b(x_b)=p_O(x_O)$ then clearly $\mi_\pi(x_O)=-(|\pi|-1) \log p_O(x_O)$ and the least upper bound is achieved. If on the other hand $\mi_\pi(x_O)=-(|\pi|-1) \log p_O(x_O)$ then 
\begin{alignat}{2}
 && \log \frac{p_O(x_O)}{\prod_{b \in \pi} p_b(x_b)} &=  -(|\pi|-1) \log p_O(x_O)\\
\Leftrightarrow\;\;\;\;\;\;&& \log \frac{p_O(x_O)}{\prod_{b \in \pi} p_b(x_b)} &=\log \frac{p_O(x_O)}{p_O(x_O)^{|\pi|}} \\
\Leftrightarrow\;\;\;\;\;\;&& \prod_{b \in \pi} p_b(x_b) &= p_O(x_O)^{|\pi|},
 \end{alignat}
and because $p_b(x_b) \geq p_O(x_O)$ (\cref{eq:blockprobineq}) any deviation of any of the $p_b(x_b)$ from $p_O(x_O)$ leads to $\prod_{b \in \pi} p_b(x_b) > p_O(x_O)^{|\pi|}$ such that for all $b \in \pi$ we must have $p_b(x_b)=p_O(x_O)$.
\item[ad \ref{thm:slibound3}] By definition for any $b \in \pi$ we have $b \subseteq O$ such that $x_b$ always occurs if $x_O$ occurs. Now assume $x_b$ occurs and $x_O$ does not occur. In that case there is a positive probability for an STP $(x_b,\bar{x}_{O \bs b})$ with $\bar{x}_{O \bs b} \neq x_{O \bs b}$ i.e.\ $p_O(x_b,\bar{x}_{O \bs b}) >0$. Recalling \cref{eq:blockprob} we then see that
\begin{align}
 p_b(x_b) &= p_O(x_O) + \sum_{\bar{x}_{O\bs b} \neq x_{O\bs b}} p_O(x_b,\bar{x}_{O\bs b}) \\
          &> p_O(x_O).
\end{align}
which contradicts the fact that $p_b(x_b)=p_O(x_O)$ so $x_b$ cannot occur without $x_O$ occurring as well.
\end{description}
\end{proof}

Remarks:
\begin{itemize}
\item Note that this is the least upper bound for Bayesian networks in general. For a specific Bayesian network there might be no STP that achieves this bound.
 \item So the least upper bound of SLI $\mi_\pi(x_O)$ is the self-information $- \log p(x_O)$ of the STP $x_O$ multiplied by one less than the cardinality $|\pi|$ of the partition.
 \item In other words, the maximally possible SLI increases with the improbability of the STP and the number of parts that it is split into. 
 \item For an STP $x_O$ that achieves the least upper bound of SLI, the occurrence of any part $x_b$ indicated by the partition $\pi$ of $O$ implies the occurrence of the entire STP. 
 \item Using this least upper bound it is easy to derive a least upper bound for the SLI of an STP $x_O$ across all partitions $|\pi|$. We just have to note that $|\pi| \leq |O|$. This leads directly to the next statement.
 \item Further down we will use the least upper bound for specific partitions in order to normalise the SLI.
\end{itemize}

\begin{thm}
 For any Bayesian network $\Xv$ and STP $x_O$ the least upper bound of the SLI with respect to arbitrary partitions is 
 
\begin{equation}
 \max_{\pi} \mi_{\pi}(x_0) \leq -(|O|-1) \log p_O(x_O).
\end{equation} \end{thm}
\begin{proof}
 Follows from \cref{thm:slibound} and the fact that for an arbitrary partition $\pi$ of $O$ we have $|\pi| \leq |O|$.
\end{proof}
%

\begin{thm}[Deterministic least upper bounds]
\label{thm:detsup}
 For any deterministic Bayesian network (\cref{def:detbn}) $\Xv$ with uniform initial distribution $p_{V_0}$ we find the following bounds.
\begin{thmlist}
 \item \label{thm:detsup1}The least upper bound of the SLI with respect to partition $\pi$ for a given STP $x_O$ is 
 \begin{equation}
  \mi_{\pi}(x_O) \leq -(|\pi|-1) \log \frac{N(x_O)}{|\X_{V_0}|}
 \end{equation} 
 \item \label{thm:detsup2}The least upper bound of the SLI for an STP $x_O$ across all partitions is 
 \begin{equation}
  \max_{\pi} \mi_{\pi}(x_0) \leq -(|O|-1) \log \frac{N(x_O)}{|\X_{V_0}|}
 \end{equation} 
 \item \label{thm:detsup3}The least upper bound of the SLI with respect to partitions $\pi$ of cardinality $|\pi|$ across all STP $x_O$ is
 \begin{equation}
  \max_{x_O} \mi_{\pi}(x_O) \leq (|\pi|-1) \log |\X_{V_0}|
 \end{equation} 
 \item \label{thm:detsup4}The least upper bound of the SLI across all partitions and all STP is
 \begin{equation}
  \max_{\pi} \max_{x_O} \mi_{\pi}(x_O) \leq (|V|-1) \log |\X_{V_0}|
 \end{equation} 
 
 \end{thmlist}

\end{thm}
\begin{proof}
 \begin{description}
  \item[ad \ref{thm:detsup1}] Follows directly from \cref{thm:slibound} by replacing $p_O(x_O)$ with $\frac{N(x_O)}{|\X_{V_O}|}$ as shown in \cref{eq:detprob}.
  \item[ad \ref{thm:detsup2}] Follows from \cref{thm:detsup1} and $|\pi| \leq |O|$ for all partitions $\pi$ of $|O|$.
  \item[ad \ref{thm:detsup3}] Follows from \cref{thm:detsup1} and $N(x_O) \geq 1$ for all $x_O$.
  \item[ad \ref{thm:detsup4}] Follows from \cref{thm:detsup3} and $|\pi| \leq |V|$ for all partitions $\pi$ of all $O \subseteq V$.
 \end{description}
\end{proof}

Remarks:
\begin{itemize}
 \item Again these are tight upper bounds among all deterministic Bayesian networks with uniform initial distribution. Particular instances of such Bayesian networks may not contain any STP which achieve these bounds. 
\item  Note that the first term in \cref{eq:detmipi} corresponds to the tight upper bound in \cref{thm:detsup3}. So $\mi_{\pi}(x_O)$ is just this upper bound minus the second term which is always positive as $\prod_b N(x_b) \geq N(x_O)$.
 
\end{itemize}

\subsection{Negative SLI}
This section shows that SLI of an STP $x_O$ with respect to partition $\pi$ can be negative \textit{independent} of the the probability of $x_O$ (as long as it is not $1$) and the cardinality of the partition (as long as that is not $1$). This is not important for the rest of the thesis but is of technical interest in its own right. It also shows how to get negative SLI at all which may not be obvious. 

\begin{thm}
For any given probability $q<1$ and cardinality $|\pi|>1$ of a partition $\pi$ there exists an STP $x_O$ in a Bayesian network $\Xv$ such that $q = p_O(x_O)$ and 
\begin{equation}
  \mi_\pi(x_O)<0.
\end{equation} 
\end{thm}
\begin{proof}
  We construct the probability distribution $p_O:\X_O \rightarrow [0,1]$ and ignore the behaviour of the Bayesian network $\Xv$ outside of $O \subseteq V$. In any case $\{X_i\}_{i \in O}$ is also by itself a Bayesian network. We define (see remarks below for some intuitions behind these definitions):
  \begin{thmlist}
 \item \label{thm:negsli1} for all $i \in O$ let $|\X_i| = n$
 \item \label{thm:negsli2}for every block $b \in \pi$ let $|b| = \frac{|O|}{|\pi|}$,
 
\item \label{thm:singleblockfailures} for $\bar{x}_O \in \X_O$ let:
\begin{equation}
        p_O(\bar{x}_O):=\begin{cases}
                          q &\text{ if } \bar{x}_O=x_O,\\
                          \frac{1-q-d}{\sum_{b \in \pi} |\neg(x_b)|} & \text{ if } \exists c \in \pi \text{ s.t. } \bar{x}_{O\bs c}=x_{O \bs c} \wedge \bar{x}_c \neq x_c, \\
                          \frac{d}{|\neg(x_O)|} & \text{ if } \bar{x}_O \in \neg(x_O), \\
                          0 & \text{ else}.
                        \end{cases}
      \end{equation} 

  \end{thmlist}
  Then we can calculate the SLI. First note that according to \ref{thm:negsli1} and \ref{thm:negsli2} we have $|\X_b| = |\X_c|$ for all $b,c \in \pi$ and therefore also $|\neg(x_b)|=|\neg(x_c)|$ for all $b,c \in \pi$. So let $m:=|\neg(x_b)|$. Then note that according to \ref{thm:singleblockfailures} for all $b \in \pi$ 
  
\begin{align}
  \sum_{\bar{x}_{O \bs b} \neq x_{O \bs b}} p_O(x_b,\bar{x}_{O \bs b}) &= \sum_{c \in \pi \bs b} \sum_{\bar{x}_c \neq x_c} p_O(x_b,x_{O\bs(b \cup c)},\bar{x}_c) \\
  &= \sum_{c \in \pi \bs b} \sum_{\bar{x}_c \neq x_c} \frac{1-q-d}{\sum_{b \in \pi} |\neg(x_b)|}  \\
  &= \sum_{c \in \pi \bs b} \sum_{\bar{x}_c \neq x_c} \frac{1-q-d}{m |\pi|}  \\
  &= \sum_{c \in \pi \bs b} \frac{1-q-d}{m |\pi|} |\neg(x_c)|   \\
 &= \frac{|\pi|-1}{|\pi|} (1-q-d) \label{eq:minsliblockprobcontribution}
 \end{align}
Plug this into the SLI definition:
  \begin{align}
    \mi_\pi(x_O)&= \log \frac{p_O(x_O)}{\prod_{b \in \pi} p_b(x_b)} \\
    &= \log \frac{q}{\prod_{b \in \pi} q + \sum_{\bar{x}_{O \bs b} \neq x_{O \bs b}} p_O(x_b,\bar{x}_{O \bs b})} \\
    &= \log \frac{q}{\prod_{b \in \pi} q + \frac{|\pi|-1}{|\pi|} (1-q-d)} \\
    &= \log \frac{q}{\left(1- \frac{1-q}{|\pi|}\right)^{|\pi|}}. \label{eq:minsliachieved}
  \end{align}
 If we now set $d=0$ then we can use Bernoulli's inequality\footnote{We thank \citet{vaneitzen_mathstack_2016} for pointing this out. An example reference for Bernoulli's inequality is \citet{bullen_handbook_2003}.} to prove that this is negative for $0<q<1$ and $|\pi|\geq 2$. Bernoulli's inequality is
 \begin{equation}
   (1+x)^n \geq 1 + n x
 \end{equation}
 for $x \geq -1$ and $n$ a natural number. Replacing $x$ by $-(1-q)/|\pi|$ we see that
 \begin{align}
   \left(1- \frac{1-q}{|\pi|}\right)^{|\pi|} > q
 \end{align}
such that the argument of the logarithm is smaller than one which gives us negative SLI.

\end{proof}
Remarks:
\begin{itemize}
  \item The construction used to proof this theorem with the achieved value in \cref{eq:minsliachieved} is also our best candidate for a tight lower bound of SLI for a given $p_O(x_O)$ and $|\pi|$. However, we have not been able to prove this yet.
  \item The construction equidistributes the probability $1-q$ (left to be distributed after $q$ is chosen) to the STPs $\bar{x}_O$ that are \textit{almost} the same as the STP $x_O$. They are almost the same in a precise sense because they only differ in only one of the blocks of $\pi$ they differ by as little as can possibly be resolved/revealed by the partition $\pi$. 
  \item In order to achieve the negative SLI of \cref{eq:minsliachieved} the requirement is only that \cref{eq:minsliblockprobcontribution} is satisfied. Our construction shows one way how this can be achieved.
  \item For a pattern and partition such that $|O|/|\pi|$ is not a natural number, the same bound might still be achieved however a little extra effort has to go into the construction \ref{thm:singleblockfailures} such that \cref{eq:minsliblockprobcontribution} still holds. This is not necessary for our purpose here as we only want to show the existence of patterns that with negative SLI.  
  \item An interpretation of the construction is that STPs which either occur as a whole or (with uniform probability) missing exactly one part always have negative SLI. 
\end{itemize}

\subsection{Normalised specific local integration}
Here we present a way to employ the least upper bound to define a normalised version of SLI. This notion is not important for the rest of this thesis.

\begin{mydef}[Normalised specific local integration]
\label{def:nmi}
 The normalised specific local integration is just the specific normalised integration $\mi_\pi(x_O)$ divided by the least upper bound for the STP $x_O$ and the partition $\pi$:
 \begin{equation}
  \nmi_\pi(x_O):=\frac{\mi_\pi(x_O)}{-(|\pi|-1) \log p_O(x_O)}.
 \end{equation} 
 The value of $\nmi_\pi(x_O)$ is in the the interval $(-\infty, 1]$.
\end{mydef}

\begin{thm}
 For any Bayesian network, any STP $x_O$ and any partition $\pi$
 \begin{equation}
  \nmi_\pi(x_O) \leq 1.
 \end{equation} 
\end{thm}
\begin{proof}
 Follows from \cref{thm:slibound} and the definition of $\nmi_\pi(x_O)$.
\end{proof}

Remarks:
\begin{itemize}
\item The normalised SLI gives us a measure of integration which is independent of the cardinality of the partition. This means we can compare the specific local integrations of a STP across partitions $\pi$ of different cardinalities. 
\item At the same time the normalised SLI also compensates for differences in the self-information $-\log p_O(x_O)$ of the STP $x_O$. This self-information can be interpreted as a measure of the ``size'' of the STP. For a given STP $x_O$ this size of course does not change across the partitions. However, dividing by it allows to also compare the degree of SLI across STPs of differing sizes. 

\end{itemize}

\subsection{Difference of SLI with respect to different partitions}

This section present some algebraic properties of SLI. An alternative expression for SLI, the difference between the SLI with respect to different partitions of the same STP,  and between the SLI with respect to different partitions of different STP. It is presented here for reference and used only in some of the proofs that follow. 

\begin{mydef}
Given a partition $\pi$ of a set $V$ and a subset $A \subseteq V$ we define the \textit{restricted partition} $\pi\rvert_A$ of $\pi$ to $A$ via:
\begin{equation}
 \pi\rvert_A:= \{b \cap A: b\in \pi\}.
\end{equation}
Conversely, if $\xi$ is a partition of $A \subseteq V$ and $\pi$ a partition of $V$ and $\pi\rvert_A=\xi$ we call $\pi$ an \textit{extension of $\xi$ to $V$}.
\end{mydef}

\begin{thm}
Given any particular total order $(b_1,...,b_n)$ of the blocks of a partition $\pi$ of $O \subseteq V$ with $|\pi|=n$ we can always write the SLI as a sum over the blocks in the partition:
\begin{equation}
 \mi_\pi(x_O)= \sum_{i=1}^{n-1} \log \frac{p(x_{b_{i+1}}|x_{\bigcup_{j=1}^i b_j})}{p_{b_{i+1}}(x_{b_{i+1}})}
\end{equation}  
\end{thm}
\begin{proof}
 Follows directly from the chain rule of probability and the properties of the logarithm.
\end{proof}

\begin{mydef}
Given two partitions $\pi,\xi$ of $O \subseteq V$ and an STP $x_O$ then we define the difference $\dmi^\pi_\xi(x_O)$ of the respective specific local integrations via:
\begin{equation}
 \dmi^\pi_\xi(x_O):=\mi_\pi(x_O) - \mi_\xi(x_O).
\end{equation} 
\end{mydef}

\begin{thm}
\label{thm:dmi}
 Given three partitions $\pi,\xi,\rho$ of $O \subseteq V$ and an STP $x_O$. Then:
\begin{thmlist}
 \item \begin{equation}
 \dmi^\pi_\xi(x_O) = \log \frac{\prod_{a \in \xi} p_a(x_a)}{\prod_{b \in \pi} p_b(x_b)}.
\end{equation}
\item \label{thm:dmiselfinfo} \begin{equation}
 \dmi^\pi_\xi(x_O) = \sum_{b \in \pi} \log \frac{1}{p_b(x_b)} - \sum_{a \in \xi} \log \frac{1}{p_a(x_a)}.
\end{equation}
\item \label{thm:dmirefine} If $\pi \lpreeq \xi$ then:
\begin{equation}
 \dmi^\pi_\xi(x_O) = \sum_{a \in \xi} \mi_{\pi\rvert_a}(x_a).
\end{equation} 
\item \label{thm:dmimi} If $\rho \lpreeq \pi$ and $\rho \lpreeq \xi$ i.e.\ $\rho$ is a lower bound of $\pi$ and $\xi$ then
\begin{align}
 \dmi^\pi_\xi(x_O) &=  \sum_{a \in \xi} \mi_{\rho\rvert_a}(x_a) - \sum_{b \in \pi} \mi_{\rho\rvert_b}(x_b) \\
 &=\dmi^\rho_\xi(x_O) -\dmi^\rho_\pi(x_O).
\end{align} 
\end{thmlist}
\end{thm}
\begin{proof}
 Follows straightforwardly from the definitions and properties of the logarithm.
\end{proof}

Remarks:
\begin{itemize}
 \item \cref{thm:dmiselfinfo} says that the difference between the specific local integrations $\dmi^\pi_\xi(x_O)$ of two arbitrary partitions $\pi,\xi$ is equal to the difference of the according sums over the self-informations of the blocks in each partition. 
 \item \cref{thm:dmirefine} says that the difference between the specific local integrations $\dmi^\pi_\xi(x_O)$ of a refinement $\pi$ of a partition $\xi$ and the partition $\xi$ itself is the sum over the specific local integrations $\mi_{\pi\rvert_a}(x_a)$ of each of the blocks $a \in \xi$ of the original partition $\xi$ with respect to their refinement $\pi\rvert_a$ due to $\pi$.
 \item \cref{thm:dmimi} says that the difference between the specific local integrations $\dmi^\pi_\xi(x_O)$ of two arbitrary partitions $\pi,\xi$ is also equal to the negative difference between the sums over the SLI of each of their blocks with respect to a partition $\rho$ that refines both $\pi$ and $\xi$. 
\end{itemize}

\begin{thm} \label{thm:misubset}
 Given $S \subseteq O \subseteq V$ as well as a partition $\pi$ of $O$ and a partition $\xi$ of $S$ we have:
 \begin{thmlist}
\item \label{thm:misubsetrelation}
 \begin{equation}
 \mi_\pi(x_O)=\log \frac{p(x_{O\bs S}|x_S)}{\prod_{b \in \pi} p(x_{b\bs S}|x_{b \cap S})} + \mi_\xi(x_S) + \dmi^{\pi\rvert_S}_\xi(x_S).
\end{equation} 
\item \label{thm:miextension}
If we set $\xi = \pi\rvert_S$ then:
\begin{equation}
 \mi_\pi(x_O)=\log \frac{p(x_{O\bs S}|x_S)}{\prod_{b \in \pi} p(x_{b\bs S}|x_{b \cap S})} + \mi_{\pi\rvert_S}(x_S).
\end{equation} 
 \end{thmlist}
\end{thm}

\begin{proof}
 We will use here that $p(x_\emptyset) = p(x_\emptyset |x_O) = p(\emptyset)=1$ and $p(x_O|x_\emptyset)=p(x_O)$ for any set $O \in V$ and STP $x_O$. This is in accordance with probability theory. Then for \ref{thm:misubsetrelation}:
 \begin{align}
 \mi_\pi(x_O) &= \log \frac{p_O(x_O)}{\prod_{b \in \pi} p_b(x_b)} \\
 &= \log \frac{p(x_{O\bs S}|x_S) p_S(x_S)}  {\prod_{b \in \pi} p(x_{b \bs S}|x_{b \cap S}) p_{b \cap S}(x_{b\cap S})} \\
 &= \log \frac{p(x_{O\bs S}|x_S)}  {\prod_{b \in \pi} p(x_{b \bs S}|x_{b \cap S}) } + \log \frac{p_S(x_S)}{\prod_{b \in \pi} p_{b \cap S}(x_{b\cap S})}\\
&= \log \frac{p(x_{O\bs S}|x_S)}  {\prod_{b \in \pi} p(x_{b \bs S}|x_{b \cap S}) } + \log \frac{p_S(x_S)}{\prod_{c \in \pi\rvert_S} p_c(x_c)}\\
&= \log \frac{p(x_{O\bs S}|x_S)}  {\prod_{b \in \pi} p(x_{b \bs S}|x_{b \cap S}) } + \mi_\xi(x_S) + \dmi^{\pi\rvert_S}_\xi(x_S).
\end{align}
Then \ref{thm:miextension} follows by setting $\xi = \pi\rvert_S$.
\end{proof}

Remarks:
\begin{itemize}
 \item \cref{thm:dmi} follows from \cref{thm:misubset} by setting $S = O$.
 \item \cref{thm:miextension} says that an STP $x_S$ with vanishing or negative $\mi_{\xi}(x_S)$ can be part of an STP $x_O$ with $S \subset O$ such that an extension of $\xi$ to partition $\pi$ of $O$ can have positive $\mi_\pi(x_O)$. 
\end{itemize}

\section{Complete local integration}
\label{sec:cli}
Complete local integration (CLI) is an important concept in this thesis as positive CLI will form the criterion distinguishing arbitrary STPs from \textit{entities} in multivariate Markov chains (see \cref{sec:entinmvmc}). 

\begin{mydef}[(Complete) local integration]
\label{def:ci}
 Given a Bayesian network $\Xv$ and an STP $x_O$ of this network the \textit{complete local integration $\ci(x_O)$ of $x_O$} is the minimum SLI over the non-unit partitions $\pi \in \Latt(O) \bs \lunit_O$:
 \begin{equation}
  \label{eq:ci}
  \ci(x_O):= \min_{\pi \in \Latt(O) \bs \lunit_O} \mi_\pi(x_O).
 \end{equation}
 We call an STP $x_O$ \textit{completely locally integrated} if $\ci(x_O)>0$. 
\end{mydef}

Remarks:
\begin{itemize}
\item The reason for excluding the unit partition $\lunit_O$ of $\Latt(O)$ (where $\lunit_O =\{O\}$ see \cref{def:lunit}) is that with respect to it every STP has $\mi_{\lunit_O}(x_O)=0$.  
 \item The CLI is the SLI of $x_O$ with respect to the partition with respect to which $x_O$ is least integrated. Maybe more clearly, it is the SLI of $x_O$ with respect to the partition that disintegrates $x_O$ the most. The same idea is also employed by \citet{tononi_measuring_2003,tononi_information_2004,balduzzi_integrated_2008} in a non-local setting. It is known as the \textit{weakest link approach} \citep{ay_information_2015} to dealing with multiple levels of integration. We note here that this is not the only approach that is being discussed. Another approach is to look at weighted averages of all integrations. For a further discussion of this point in the case of non-local integration (or complexity which, on a global level, may well be the same thing) see \citet{ay_information_2015} and references therein. A full analysis of which approach is best suited for the local integration measure presented here is beyond the scope of this thesis. 
\end{itemize}

\section{Disintegration}
\label{sec:disint}
In this section we define the disintegration hierarchy and its refinement-free version. We then prove the disintegration theorem which is the main formal result of this thesis. It exposes a connection between partitions minimising the SLI of a trajectory and the CLI of the blocks of such partitions. More precisely for a given trajectory the blocks of the finest partitions among those leading to a particular value of SLI consist only of completely locally integrated blocks. Conversely \textit{each} completely locally integrated STP is a block in such a finest partition among those leading to a particular value of SLI. The theorem therefore reveals the special role of STPs with positive CLI with respect to an entire trajectory of the system. For our purposes this theorem allows further interpretations of the measure of CLI which will be discussed in \cref{sec:identity}. We believe however that it will also be of general interest in the study of complex systems fore example due to the relation of SLI and CLI to measures of complexity like multi-information and local information dynamics (\cref{sec:formallyrelated}).

\begin{mydef}[Disintegration hierarchy]
\label{def:dishier}
  Given a Bayesian network $\Xv$ and a trajectory $x_V \in \X_V$, the \textit{disintegration hierarchy of $x_V$} is the set $\dis(x_V)=\{\dis_1,\dis_2,\dis_3,...\}$ of sets of partitions of $x_V$ with:
  \begin{thmlist}
    \item 
    \begin{equation}
      \dis_1(x_V) := \argmin_{\pi \in \Latt(V)} \mi_\pi(x_V)
    \end{equation}
\item and for $i > 1$:
  \begin{equation}
    \dis_i(x_V) := \argmin_{\pi \in \Latt(V) \bs \dis_{\prec i}(x_V)} \mi_\pi(x_V).
  \end{equation} 
  \end{thmlist}
where $\dis_{\prec i}(x_V) := \bigcup_{j<i} \dis_j(x_V)$. We call $\dis_i(x_V)$ the $i$-th \textit{disintegration level}.
\end{mydef}
Remark:
\begin{itemize}
  \item Note that $\argmin$ returns all partitions that achieve the minimum SLI. 
  \item Since the Bayesian networks we use are finite, the partition lattice $\Latt(V)$ is finite, the set of attained SLI values is finite, and the number $|\dis|$ of disintegration levels is finite.
  \item In most cases the Bayesian network contains some symmetries among their mechanisms which cause multiple partitions to attain the same SLI value.
  \item For each trajectory $x_V$ the disintegration hierarchy $\dis$ then partitions the elements of $\Latt(V)$ into subsets $\dis_i(x_V)$ of equal SLI. The levels of the hierarchy have increasing SLI.  
\end{itemize}

\begin{mydef}
  Let $\Latt(V)$ be the lattice of partitions of set $V$ and let $\mathfrak{E}$ be a subset of $\Latt(V)$. Then for every element $\pi \in \Latt(V)$ we can define the set 
  \begin{equation}
\mathfrak{E}_{\lpre \pi} := \{\xi \in \mathfrak{E} : \xi \lpre \pi\}.
\end{equation} 
That is $\mathfrak{E}_{\lpre \pi}$ is the set of partitions in $\mathfrak{E}$ that are refinements of $\pi$. 
\end{mydef}

\begin{mydef}[Refinement-free disintegration hierarchy]
  \label{def:rfreedishier}
  Given a Bayesian network $\Xv$, a trajectory $x_V \in \X_V$, and its disintegration hierarchy $\dis(x_V)$ the \textit{refinement-free disintegration hierarchy of $x_V$} is the set $\dis^\lmin(x_V)=\{\dis^\lmin_1,\dis^\lmin_2,\dis^\lmin_3,...\}$ of sets of partitions of $x_V$ with:
  \begin{thmlist}
    \item 
    \begin{equation}
    \label{eq:dislmin1}
      \dis^\lmin_1(x_V) := \{\pi \in \dis_1(x_V):\dis_1(x_V)_{\lpre \pi}=\emptyset\},
    \end{equation}
\item and for $i > 1$:
  \begin{equation}
    \dis^\lmin_i(x_V) := \{\pi \in \dis_i(x_V):\dis_{\prec i}(x_V)_{\lpre \pi}=\emptyset\}
  \end{equation} 
  \end{thmlist}
\end{mydef}
Remark:
\begin{itemize}
  \item Each level $\dis^\lmin_i(x_V)$ in the refinement-free disintegration hierarchy $\dis^\lmin(x_V)$ consists only of those partitions that neither have refinements at their own nor at any of the preceding levels. So each partition that occurs in the refinement-free disintegration hierarchy at the $i$-th level is a finest partition that achieves such a low level of SLI or such a high level of disintegration. 
  \item As we will see below, the blocks of the partitions in the refinement-free disintegration hierarchy are the main reason for defining the refinement-free disintegration hierarchy. 
\end{itemize}

%

\begin{thm}[Disintegration theorem]
\label{thm:disintegration}
Let $\Xv$ be a Bayesian network, $x_V \in \X_V$ one of its trajectories, and $\dis^\lmin(x_V)$ the associated refinement-free disintegration hierarchy. 

\begin{thmlist}
\item \label{thm:disintforward} Then for every $\dis^\lmin_i(x_V) \in \dis^\lmin(x_V)$ we find for every $b \in \pi$ with $\pi \in \dis^\lmin_i(x_V)$ that there are only the following possibilities:
\begin{enumerate}
  \item $b$ is a singleton, i.e.\ $b = \{i\}$ for some $i \in V$, or
  \item $x_b$ is completely locally integrated, i.e.\ $\ci(x_b) > 0$.
\end{enumerate}
\item\label{thm:disintbackward} Conversely, for any completely locally integrated STP $x_A$, there is a partition $\pi^A \in \Latt(V)$ and a level $\dis^\lmin_{i^A}(x_V) \in \dis^\lmin(x_V)$ such that $A \in \pi^A$ and $\pi^A \in \dis^\lmin_{i^A}(x_V)$.
\end{thmlist}

%
%
%
\end{thm}
\begin{proof}
\begin{description}
  \item[ad \ref{thm:disintforward}] 
We prove the theorem by contradiction. For this assume that there is block $b$ in a partition $\pi \in \dis^\lmin_i(x_V)$ which is neither a singleton nor completely integrated. 
  Let $\pi \in \dis^\lmin_i(x_V)$ and $b \in \pi$. Assume $b$ is not a singleton i.e.\ there exist $i\neq j \in V$ such that $i \in b$ and $j \in b$. 
  Also assume that $b$ is not completely integrated i.e.\ there exists a partition $\xi$ of $b$ with $\xi \neq \lunit_b$ such that $\mi_\xi(x_b)\leq 0$. Note that a singleton cannot be completely locally integrated as it does not allow for a non-unit partition. So together the two assumptions imply $p_b(x_b) \leq \prod_{d \in \xi} p_d(x_d)$ with $|\xi|>1$. But then 
\begin{align}
  \mi_{\pi}(x_V)&= \log \frac{p_V(x_V)}{p_b(x_b) \prod_{c \in \pi\bs b} p_c(x_c)} \\
  &\geq \log \frac{p_V(x_V)}{\prod_{d \in \xi} p_d(x_d) \prod_{c \in \pi\bs b} p_c(x_c)}
\end{align}
We treat the cases of ``$>$'' and ``$=$'' separately. First, let
\begin{equation}
   \mi_{\pi}(x_V) = \log \frac{p_V(x_V)}{\prod_{d \in \xi} p_d(x_d) \prod_{c \in \pi\bs b} p_c(x_c)}.
\end{equation} 
Then we can define $\rho:=(\pi \bs b) \cup \xi$ such that 
\begin{thmlist}
  \item $\mi_\rho(x_V) = \mi_\pi(x_V)$ which implies that $\rho \in \dis_i(x_V)$ because $\pi \in \dis_i(x_V)$, and
  \item $\rho \lpre \pi$ which contradicts $\pi \in \dis^\lmin_i(x_V)$.
\end{thmlist}
Second, let
\begin{equation}
   \mi_{\pi}(x_V) > \log \frac{p_V(x_V)}{\prod_{d \in \xi} p_d(x_d) \prod_{c \in \pi\bs b} p_c(x_c)}.
\end{equation}
Then we can define $\rho:=(\pi \bs b) \cup \xi$ such that 
\begin{equation}
  \mi_{\rho}(x_V) < \mi_{\pi}(x_V),
\end{equation} 
which contradicts $\mi_{\pi}(x_V) \in \dis^\lmin_i(x_V)$.
\item[ad \ref{thm:disintbackward}] Let $\pi^A:=\{A\} \cup \{\{j\}\}_{j \in V \bs A}$. Since $\pi^A$ is a partition of $V$ it is an element of some disintegration level $\dis_{i^A}$. Then partition $\pi^A$ is also an element of the refinement free disintegration level $\dis^\lmin_{i^A}(x_V)$ as we will see in the following. This is because any refinements must (by construction of $\pi^A$ break up $A$ into further blocks which means that the local specific integration of all such partitions is higher. Then they must be at lower disintegration level $\dis_k(x_V)$ with $k \geq i^A$.  Therefore $\pi^A$ has no refinement at its own or a higher disintegration level. More formally, let $\xi \in \Latt(V),\xi \neq \pi^A$ and $\xi \lpre \pi^A$ since $\pi^A$ only contains singletons apart from $A$ the partition $\xi$ must split the block $A$ into multiple blocks $c \in \xi\rvert_A$. Since $\ci(x_A) >0$ we know that 
\begin{equation}
  \mi_{\xi\rvert_A}(x_A) = \log \frac{p_A(x_A)}{\prod_{c \in \xi\rvert_A}p_c(x_c)} > 0
\end{equation} 
so that $\prod_{c \in \xi\rvert_A}p_c(x_c) < p_A(x_A)$ and
\begin{align}
  \mi_{\xi}(x_V)& = \log \frac{p_V(x_V)}{\prod_{c \in \xi\rvert_A}p_c(x_c) \prod_{i \in V \bs A} p_i(x_i)} \\
  &> \log \frac{p_V(x_V)}{p_A(x_A) \prod_{i \in V \bs A} p_i(x_i)} \\
  &= \mi_{\pi^A}(x_V).
\end{align} 
Therefore $\xi$ is on a disintegration level $\dis_k(x_V)$ with $k>i^A$, but this is true for any refinement of $\pi^A$ so
$\dis_{\prec i^A}(x_V)_{\lpre \pi^A}=\emptyset$ and $\pi^A \in \dis^\lmin_{i^A}(x_V)$.
\end{description}
\end{proof}

\section{Symmetries and STPs}
\label{sec:symm}
In this section we present the behaviour of SLI under permutations of the nodes in the Bayesian network. The behaviour of SLI under such operations can be used to explain the appearance of identical disconnected components on the same disintegration levels in the disintegration hierarchy. We we will see this for simple example systems in \cref{sec:exstp}. The behaviour under transformations like the permutations is also and important property of formal objects in general and can serve as the starting point for further investigations. In order to be able to express the behaviour of SLI under permutations we first have to define the behaviour of STPs and their probabilities under symmetry operations. 

We first define terminology for dealing with subgroups of the symmetric group which is the group of all permutations of a set. In particular we will often restrict the permutations to those that only permute nodes within a subset of the Bayesian network. This is relevant for example if we are dealing with a driven multivariate Markov chain where the driven random variables may be permuted freely among each other but not with the driving random variables. 

Then we define the group actions of such permutations on STPs, partitions, and probabilities of STPs and show that they are indeed group actions\footnote{Group actions have nothing to do with the actions of agents that are important in the conceptual part of this thesis.}. We then state clearly what we mean by symmetries of STPs, partitions, and probabilities of STPs. After two helper theorems we finally come to the SLI symmetry theorems (\cref{thm:symmpart,thm:symmpartcor}). The first establishes the behaviour of SLI of STPs $x_A$ under permutations that are symmetries of the probability distribution over $X_A$ i.e.\ over the nodes that are occupied by the STP. The second then establishes the conditions under which the SLI stays invariant under such permutations. These conditions will be used in our example in \cref{sec:exstp}. We also anticipate that they can be used to establish further theorems about SLI and CLI in particularly symmetric systems. However this is beyond the scope of this thesis.    
\subsection{Symmetric group terminology}

\begin{mydef}
  Let $V$ be a finite set. 
  \begin{thmlist}
  \item A \textit{permutation of $V$} is a bijective function $g:V \rightarrow V$. \\
  \item The set of all permutations together with 
  function composition $(g_1 \circ g_2)(i) := g_1(g_2(i))$,
form a group called the \textit{symmetric group} $\symg_V$. \\
\item A \textit{subgroup} of $\symg_V$ is any subset $\gr{G} \subseteq \symg_V$ such that for all $g_1,g_2 \in \gr{G}$ we have $g_1 \circ g_2 \in \gr{G}$ and for every $g \in \gr{G}$, $g^{-1} \in \gr{G}$. \\
\item Given a subset $A \subseteq V$ and permutation $g \in \symg_V$ define:
  \begin{equation}
    g(A):=\{g(i) : i \in A\}.
  \end{equation} \\
  \item Given a subgroup $\gr{G} \subseteq \symg_V$ and an element $i \in V$ define \textit{the orbit of $i$ under $\gr{G}$} as the set:
  \begin{equation}
    \gr{G}(i):=\{g(i): g \in \gr{G}\}.
  \end{equation} 
  \item Given a subgroup $\gr{G} \subseteq \symg_V$ and a subset $A \subseteq V$ define 
  \begin{equation}
    \gr{G}(A):= \{g(i): g \in \gr{G}, i \in A\}.
  \end{equation} \\
  \item A subset $A \subseteq V$ is an \textit{ invariant subset or invariant under the action of group $\gr{G}$ } if 
  \begin{equation}
    \gr{G}(A) = A.
  \end{equation} \\
  \item A subset $A \in V$ is a \textit{fixed subset or fixed under the action of group $\gr{G}$} if for every $g \in \gr{G}$ and $i \in A$
  \begin{equation}
    g(i) = i.
  \end{equation} \\
  \item For any a subset $A \in V$ let $\symg_A \subseteq \symg_V$ denote the subgroup of permutations such that $V\bs A$ is a fixed subset of $\symg_A$. I.e.\ for $g \in \symg_A$ we have $g(i)=i$ for all $i \in V \bs A$.
  \item Let $V$ be a Cartesian product $V = J \times T$ and let $\gr{G}_1$ be a subgroup of $\symg_J$ and $\gr{G}_2$ a subgroup of $\symg_T$. Then we can form the group $\gr{G}_1 \times \gr{G}_2$ which is a subgroup of $\symg_V$ by defining for any $g_1 \in \gr{G}_1$,$g_1 \in \gr{G}_2$ that $(g_1,g_2)(j,t) = (g_1(j), g_2(t))$.
  \end{thmlist}
\end{mydef}
Remark:
\begin{itemize}
  \item Note that for all permutations $g:V\rightarrow V$ we have $g(V)=V$, but in general for $A \subset V$ we may have either $g(A) \neq A$ or $g(A) = A$. 
  \item Every subgroup contains the identity $\id$ of $\symg_V$.
\end{itemize}

\subsection{Actions of the symmetric group on patterns, partitions, and probabilities}

\begin{mydef}
  Given a Bayesian network $\Xv$, a subset $A \in V$ with $\X_i = \X_j$ for all $i,j \in A$, permutations $g,h \in \symg_A$, and a pattern $x_A \in \X_A \corres \stp{A}$ define the following.
\begin{thmlist}
\item For individual $i \in A$
\begin{equation}
     (X_i = x_i)^g := (X_i = x_{g(i)})
  \end{equation}
  we also write $x^g_i$ where there is no danger of confusion.
\item Furthermore
\begin{equation}\label{def:hinsert}
  (X_i = x_{g(i)})^h := (X_i = x_{g(h(i))})
\end{equation} 
 we also write $(x^g)^h_i$ where there is no danger of confusion.
\item For $B \subseteq A$ the STP $x_A$
\begin{equation}
    \{X_i = x_i\}^g_{i \in B} := \{(X_i = x_i)^g\}_{i \in B}
  \end{equation}
  we also write $x^g_B$ where there is no danger of confusion.

\end{thmlist}

  \end{mydef}
Remarks:
\begin{itemize}
  \item Note that we require the state spaces of all the random variables in $A$ to be equal in order for $x^g_i$ to be well defined for all $i \in A$ and all $x_i \in \X_i$.
  \item In words the pattern $\{X_i = x_i\}^g_{i \in B}$ fixes the random variables at $i \in B$ to the values that $x_A$ defines at $g(i)$. Since the state spaces are identical by assumption this is well defined. 

\item These definitions are based on the full notation of patterns because the shorthand notation does not afford the necessary expressiveness. For calculations in the rest of the section we will often resort to the full notation but at the same time try to use the visually less demanding shorthand where possible.
  \item A simple example: let $V=A= \{1,2\}$ and $g(1)=2,g(2)=1$ then: 
  \begin{align}
\{X_1=x_1,X_2=\bar{x}_2\}^g &= \{(X_1=x_1)^g,(X_2=\bar{x}_2)^g\} \\
&= \{X_1=\bar{x}_{g(1)},X_2=x_{g(2)}\}\\
&=\{X_1=\bar{x}_2,X_2=x_1\}.
\end{align}

We used the bar over $\bar{x}_2$ to highlight the movement of the values. This will be useful when we look at marginalisations later. Note that $(X_1=x_1)^g$ does not reflect the value that $g(1)$ maps to (there is no bar over the $x$) but $X_1=\bar{x}_{g(1)}$ does. This means that for the latter notation we already have to know the result of $g(1)$ in order to know whether to put a bar over $x$ or not.
\item These definition provide a slight adaptation of the definition of ``$x^g$'' in \citet{ceccherinisilberstein_cellular_2009}. The reason for the counter intuitive rule in \ref{def:hinsert} also stems from the close relation of the present definition to the idea behind the definition in that publication. In their case $x$ is a function acting on the indices and $x^g$ is defined as the function taking as argument $g(i)$. Consequently, the function $(x^g)^h$ takes as argument $g(h(i))$. In our case $x$ is not a function (and cannot directly be made into one as the state spaces of the random variables in our Bayesian network may differ unlike in \citet{ceccherinisilberstein_cellular_2009}) so that we emulate similar behaviour with the above rule. 
  \end{itemize}

\begin{mydef}[Action on a STP]
\label{def:actonstp}
  Given a Bayesian network $\Xv$ a subset $A \in V$ with $\X_i = \X_j$ for all $i,j \in A$, a permutation $g \in \symg_A$, a pattern $x_A$, and a subset $C \in A$ define the \textit{action of $g$ on $x_C$}, as:
  In full notation:
  \begin{equation}
    \tilde{g}\stp{C}  := \gdstp{C}{g^{-1}}.
  \end{equation}
  In short notation:
  \begin{equation}
    \tilde{g} x_C := x^{g^{-1}}_C.
  \end{equation} 
  If not necessary we write just $g$ instead of $\tilde{g}$.
\end{mydef}
Remark:
\begin{itemize}
\item We use $\tilde{g}$ here to indicate that this is another object than the permutation $g$ which is a function on index sets. For the subsequent proof that the above definition is an action this distinction is necessary. Beyond this proof it is always clear that it is another object from the context.
  \item The result of the action of $g$ on $x_C$ is then another pattern $\bar{x}_C$ where the new value $\bar{x}_i$ of the random variable $X_i$ at node $i$ is now the value $x_j=x_{g^{-1}(i)}]$ originally fixed for the random variable $X_j$ at node $j = g^{-1}(i)$. One can think of this construction in analogy to ``shifting a function $f:\X\rightarrow \Y$ to the right'' by a constant $d$ by defining $\bar{f}(x):=f(x-d)$. In order to get the function to move in the positive direction by $d$ we its negative $-d$ to the argument. This has the desired effect. Similarly to transform the STP by $g$ we act on the indices with its inverse $g^{-1}$. This is common practice in defining group actions, for a similar construction see \citet{ceccherinisilberstein_cellular_2009}.
  \item It might be redundant to define the action of $g x_C$ on top of the previous definition of $x^g_C$ which is an equivalent construction. The detour is presented here because it allows for the use of the standard construction of the action on probabilities in \cref{def:actonprob}.
\end{itemize}

\begin{thm}
  Given a Bayesian network $\Xv$ a subset $A \in V$ with $\X_i = \X_j$ for all $i,j \in A$, the action on STPs of \cref{def:actonstp} is a group action of the group $\symg_A$ on the set of all STP $\bigcup_{B\subseteq A}\X_B$. This means that for all $g,h \in \symg_A$ and all $x_C \in \bigcup_{B\subseteq A}\X_B$ we have
  \begin{thmlist}
   \item \label{thm:actonstpisact1} \begin{equation}
\tilde{g} x_C \in \bigcup_{B\subseteq A}\X_B,                                               \end{equation} 
   \item \label{thm:actonstpisact2} 
   \begin{equation}
     \tilde{h} (\tilde{g} x_C) = \widetilde{(h \circ g)} x_C,
   \end{equation} 
   \item \label{thm:actonstpisact3} \begin{equation}
\tilde{\id} x_C = x_C.                      \end{equation} 
 \end{thmlist}

\end{thm}
\begin{proof}
  \begin{description}
    \item[ad \ref{thm:actonstpisact1}] Note that $\tilde{g} x_C = \gdstp{C}{g^{-1}}$. Recall $\X_i = \X_j$ for all $i,j \in A$ and $g \in \symg_A$ such that $g^{-1}\in \symg_A$. Then we know that for all $i \in C$ $x_{g^{-1}(i)} \in \X_{g^{-1}(i)}=\X_i$ thus $\bar{x}_C$ with $\bar{x}_i = x_{g^{-1}(i)}$ is a pattern $\bar{x}_C \in \X_C \subseteq \bigcup_{B \subseteq A} \X_B$.
    \item[ad \ref{thm:actonstpisact2}] In the full notation we have:
     \begin{align}
    \tilde{h} (\tilde{g} x_C) &= \tilde{h} \gustp{C}{g^{-1}}\\
    &=\tilde{h} \{(X_i = x_i)^{g^{-1}}\}_{i \in C}\\
    &=\{X_i = x_{g^{-1}(i)}\}^{h^{-1}}_{i \in C}\\
    &=\{(X_i = x_{g^{-1}(i)})^{h^{-1}}\}_{i \in C}\\
    &= \{(X_i = x_{(g^{-1}(h^{-1})(i)}))\}_{i \in C}\\
    &= \{(X_i = x_{(g^{-1} \circ h^{-1})(i)})\}_{i \in C}\\
    &= \{(X_i = x_{(h \circ g)^{-1}(i)})\}_{i \in C}\\
    &= \{(X_i = x_i)\}^{(h \circ g)^{-1}}_{i \in C} \\
&= \widetilde{(h \circ g)} \{(X_i = x_i)\}_{i \in C} \\
&= \widetilde{(h \circ g)} x_C.
\end{align} 
  \item[ad \ref{thm:actonstpisact3}] Note $\tilde{\id} x_C = x^\id_C = \gdstp{C}{\id}=x_C$.
  \end{description}
\end{proof}

\begin{mydef}[Action on a partition]
\label{def:actonpart}
  Given a set $V$, a subset $A \subseteq V$, a partition $\pi \in \Latt(A)$, and a permutation $g \in \symg_A$ define the \textit{action of $g$ on $\pi$} by
  \begin{equation}
    \hat{g} \pi := \{g(b) \subseteq V : b \in \pi\}.
  \end{equation}
  Again if it is not necessary we just write $g \pi$ instead of $\hat{g} \pi$. 
\end{mydef}
\begin{thm}
  Given a set $V$, a subset $A \subseteq V$ a partition $\pi \in \Latt(A)$, and a permutation $g \in \symg_A$, the action on partitions of \cref{def:actonpart} is a group action of $\symg_A$ on the set of all partitions $\pi \in \Latt(A)$. This means that for all $g,h \in \symg_A$ and all $\pi \in \Latt(A)$ we have
 \begin{thmlist}
   \item \label{thm:actonpartisact1} \begin{equation}
\hat{g} \pi \in \Latt(A)                                               \end{equation} 
   \item \label{thm:actonpartisact2} \begin{equation}
\hat{h} (\hat{g} \pi) =\widehat{(h \circ g)} \pi                                     \end{equation} 
   \item \label{thm:actonpartisact3} \begin{equation}
\id \pi = \pi                      \end{equation} 
 \end{thmlist}
\end{thm}

\begin{proof}
  \begin{description}
    \item[ad \ref{thm:actonpartisact1}] Note that since $g \in \symg_A$ we have $g(i) \in A$ for every $i \in A$. Therefore $\hat{g} \pi = \{g(b) \subseteq V : b \in \pi\}=\{g(b) \subseteq A : b \in \pi\}$, so all blocks of $\pi$ are mapped to subsets of $A$. 
    
    To show that $\hat{g} \pi$ is a partition we need to show two things. First, that for all $b_1,b_2 \in \pi$ we have $g(b_1) \cap g(b_2) = \emptyset$. Note $b_1 \cap b_2 = \emptyset$ by assumption and $g:V \rightarrow V$ is injective (even bijective) so that we have for all $i,j \in V$, $g(i) \neq g(j)$ if $i \neq j$. Then for all $i \in b_1,j \in b_2$ always $i \neq j$ and $g(i) \neq g(j)$ so $g(b1) \cap g(b_2) = \emptyset$. 
    
    Second, we have to show that $\bigcup_{c \in \hat{g} \pi} c = A$ which follows from surjectivity of $g$. Pick any $j \in A$ then there exists $i \in A$ such that $g(i) = j$. Since $\pi$ is a partition of $A$, for each such $i \in A$ there is a block $b \in \pi$ with $i \in \pi$ such that $j \in \hat{g} \pi$.  
    \item[ad \ref{thm:actonpartisact2}] Note:
     \begin{align}
    \hat{h} (\hat{g} \pi) &= \hat{h} \{g(b) \subseteq A : b \in \pi\} \\ 
    &= \{h(g(b)) \subseteq A : b \in \pi\}\\
    &= \{(h\circ g)(b) \subseteq A : b \in \pi\}\\
    &= \widehat{(h \circ g)}\pi
  \end{align} 
  \item[ad \ref{thm:actonpartisact3}] Note $\hat{\id} \pi = \{\id(b) \subseteq A : b \in \pi\} = \pi$.
  \end{description}
\end{proof}

\begin{thm}
\label{thm:refinementpreserve}
  Given a set $V$, a subset $A \subseteq V$, partitions $\pi, \xi \in \Latt(A)$, and a permutation $g \in \symg_A$ the action of $g$ on the partitions preserves the refinement relation. Formally:
  \begin{equation}
    \pi \lpreeq \xi \Leftrightarrow g\pi \lpreeq g\xi.
  \end{equation} 
\end{thm}
\begin{proof}
  From $i \equiv_\pi j$ which just means that there exists $b in \pi$ with $i,j \in b$ we get $g(i) \equiv_{g\pi} g(j)$ since membership of sets is preserved i.e.\ if $i \in b$ then $g(i) \in g(b)$. So if $i \equiv_\pi j \Rightarrow i \equiv_\xi j$ then also $g(i) \equiv_{g\pi} g(j) \Rightarrow g(i) \equiv_{g\xi} g(j)$. So by \cref{def:refinement} $g\pi \lpreeq g\xi$. 
\end{proof}
Remark:
\begin{itemize}
  \item Visually, this means that the Hasse diagram of a set of transformed partitions is the same as that of the non-transformed partitions.
\end{itemize}
\begin{thm}
\label{thm:cardinalitypreserve}
  Given a set $V$, a subset $A \subseteq V$, partitions $\pi \in \Latt(A)$, and a permutation $g \in \symg_A$ the action of $g$ on the partitions preserves the cardinality $|\pi|$. Formally:
  \begin{equation}
    |\pi| = |g\pi|.
  \end{equation} 
\end{thm}
\begin{proof}
We have seen that $g \pi$ is a partition as well. By definition it is only composed out of the images of the blocks, so there cannot be more blocks in $g \pi$ than in $\pi$. We also know that $g$ is injective so no two elements can be mapped to the same element and by extension no tow blocks can be mapped to the same block. Therefore there cannot be fewer blocks in $g \pi$ than in $\pi$.
\end{proof}

\begin{mydef}[Action on probability distributions]
\label{def:actonprob}
 Given a Bayesian network $\Xv$ a subset $A \in V$ with $\X_i = \X_j$ for all $i,j \in A$, and any permutation $g \in \symg_A$ define the \textit{action of $g$ on the probability distribution $p_V:\X_V \rightarrow \X_V$} by setting for  
 for each $x_V \in \X_V$:
 \begin{align}
   (\check{g} p_V)(x_V):&= \Pr( g^{-1} \stp{V} ) \\
   &= \Pr(\gdstp{V}{g}).
 \end{align}  
 If there is no danger of confusion we also write in short notation of the patterns $(\check{g} p_V)(x_V)= p_V(x^g_V)$. Also if not necessary we will write $(g p_V)$ instead of $(\check{g} p_V)$.
\end{mydef}
Remark:
\begin{itemize}
  \item So the probability distribution $\check{g} p_V$ resulting from the action of $g$ on $p_V$ assigns $x_V$ the probability that was originally assigned to $g^{-1} x_V$. The latter is the trajectory $x^g_V$ that we obtain by moving the values $x_{g(i)}$ to nodes $i$. 
  \item The action of $g$ on $p_V$ as defined here corresponds to the usual way group actions are defined on functions. Namely by making the inverse $g^{-1}$ act on the argument to the function.
  \item We only require that the state spaces of the random variables $\{X_i\}_{i \in A}$ whose indices $g$ does not keep fixed are equal. This allows us to deal with situations where some random variables have different state spaces or do not exhibit the same symmetric structure as those in $A$. This can be useful in the case of driven Markov chains.
\end{itemize}

%

\begin{thm}
  Given a Bayesian network $\Xv$ a subset $A \in V$ with $\X_i = \X_j$ for all $i,j \in A$, and a permutation $g \in \symg_A$, the action on probability distributions of \cref{def:actonprob} is a group action of $\symg_A$ on the set of all probability distributions $p_V \in \Pset(\X_V)$. This means that for all $g,h \in \symg_A$ and all $p_V$ we have
 \begin{thmlist}
   \item \label{thm:actonprobisact1} \begin{equation}
\check{g} p_V \in \Pset(\X_V)                                              \end{equation} 
   \item \label{thm:actonprobisact2} \begin{equation}
\check{h} (\check{g} p_V) =\widecheck{(h \circ g)} p_V                                     \end{equation} 
   \item \label{thm:actonprobisact3} \begin{equation}
\check{\id} p_V = p_V                      \end{equation} 
 \end{thmlist}
\end{thm}

\begin{proof}
  \begin{description}
    \item[ad \ref{thm:actonprobisact1}] First, note that since $g \in \symg_A$ so that $g^{-1} \in \symg_A$ and $\X_k = \X_l$ for all $k,l \in A$ we have $x_{g^{-1}(i)} \in \X_i$ for all $i \in A$. Also because of $g^{-1} \in \symg_A$ we have $g^{-1}(i)=i$ for $i \in V\bs A$, so actually $x_{g^{-1}(i)} \in \X_i$ for all $i \in V$. This makes $x^g_V$ a valid pattern.
    
    Second, show that for all $x_V \in \X_V$ we have $(\check{g} p_V)(x_V) \in [0,1]$. This follows from $x^g_V$ being a valid pattern for each $x_V$ and the probability of any valid pattern being in the interval $[0,1]$.
    
    Third, we show that the transformed distribution $\check{g} p_V$ is normalised. Since the action of $g$ on $x_V$ realises a bijective function on $\X_V$ ($g$ has an inverse, and its inverse has an inverse) we can substitute $x_V$ with $\check{g} x_V$ as in the following:\footnote{We use $\bigoplus$ here to denote that for each $j$ there is a sum over $x_j$. This is to keep the notation somewhat short, and clear.}
    \begin{align}
      \sum_{x_V \in \X_V} (\check{g} p_V)(x_V) &= \bigoplus_{j \in A} \sum_{x_j \in \X_j} \Pr\left(g^{-1} \stp{V}\right)\\
      &= \bigoplus_{j \in A} \sum_{x_{g^{-1}(j)} \in \X_j} \Pr\left(g^{-1} \gdstp{V}{g^{-1}}\right)\\
      &= \bigoplus_{j \in A} \sum_{x_j \in \X_j} \Pr\left(g^{-1} \gdstp{V}{g^{-1}}\right)\\
      &= \bigoplus_{j \in A} \sum_{x_j \in \X_j} \Pr\left(g^{-1}( g \stp{V})\right)\\
      &= \bigoplus_{j \in A} \sum_{x_j \in \X_j} \Pr\left(\stp{V}\right)\\
      &= 1.
    \end{align}
    \item[ad \ref{thm:actonprobisact2}] Note, for all $x_V$:
     \begin{align}
    (\check{h} (\check{g} p_V))(x_V) &= (\check{g} p_V)(h^{-1} x_V) \\ 
    &= p_V(g^{-1}(h^{-1} x_V)) \\ 
    &= p((g^{-1} \circ h^{-1}) x_V) \\ 
    &= p((h \circ g)^{-1} x_V) \\ 
    &= (\widecheck{(h \circ g)}p_V)(x_V).
  \end{align} 
  \item[ad \ref{thm:actonprobisact3}] For all $x_V \in \X_V$, we have $\check{\id} p_V(x_V) = p_V(\id x_V)=p_v(x_V)$.
  \end{description}
\end{proof}

\begin{mydef}[Symmetries]
\label{def:symmetries}
  Given a Bayesian network $\Xv$ a subset $A \in V$ with $\X_i = \X_j$ for all $i,j \in A$, a permutation $g \in \symg_A$, a pattern $x_A$, a subset $C \in A$, and a partition $\pi$ of $C$.  
  \begin{thmlist}
\item We say \textit{$g$ is a symmetry of $x_C$} or \textit{$x_C$ is invariant under $g$} if
  \begin{equation}
    g x_C = x_C,
  \end{equation}
  and call the group of all symmetries of $x_C$ the \textit{symmetry group of $x_C$}. 
\item We say \textit{$g$ is a symmetry of $\pi$} or \textit{$\pi$ is invariant under $g$} if
  \begin{equation}
    g \pi = \pi,
  \end{equation}
  and call the group of all symmetries of $\pi$ the \textit{symmetry group of $\pi$}.  
\item We say \textit{$g$ is a symmetry of $p_C:\X_C \rightarrow [0,1]$} or \textit{$p_C$ is invariant under $g$} if
  \begin{equation}
    g p_C = p_C,
  \end{equation}
  and call the group of all symmetries of $p_C$ the \textit{symmetry group of $p_C$}.  
\end{thmlist}
\end{mydef}
Remark:
\begin{itemize}
  \item We can only have $g x_C = x_C$ or $g \pi = \pi$ if $gC=C$. 
\end{itemize}

\subsection{Transformation of SLI for invariant probability distributions}

\begin{thm}
\label{thm:actonpstp}
Given a Bayesian network $\Xv$ a subset $A \in V$ with $\X_i = \X_j$ for all $i,j \in A$, a permutation $g \in \symg_A$, and a subset $B \subseteq A$ we have for every $x_B \in \X_B$:
 \begin{equation}
   (g p_B)(x_B)= \Pr\left( \gdstp{g^{-1}(B)}{g}\right). 
 \end{equation}   
We can also write this as 
\begin{equation}
(g p_B)(x_B)= p_{g^{-1}(B)} ( x^g_{ g^{-1}(B)}).
\end{equation}  

Furthermore, if $B = A$ we get:
\begin{equation}
(g p_A)(x_A)= p_A (x^g_A).
\end{equation} 
\end{thm}
\begin{proof}
  By definition we have:
  \begin{align}
     (g p_B)(x_B) &= \bigoplus_{j \in V \bs B} \sum_{\bar{x}_j \in \X_j} \Pr\left( g^{-1} \left( \stp{B},\barstp{V \bs B} \right)\right)\\
     &= \bigoplus_{j \in V \bs B} \sum_{\bar{x}_j \in \X_j} \Pr\left(\gustp{B}{g},\bargustp{A \bs B}{g},\barstp{V \bs A} \right) \\
     &=\bigoplus_{j \in A \bs B} \sum_{\bar{x}_j \in \X_j} \Pr\left( \gustp{B}{g},\bargustp{A \bs B}{g}\right) \\
     \begin{split}&=\bigoplus_{j \in A \bs B} \sum_{\bar{x}_j \in \X_j} \Pr\left( \gustp{g^{-1}(B) \cap B}{g}, \right.\\
     &\phantom{\bigoplus_{j \in A \bs B} \sum_{\bar{x}_j \in \X_j} \Pr\left(\right.} \bargustp{g^{-1}(B) \bs B}{g}, \\
     &\phantom{\bigoplus_{j \in A \bs B} \sum_{\bar{x}_j \in \X_j} \Pr\left(\right.} \bargustp{g^{-1}(A \bs B) \cap (A \bs B)}{g}, \\
     &\phantom{\bigoplus_{j \in A \bs B} \sum_{\bar{x}_j \in \X_j} \Pr\left(\right.}\left.\gustp{g^{-1}(A \bs B) \bs (A \bs B)}{g}\right) \\
\end{split}\\
     \begin{split}&=\bigoplus_{j \in g^{-1}(A \bs B)} \sum_{\bar{x}_j \in \X_j} \Pr\left( \gdstp{g^{-1}(B) \cap B}{g}, \right.\\
     &\phantom{\bigoplus_{j \in A \bs B} \sum_{\bar{x}_j \in \X_j} \Pr\left(\right.} \gdstp{g^{-1}(B) \bs B}{g}, \\
     &\phantom{\bigoplus_{j \in A \bs B} \sum_{\bar{x}_j \in \X_j} \Pr\left(\right.} \bargdstp{g^{-1}(A \bs B) \cap (A \bs B)}{g}, \\
     &\phantom{\bigoplus_{j \in A \bs B} \sum_{\bar{x}_j \in \X_j} \Pr\left(\right.}\left.\bargdstp{g^{-1}(A \bs B) \bs (A \bs B)}{g}\right) \\
\end{split}\\
     &=\bigoplus_{j \in A \bs B} \sum_{\bar{x}_j \in \X_j} \Pr\left( \gdstp{g^{-1}(B)}{g}, \bargdstp{g^{-1}(A \bs B)}{g}\right) \\
     &= \Pr\left( \gdstp{g^{-1}(B)}{g}\right) \\
     &= \Pr\left(g^{-1} \stp{g^{-1}(B)}\right) \\
&=: p_{g^{-1}(B)} ( x^g_{ g^{-1}(B)}).
\end{align}
Note the movement of the bar and with it the change of summation indices from $A\bs B$ to $g^{-1}(A \bs B)$.


Since $g \in \symg_A$ we have $g^{-1}(A) = A$ such that if $B=A$ it follows immediately that 
\begin{equation}
   p_{g^{-1}(A)} ( x^g_{ g^{-1}(A)}) = p_A(x^g_A).
\end{equation} 
\end{proof}
Remark:
\begin{itemize}
\item The main advantage of the full notation is the clear separation between the action of $g$ on the index of the assigned values i.e.\ $x_i \mapsto x_{g(i)}$ and the action on the sets of indices that are affected by a pattern i.e.\ $B \mapsto g(B)$. 
In the shorthand notation this separation may easily get lost during manipulations of the equations. 
  \item This theorem shows the effect of the action of a group element $g \in \symg_A$ on the (marginalised) probability of a pattern within $A$. Marginalisation is used in the usual way but now for the transformed probability distribution $g p_V$ instead of for $p_V$. 
\end{itemize}

\begin{thm}
  Given a Bayesian network $\Xv$, a subset $A \subset V$ of the index set $V$, and a group of permutations $\gr{G} \subseteq \symg_A$. If for all $g \in \gr{G}$ we have:
  \begin{equation}
    g p_V = p_V,
  \end{equation}
  then for all $g \in \gr{G}$ also:
  \begin{equation}
    g p_A = p_A.
  \end{equation} 
\end{thm}
\begin{proof}
   \begin{align}
     (g p_A)(x_A) &= \bigoplus_{j \in V \bs A} \sum_{\bar{x}_j \in \X_j} \Pr\left( g^{-1} \left( \stp{A},\barstp{V \bs A} \right)\right)\\
         &= \bigoplus_{j \in V \bs A} \sum_{\bar{x}_j \in \X_j} \Pr\left(\gustp{A}{g},\bargustp{V \bs A}{g} \right) \\
         &= \bigoplus_{j \in V \bs A} \sum_{\bar{x}_j \in \X_j} \Pr\left(\gustp{A}{g},\barstp{V \bs A} \right) \\
         &= \bigoplus_{j \in V \bs A} \sum_{\bar{x}_j \in \X_j} \Pr\left(\stp{A},\barstp{V \bs A} \right) \\
         &= p_A(x_A).
     \end{align}    
\end{proof}
Remark:
\begin{itemize}
  \item This theorem shows that if we find subgroups of $\symg_V$ affecting only a subset $A \subseteq V$ of the nodes in Bayesian network (i.e.\ any subgroup of $\symg_A$) and we have $g p_V = p_V$ i.e.\ the joint probability distribution over $X_V$ is invariant with respect to the action of this subgroup, then the joint probability distribution $p_A$ over the subset of nodes is also invariant with respect to the action of this group.   
\end{itemize}

\begin{thm}[SLI symmetry theorem]
\label{thm:symmpart}
Given a Bayesian network $\Xv$, a subset $A \subset V$ of the index set $V$, and a group of permutations $\gr{G} \subseteq \symg_A$ such that for all $g \in \gr{G}$ we have:
\begin{equation}
  g p_A = p_A
\end{equation} 
then for all $x_A \in \X_A$ and all $g \in \gr{G}$:
\begin{equation}
  \mi_{g \pi}(x_A) = \mi_\pi(x^g_A)
\end{equation}
\end{thm}
\begin{proof}
Note first:
\begin{align}
    \mi_{g \pi}(x_A)&= \log \frac{p_A(x_A)}{\prod_{b \in g \pi} p_b(x_b)}\\
    &=\log \frac{p_A(x_A)}{\prod_{c \in \pi} p_{g(c)} (x_{g(c)})}. \label{eq:symmpart}
  \end{align}
Now look at the block probabilities individually and use the invariance of $p_A$:
\begin{align}
  p_{g(c)} (x_{g(c)})&= \sum_{\bar{x}_{V\bs g(c)}} p_V(x_{g(c)}, \bar{x}_{A\bs g(c)},\bar{x}_{V\bs A}) \\
  &=\bigoplus_{j \in V \bs g(c)} \sum_{\bar{x}_j \in \X_j} \Pr\left( \stp{g(c)},\barstp{V \bs g(c)}\right) \\
  &=\bigoplus_{j \in V \bs g(c)} \sum_{\bar{x}_j \in \X_j} \Pr\left( g^{-1} \left( \stp{g(c)},\barstp{V \bs g(c)} \right)\right) \\
  &=(g p_{g(c)})(x_{g(c)}) \\
  &=Pr\left(\gdstp{g^{-1}(g(c))}{g}\right) \\
  &=Pr\left(\gdstp{c}{g} \right)\\
  &=p_c(x^g_c).\label{eq:symmblock}
\end{align}
Plugging the block probabilities and $p_A(x_A) =p_A(x^g_A)$ into \cref{eq:symmpart} concludes the proof.
\end{proof}
Remark:
\begin{itemize}
\item This theorem concerns the reaction of the SLI to a transformation of the partition by a permutation under which the probability distribution is invariant.
\item Note that we can substitute $x_A \mapsto x^{g^{-1}}_A$ and get $\mi_\pi(x_A) = \mi_{g \pi}(g x_A)$. 
\end{itemize} 

\begin{thm}[SLI symmetry corollary]
\label{thm:symmpartcor}
Under the assumptions of \cref{thm:symmpart}, let $\gr{H}$ be a subgroup of $\gr{G}$.
\begin{thmlist}
  \item \label{thm:ontrapartsymm1} If for some $x_A$ we have for all $h \in \gr{H}$: 
  \begin{equation}
x^h_A = x_A  \end{equation} 
  then for all $h \in \gr{H}$:
\begin{equation}
  \mi_{h \pi}(x_A)=\mi_\pi(x_A).
\end{equation} 
\item \label{thm:ontrapartsymm2} If $x^h_A \neq x_A$ but for all $b \in \pi$ we have 
\begin{equation}
p_b(x^h_b) = p_b(x_b) 
\end{equation} 
then for all $g \in \gr{G}$:
\begin{equation}
\mi_{g\pi}(x_A)=\mi_\pi(x_A).                             
\end{equation} 
\item \label{thm:ontrapartsymm3} If $x^g_A \neq x_A$ and there exists a $b \in \pi$ with $p_b(x^g_b) \neq p_b(x_b)$, but for all $g \in \gr{G}$ we have 
\begin{equation}
\prod_{b \in \pi} p_b(x^g_b) = \prod_{b \in \pi} p_b(x_b)
\end{equation}   
then for all $g \in \gr{G}$:
\begin{equation}
\mi_{g\pi}(x_A)=\mi_\pi(x_A).                             
\end{equation} 
\end{thmlist}
\end{thm}
\begin{proof}
  \begin{description}
    \item[ad \ref{thm:ontrapartsymm1}:] Follows directly from \cref{thm:symmpart} by plugging in $x^h_A=x_A$ which also implies $x^h_b = x_b$ for all $b \in \pi$.
    \item[ad \ref{thm:ontrapartsymm2}:] Note that $p(x^h_A)=p(x_A)$ since $h p_A = p_A$ and $(h p_A)(x_A) = p_A(x^h_A)$ according to  \cref{thm:actonpstp}. Then:
    \begin{equation}
      \mi_{h\pi}(x_A) = \log \frac{p_A(x^h_A)}{\prod_{b \in \pi} p_b(x^h_b)} = \log \frac{p_A(x_A)}{\prod_{b \in \pi} p_b(x_b)}.
    \end{equation}
    Where we used that $p_b(x^h_b)=p_b(x_b)$ for all $b \in \pi$ by assumption.
    \item[ad \ref{thm:ontrapartsymm3}:] Just like for \ref{thm:ontrapartsymm2} we have
        \begin{equation}
      \mi_{h\pi}(x_A) = \log \frac{p_A(x_A)}{\prod_{b \in \pi} p_b(x^h_b)} = \log \frac{p_A(x_A)}{\prod_{b \in \pi} p_b(x_b)},
    \end{equation}
    where, this time we used that $\prod_{b \in \pi} p_b(x^h_b) = \prod_{b \in \pi} p_b(x_b)$ by assumption.
  \end{description}
\end{proof}

Remarks:
\begin{itemize}
\item $\gr{H}$ needs to be a subgroup of $\gr{G}$ because else we don't have $p_{h(b)}(x_{h(b)})=p_b(x^h_b)$. Recall that this only holds for elements $g \in \gr{G}$ because for those $g p_A =p_A$ and only in that case we have \cref{eq:symmblock}. 
  \item These equalities all concern the consequences of \cref{thm:symmpart} for partitions of a single STP $x_A$. The consequences across differing trajectories are not covered.
  \item The three statements can be seen to describe three levels of conditions which imply the equality of SLI. In all three cases, due to the fact that $\gr{G}$ leaves $p_A(x_A)$ invariant we have $p_{g b}(x_{g b})=p_b(x^g_b)$ for all $b \in \pi$. In the first case and on what could be called the lowest level if a subgroup of $\gr{G}$ also leaves $x_A$ invariant then $x^h_b=x_b$ and therefore $p_b(x^h_b)=p_b(x_b)$ for all $b \in \pi$. 
  
  The second states that if there is no invariance of $x_A$ then for some other reason the probability distribution $p_A$ may still be such that all block probabilities are invariant i.e.\ $p_b(x^h_b)=p_b(x_b)$ for all $b \in \pi$ and we still get equal SLI.
  
  The third then shows that if there is no invariance of $x_A$ or of all the $p_b$ then it may still be the case that the product $\prod_{b \in \pi} p_b(x^h_b) = \prod_{b \in \pi} p_b(x_b)$ being equal even if the individual terms differ. 
\end{itemize}

\section{Symmetries and Markov chains}
\label{sec:symmmc}
Here, we state and prove in our own notation three theorems on symmetries and Markov chains. We first define what we mean by a symmetry of a Markov matrix. These are purely spatial permutations that commute with the Markov matrix. We then show the consequence for the individual entries of the Markov matrix of such a symmetry. Then we show that a symmetry of the initial distribution which is also a symmetry of the Markov matrix becomes a symmetry of the joint probability distribution of the entire Bayesian network i.e.\ of the probability distribution over the trajectories. This theorem is well known and only presented for quick reference since we use it in \cref{sec:mcconstsymmetries}. Finally, we look at the special case where only some of the spatial indices are permuted and (for example) the driving variables are left alone. In that case, as long as the interactions with the driving variables obey the symmetry as well (in the sense established by the theorem) the symmetry of an initial distribution is also extended to the joint probability distribution over the entire network. This theorem is provided as a connection to simulations of driven systems which exhibit life-like behaviour. If the drive obeys the condition presented here the SLI symmetry theorems can also be used for such systems and not only for cellular automata. We give a short example of a thermostat-like system in the remark. For the rest of the thesis, this theorem is not important.

\begin{mydef}[Spatial symmetries of Markov matrices]
  Let $\Xv$ be a multivariate Markov chain with $V = J \times T$ and let $g \in \symg_{J} \times \{\id\}$.\footnote{Recall that $\id$ indicates the identity element of a group.}
Then we say \textit{$g$ is a symmetry of $P_{t+1}$} or \textit{$P_{t+1}$ is invariant under $g$} if for all 
probability distributions $p_{V_t}:\X_{V_t} \rightarrow [0,1]$ we have
\begin{equation}
\label{eq:gPcommute}
 gp_{V_{t+1}} = g (P_{t+1} p_{V_t}) = P_{t+1} (g p_{V_t}),
\end{equation}
and call the group of all symmetries of $P_{t+1}$ the \textit{symmetry group of $P_{t+1}$}.
\end{mydef}

\begin{thm}
\label{thm:symmP}
   Let $\Xv$ be a multivariate Markov chain with $V = J \times T$.
   \begin{thmlist}
     \item \label{thm:symmP1} Let $\gr{G}$ be a subgroup of the symmetry group of $P_{t+1}$. Then for every $g \in \gr{G}$:
   \begin{equation}
\label{eq:symmP}
  p_{V_{t+1}}(x^g_{V_{t+1}}|x^g_{V_t}) = p_{V_{t+1}}(x_{V_{t+1}}|x_{V_t})
\end{equation} 
which is equivalent to 
\begin{equation}
\label{eq:symmP2}
  p_{V_{t+1}}(x^g_{V_{t+1}}|\hat{x}_{V_t}) = p_{V_{t+1}}(x_{V_{t+1}}|\hat{x}^{g^{-1}}_{V_t}).
\end{equation}
\item \label{thm:symmP2}    
Conversely, if for all elements $g$ of a subgroup $\gr{G}$ of $\symg_J \times \{\id\}$ we have
\begin{equation}
  p_{V_{t+1}}(x^g_{V_{t+1}}|\hat{x}_{V_t}) = p_{V_{t+1}}(x_{V_{t+1}}|\hat{x}^{g^{-1}}_{V_t}).
\end{equation}
then $\gr{G}$ is a symmetry group of $P_{t+1}$.
   \end{thmlist}
\end{thm}
\begin{proof}
\begin{description}
  \item[ad \ref{thm:symmP1}:] 
  By assumption \cref{eq:gPcommute} holds for any choice of $p_{V_t}$ so we choose for arbitrary $\hat{x}_{V_t}\in \X_{A_t}$
\begin{equation}
  p_{V_t}(x_{V_t}) = \delta_{\hat{x}_{V_t}}(x_{V_t}),
\end{equation} 
and plug this first into the left hand side of \cref{eq:gPcommute}:
\begin{align}
  (g p_{V_{t+1}})(x_{V_{t+1}}) &= p_{V_{t+1}}(x^g_{V_{t+1}}) \\
  &=\sum_{\bar{x}_{V_t}} p_{V_{t+1}}(x^g_{V_{t+1}}|\bar{x}_{V_t}) p_{V_t}(\bar{x}_{V_t}) \\
  &=\sum_{\bar{x}_{V_t}} p_{V_{t+1}}(x^g_{V_{t+1}}|\bar{x}_{V_t}) \delta_{\hat{x}_{V_t}}(\bar{x}_{V_t}) \\
  &= p_{V_{t+1}}(x^g_{V_{t+1}}|\hat{x}_{V_t}).
\end{align}
Then into the right hand side of \cref{eq:gPcommute}:
\begin{align}
  (P_{t+1} (g p_{V_t}))(x_{V_{t+1}})&=\sum_{\bar{x}_{V_t}} p_{V_{t+1}}(x_{V_{t+1}}|\bar{x}_{V_t}) p_{V_t}(\bar{x}^g_{V_t}) \\
  &=\sum_{\bar{x}_{V_t}} p_{V_{t+1}}(x_{V_{t+1}}|\bar{x}_{V_t}) \delta_{\hat{x}_{V_t}}(\bar{x}^g_{V_t}) \\
  &= p_{V_{t+1}}(x_{V_{t+1}}|\hat{x}^{g^{-1}}_{V_t}),
\end{align}
where we used that $\hat{x}_{V_t}=\bar{x}^g_{V_t} \Leftrightarrow \hat{x}^{g^{-1}}_{V_t}=\bar{x}_{V_t}$. So we have for all $x_{V_{t+1}}$ and all $\hat{x}_{V_t}$:
\begin{equation}
  p_{V_{t+1}}(x^g_{V_{t+1}}|\hat{x}_{V_t}) = p_{V_{t+1}}(x_{V_{t+1}}|\hat{x}^{g^{-1}}_{V_t}).
\end{equation} 
Now substitute $\hat{x}_{V_t} \mapsto x^g_{V_t}$ to get:
\begin{equation}
  p_{V_{t+1}}(x^g_{V_{t+1}}|x^g_{V_t}) = p_{V_{t+1}}(x_{V_{t+1}}|x_{V_t}).
\end{equation}

\item[ad \ref{thm:symmP2}:]
For the converse note that if \cref{eq:symmP2} holds for an element $g$ of a subgroup $\gr{G}$ of $\symg_J \times \{\id\}$:
\begin{align}
  (g p_{V_{t+1}})(x_{V_{t+1}}) &=\sum_{\bar{x}_{V_t}} p_{V_{t+1}}(x^g_{V_{t+1}}|\bar{x}_{V_t}) p_{V_t}(\bar{x}_{V_t}) \\
  &=\sum_{\bar{x}_{V_t}} p_{V_{t+1}}(x_{V_{t+1}}|\bar{x}^{g^{-1}}_{V_t}) p_{V_t}(\bar{x}_{V_t}) \\
  &=\sum_{\bar{x}^g_{V_t}} p_{V_{t+1}}(x_{V_{t+1}}|\bar{x}_{V_t}) p_{V_t}(\bar{x}^g_{V_t}) \\
  &=\sum_{\bar{x}_{V_t}} p_{V_{t+1}}(x_{V_{t+1}}|\bar{x}_{V_t}) p_{V_t}(\bar{x}^g_{V_t}) \\
  &= (P_{V_{t+1}}(g p_{V_t}))(x_{V_{t+1}}).
\end{align}
Where we used that we sum over all elements of $\X_{V_t}$ so changing the $\bar{x}_{V_t}$ to $\bar{x}^g_{V_t}$ cannot change the result.
\end{description} 
\end{proof}

\begin{thm}[Extension of symmetries of Markov matrices to the whole Markov chain]
\label{thm:spatialsymm}
Let $\Xv$ be a multivariate Markov chain with $V = J \times T$ and let $\gr{G}$ be a subgroup of $\symg_{J} \times \{\id\}$. 
If for all $t \in T$ a group $\gr{G}$ is a subgroup of the symmetry group of the Markov matrix $P_t$ and also a subgroup of the symmetry group of the initial distribution $p_{V_0}:\X_{V_0} \rightarrow [0,1]$ then for all $g \in \gr{G}$
\begin{equation}
  g p_V = p_V
\end{equation}
and $\gr{G}$ is also a subgroup of the symmetry group of $p_V$.
\end{thm}
\begin{proof}
First note that for all $g \in \symg_V$
\begin{equation}
  g p_V = g p_{V_0,V_1,...,V_T},
\end{equation} 
so
\begin{align}
\label{eq:wholebn}
  g p_V(x_V) &= g p_{V_0,V_1,...,V_T}(x_{V_0},x_{A_1},...,x_{V_T})\\
	     &= p_{V_0,V_1,...,V_T}(x^g_{V_0},x^g_{A_1},...,x^g_{V_T})\\
	     &= \prod_{t=0}^T p_{V_{t+1}}(x^g_{V_{t+1}}|x^g_{V_t}) p_{V_0}(x^g_{V_0}).
\end{align}
Now if $\gr{G}$ is a subgroup of the symmetry group of $P_{t+1}$ and $g \in \gr{G}$ we have from \cref{thm:symmP}:  
\begin{equation}
  p_{V_{t+1}}(x^g_{V_{t+1}}|x^g_{V_t}) = p_{V_{t+1}}(x_{V_{t+1}}|x_{V_t}).
\end{equation} 
Plug this into \cref{eq:wholebn} to get:
\begin{align}
  g p_V(x_V) &= \prod_{t=0}^T p_{V_{t+1}}(x^g_{V_{t+1}}|x^g_{V_t}) p_{V_0}(x^g_{V_0})\\
   &=\prod_{t=0}^T p_{V_{t+1}}(x_{V_{t+1}}|x_{V_t}) p_{V_0}(x^g_{V_0}).
\end{align}
Since additionally $\gr{G}$ is a subgroup of the symmetry group of $p_{V_0}$ we have $g p_{V_0} = p_{V_0}$ and therefore arrive at $g p_V= p_V$. 
\end{proof}
Remark:
\begin{itemize}
  \item In words this theorem state that if the transition matrix commutes with the action of a group that permutes only indices within time slices (i.e.\ spatial indices) and the initial distribution is invariant with respect to such permutations then the joint probability distribution $p_V$ over the entire Bayesian network is invariant with respect to such permutations.
  \item If the spatial permutation leaves a set $B \subset J$ fixed we can treat this as a special case of this theorem. We will do this next.
\end{itemize}

\begin{thm}[Spatial symmetries of driven multivariate Markov chains]
\label{thm:drivensymm}
Let $\Xv$ be a driven Markov chain with index set $V=J \times T$ and $J = A \cup B$ with $A \cap B =\emptyset$. Here $B$ indicates the driving random variables and $A$ the driven ones. Also let $\gr{G}$ be a subgroup of $\symg_A \times \{\id\}$. 
If for all $g=(h,\id) \in \gr{G}$, and all $x_{A,t},x_{A,t+1},x_{B,t},x_{B,t+1}$ we have 
\begin{equation}
\label{eq:drivenchainsymmP}
 \begin{split}p_{A,t+1}(x_{A,t+1}|x_{B,t+1},x_{A,t})& p_{B,t+1}(x_{B,t+1}|x_{A,t},x_{B,t}) \\
    =& p_{A,t+1}(x^g_{A,t+1}|x_{B,t+1},x^g_{A,t}) p_{B,t+1}(x_{B,t+1}|x^g_{A,t},x_{B,t})
    \end{split}
\end{equation} 
and we are given an initial distribution $p_{A_0,B_0}:\X_{A_0}\times \X_{B_0} \rightarrow [0,1]$ with
\begin{equation}
  g p_{A_0,B_0} = p_{A_0,B_0},
\end{equation} 
then
\begin{equation}
  g p_V = p_V.
\end{equation} 
\end{thm}
\begin{proof}
First note for a driven Markov chain we have (see \cref{def:mvdrivenchain})
\begin{equation}
p_{V_{t+1}}(x_{V_{t+1}}|x_{V_t})=p_{A,t+1}(x_{A,t+1}|x_{B,t+1},x_{A,t}) p_{B,t+1}(x_{B,t+1}|x_{A,t},x_{B,t}).
\end{equation} 
Then we get for $g \in \symg_A \times \{\id\}$:
\begin{equation}
  p_{V_{t+1}}(x^g_{V_{t+1}}|x^g_{V_t}) =p_{A,t+1}(x^g_{A,t+1}|x_{B,t+1},x^g_{A,t}) p_{B,t+1}(x_{B,t+1}|x^g_{A,t},x_{B,t})
\end{equation} 
so condition \cref{eq:drivenchainsymmP} is equivalent to \cref{eq:symmP} for $g \in \symg_A \times \{\id\}$. This means that if 
\begin{equation}
  g p_{V_0} = g p_{A\cup B,0} =g p_{J,0} = p_{V_0},
\end{equation} 
we get $ g p_V = p_V.$
\end{proof}
Remarks:
\begin{itemize}
  \item If $B_t = \emptyset$ this reduces to \cref{thm:spatialsymm}.
  \item Here we have limited the action of the symmetry group to the driven spatial random variables. The condition just state that as long as the interaction with the driving random variables is invariant with respect to the permutations of the driven random variables, the symmetry of an initial distribution is maintained throughout the entire Bayesian network. 
  \item A possible example system is where the drive depends on the average of all states of the nodes in $A_t$. Such an average is invariant under any permutation. For example let $\X_{B,t+1}=\{0,1\}$ and for all $j \in A_t$, $\X_j =\{0,1\}$ and define (for all $t \in T$:
  \begin{equation}
    p_{B,t+1}(x_{B,t+1}|x_{A,t},x_{B,t}) = \begin{cases}
                                                   1 & \text{ if } x_{B,t+1} =1 \wedge \sum_{j \in A} x_{j,t} \leq |A_t|/2 \\
                                                   1 & \text{ if } x_{B,t+1} =0 \wedge \sum_{j \in A} x_{j,t} > |A_t|/2 \\
                                                   0 & \text{ else,}
                                                 \end{cases}
  \end{equation} 
  such that $x_{B,t+1} \in \{0,1\}$ depends on whether more than half of the variables in $A_t$ have value $1$ or not. It is straightforward to check that then $p_{B,t+1}(x_{B,t+1}|x_{A,t},x_{B,t}) =p_{B,t+1}(x_{B,t+1}|x^g_{A,t},x_{B,t})$ for any $g \in \symg_{A,t}$.
  The influence on the nodes in $A_{t+1}$ also has to be symmetric, which can easily be achieved by setting all mechanisms of driven nodes (those in $A$) equal (just like in the case of spatial homogeneity, see \cref{def:homogeneity}, but only among the driven nodes) e.g.\ let for simplicity $\pa(j,t+1)=(j,t)$, $\epsilon \in [0,1]$, and set for all $j \in A$ and $x_{B,t+1}$ 
  \begin{equation}
    p_{j,t+1}(x_{j,t+1}|x_{B,t+1},x_{j,t}) = \begin{cases}
                                               \epsilon & \text{ if } x_{j,t+1} = x_{B,t+1} \\
                                               1-\epsilon & \text{ if } x_{j,t+1} = x_{j,t}. \\
%
                                             \end{cases}
  \end{equation} 
  Then, if less (more) than half of the variables in $A$ have value $1$, each variable is switched to $1$ ($0$) with probability $\epsilon$ ($1- \epsilon$) and else stays the same. This is in effect similar to a thermostat keeping the amount of ones among the nodes in $A$ around $|A|/2$.
\end{itemize}

\chapter{Agents within Markov chains}
\label{ch:agents}
This chapter constitutes the conceptual part of this thesis. We present here an avenue for a fully formal definition of agents. Taking our cue from the literature, we extract a list of notions that, if formally defined, would suffice for a formal agent definition. We do not arrive at a full definition. For goal-directedness, which would complete such a definition we make no proposal. For the other notions we propose definitions and motivate these in each case. Finally, we connect the resulting proto-agents (lacking goal-directedness) to the existing formal model of agent-environment system called the perception-action loop.

%
%
%
In more detail the chapter contains the following.
\begin{itemize}
  \item In \cref{sec:agentdef} we give a working definition of agents in accordance with the literature, this stipulates agents as entities contained within systems, the entities must exhibit actions, perception, and goal-directedness. 
  \item In \cref{sec:mcchoice} we restrict ourselves to a class of target systems that we want to define agents for, i.e.\ (driven) multivariate Markov chains.
  \item In \cref{sec:entinmvmc} we present an arguments for three phenomena that should be possible (and not be precluded) under a general agent definition: compositionality, degree of freedom traversal, and counterfactual variation. We also propose spatiotemporal patterns (STPs) as structures that can exhibit these three features within multivariate Markov chains.
  \item In \cref{sec:identity} we highlight the problem of selecting entities among all STPs i.e.\ which we call the problem of identity. We also propose completely locally integrated STPs as one possible solution.
  \item In \cref{sec:actions,sec:perceptions} we present definitions of actions and perceptions that are suitable for any notion of entities that is based on STPs.
  \item In \cref{sec:palooprelation} we show that our proposed notions of action and perception can be seen as generalisations of existing notions in the perception-action loop literature.
\end{itemize}

The original contributions contained in this chapter are:
\begin{itemize}
  \item An argument (via compositionality, degree of freedom traversal, and counterfactual variation) for a STP-based definition of entities. 
\item The abstraction of entity-sets which enables the formal connection to perception-action loop.
\item A tentative\footnote{For some context on what we mean by ``tentative'' see \cref{ch:agents}.} formal definition of entities as completely locally integrated STPs.
  \item A tentative formal definition of action for arbitrary entity-sets.
  \item A classification of actions into value actions and extent actions.
  \item A tentative formal definition of perception for arbitrary entity-sets. 
  \item An exposition of the role of non-interpenetration of entity-sets in perception. Namely, it makes perception naturally unique.
  \item The formal exposition of the connection of the action definition to non-heteronomy of \citet{bertschinger_autonomy_2008} in the perception-action loop.
  \item The formal exposition of the way the perception definition specialises to the perception-action loop.
  \item A construction of a conditional probability distribution (the branch-morph, including branching partition) over the futures of entities which allows the definition of perception.
  \item Proof that the condition on co-perception environments is not stronger than the assumptions about environment states inherent in the perception-action loop.
\end{itemize}
The formal definitions of entities, actions, and perceptions are only tentative. We will establish in this thesis that they are compatible, i.e.\ the defined actions and perceptions apply to the defined entities. We also establish in \cref{sec:exampleconnection,sec:actperceptexamples} that they fulfil certain expectations and the requirements we propose in this thesis. They remain tentative for three reasons (in increasing generality): 
\begin{enumerate}
  \item There are still some open questions regarding their interpretation in certain cases. See \cref{sec:exampleconnection,sec:actperceptexamples,sec:exdiscussionoutlook}.
  \item There are further phenomena that should not be precluded by agent definitions that we have not investigated yet such as death, birth, growth, and replication.
  \item A final formal definition of agents should also be empirically grounded. For this it should correctly predict the conditions for the occurrence of agents in some system. For this it is in turn necessary that the occurring agents are more or less universally acknowledged examples of agents. The best examples of such agents are humans but some researchers are willing to attribute agency to much simpler living organisms like bacteria. In that case one can imagine that a formal definition of agents could in principle get empirical justification: Say we can formally model beakers filled with suitable chemicals and their dynamics. A formal agent definition would be empirically justified if it can be used to reliably predict the conditions (external influence and initial conditions) that lead to the occurrence of agents e.g.\ the emergence of a bacterium within the beaker. This scenario can (in principle only) also be extended to the case where only humans are agents. In practice this seems further magnitudes more unlikely than the scenario involving the bacterium.

  Another and possibly simpler path to recognition of an agent definition would be if it was able to predict which formal systems that can be simulated contain agents and where such simulations turn out to be ``convincing enough'' to the research community. Currently neither of these scenarios has been realised with our formal agent definition. Neither scenario is also likely to happen in the near future. A classification of formal systems according to their capacity to contain agents will require not only further mathematical tools but also the right guess with respect to the agent definition. This thesis presents one such guess and develops some associated new mathematical tools. Whether either proposal will be part of a final and acknowledged formal definition of agents is work for the future. 
\end{enumerate}

\section{A working definition of agents}
\label{sec:agentdef}
Conceptually, agent definitions are not particularly controversial in the literature. As \citet{barandiaran_defining_2009} have argued a rough consensus is that an agent is ``at least, \textit{a system doing something by itself according to certain goals or norms within a specific environment}'' \citep[p.2, italics from the original]{barandiaran_defining_2009}. In their subsequent discussion they highlight the necessity for a ``distinguished entity'' to exist in order to take the role of the ``system doing something''. For the moment this almost suffices for our purposes. 
It is not the main goal of this work to question the concept of agents but instead to contribute to its formalisation within a preselected class of formal/artificial systems. 

To get a concise working definition we insert the ``distinguished entity'' into the consensus definition above and reformulate it slightly:

\begin{mydef}[Agent working definition]
\label{def:agent}
An \textit{agent} is a distinguished entity contained within a strictly larger system exhibiting perception, action, and goal-directedness.
\end{mydef}

Instead of referring to a ``specific environment'' we only require a larger system that encompasses/contains the agent as well as something more. The ``something more'' can certainly take the form of an environment and, conversely, if there is an environment the ``larger system'' can always be defined as the agent together with its environment. If anything our definition is therefore more general than that of \citeauthor{barandiaran_defining_2009} though this is not the main purpose of this formulation. The main purpose is to highlight the existence of the larger system which \textit{contains} the agent entirely. It is important to remember that this existence is not a new requirement but is present already in the prevalent concept of agents via the reference to an ``environment''. 
Also note that when we require that the distinguished entity is contained within a larger system, we mean the entity cannot be \textit{everything that exists} for any amount of time. By ``larger'' we therefore mean the system is larger during and throughout the agent's existence/presence and not ``larger'' only in the temporal domain. 

It is worth highlighting two more things about how we interpret our working definition. The first is that an agent is a ``distinguished entity'' in so far that it exhibits actions, perception, and goal-directedness. These three properties \textit{distinguish} an entity that is an agent among all entities. This leads directly to the second point: the set of agents is a (set theoretical) \textit{subset} of the set of entities contained in the larger system. From this it follows that the set of entities has to be defined in such a way that it encompasses all agents. We will come back to this in \cref{sec:entinmvmc}.

That we choose to use ``actions'' instead of ``doing something by itself'' and ``goal-directedness'' instead of ``according to certain goals or norms'' should not be over-interpreted. We consider these expressions as interchangeable. We also included perception in the definition because it is (sometimes via ``interaction'') a very common requirement in the literature \citep[e.g.][]{maes_modeling_1993,beer_dynamical_1995,smithers_autonomous_1995,franklin_is_1997,christensen_autonomy_2000}. Furthermore, \citet{barandiaran_defining_2009} in their more detailed discussion of the requirements for agency refer to ``interactional asymmetry'' instead of ``doing something''. The use of interaction suggests that they also agree with a requirement of perceptions.  

We also want to draw attention to the fact that \cref{def:agent} is still a very weak definition of agents. Especially when discussing biological agents i.e. living organisms, further requirements are common. One such requirement concerns the relation between the goals and the agent. In the definition above this relation is arbitrary, any goals are valid for any agent. Stronger definitions require the goals to be somewhat intrinsic to the agent. This can mean that goals must be in the agent's own interest \citep[e.g.][]{franklin_is_1997,kauffman_investigations_2000} for example ensure its survival/existence \citep[cf.\ constitutive autonomy e.g.][]{froese_enactive_2009}.

Primary examples of agents are living organisms, the higher organised they are the less controversial the claim that they are indeed agents. In the end it cannot be denied, for example, that humans are agents. Many authors agree that bacteria already qualify as (sometimes called minimal) agents \citep{christensen_autonomy_2000,kauffman_emergence_2006,froese_enactive_2009,barandiaran_defining_2009}. Due to the supposed lack of representational capabilities others do disagree \citep[see][for references]{sep-agency}. We take the point of view in line with our definition of agents above that bacteria and all living organisms are agents. For the purpose of this thesis it is not essential to make a final commitment on these matters. It would suffice for our arguments if we would use a more restrictive definition, e.g.\ that only humans are agents. Whenever we speak of living organisms in the following the inclined reader might then just replace this by ``humans'' and the same or similar arguments still hold. Apart from living organisms, other examples of agents are robots, and more controversially, societies, companies, and nation states.

\section{Multivariate Markov chains as a class of systems containing agents}
\label{sec:mcchoice}
In order to transform the working \cref{def:agent} into a formal definition we have to formally define every one of the terms mentioned there. The most fundamental term in \cref{def:agent} is the ``larger system''. The distinguished entities must be ``parts of'' the larger system so to define those we need to define the larger system first. Perception, action, and goal-directedness can then be defined once we have well defined distinguished entities. 

As the class of larger systems we choose finite multivariate Markov chains (see \cref{def:mvmarkov,def:mvdrivenchain} as well as \cref{fig:multimarkovchain2,fig:drivenmultimc2}).

\begin{figure}
\begin{center}
  \begin{center}
\begin{tikzpicture}[transform shape,->,>=stealth,shorten >=2pt,auto,thick]
    \pgfmathtruncatemacro{\cols}{4}
    \pgfmathtruncatemacro{\firstcol}{0}
    \pgfmathtruncatemacro{\rows}{5}
    \pgfmathtruncatemacro{\fadingcols}{1}
    \pgfmathsetlengthmacro{\vdist}{1cm}
    \pgfmathsetlengthmacro{\hdist}{.2\textwidth}

    \pgfmathtruncatemacro{\secondcol}{\firstcol+1}
    \pgfmathtruncatemacro{\fullcols}{\cols-\fadingcols}
        
    \node (N-1-\firstcol) [] {$X_{1,0}$};

    \foreach \y in {\secondcol,...,\fullcols}{%
       \pgfmathtruncatemacro{\yonleft}{\y-1}
       \node (N-1-\y) [node distance=\hdist,right of=N-1-\yonleft] {$X_{1,\y}$};
    }
  \foreach \x in {2,...,\rows}{%
     \pgfmathtruncatemacro{\xabove}{\x-1}
     \foreach \y in {\firstcol,...,\fullcols}{%
       \node (N-\x-\y) [below of=N-\xabove-\y] {$X_{\x,\y}$};
     }
  } 
  
  \node (space) [node distance=1cm, left of=N-2-\firstcol, rotate=270] {degrees of freedom (DOFs) $\rightarrow$};

  \node (time) [node distance=1cm, below of=N-\rows-\firstcol, xshift=-.5cm] {time $\rightarrow$};

  \pgfmathtruncatemacro{\rowsminusone}{\rows-1}
  \foreach \x in {2,...,\rowsminusone}{%
    \foreach \y in {\secondcol,...,\fullcols}{%
	\pgfmathtruncatemacro{\yonleft}{\y-1}
	\pgfmathtruncatemacro{\xabove}{(Mod(\x-1-1,\rows)+1)}
	\pgfmathtruncatemacro{\xbelow}{Mod(\x-1+1,\rows)+1}
        \draw (N-\xabove-\yonleft) edge (N-\x-\y);
        \draw (N-\x-\yonleft) edge (N-\x-\y);
        \draw (N-\xbelow-\yonleft) edge (N-\x-\y);
    }
  }
  \pgfmathtruncatemacro{\x}{1}
  \foreach \y in {\secondcol,...,\fullcols}{%
	\pgfmathtruncatemacro{\yonleft}{\y-1}
	\pgfmathtruncatemacro{\xbelow}{2}
        \draw (N-\x-\yonleft) edge (N-\x-\y);
        \draw (N-\xbelow-\yonleft) edge (N-\x-\y);
  }
  \pgfmathtruncatemacro{\x}{\rows}
  \foreach \y in {\secondcol,...,\fullcols}{%
    \pgfmathtruncatemacro{\yonleft}{\y-1}
    \pgfmathtruncatemacro{\xabove}{4}
    \draw (N-\xabove-\yonleft) edge (N-\x-\y);
    \draw (N-\x-\yonleft) edge (N-\x-\y);
  }

    \foreach \y in {\cols}{%
       \pgfmathtruncatemacro{\yonleft}{\y-1}
       \node (N-1-\y) [node distance=\hdist,right of=N-1-\yonleft] {};
    }
  \foreach \x in {2,...,\rows}{%
     \pgfmathtruncatemacro{\xabove}{\x-1}
     \foreach \y in {\cols}{%
       \node (N-\x-\y) [below of=N-\xabove-\y] {};
     }
  }

  \pgfmathtruncatemacro{\rowsminusone}{\rows-1}
  \foreach \x in {2,...,\rowsminusone}{%
    \foreach \y in {\cols}{%
	\pgfmathtruncatemacro{\yonleft}{\y-1}
	\pgfmathtruncatemacro{\xabove}{(Mod(\x-1-1,\rows)+1)}
	\pgfmathtruncatemacro{\xbelow}{Mod(\x-1+1,\rows)+1}
        \draw (N-\xabove-\yonleft) edge[dotted,-] (N-\x-\y);
        \draw (N-\x-\yonleft) edge[dotted,-] (N-\x-\y);
        \draw (N-\xbelow-\yonleft) edge[dotted,-] (N-\x-\y);
    }
  }
  \pgfmathtruncatemacro{\x}{1}
  \foreach \y in {\cols}{%
	\pgfmathtruncatemacro{\yonleft}{\y-1}
	\pgfmathtruncatemacro{\xbelow}{2}
        \draw (N-\x-\yonleft) edge[dotted,-] (N-\x-\y);
        \draw (N-\xbelow-\yonleft) edge[dotted,-] (N-\x-\y);
  }
  \pgfmathtruncatemacro{\x}{\rows}
  \foreach \y in {\cols}{%
    \pgfmathtruncatemacro{\yonleft}{\y-1}
    \pgfmathtruncatemacro{\xabove}{4}
    \draw (N-\xabove-\yonleft) edge[dotted,-] (N-\x-\y);
    \draw (N-\x-\yonleft) edge[dotted,-] (N-\x-\y);
  }
  



\end{tikzpicture}
\end{center}
  \caption{First time steps of the Bayesian network representing a multivariate Markov chain $\Xt$. The shown edges are just an example, any two nodes within the same or subsequent columns can be connected.}
  \label{fig:multimarkovchain2}
\end{center}
\end{figure}
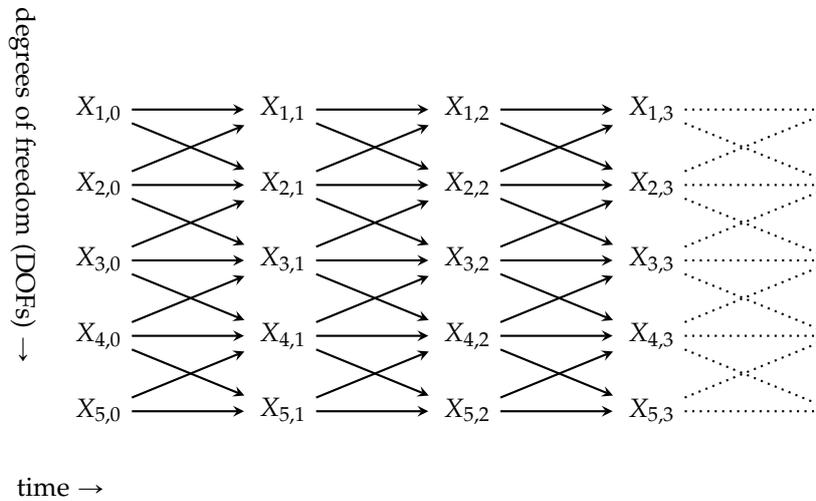

\begin{figure}
\begin{center}
  \begin{center}
\begin{tikzpicture}[transform shape,->,>=stealth,shorten >=2pt,auto,thick]
    \pgfmathtruncatemacro{\cols}{4}
    \pgfmathtruncatemacro{\firstcol}{0}
    \pgfmathtruncatemacro{\rows}{3}
    \pgfmathtruncatemacro{\drivingrows}{1}
    \pgfmathtruncatemacro{\fadingcols}{1}
    \pgfmathsetlengthmacro{\vdist}{1cm}
    \pgfmathsetlengthmacro{\hdist}{.2\textwidth}
    \pgfmathsetlengthmacro{\drivedist}{.1\textwidth}

    \pgfmathtruncatemacro{\secondcol}{\firstcol+1}
    \pgfmathtruncatemacro{\fullcols}{\cols-\fadingcols}
    \pgfmathtruncatemacro{\firstdrivingrow}{\rows+1}
    \pgfmathtruncatemacro{\totalrows}{\rows+\drivingrows}
    
    \node (N-1-\firstcol) [] {$X_{1,0}$};

    \foreach \y in {\secondcol,...,\fullcols}{%
       \pgfmathtruncatemacro{\yonleft}{\y-1}
       \node (N-1-\y) [node distance=\hdist,right of=N-1-\yonleft] {$X_{1,\y}$};
    }

  \foreach \x in {2,...,\rows}{%
     \pgfmathtruncatemacro{\xabove}{\x-1}
     \foreach \y in {\firstcol,...,\fullcols}{%
       \node (N-\x-\y) [below of=N-\xabove-\y] {$X_{\x,\y}$};
     }
  } 

    \pgfmathtruncatemacro{\firstdriverow}{\rows+1}
    \node (N-\firstdriverow-\firstcol) [node distance=\drivedist, below of=N-\rows-\firstcol] {$Y_{0}$};

    \foreach \y in {\secondcol,...,\fullcols}{%
       \pgfmathtruncatemacro{\yonleft}{\y-1}
       \node (N-\firstdriverow-\y) [node distance=\hdist,right of=N-\firstdriverow-\yonleft] {$Y_{\y}$};
    }
  

  \node (space) [node distance=1cm, left of=N-2-\firstcol, rotate=270] {degrees of freedom (DOFs) $\rightarrow$};

  \node (time) [node distance=1cm, below of=N-\totalrows-\firstcol, xshift=-.5cm] {time $\rightarrow$};

  \pgfmathtruncatemacro{\rowsminusone}{\rows-1}
  \foreach \x in {2,...,\rowsminusone}{%
    \foreach \y in {\secondcol,...,\fullcols}{%
	\pgfmathtruncatemacro{\yonleft}{\y-1}
	\pgfmathtruncatemacro{\xabove}{(Mod(\x-1-1,\rows)+1)}
	\pgfmathtruncatemacro{\xbelow}{Mod(\x-1+1,\rows)+1}
        \draw (N-\xabove-\yonleft) edge (N-\x-\y);
        \draw (N-\x-\yonleft) edge (N-\x-\y);
        \draw (N-\xbelow-\yonleft) edge (N-\x-\y);
    }
  }
  \pgfmathtruncatemacro{\x}{1}
  \foreach \y in {\secondcol,...,\fullcols}{%
	\pgfmathtruncatemacro{\yonleft}{\y-1}
	\pgfmathtruncatemacro{\xbelow}{2}
        \draw (N-\x-\yonleft) edge (N-\x-\y);
        \draw (N-\xbelow-\yonleft) edge (N-\x-\y);
  }
  \pgfmathtruncatemacro{\x}{\rows}
  \foreach \y in {\secondcol,...,\fullcols}{%
    \pgfmathtruncatemacro{\yonleft}{\y-1}
    \pgfmathtruncatemacro{\xabove}{4}
    \draw (N-\xabove-\yonleft) edge (N-\x-\y);
    \draw (N-\x-\yonleft) edge (N-\x-\y);
  }
  
  \pgfmathtruncatemacro{\rowsminusone}{\rows-1}
  \foreach \y in {\secondcol,...,\fullcols}{%
	\pgfmathtruncatemacro{\yonleft}{\y-1} 
	\pgfmathtruncatemacro{\yonright}{\y+1}
	\draw (N-\firstdrivingrow-\yonleft) edge (N-\firstdrivingrow-\y);
    \foreach \x in {1,...,\rows}{%
%
        \draw (N-\x-\yonleft) edge (N-\firstdrivingrow-\y);
        \draw (N-\firstdrivingrow-\yonleft) edge[bend right=45] (N-\x-\yonleft);
        
    }
  }

    \foreach \y in {\cols}{%
       \pgfmathtruncatemacro{\yonleft}{\y-1}
       \node (N-1-\y) [node distance=\hdist,right of=N-1-\yonleft] {};
    }
  \foreach \x in {2,...,\rows}{%
     \pgfmathtruncatemacro{\xabove}{\x-1}
     \foreach \y in {\cols}{%
       \node (N-\x-\y) [below of=N-\xabove-\y] {};
     }
  } 
  \node (N-\firstdrivingrow-\cols) [node distance=\drivedist, below of=N-\rows-\cols] {};
  
    	\pgfmathtruncatemacro{\yonleft}{\cols-1} 
  	\pgfmathtruncatemacro{\y}{\cols} 
	\draw (N-\firstdrivingrow-\yonleft) edge[dotted,-] (N-\firstdrivingrow-\y);
    \foreach \x in {1,...,\rows}{%
%
        \draw (N-\x-\yonleft) edge[dotted,-] (N-\firstdrivingrow-\y);
        \draw (N-\firstdrivingrow-\yonleft) edge[bend right=45] (N-\x-\yonleft);
        
    }

  \pgfmathtruncatemacro{\rowsminusone}{\rows-1}
  \foreach \x in {2,...,\rowsminusone}{%
    \foreach \y in {\cols}{%
	\pgfmathtruncatemacro{\yonleft}{\y-1}
	\pgfmathtruncatemacro{\xabove}{(Mod(\x-1-1,\rows)+1)}
	\pgfmathtruncatemacro{\xbelow}{Mod(\x-1+1,\rows)+1}
        \draw (N-\xabove-\yonleft) edge[dotted,-] (N-\x-\y);
        \draw (N-\x-\yonleft) edge[dotted,-] (N-\x-\y);
        \draw (N-\xbelow-\yonleft) edge[dotted,-] (N-\x-\y);
    }
  }
  \pgfmathtruncatemacro{\x}{1}
  \foreach \y in {\cols}{%
	\pgfmathtruncatemacro{\yonleft}{\y-1}
	\pgfmathtruncatemacro{\xbelow}{2}
        \draw (N-\x-\yonleft) edge[dotted,-] (N-\x-\y);
        \draw (N-\xbelow-\yonleft) edge[dotted,-] (N-\x-\y);
  }
  \pgfmathtruncatemacro{\x}{\rows}
  \foreach \y in {\cols}{%
    \pgfmathtruncatemacro{\yonleft}{\y-1}
    \pgfmathtruncatemacro{\xabove}{4}
    \draw (N-\xabove-\yonleft) edge[dotted,-] (N-\x-\y);
    \draw (N-\x-\yonleft) edge[dotted,-] (N-\x-\y);
  }
  



\end{tikzpicture}
\end{center}
  \caption{First time steps of the Bayesian network representing a multivariate process $\Xv$ driven by a process $\Yt$. Note that the process $\Yt$ can also be multivariate, but this would further clutter the graph. Also note that not all edges depicted here must be present. Here, each random variable in each time-slice of the driven process is influenced by the driving process and influences it.  
  }
  \label{fig:drivenmultimc2}
\end{center}
\end{figure}

One immediate consequence of choosing a well defined class of systems is that it forces us to construct all other notions from those well defined for this class of systems. This has the advantage that it greatly restricts the concepts to consider for the definitions. The disadvantage is that if the choice is a bad choice we are destined to fail. A bad choice here would mean that a useful notion of agents is impossible within our choice of larger systems. In this section we therefore explain what motivates our choice of finite Markov chains as the ``larger systems''.


%

Since we have required that agents are parts of larger systems in \cref{def:agent} this system must in one way or another \textit{contain} the agent candidates. As a choice for the class of systems that represent the larger systems we should then use systems for which it is plausible that they can contain agents. At the same time want to start with the \textit{simplest} class of systems that shows at least some promise or is not easily dismissed. 

This desire for simplicity is due to two factors. First, in artificial life we are mainly interested in the principles that allow the occurrence of agents/life within a system and not in the precise description of actual agents/living organisms. Second, choosing simpler systems greatly reduces the technical burden so that the concepts play a more prominent role. 

The main factor in choosing a finite system is that implementations of the systems under consideration in computer simulations are of great interest in artificial life. Such simulations are restricted to discrete and finite systems. Continuous systems can be approximated, but the approximations are in the end finite again so such approximations are included in the class of finite systems. 

As living organisms are our prime example of agents it would be straightforward to choose systems which resemble or model systems that contain living organisms. The safest bet is then to use models of the universe as a whole. By definition the universe contains living organisms entirely. 

Another reasonably safe bet is the entire geosphere, by which we mean the planet earth together with its atmosphere and the exchange of radiation with the sun and the rest of the universe. Similarly there are smaller subsystems of the geosphere that can contain living organisms like ponds, tidal pools, and other ecosystems.


Realistic models of the universe or the geosphere are continuous, use quantum mechanics \citep[e.g. as in][]{saitta_miller_2014}, relativistic mechanics, or even more involved theories. 


%
However, for the sake of simplicity and finiteness we abstract away from the more realistic continuous, relativistic, or quantum dynamical systems to finite multivariate Markov chains \cref{def:mvmarkov,def:mvdrivenchain}. 

This class of systems contains synchronous finite cellular automata like the game of life \citep{conway_game_1970}. These automata can be seen as discretised versions of field theories \citep{shalizi_what_2003} and have successfully been used to model physical systems \cite{chopard_cellular_2009}. 

Multivariate Markov chains can also be used to approximate particle-based systems if we use the random variables to represent the positions and momenta of the particles. An interesting recent system with life-like behaviour which falls into this class is \cite{schmickl_how_2016}. 

Since we also include driven multivariate Markov chains our considerations also extend to reaction-diffusion systems. Such systems are also frequently used to model biological phenomena \citep{turing_chemical_1952}, as well as individualised and metabolising structures \citep{virgo_thermodynamics_2011,froese_motility_2014,bartlett_emergence_2015,bartlett_precarious_2016}. 

Last but not least, the driven multivariate Markov chains can be used to approximate/simulate systems obeying the (multivariate) Langevin equation (as a discretised version of the associated Fokker-Planck equation). Such systems underlie recent investigations into the physics of cell replication \citep{england_statistical_2013} and adaptation \citep{perunov_statistical_2014} as well as a theory of life \citep{friston_life_2013}. They are also used in \citet{still_thermodynamics_2012} to study advantages of prediction, perception and action for thermodynamic efficiency. Similarly \citet{sagawa_thermodynamics_2012} studies the increased work extraction due to feedback control, which can be seen as perception and action as well. \citet{kondepudi_end-directed_2015} studies a form of goal-directedness in a driven system. 

Therefore we believe that multivariate (driven) Markov chains are a reasonable choice for a first class of systems to develop an agent definition for. Our hope is that these system are powerful enough to contain agents. However, if we find that they are not then we may at least find out why they are not. At the current state of this research this question is still undecided.

\section{Entities in multivariate Markov chains that can be agents}
\label{sec:entinmvmc}
As mentioned in \cref{sec:agentdef} all agents are entities. After choosing multivariate Markov chains as the class of containing systems we have to define entities in this class of systems. For this purpose we propose to use subsets of STPs in general and completely integrated STPs (\cref{def:ci}) in particular. Employing STPs to represent entities is already implicit in \citep{beer_cognitive_2014,beer_characterizing_2014}. The notion of completely integrated STPs and the proposal of using them as entities in an agent definition are two of the main original contributions of this thesis. 
The section is loosely based on our own publication \citet{biehl_towards_2016}. 

Formally, the first goal of this section is to establish that the set of entities $\Ent(\Xv)$ for any given (driven) multivariate Markov chain\footnote{We will not explicitly mention ``driven'' in the following. We will also refer to the Markov chain $\Xv$ without explicitly mentioning driving or driven random variables. The process $\Xv$ should be seen as the process of interest that may or may not be driven by some other process whose dynamics are ignored. For the purpose of this chapter whether $\Xv$ is driven or not makes no difference.}  $\Xv$ should be a subset of the STPs of $\Xv$ i.e.\
\begin{equation}
\label{eq:entset1}
  \Ent(\Xv) \subseteq \bigcup_{O \subseteq V} \X_O.
\end{equation} 
Using subsets of STPs is in contrast (i.e.\ \textit{not} equivalent) to using subsets of random variables i.e.\ 
\begin{equation}
  \Ent(\Xv) \subseteq 2^V
\end{equation} 
as entities. The latter are often implicitly used in the literature. The arguments for \cref{eq:entset1} are to a large degree independent of arguments that concern the exact determination of \textit{which} subset $\Ent(\Xv)$ should correspond to. Consequently, there may be different notions of agents based on different choices of the exact subset. In order to accommodate this we introduce the notion of \textit{entity sets} which later allow us to define actions and perception independent of the exact choice of $\Ent(\Xv)$. The exact choice is the subject of the second part of this section and the problem of identity. There we will motivate the choice of completely integrated STPs 
\begin{equation}
  \Ent(\Xv)= \{x_O \in \bigcup_{O \subseteq V} \X_O : \ci(x_O)>0\}.
\end{equation} 

Note that this section does not contain a rigorous derivation of the necessity to choose the entities as we propose. We merely present heuristic arguments which speak for this choice. The main tool in this endeavour is the following argument.

As already mentioned in \cref{sec:agentdef} the set of entities for a given larger system has to (at least) encompass all agents within the system. In other words the definition of entities must not exclude structures which might be agents. Now say that there are phenomena or properties that are known to be exhibited by some (possibly not all) agents. Say that  furthermore there is an entity definition which implies that these phenomena or properties are \textit{impossible} for entities. Then we must reject this entity definition on the grounds that it cannot encompass all agents since it \textit{precludes} these phenomena or properties. 
In the following this argument will be employed multiple times and referred to as the \textit{non-preclusion argument}. Note that we cannot require all phenomena that are exhibited by some agents to be exhibited by all entities of an entity definition. This would lead to a small and possibly empty subset of agents. We can however require that all phenomena that are exhibited by some agents are not-precluded by the entity definition. In this way every phenomenon that is exhibited by some agent can be turned into a condition on entity definitions.

%

We illustrate our arguments for choosing entities within the class of multivariate Markov chains using the popular example of a glider in the game of life cellular automaton. The glider is not necessarily a life-like structure, but it already exhibits the three phenomena that we will further discuss in this thesis: 
\begin{enumerate}
  \item compositionality,
  \item degree of freedom traversal,
  \item counterfactual variation. 
\end{enumerate}
These phenomena are also exhibited by more life-like structures in less well known examples of (driven) multivariate Markov chains. We refer the reader to the motile and interacting reaction-diffusion spots in \citet{virgo_thermodynamics_2011,froese_motility_2014}, different reaction-diffusion spots in \citet{bartlett_emergence_2015,bartlett_precarious_2016}, and the particle-based cell-like structures in \citet{schmickl_how_2016}. In the following when we refer to ``other life-like structures'' we refer to these examples. 

Note that there are further phenomena of living organisms and life-like systems that should not be precluded by an entity definition. Examples of such phenomena are birth, death, growth, and replication. The investigation of these is beyond the scope of this thesis.

In \cref{sec:compo,sec:cfvar,sec:doftraversal} we discuss each of the three phenomena above separately. For each we will also note that they seem plausible for real living organisms. We then invoke in each case the non-preclusion argument and require that a definition of entities in multivariate Markov chains should allow structures that exhibit this phenomenon.  In the course of these arguments we settle for STPs as the superset of entities and state this explicitly in \cref{sec:entdef}. There we will also define compositionality, degree of freedom traversal, and counterfactual variation formally. This leaves open the problem of selecting entities among all STPs which is the problem of identity discussed in \cref{sec:identity}.


\subsection{Compositionality of entities}
\label{sec:compo}
The compositionality of entities refers to the possibility that life-like structures are composite of multiple parts. This can be separated into two kinds of compositionality, spatial compositionality and temporal compositionality. We first discuss spatial compositionality.

In the example of the glider we observe that a glider is not just a single cell or the state of a single cell (e.g.\ black or white). In order for a glider to occur multiple cells that are in a particular arrangement have to have particular states at some time-step. The glider is therefore a (spatially) composite structure. The same is true for other life-like structures. 

In reaction-diffusion systems \citep[e.g.][]{froese_motility_2014,bartlett_emergence_2015} the individualised spots occupy a contiguous bounded region in a two dimensional plane. Each position in the reaction-diffusion system is a random variable that indicates the concentrations of the involved chemicals at this position. A single position or a single set of concentrations at a position does not constitute a spot or life-like structure these are composite of all the concentrations in an area. So reaction-diffusion spots are also spatially composite structures.

In the particle-based system of \citet{schmickl_how_2016} the life-like structures are spores or cells. These are composite of particles. In this case each particle $j$ has three degrees of freedom, two positions $x_{j,t},y_{j,t}$ and heading $\phi_{j,t}$. Each degree of freedom can represented by a random variable. The union of these degrees of freedom over all particles and all times form the random variables of the multivariate Markov chain. A spore or cell occurs if these random variables stand in particular relation to each other. Without going into further details, a necessary condition is that the positions of multiple particles must be (in some sense) close to each other. A single particle is not a cell or a spore so again we find that life-like structures are spatially composite structures. 

Finally, living organisms (presumably composite of molecules) are also generally seen as composite structures. 

So by the non-preclusion argument the entities in multivariate Markov chains should include spatially composite structures. 

The glider is also a temporally composite structure. It is an essential feature of a glider that it ``moves'' which means that in general it can exist at multiple time-steps and we refer to it as the same glider. This means the glider can persist or can be composite out of parts at different time-steps.

Similarly, the life-like structures in other systems are persistent structures. Furthermore living organisms persist and are generally seen to have histories which is another indication that they are composite out of parts at different times. In \cref{sec:doftraversal} we will discuss temporal compositionality with particular attention to the possibility that the spatial parts that the glider or many life-like structures are composite of change over time. We therefore keep the discussion of temporal compositionality short.



Note that both choices of entities $\Ent(\Xv) \subseteq 2^V$ and $\Ent(\Xv) \subseteq \bigcup_{O \subseteq V} \X_O$ can represent composite structures in the form of sets of random variables or sets of values of random variables respectively.

\subsection{Degree of freedom traversal of entities}
\label{sec:doftraversal}
Degree of freedom traversal refers to the possibility that life-like structures maintain a form of identity while exchanging the spatial parts they are made of. 

The glider in the game of life ``moves'' in one of four possible directions. As it moves the cells that it occupies change at every time step. Nonetheless we speak of the \textit{same} glider even when none of the cells it occupied in one configuration are still occupied several time steps later. Intuitively then the glider maintains its identity along its path. We call the maintenance of identity under exchange of the spatial occupied cells or spatial occupied random variables \textit{degree of freedom traversal}. Note that a block in the game of life does not exhibit such degree of freedom traversal.

Other life-like structures in multivariate Markov chains also exhibit degree of freedom traversal. In reaction-diffusion systems \citep[e.g.][]{froese_motility_2014,bartlett_emergence_2015} the individualised spots travel through the two dimensional plane and thereby occupy changing spatial regions over time. As mentioned before, each position in a reaction diffusion system indicates a random variable representing the concentrations of the involved chemicals at this position. So reaction-diffusion spots also exhibit degree of freedom traversal.

In the particle-based system of \citet{schmickl_how_2016} particles can be seen to jump into and out of the spores and cells which intuitively maintain their identity throughout. As mentioned before each particle corresponds to three degrees of freedom/random variables. The ``spatial''
random variables that are occupied by a spore or cell are those of the particles that it is formed by. This means that when a particle jumps in or out of the spore or cell the random variables occupied by this structure change. Therefore these structures also exhibit degree of freedom traversal.  

Furthermore, we also see real living organisms as maintaining their identity while exchanging the parts they are made of. The molecules that a cell is made of change during its lifetime. So if a cell (or a larger living organism) is seen as a particular configuration of molecules then these also exchange the parts they are made of. 

Together these observations suggest that degree of freedom traversal is exhibited by life-like structures in Markov chains. According to the non-preclusion argument a definition of entities should therefore allow the possibility of entities that traverse degrees of freedom. 

Both candidates for entities mentioned in \cref{sec:compo} can represent degree of freedom traversal. The subsets of random variables $\Ent(\Xv) \subseteq 2^V$ also contain sets of random variables that differ from time-step to time-step. This is possible because we use the ``time-unrolled'' Bayesian network formulation of multivariate Markov chains where each time step has its own set of random variables $\{X_i\}_{i \in V_t}$ for time-slice $V_t$. A subset of $V$ can then combine any subsets of the time-slices at different times. 

The subsets of STPs $\Ent(\Xv) \subseteq \bigcup_{O \subseteq V} \X_O$ clearly also contain STPs that fix different random variables at different times since they can fix the random variables in any subset of $V$. 

We note here again that the choice of entity sets $\Ent(\Xv) \subseteq 2^V$ was used in \citet{krakauer_information_2014} to construct a notion of what is called individuals in this work. This work also deals with degree or freedom traversal of these individuals. However, the phenomenon of life-like structures discussed in \cref{sec:cfvar} suggests that this construction precludes certain structures that we would like to include. This is also the case for the perception-action loop (\cref{sec:paloopformal}) where the agent is represented by the set of random variables $\Mt$.


%

\subsection{Counterfactual variation of entities}
\label{sec:cfvar}
Counterfactual variation of structures refers to the possibility that the life-like structures within a multivariate Markov chain differ from one \textit{trajectory} to another. The difference or variation can take two forms:
\begin{enumerate}
  \item variation in value,
  \item variation in extent.
\end{enumerate}
We could add ``variation in existence'' but this is a special case of variation in extent. Before we give a formal definition of counterfactual variation we can already state a few observations about gliders. First, a glider can occur in one trajectory of the cellular automaton and not occur in a second one (e.g.\ if all cells are white in the second one). This would be a variation in existence. It can be seen as a variation in extent since its extent in the second trajectory is zero. 

Second, a glider can occur in one place and move in one direction in the course of one trajectory and occur in another place and move in another direction in the course of a second trajectory. This is also a variation in extent if the cells that the two gliders occupy in the two trajectories along their path do not completely coincide. 

\begin{figure}
\begin{center}
  \input{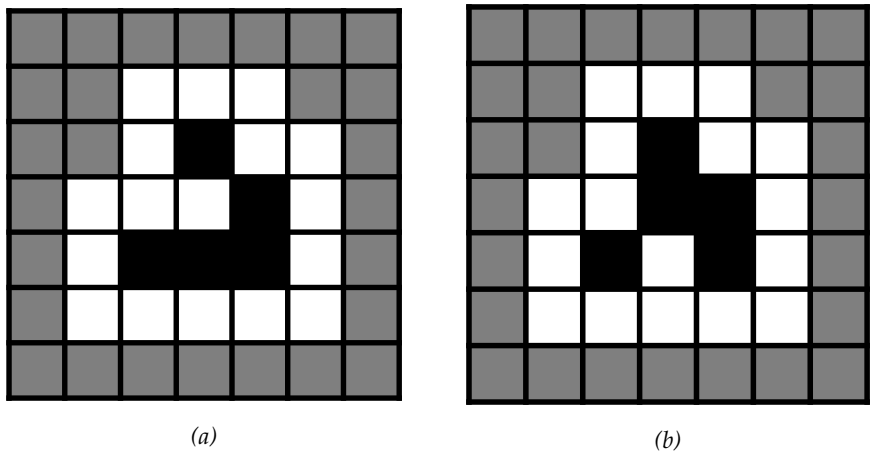}
  \caption{Counterfactual variation in value exhibited by the spatial patterns of two gliders in two different trajectories of a game of life cellular automaton. Both grids depict a single time-step of the two dimensional cellular automaton and therefore a special case of STPs without temporal extension. The cells that are not occupied by the patterns are grey, occupied cells are white or black according to the glider configuration. The extra layer of white cells around the black cells of the gliders is in accordance with the extent of gliders as derived by \citet{beer_cognitive_2014}. The two particular configurations shown here are also presented there. We see that the cells that are occupied coincide in both cases. The only difference between the two glider configurations are the values of two cells. The one right in the centre of the grid and the cell just below it have switched their values. This shows that there can be a glider in one trajectory and a different glider in another trajectory with both having identical extent. This is a counterfactual variation in value.}
  \label{fig:glidersvaluevariation}
\end{center}
\end{figure}

Third, a variation in value occurs if in two trajectories the structures/gliders occupy exactly the same cells but the cells they occupy have differing values. For gliders this can happen if we look at trajectories that are only one time-step long \cref{fig:glidersvaluevariation}. 

Note that if two glider in two trajectories have the same values and extent we consider it them the same glider.

%

Note that it is a non-trivial question which cells/random variables the glider actually occupies. A detailed discussion can be found in \citet{beer_cognitive_2014}. The example of counterfactual variation in value in \cref{fig:glidersvaluevariation} is based on the characterisation of a glider as a STP by \citeauthor{beer_cognitive_2014}.

We have seen that gliders are structures that exhibit counterfactual variation of the two kinds. 
For real living organisms it is impossible to say with certainty whether they exhibit counterfactual variation. We have no access to counterfactual trajectories of the universe. However, due to the symmetries of the laws of physics it is quite plausible that living organisms also exhibit counterfactual variation. Assume we have two identical aquarium containing a different fish each. The laws of physics suggest that the fish in one aquarium could also be in the same place in the other aquarium and similarly for the other fish. This would be counterfactual variation in extent of the fish within either aquarium. 

Together with evidence from other life-like structures we therefore consider it justified to invoke the non-preclusion argument and require that counterfactual variation is possible for entities within multivariate Markov chains. Therefore it should be possible for entities to occur within single trajectories and not occur in others. This suggests STPs as candidates for entities i.e.\ $\Ent(\Xv) \subseteq \bigcup_{A \subseteq V} \X_A$. 


Conversely, this throws out the possibility that entities are \textit{by definition} structures that do not vary with the trajectories. In particular this excludes the possibility to define entities as random variables, sets of random variables, or stochastic processes i.e.\ $\Ent(\Xv) \nsubseteq 2^V$ since these do not vary with the trajectories. A subset of random variables $X_A$ where $A \subseteq V$ is not dependent on the trajectories, it only takes different \textit{values} in different trajectories. The different values are just STPs however. Note that this does not mean that it is not justified in \textit{particular} cases to represent entities by sets of random variables. In particular, if it happens that every STP $x_A \in \X_A$ is an entity then it could make sense to speak of these STPs as different realisations of one thing / an individual / an ``entity'' \footnote{The quotations only indicate that we will reserve the term \textit{entity} for STP based entities in the rest of this thesis.}. For example, this is the assumption for the agent process $\Mt$ in the perception-action loop (see \cref{sec:paloopformal}) and will be discussed further in \cref{sec:perceptions,sec:palooprelation}.

We then settle for STPs as the candidates for entities i.e.\ entities are seen as special cases of STPs. For a given multivariate Markov chain entities are therefore a subset of the STPs. We fix this assumption formally in the next section where we also give a formal definition of compositionality, degree of freedom traversal, and counterfactual variation.


\subsection{Definition of entity sets via STPs}
\label{sec:entdef}
In accordance with \cref{sec:cfvar} we define the set of entities in a multivariate Markov chains to be a subset of the STPs. We call this subset the entity set.

\begin{mydef}[Entity set]
\label{def:entset}
 Given a multivariate Markov chain $\Xv$ with index set $V=J \times T$ an \textit{entity set} $\Ent(\Xv)$ is a subset of the set of all STPs $\bigcup_{A \subseteq V} \X_A$ i.e.\
 \begin{equation}
   \Ent(\Xv) \subseteq \bigcup_{A \subseteq V} \X_A.
 \end{equation} 
\end{mydef}
\begin{itemize}
  \item We expect that it is useful to require certain algebraic properties from entity sets. However, this is beyond the scope of this thesis.
\end{itemize}


Using entity sets we can also define compositionality, degree of freedom traversal, and counterfactual variation formally. To connect this with the example of a glider recall that a trajectory $x_V \in \X_V$ of a multivariate Markov chain is a STP that extends throughout the entire Bayesian network. In other words a trajectory occupies all random variables in the Markov chain. The multivariate Markov chain describing a cellular automaton assigns each cell $j \in J$ (where $J$ is a two dimensional grid of cells) at each time step $t \in T$ a random variable $X_{j,t}$. The random variables in the multivariate Markov chain are then indexed by $V = J \times T$ and a trajectory is a STP $x_V = x_{J,T}$. Structures like the glider (and similar life-like structures) occupy/fix subsets of the random variables in the multivariate Markov chain. They can therefore be described by STPs.

We can then define spatial and temporal compositionality, first for STPs, then for entity sets.
\begin{mydef}[Composite STPs]
  Given a multivariate Markov chain $\Xv$ we say a STP $x_A$ is 
  \begin{thmlist}
    \item \textit{spatially composite} or \textit{extended} if it has a time-slice $x_{A_t}$ occupying more than one random variable i.e.\ if there exists $t$ with $|A_t| > 1$,
    \item \textit{temporally composite} or \textit{extended} if it has more than one non-empty time-slice i.e.\ if $|\{t \in T: A_t \neq \emptyset\}|>1$.
  \end{thmlist}
  If $x_A$ is spatially and temporally extended we say it is \textit{spatiotemporally composite} or \textit{extended}.
\end{mydef}

\begin{mydef}[Compositionality]
  Given a multivariate Markov chain $\Xv$ the entity set $\Ent(\Xv)$ satisfies compositionality if it contains a composite STP.
\end{mydef}

Similarly for degree of freedom traversal.

\begin{mydef}[Degree of freedom traversing STPs]
  Given a multivariate Markov chain $\Xv$ with index set $V=J \times T$ we say a STP $x_A$ traverses degrees of freedom if there are two time-slices that occupy random variables with different spatial indices i.e.\ if there exists $t,s \in T$ with $t \neq s$ such that
  \begin{equation}
    \{j \in J: (j,t) \in A_t \} \neq \{j \in J: (j,s) \in A_s \}. 
  \end{equation} 
\end{mydef}

\begin{mydef}[Degree of freedom traversing entity set]
  Given a multivariate Markov chain $\Xv$ the entity set $\Ent(\Xv)$ is degree of freedom traversing if it contains a degree of freedom traversing STP.
\end{mydef}
Remark:
\begin{itemize}
  \item The indices of a multivariate Markov chain may be renamed such that the property of degree of freedom traversal vanishes from the entity set. We accept this caveat here. If needed the notion of degree of freedom traversal can be strengthened by requiring that no such index renaming removes the property. Note that if there are two entities, one that doesn't traverse degrees of freedom and one that does and both occupy the same degree of freedom at some time $t$ then they make it impossible to rename the indices at all times where the two entities differ. This is the case for the two gliders of \cref{fig:glidersvaluevariation}. Also, since the renaming of indices is rarely practically done or considered in case of cellular automata (it leads to complex update rules) and other systems exhibiting life-like phenomena our simple notion of degree of freedom traversal is sufficient for the purpose of this thesis. 
\end{itemize}

Finally, we define counterfactual variation by first defining variation (or difference) in value and extent:

\begin{mydef}[Variation of STPs]
  Given a multivariate Markov chain $\Xv$. Two STPs $x_A, \bar{x}_B$ \textit{differ} or \textit{vary}
    \begin{thmlist}
    \item \textit{in value} if $A = B$ and there exists $i \in A$ with $x_i \neq \bar{x}_i$,
    \item \textit{in extent} if $A \neq B$.
    \item \textit{in value and extent} if $A \neq B$ and there exists $i \in A \cap B$ with $x_i \neq \bar{x}_i$.
  \end{thmlist}
  We just say $x_A, \bar{x}_B$ \textit{differ} or \textit{vary} if any of the above are true. Else we say they are \textit{identical} or \textit{equal}.
\end{mydef}
Remark:
\begin{itemize}
  \item We will encounter the difference in value and extent again when we define actions for agents based on STPs in \cref{sec:actions}. There we can distinguish between actions in value and actions in extent.
\end{itemize}

We then have counterfactual variation if the set of all entities in one trajectory differs from the set of entities in another.

\begin{mydef}[Counterfactually varying entity set]
\label{def:cfvar}
  Given a multivariate Markov chain $\Xv$ the entity set $\Ent(\Xv)$ exhibits 
  \begin{thmlist}
  \item \textit{counterfactual variation} if there are two trajectories $x_V,\bar{x}_V \in \X_V$ with $x_V \neq \bar{x}_V$ such that the set of entities that occur in each are not equal i.e.\
  \begin{align}
  \{\hat{x}_A \in \Ent(\Xv):\hat{x}_A=x_A\} &= \{\hat{x}_A \in \Ent(\Xv):\hat{x}_A=\bar{x}_A\}
  \end{align} 
  \item \textit{counterfactual variation in value only} if for any two trajectories $x_V,\bar{x}_V \in \X_V$ set of entities that occur in each only differ in value which means that they all occupy the same sets of random variables i.e.\
  \begin{align}
  \{A \subseteq V: \exists \hat{x}_A \in \Ent(\Xv),\hat{x}_A=x_A\} &= \{A \subseteq V: \exists \hat{x}_A \in \Ent(\Xv),\hat{x}_A=\bar{x}_A\}  
  \end{align}
  \item \textit{counterfactual variation in extent} if it exhibits counterfactual variation but \textit{not} counterfactual variation in value only.
\end{thmlist}
  \end{mydef}

In \cref{sec:exampleconnection} we will see that the entity set we propose in \cref{sec:cliasentities} exhibits all three of these phenomena. Next we turn our attention towards choosing the right entity set from among all STPs.
%
%


\subsection{The problem of identity}
\label{sec:identity}
\subsubsection{General considerations}
Roughly, the problem of identity is the problem of determining which structures within a system form a (possibly) composite entity and which structures don't. We have already mentioned the maintenance of identity that we attribute to gliders and other life-like systems in \cref{sec:cfvar}. Since we have now decided on entities as subsets of STPs we can now look at identity more closely.  

We can also state the problem of identity formally.
\begin{mydef}
  Given a multivariate Markov chain $\Xv$ the problem of identity is the problem of deciding on a particular entity set $\Ent(\Xv)$.
\end{mydef}
This immediately suggests the trivial solutions of choosing all STPs as entities:
\begin{equation}
  \Ent(\Xv)=\bigcup_{A \subseteq V} \X_A.
\end{equation} 
This trivial solution is akin to what is called ``unrestricted mereological composition'' in philosophy \cite{gallois_identity_2015}. Every combination of spatial, temporal, and STPs is an entity (or object in the case of unrestricted mereological composition). This includes all trajectories whether they are possible under the mechanisms of the multivariate Markov chain or not.

However, our intuition is that some STPs are more entity-like than others. We think that a glider or a block in the game of life is more of an entity than, for example, half of a glider together with half a block far away in the grid. We also think that a glider pattern at time $t$ and together with the subsequent glider pattern at time $t+1$ are more of an entity than the glider at time $t$ and the far away block at time $t+1$ taken together as one STP. 
Similarly, in the real world we also see differences between the degrees to which certain subsets of the world are entities. An animal's leg together with the rest of its body forms more of an entity than its leg together with part of a nearby (or indeed far away) tree trunk. Living organisms account for a large number of examples but also some non-biological structures at least seem like entities e.g.\ hurricanes, tornadoes, etc., (soap) bubbles, and maybe human created artefacts.

The question is then what makes some composite structures more ``entity-like'' than others. An answer would be that there is a special relation that holds between the parts of ``entity-like'' structures that does not hold (or holds to a lesser degree) for other structures. In the case of STPs in multivariate Markov chains we should then look for relations between the parts of the STPs. Here, the different parts of STPs are related (or unrelated) due to the dynamics of the system. These dynamics are defined via the mechanisms of the multivariate Markov chain and generate the probability distribution over the entire chain. In this sense the probability distribution over the chain contains all information about relations between STPs (and their parts since the parts are again STPs). This suggests using the probabilities to formulate a quantitative condition, relation, or measure that given a STPs tells us in how far it constitutes and entity.

\subsubsection{Completely locally integrated STPs as entities}
\label{sec:cliasentities}
There are certainly multiple candidates for such a measure. It is beyond the scope of this thesis to provide a systematic comparison between multiple such candidates. We only propose complete local integration as on instance of such a measure. The formal structure of complete local integration has been investigated in \cref{sec:slicli} and we provide examples of this structure in \cref{sec:exstp}. In future work we will investigate other identity measures and also plan to try more axiomatic approaches. A starting point for an axiomatic approach would be the questions raised in the philosophical discussion on location and serology \citep{sep-location-mereology}. One such question is whether entities should be allowed to overlap or interpenetrate\footnote{Note that in \cref{sec:perceptions} we will also see that non-interpenetration of entities allows a uniquely defined notion of perception. A uniquely defined notion of perception in the case of interpenetrating entities still eludes us.}. Choosing an answer could be turned into an axiom which might restrict the possible identity measures. However here we only present some intuitions behind the first candidate for an identity measure: complete local integration.

For this let us treat the problem of identity as a combination of
\begin{enumerate}
  \item spatial identity and
  \item temporal identity
\end{enumerate}
for the moment. In the end we will propose a solution which makes no distinction between these two aspects. We note here that conceiving of entities (or objects) as composite of spatial and temporal parts as we do in this thesis is referred to as four-dimensionalism or perdurantism in philosophical discussions \citep[see e.g.][]{sep-temporal-parts}. The opposing view holds that entities are spatial and endure or persist over time. This view is called endurantism. Here we will not go into the details of this discussion.

The main intuition behind complete local integration is that every part of an entity should make every other part more probable. 

This seems to hold for example for the spatial identity of living organisms. Isolated parts of living organisms are \textit{rare} compared to whole living organisms. For example it is rare to see only an arm without the rest of a human body attached compared to seeing an arm with the rest of a human body attached. The body seems to make the existence of the arm more probable and vice versa. This seems to hold for all living organisms but also for some non-living structures. The best example of a non-living structure we know of for which this is obvious are soap bubbles\footnote{We thank Eric Smith for pointing out this example.}. Half soap bubbles (or thirds, quarters,...) only ever exist for split seconds whereas entire soap bubbles can persist for up to minutes. Any part of a soap bubble seems to make the existence of the rest more probable. Similarly, parts of hurricanes or tornadoes are rare. 
So what about spatial parts of structures that are not so entity-like? As a crude approximation we can think about whether a  monkey's leg makes a part of a tree trunk more probable/common (than the part of the tree trunk is by itself). In fact there might be a slight increase of the probability of a part of a tree trunk in the universe if there is a monkey's leg in the universe. However, surely the monkey's leg has much more positive influence on the probability of the existence of the rest of the monkey. Similarly, the part of the tree trunk is more a part of the tree in this sense than of the monkey's leg. 
These arguments concerned the spatial identity problem. However, for temporal identity similar arguments hold. The existence of a living organism at one point in time makes it more probable that there is a living organism (in the vicinity) at a subsequent (and preceding) point in time. If we look at structures that are not entity-like with respect to the temporal dimension we find a different situation. A part of a tree trunk at some instance of time does not make the existence of a monkey's leg at a subsequent instance much more probable. It makes the existence of a tree at a subsequent instance much more probable. So the part of the tree trunk seems to be more a temporal instance/part of the tree than of the monkey's leg.
For STPs we can easily formalise such intuitions. We required that for an entity every part of the structure, which is now a STP $x_O$, makes every other part more probable. A part of a STP is a STP $x_b$ with $b \subset O$. If we require that every part of a STP makes every other part more probable then we can write that $x_O$ is an entity if:
\begin{equation}
  \min_{b \subset O} \frac{p_{O\bs b}(x_{O\bs b}|x_b)}{p_{O\bs b}(x_{O \bs b})} > 1.
\end{equation} 
This is equivalent to 
\begin{equation}
  \min_{b \subset O} \frac{p_O(x_O)}{p_{O\bs b}(x_{O \bs b}) p_b(x_b)} > 1.
\end{equation} 
If we write $\Latt_2(O)$ for the set of all bipartitions of $O$ we can rewrite this further as
\begin{equation}
\label{eq:biparts}
  \min_{\pi \in \Latt_2(O)} \frac{p_O(x_O)}{\prod_{b \in \pi} p_{b}(x_b)} > 1.
\end{equation} 
We can interpret this form as requiring that for every possible partition $\pi \in \Latt_2(O)$ into two parts $x_{b_1},x_{b_2}$ the probability of the whole STP $x_O=(x_{b_1},x_{b_2})$ is bigger than its probability would be if the two parts were independent. To see this, note that if the two parts $x_{b_1},x_{b_2}$ were independent we would have
\begin{equation}
  p_O(x_O)=:p_{b_1,b_2}(x_{b_1},x_{b_2})=p_{b_1}(x_{b_1})p_{b_2}(x_{b_2}).
\end{equation} 
Which would give us 
\begin{equation}
  \frac{p_O(x_O)}{\prod_{b \in \pi} p_{b}(x_b)} =1
\end{equation} 
for this partition.

From this point of view the choice of bipartitions only seems arbitrary. For example, the existence a partition $\xi$ into three parts such that 
\begin{equation}
  p_O(x_O) = \prod_{c \in \xi} p_c(x_c)
\end{equation} 
seems to suggest that the STP $x_O$ is not an entity but instead composite of three parts. We can therefore generalise \cref{eq:biparts} to include all partitions $\Latt(O)$ (see \cref{def:partitionlattice}) of $O$ except the unit partition $\lunit_O$ (\cref{def:lunit}). Then we would say that $x_O$ is an entity if
\begin{equation}
\label{eq:nolog}
  \min_{\pi \in \Latt(O)\bs \lunit_O} \frac{p_O(x_O)}{\prod_{b \in \pi} p_{b}(x_b)} > 1.
\end{equation} 
This measure already results in the same entities as the measure we propose. 

However, in order to connect with information theory, log-likelihoods, and related literature we formally introduce the logarithm into this equation. For this we use the definition of specific local integration $\mi_\pi(x_O)$ of a STP $x_O$ with respect to a partition $\pi \in \Latt(O)$ (see \cref{def:mipi}) as
\begin{equation*}
 \mi_\pi(x_O):= \log \frac{p_O(x_O)}{\prod_{b \in \pi} p_b(x_b)}. \tag{\ref{eq:mipi} revisited}
\end{equation*} 
Then according to \cref{def:ci} we can write the complete local integration $\ci(x_O)$ of a STP as
\begin{equation*}
  \ci(x_O):= \min_{\pi \in \Latt(O) \bs \lunit_O} \mi_\pi(x_O).\tag{\ref{eq:ci} revisited}
 \end{equation*}
Finally, we can define \textit{$\ci$-entities} as those STPs that are completely locally integrated.

\begin{mydef}[$\ci$-entity]
Given a multivariate Markov chain $\Xv$ a STP $x_O$ is a $\ci$-entity if
\begin{equation}
\label{eq:entity}
  \ci(x_O) > 0.
\end{equation}   
\end{mydef}
The $\ci$-entity-set $\Ent_\ci(\Xv)$ is then defined as follows.
\begin{mydef}[$ci$-entity-set]
\label{def:cientset}
Given a multivariate Markov chain $\Xv$ the $\ci$-entity-set is the entity-set
\begin{equation}
  \Ent_\ci(\Xv) := \{x_O \in \bigcup_{A \subseteq V} \X_A: \ci(x_O) > 0\}.
\end{equation} 
\end{mydef}
Note, that due to the disintegration theorem (\cref{thm:disintegration}) $\Ent_\ci(\Xv)$ contains the same elements as the  union of the refinement-free disintegration hierarchies over all trajectories:
\begin{equation}
  \Ent_\ci(\Xv) = \bigcup_{x_V \in \X_V} \dis^\lmin(x_V).
\end{equation} 


\subsubsection{Interpretations and relations}
\label{sec:interpret}
The notion of $\ci$-entities can be interpreted in multiple ways. The introduction of the logarithm into our formalism might seem arbitrary. However, it leads to connections to other considerations especially in information theory and inference. Here we list some of these connections.  

\begin{itemize}
  \item A first consequence of introducing the logarithm is that we can now formulate the condition of \cref{eq:entity} analogously to an old phrase attributed to Aristotle that ``the whole is more than the sum of its parts''. In our case this would need to be changed to ``the $\log$-probability of the (spatiotemporal) whole is greater than the sum of the $\log$-probabilities of its (spatiotemporal) parts''. This can easily be seen by rewriting \cref{eq:mipi} as:
\begin{equation}
  \mi_\pi(x_O)= \log p_O(x_O) - \sum_{b \in \pi} \log p_b(x_b).
\end{equation} 

\item
Another side effect of using the logarithm is that we can interpret \cref{eq:entity} in terms of the surprise value (also called information content) $-\log p_O(x_O)$ \cite{mackay_information_2003} of the STP $x_O$ and the surprise value of its parts with respect to any partition $\pi$. Rewriting \cref{eq:mipi} using properties of the logarithm we get:
\begin{equation*}
 \label{eq:surprise}
 \mi_\pi(x_O)= \sum_{b \in \pi} (-\log p_b(x_b)) - (- \log p_O(x_O)).
\end{equation*} 
Interpreting \cref{eq:entity} from this perspective we can then say that a STP is an entity if the sum of the surprise values of its parts is larger than the surprise value of the whole. 

\item
With respect to hypothesis testing, we can view the product probability $\prod_{b \in \pi} p_b(x_b)$ with respect to partition $\pi$ as the probability of $x_O$ associated with the hypothesis that the parts $x_b$ are stochastically independent. Let us call this hypothesis $\mathcal{H}_\pi$. Then we can write:
\begin{equation}
  p(x_O|\mathcal{H}_\pi):=\prod_{b \in \pi} p_b(x_b).
\end{equation} 
Similarly, we can view the joint probability $p_O(x_O)$ as the probability of $x_O$ under the hypothesis that the full joint probability is needed. Let us write $\mathcal{H}_{\lunit}$ for this hypothesis and define accordingly:
\begin{equation}
   p(x_O|\mathcal{H}_\lunit):=p_O(x_O).
\end{equation} 
The occurrence of $x_O$ is then said to provide what is called the ``weight of evidence in favour of $\mathcal{H}_\lunit$'' \citep{mackay_information_2003} defined by
\begin{equation}
  \log \frac{p(x_O|\mathcal{H}_\lunit)}{p(x_O|\mathcal{H}_\pi)} > 0.
\end{equation} 
So in this terminology a completely locally integrated STP $x_O$ provides evidence in favour of $\mathcal{H}_\lunit$ compared to \textit{each} hypothesis $\mathcal{H}_\pi$, $\pi \in \Latt(O)\bs \lunit$ that supposes it is composite of stochastically independent parts.

\item 
In coding theory, the Kraft-McMillan theorem \citep{cover_elements_2006} tells us that the optimal length (in a uniquely decodable binary code) of a code word for an event $x$ is $l(x)=-\log p(x)$ if $p(x)$ is the \textit{true} probability of $x$. If the encoding is not based on the true probability of $x$ but instead on a different probability $q(x)$ then the difference between the optimal code word length and the chosen code word length is 
\begin{equation}
-\log q(x) -(- \log p(x)) = \log \frac{p(x)}{q(x)}. 
\end{equation} 
Then we can interpret the specific local integration as a difference in code word lengths. Say we want to encode what occurs at the nodes/random variables indexed by $O$ i.e.\ we encode the random variable $X_V$. We can encode every event (now a STP) $x_O$ based on $p_O(x_O)$. Let's call this the \textit{joint code}. Given a partition $\pi \in \Latt(O)$ we can also encode every event $x_O$ based on its product probability $\prod_{b \in \pi_O} p_b(x_b)$. Let's call this the \textit{product code with respect to $\pi$}. For a particular event $x_O$ the difference of the code word lengths between the joint code and the product code with respect to $\pi$ is then just the specific local integration with respect to $\pi$.
%

Complete local integration then requires that the joint code code word is shorter than all possible product code code words. This means there is no partition with respect to which the product code for the STP $x_O$ has a shorter code word than the joint code. So entities are STPs that are shorter to encode with the joint code than a product code.
%

\item
We can relate our measure of identity to other measures in information theory. For this we note that the expectation value of specific local integration with respect to a partition $\pi$ is the multi-information $\mathcal{I}_\pi(X_O)$ \cite{mcgill_multivariate_1954,amari_information_2001} with respect to $\pi$, i.e.
\begin{align}
  \mathcal{I}_\pi(X_O):&=\sum_{x_O \in \X_O} p_O(x_O) \log \frac{p_O(x_O)}{\prod_{b\in \pi} p_b(x_b)} \\
  &=\sum_{x_O \in \X_O} p_O(x_O) \mi_\pi(x_O).
\end{align} 
The multi-information plays a role in measures of complexity and information integration \citep{ay_information_2015}. The generalisation from bipartitions to arbitrary partitions is applied to expectation values similar to the multi-information above in \citet{tononi_information_2004}. The relations of our localised measure (in the sense of \cite{lizier_local_2012}) to multi-information and information integration measures also motivates the name specific \textit{local} \textit{integration}. Relations to these measures will be studied further in the future. Here we note that these are not suited for measuring identity of STPs since they are properties of the random variables $X_O$ and not the values $x_O$. 

\item 
Using the disintegration theorem (\cref{thm:disintegration}) results in yet another point of view. The theorem states that for each trajectory $x_V \in \X_V$ of a multivariate Markov chain the refinement-free disintegration hierarchy only contains completely integrated STPs i.e.\ it only contains $\ci$-entities. It also contains all $\ci$-entities that occur in that trajectory. The disintegration hierarchy is obtained by sorting the partitions $\pi \in \Latt(V)$ of the trajectory $x_V$ according to increasing specific local integration $\mi_\pi(x_V)$ of $x_V$. This results in the disintegration levels $\dis_i(x_V)$ with $\dis_1(x_V)$ containing the partitions with the least specific local integration. To get to the refinement-free version of the disintegration hierarchy, we remove all partitions from each level $\dis_i(x_V)$ that either have a refinement at that level or have a refinement at a lower level $\dis_j(x_V)$ with $j < i$. 
A partition in the refinement-free disintegration hierarchy is always a minimal/finest partition (\cref{def:minandmaxelements}) reaching such a low specific local integration.

Each $\ci$-entity is then a block $x_c$ with $c \in \pi$ of a partition $\pi \in \dis^\lmin(x_V)$ for some trajectory $x_V \in \X_V$ of the multivariate Markov chain. 

Let us recruit the interpretation from coding theory above. If we want to find the optimal encoding for the entire multivariate Markov chain $\Xv$ this means finding the optimal encoding for the random variable $X_V$ whose values are the trajectories $x_V \in \X_V$. The optimal code has the code word lengths $-\log p_V(x_V)$ for each trajectory $x_V$. The partitions in the lowest level $\dis^\lmin_1(x_V)$ in the refinement-free disintegration hierarchy for $x_V$ have minimal specific local integration i.e.\
\begin{equation}    
   \mi_\pi(x_V) = \log \frac{p_V(x_V)}{\prod_{c \in\pi} p_c(x_c)} 
\end{equation} 
is minimal among all partitions. At the same time these partitions are the finest partitions that achieve this low specific local integration. This implies on the one hand that the code word lengths of the product codes associated to these partitions are the shortest possible for $x_V$ among all partitions. On the other hand these partitions split up the trajectory in as many parts as possible while generating these shortest code words. In this combined sense the partitions in $\dis^\lmin_1(x_V)$ generate the ``best'' product codes for the particular trajectory $x_V$. 

Note that the \textit{expected code word length} of the product code:
\begin{equation}
  \sum_{x_V \in \X_V} p_V(x_V) (-\log \prod_{c \in \pi} p_c(x_c)) 
\end{equation} 
which is the more important measure for encoding in general, might not be short at all. The product codes based on partitions in $\dis^\lmin_1(x_V)$ are specifically adapted to assign a short code word to $x_V$ i.e.\ to a single trajectory or story of this system. They are constructed/forced to describe $x_V$ as a composition of stochastically independent parts. More precisely they are constructed in the way that would be optimal for stochastically independent parts. The parts themselves are chosen to minimise $\mi_\pi(x_V)$ for $x_V$.

Nonetheless, the product codes exist (they can be generated using Huffman coding or arithmetic coding \cite{cover_elements_2006} based on the product probability) and are uniquely decodable. What would they be useful for? Say for some reason the trajectory $x_V$ is more important than any other and that we want to ``tell its story'' as a story of as many as possible (stochastically) independent parts (that are maybe not really stochastically independent) i.e.\ we wanted to encode the trajectory \textit{as if it were} a combination of as many as possible stochastically independent parts/events. And because $x_V$ is more important than all other trajectories we wanted the code word for $x_V$ to be the shortest possible. Then we would use the product codes of partitions in the refinement-free disintegration hierarchy because those combine exactly these two conditions. The pseudo-stochastically-independent parts would then be the blocks of these partitions which according to the disintegration theorem are exactly the $\ci$-entities occurring in $x_V$.

On a very speculative note we mention that the trajectory/history that we (real living humans) live in is more important to us than all other possible trajectories of our universe (if there are any). What happens in this trajectory needs to be communicated more often than what happens in counterfactual trajectories. Furthermore a good reason to think of a system as composite of as many parts as possible is that this reduces the number of parameters that need to be learned which in turn improves the learning speed \citep[see e.g.][]{kolchinsky_prediction_2011}. So the entities that mankind has partitioned our history into might somehow serve a purpose related to the product codes generated form partitions of the refinement-free disintegration hierarchy of our universe. 

Recall that this kind of product code is not the optimal code in general (which would be the one with shortest expected code word length). It is possibly more of a naive code that does not require deep understanding of the dynamical system but instead can be learned fast and works. The language of physics for example might be more optimal in the sense of shortest expected code word lengths reflecting a desire to communicate efficiently about all counterfactual possibilities as well. 
\end{itemize}
This concludes the motivation of our proposal to use completely locally integrated patterns as entities in multivariate Markov chains. Next we will present definitions of actions and perceptions.

\section{Entity action}
\label{sec:actions}

Here we define a concept of actions for a given entity set (\cref{def:entset}) in a multivariate Markov chain. First we discuss some challenges that arise when trying to define actions within such rigidly defined systems (\cref{sec:contrast}). Then we motivate our approach to actions (\cref{sec:background}). In \cref{sec:actiondef}, we finally present the formal definition of actions for entities in multivariate Markov chains. Conceptually, this section is loosely related to our own publication \citet{biehl_apparent_2015} but the formal setting is different. The formal definition of actions in \cref{sec:actiondef} to our knowledge is the first of its kind. This is to say that it is the first formal definition of actions that is applicable to individuals/entities within multivariate Markov chains (including dynamical system, cellular automata etc.). 

In order to avoid confusion we will refer to individuals whenever we speak of entities that are not necessarily elements of an entity set in the technical sense of \cref{def:entset}. For example, we speak of animals as individuals that can perform actions. This does not imply that there is another notion of ``individual'' which needs to be defined. We argued in \cref{sec:entinmvmc} that in the context of multivariate Markov chains such individuals correspond to entities. Outside of multivariate Markov chains we have not made such arguments and therefore use the term individual here.  


\subsection{Contrast to more common conceptions}
\label{sec:contrast}
Paraphrasing \citet{sep-action} only slightly, what distinguishes actions among events or occurrences is that they do not merely happen to individuals but rather that they are \textit{made to happen by} the individuals. 

%
This is problematic in our setting where STPs (as entities) take the role of individuals. What ``happens'' in a multivariate Markov chain are the trajectories and the STPs occurring in them. The Markov chain's dynamics are determined by its mechanisms $p_{j,t}$ with $j \in J, t \in T$. These in turn determine (possibly stochastically) what is going to happen anywhere within the chain. All mechanisms at all time-steps are fixed by the definition of the Markov chain and then cannot be altered anymore. If it is desired that mechanisms change over time then this must be decided when defining the Markov chain. Since the occurrence of STPs is an effect of these fixed mechanisms the STP cannot ``make anything occur'' within the chain. Just like the occurrence of any STP up to time $t$ is a consequence of the mechanisms so are the occurrences of STPs in the future of $t$. More formally, given any STP (be it an entity or not) $x_A$, its morph (see \cref{def:morph}) $p_{V\bs A|A}(X_{V \bs A}|x_A)$ is the probability distribution over the rest possible states $\X_{V\bs A}$ of the multivariate Markov chain given that $x_A$ occurs. By definition this morph is  
determined uniquely by the mechanisms of the chain. This means whatever ``happens'' beyond the STP $x_A$ is already determined when the Markov chain is defined. 

Therefore, it is impossible that a multivariate Markov chain contains an STP or entity that can make something happen beyond what happens anyway due to the mechanisms. This means we have to explain and define actions in a different way. 

Before we go on we should note that many accounts of actions require the actions to be in the interest of some goal or to serve some purpose \citep{sep-action}. In accordance with our working definition we view goal-directedness as a separate phenomenon and will not follow the practice of requiring such for actions themselves. In our case an entity with actions will be considered goal-directed if its actions are goal-directed in some sense. 


After these comments on what we cannot do and what we choose not to do we will now motivate our own approach. First, we give some background and observations about actions that motivate our definition. Then we present the main ideas behind it and finally state the definition. 


\subsection{Background to our concept of actions}
\label{sec:background}
We can make two observations about the common (human) usage of the term action. The first is that events called actions are usually attributed to a limited or bounded region or part of the universe e.g.\ the body of a living organism or sometimes just its brain if it has one. These parts usually contain mechanisms or configurations of matter that are either 
\begin{itemize}
 \item not directly observable to a human observer e.g.\ hidden in an opaque container, or
 \item not well understood by the human observer, or
 \item both.
\end{itemize}
These factors inevitably lead to unpredictability of such events. In other words, events that are attributed to well understood and therefore predictable mechanisms, e.g.\ sunrises, are not considered actions. 


%

Let us consider the above more closely. Historically, actions (agency) have been attributed to more things than just animals or living systems (or robots). An example of this is the attribution of natural phenomena like thunder and lightning to divine interventions in Rome in 50 BCE which was criticised by \citet{lucretius_nature_2007}. Lightning in particular was often seen as a goal-directed action by the god Zeus; the goal being to punish humans. Later, in the 19th century \citet[pp.26]{nietzsche_zur_1892} criticised the separation of ``the lightning'' (der Blitz) as a subject and the flashing light (das Leuchten) as its action (Thun). It is notable that as science progressed it was able to explain more and more phenomena without divine (or any other) interventions and without reference to any actions (or goal-directedness) at all \footnote{The term ``action'' in the ``principle of least action'' plays a major role in physics. However, this principle is used to determine trajectories of dynamical systems and has no relation to possible actions performed by parts of the system. It is therefore ignored in the discussion here.}. Also note that, the mechanics behind lightning and thunder were difficult to understand before technological and scientific advances and still are difficult to observe as they are due to electrodynamics (and happen in places that are hard to access).

Nowadays we have a mechanistic account and events like thunder and lightning are hardly considered more special than an apple falling to the ground because of gravity. Actions do remain to be attributed to animals of course. Most prominently to humans and their nervous systems. These systems coincide with the most complex known parts of the known universe i.e.\ those parts that are extremely hard to understand
. From our point of view this is not a coincidence. It is the complexity and opacity of these mechanisms that make us attribute actions to them. If we would have the sensory and computational capacity to watch and keep track of the dynamics of entire brains, we believe that it would look to us again like an apple falling to the ground.
From this point of view actions are not, beyond their possibly complex and unobserved origin, special events but may \textit{appear} as such to observers that lack the sensory and computational capacity to resolve or understand them. This suggests that for actions to occur within a system there needs to be both observers and corresponding mechanisms that exceed the capacity of those observers to resolve them i.e.\ see them as mere consequences of the dynamical law. Note that the observers might themselves be such opaque mechanisms for other observers and for themselves. 

There may be a possibility to define actions in a fundamental way without the need to define observers first. Say there are events within the universe which are as a matter of principle not distinguished by \textit{any} observer. Then events occurring as a consequence of these events will be inexplicable for any observer. So these events will appear to be actions in general. This is the route we take below. Note that this approach remains compatible with an observer-dependent notion of actions. The ``fundamental'' actions are apparent actions for every possible observer while other events are actions for some observers and ``plainly'' predictable events for others.

What we have ignored in this discussion up to now is the role of randomness. True randomness (in the sense of stochastic independence of the event from any other event in the universe), if it exists in a universe, can never be explained, predicted, or understood. Combined with our reasoning above this suggests that all random events are actions and even fundamental actions in the sense that no observer could possibly resolve the different events that lead to the random events just because there are no different events that lead to a random event. The random event happens independently of everything else. This is also the reason why we would not like to see random events as actions. They are not the result of some indistinguishable but in-system events. Even an all-observing being external to the universe could not predict them from the internals. 

One way to avoid random events being mistaken for actions would then be to require that external observers, which are not limited by the restrictions on observability for internal observers, can predict the action events from other events that are internally unobservable. Note that this could be seen as the adaptation to our setting of the widely accepted view that actions are initiated by the agent \citep[e.g.][]{sep-agency}. In our definition of actions we do not explicitly require this. Instead, the burden to avoid the random events from being mistaken for actions is put on the choice of the entity set. In our conception of actions, actions can only be performed by entities, more precisely they can only occur as parts of larger entities. Intuitively entities are spatiotemporal-patterns whose parts are in some way connected to each other. Random events (stochastically independent from all other events) are therefore not expected to be parts of entities. If the entities in a given entity-set do not contain parts that are random events an explicit requirement of predictability is not needed. Note that, according to trivial definitions of entities (like the unrestricted mereological composition of \cref{sec:identity}) random events may be parts of larger entities. In that case we expect non-intuitive consequences anyway.




\subsection{Definition of actions for entities}
\label{sec:actiondef} 
When we want to define actions for entities the first issue we run into is that entities are already fixed STPs. They may or may not have ``acted'' within a trajectory that they occur in but once we have the entity its ``story'' is fixed. In order to define actions we therefore look at the sequence of time-slices of an entity and investigate what ``could have happened''. In the end, whenever there are counterfactual entities that could have taken the place of the entity without changes in the rest of the system and then went on a different path we will say that an action occurred. 

In more detail, first note that an action always requires the possibility of an alternative action. However, as argued before, a single entity $x_A$ occurring in trajectory $x_V$ has no alternative options since the trajectory determines everything. Therefore for an action of entity $x_A$ in trajectory $x_V$ we require the existence of alternative/counterfactual entity $y_B$ in another trajectory $y_V$. For an action to occur at time $t$ 
\begin{itemize}
  \item the entities $x_A$ and $y_B$ must occupy the same random variables at $t$ i.e.\ $A_t = B_t$,
  \item the time-slices $x_{v_t}$ and $y_{V_t}$ at $t$ of the two trajectories must coincide everywhere apart from the random variables that are occupied by the entities i.e.\ $x_{V_t \bs A_t} =y_{V_t \bs A_t}$.
\end{itemize}
It is then impossible that any observer that is in the ``environment'' $x_{V_t \bs A_t}$ of the entities can distinguish the entities because the states of all such observers are identical in both trajectories. For an action to occur the two identities must then differ at time $t+1$. We define the environment of a STP here for further use.

\begin{mydef}[Environment of an STP]
  Let $\Xv$ be a multivariate Markov chain with $V=J\times T$ and let $x_A$ be a STP. Then the environment of $x_A$ at time $t$ is the spatial pattern $x_{V_t \bs A_t}$.
\end{mydef}

%
%
%

As mentioned before we do not require that the difference at time $t+1$ is predictable from the entities during the interval. Such relations between parts of entities, if desired, must be imposed by the choice of the entity set. 


This construction may lead to the following question. According to this definition actions rely on counterfactual trajectories. However, actions as commonly understood occur all the time within the single history/trajectory that we are experiencing. Since nobody has ever experienced two alternative trajectories of our universe the question is how can this concept play a role in our conception of the world? The answer to this is that the existence of actions as we defined them will force conceptions or models of the world to incorporate them. Let us assume that humans model the universe they exist in to some degree. According to the indistinguishability requirement whenever there is an action by another entity this model will lack the data to distinguish which act will occur. In such situations it should be prepared for both acts, i.e.\ it should model both acts. So the counterfactual trajectory of the universe becomes relevant for individuals modelling their environment/world. 

Another question may concern the effect of actions according to this definition. We have not required that the actions i.e.\ the different time-slices of the counterfactual entities at time $t+1$ are distinguished by any ``observer''. While such requirements may be possible we make no such requirement here. Our definition of actions is deliberately weak. As mentioned before it is the entity set that we see as selective. In the future further notions of actions will be investigated.


%
%

We now state the definition of an action of an entity at a time $t$ in a particular trajectory formally.

\begin{mydef}[Action and co-action of an entity]
\label{def:action}
 Let $\Xv$ be a multivariate Markov chain with $V=J\times T$ and let $x_V \in \X_V$ with $p_V(x_V) > 0$. Also let $x_A$ be an entity with non-empty time-slices at $t,t+1$. Then $x_A$ \textit{performs an action $x_{A_{t+1}}$ at time $t$ in trajectory $x_V$} if there exists an entity $y_B$ with non-empty time-slices at $t,t+1$ such that 
 \begin{thmlist}
   \item $y_B$ occurs in $y_V \neq x_V$ with $p_V(y_V)>0$, 
   \item at $t$ the entities $x_A$ and $y_B$ occupy the same random variables: $B_t = A_t$,
   \item at $t$ the trajectories $x_V$ and $y_V$ are otherwise identical: $x_{V_t \bs A_t} = y_{V_t \bs A_t}$,
   \item at $t+1$ the entities are different: $x_{A_{t+1}} \neq y_{B_{t+1}}$.
 \end{thmlist}
 We also call $y_B$ a \textit{co-action entity}, $y_V$ a \textit{co-action trajectory}, and $y_{B_{t+1}}$ a \textit{co-action}.
\end{mydef}
Remark:
\begin{itemize}
  \item Note that all requirements are symmetric. Therefore, if $x_A$ performs an action $x_{A_{t+1}}$ at time $t$ in trajectory $x_V$ then also $y_B$ performs an action $y_{B_{t+1}}$ at time $t$ in trajectory $y_V$. This motivates our terminology of \textit{co}-actions.
  \item The notion of co-action entities can easily be extended to more than one co-action entity. We only have to make sure that all entities in a set of co-action entities are mutually different at $t+1$.
  \item A further requirement that we could make here would be that $y_B$ does not occur in $x_V$. This is not excluded in this definition. At time $t$ the two entities can in principle be equal $x_{A_t}=y_{B_t}$. At $t+1$ we could have $A_{t+1} \cap B_{t+1} = \emptyset$ so that even if $x_{A_{t+1}} \neq y_{B_{t+1}}$ we can have $x_{B_{t+1}} = y_{B_{t+1}}$. This requires that entities can be identical at some time $t$ and then different at some time $t+1$. We do not exclude this possibility here. It is an interesting question for further research at what level such situations should be prevented (if it should be prevented). It could be introduced as an axiom for entity sets which corresponds to prohibiting interpenetration of entities. However, it could also be a selective criterion for specific dynamics of the multivariate Markov chain. One could imagine that there is a set of dynamics obeying a certain conservation law that prevents interpenetration. The notion of $\ci$-entities does not prevent interpenetration as we see in \cref{sec:exampleconnection}.
  \item It is easy to generalise the definition of actions to situations where $x_A$ and $y_B$ must occupy the same variables for an interval of time $[t-m:t]$ before the action. In that case, the environment $x_{V_{[t-m:t]}} \bs A_{[t-m:t]}$ must also be identical during this interval.
\end{itemize}

The condition that the two acting entities differ at time $t+1$ can be fulfilled in two ways. The entities can differ in this time-slice in value or in extent.

\begin{mydef}[Value and extent actions]
  Let $\Xv$ be a multivariate Markov chain with $V=J\times T$. If $x_A$ performs an action $x_{A_{t+1}}$ at time $t$ in trajectory $x_V$ and $y_{B_{t+1}}$ is its co-action we can distinguish two special cases of actions:
  \begin{thmlist}
    \item if the actions differ in extent i.e.\ we have 
  \begin{equation}
    A_{t+1} \neq B_{t+1}
  \end{equation} 
    then we call these actions \textit{extent actions}.
    \item if the actions differ only in value i.e.\ we have 
    \begin{equation}
    A_{t+1}=B_{t+1}
    \end{equation}
    so that 
    \begin{equation}
    x_{A_{t+1}} \neq y_{A_{t+1}}.
    \end{equation}
    then we call these actions \textit{value actions}.
  \end{thmlist}
\end{mydef}
Remarks:
\begin{itemize}
   \item Value actions are a particularly weak notion of action in some sense. Since we define the action only as a difference to the co-action. The environment (or the entire future) may stay unaffected by such an ``action''. The entire morph can be identical for such actions i.e.\:
   \begin{equation}
     p_{V \bs A_{t+1}}(X_{V \bs A_{t+1}} | x_{A_{t+1}}) = p_{V \bs A_{t+1}}(X_{V \bs A_{t+1}} | y_{A_{t+1}}).
   \end{equation} 
   Note that the extent actions always have an effect since they change the random variables that are part of the entity. 
   Formally, the morphs of two different extent actions are always different because they range over different variables. Stronger definitions of value and extent actions which require for example that the environments change or 
   that their morphs differ in particular respects are also possible and may have their own merits. However, an investigation of different definitions is beyond the scope of this thesis. 
\end{itemize}

The difference between value actions and extent actions is made possible due to our definition of entities as STPs. We have argued in \cref{sec:cfvar} that entities should vary counterfactually in value and extent. An intriguing question for the future is whether the capabilities of agents to act both in value and extent are truly superior to agents that only act in value. 
With regard to the theory of computation in distributed systems by \citet{lizier_framework_2014} one can also ask whether there are computational advantages to either. 
As we will see in \cref{sec:palooprelation} probabilistic and information theoretic expressions are easy to formulate for actions in value only. However, for actions in extent this has not been done yet.

\section{Entity perception}
\label{sec:perceptions}
In this section we formally define perception for entities in multivariate Markov chains. We make no distinction here between perception, experience, and sensory input. In the tradition of modelling agent-environment systems using dynamical systems or their probabilistic generalisations stochastic processes we define perception as \textit{all effects} that the environment has on an individual/agent \citep{beer_dynamical_1995}. In contrast to previous work along this line the individuals are not modelled as a dynamical system coupled to the environment \citep{beer_dynamical_1995,der_homeokinesis_1999,ay_information-driven_2012} or a stochastic process interaction with another one \citep{klyubin_organization_2004,lungarella_methods_2005,bertschinger_autonomy_2008,seth_measuring_2010,ay_information-driven_2012}. In our case the individuals are entities i.e.\ special kinds of STPs. In order to define perception we therefore have to capture all effects of the environment on \textit{entities} or if we focus on perception of a single entity all effects on that entity.

Thinking about this we run into a similar problem as with the actions. An entity is already a fixed STP that contains all influence that it may have been subjected to. It is in this sense the \textit{result} of influence (or no influences) from its surroundings.  In order to investigate these influences we therefore have to deconstruct the entity and see how it was ``formed'' by external influences / perceptions time-slice by time-slice. 

The idea here is to use the same (or a similar) construction as in the extraction of sensor-values for the extended perception-action loop in \cref{sec:paloopformal}. 
As we have seen there this construction of sensor-values captured all influences of the environment process on the agent process. This was established by showing that the dependence on the environment can be replaced by the dependence on the sensor-values without changing the agent or environment processes. So whatever influence the environment process has on the agent process, this influence is contained in the sensor-values. Underlying this construction is the classification of the environment into classes that have identical influence on the transition of the agent process from on time-step to the next.
More precisely, the sensor-values of the extended perception-action loop are constructed as equivalence classes of environments with respect to the conditional probability distributions $p_{M_{t+1}}(.|m_t,e_t):\M_{t+1}\rightarrow [0,1]$. We defined that two environments $\hat{e}_t, \bar{e}_t \in \E_t$ at time $t$ are produce the same perception / sensor value if they induce the same conditional probability distribution over the agent's next time-step: 
  \begin{equation*}
     \hat{e}_t \equiv_{\epsilon_t} \bar{e}_t \Leftrightarrow \forall m_{t+1} \in \M_{t+1}, m_t \in \M_t : p_{M_{t+1}}(m_{t+1} |  m_t,\hat{e}_t) = p_{M_{t+1}}(m_{t+1} | m_t, \bar{e}_t).  \tag{\ref{eq:perceptionequiv} revisited}
  \end{equation*} 
Here $\epsilon_t$ is the partition induced by this equivalence relation. In this section we are interested in defining the influence of the environment on an entity. This will require a generalisation of \cref{eq:perceptionequiv} which involves some subtleties.

Before we present the generalisation let us look at a simple example of the standard construction. This will lead to a better intuition for our concept of perception. 

\subsection{Example of perception in the perception-action loop}
\label{sec:experception}
Say we have a binary agent process $\M_t = \{m^1_t,m^2_t\}$ and a ternary environment process $\E_t = 
\{e^1_t,e^2_t,e^3_t\}
$. For a given fixed value $m_t \in \M_t$ of the agent memory at $t$ each environment value $e_t \in \E_t$ then has an associated 
conditional probability distribution $p_{M_{t+1}}(.|m_t,e_t): \M_{t+1} \rightarrow [0,1]$ of the form:
  \begin{equation}   
    \begin{array}{|c|c|c|}
\hline
  & m^1_{t+1} & m^2_{t+1} \\
\hline
p_{M_{t+1}}(.|e^1_t,m_t)      & q & 1-q  \\
\hline
p_{M_{t+1}}(.|e^2_t,m_t) & r & 1-r  \\
\hline
p_{M_{t+1}}(.|e^3_t,m_t) & s & 1-s  \\
\hline
\end{array}
\end{equation}
where $q,r,s \in [0,1]$. First, assume $q=r=s$. We then have for all $m_{t+1} \in \M_{t+1}$ 
\begin{equation}
\label{eq:shitequality}
  p_{M_{t+1}}(m_{t+1}|m_t,e^1_t) = p_{M_{t+1}}(m_{t+1}|m_t,e^2_t) = p_{M_{t+1}}(m_{t+1}|m_t,e^3_t).
\end{equation} 
Since we are eventually interested in fixed realisations of entities (and not in random variables), we here drop the requirement of \cref{eq:perceptionequiv} that \cref{eq:shitequality} needs to hold for all $\bar{m}_t \in \M_t$ and consider the equivalence classes that are generated for the \textit{specific} $m_t \in \M_t$. We then get
\begin{equation}
  e^1_t \equiv_{\epsilon_t} e^2_t \equiv_{\epsilon_t} e^3_t.
\end{equation} 
This means there is only a single block in $\epsilon_t$ i.e.\ $\epsilon_t=\{\{e^1_t,e^2_t,e^3_t\}\}$. So all environments have the same influence on the next agent state $m_{t+1}$ given $m_t$. Or, equivalently, given $m_t$, no differences in the environment make a difference to $m_{t+1}$. Since we equate influence with perception here, we interpret this as saying that no perception occurs in the transitions from $m_t$ into $\M_{t+1}$.

Second, assume that $q=r\neq s$. Then $\epsilon_t=\{\{e^1_t,e^2_t\},\{e^3_t\}\}$ containing two blocks. This means that $e^1_t$ and $e^2_t$ influence the transition from $m_t$ into $\M_{t+1}$ in the same way while $e^3_t$ has a different influence. We then say that in the transition from $m_t$ into $\M_{t+1}$ there are two perceptions/sensor values corresponding to the two blocks of $\epsilon_t$. Note that while we have perception in this transition the perception is not perfect. It cannot distinguish between $e^1_t$ and $e^2_t$.

Third, assume that $q\neq r\neq s \neq q$. Then $\epsilon_t=\{\{e^1_t\},\{e^2_t\},\{e^3_t\}\}$ containing three blocks and fully resolving the environment. This means each environment influences the transition from $m_t$ into $\M_{t+1}$ differently. We then have three different perceptions in the transition from $m_t$ into $\M_{t+1}$ resolving the environment states fully.

So our notion of perception employs differences in the influence of environments on transitions from a value $m_t$ to its possible successors $\M_{t+1}$ to classify the environments. For entities we will try to use the same approach with some necessary generalisations. for this it is also helpful to note the following.   

Instead of only considering the next time-step note that we can also use the next two time-steps (or any number $r \in [t+1:n-1]$ of next time-steps\footnote{In fact the same construction can be used with any subset of the future times $[t+1:n-1]$. We will not pursue this generalisation further in this thesis.}, where $n-1$ is the last time-step in $T$) in the same way to partition the environment. For example if we consider the next two time-steps $t+1,t+2$ we can define the equivalence classes of environments via:
  \begin{equation}
     \begin{split}\hat{e}_t &\equiv_{\epsilon^2_t} \bar{e}_t \\
&\Leftrightarrow \forall m_{t+1} \in \M_{t+1}, m_t \in \M_t : \\
&\phantom{\Leftrightarrow} p_{M_{t+1},M_{t+2}}(m_{t+1},m_{t+2} |  m_t,\hat{e}_t) = p_{M_{t+1},M_{t+2}}(m_{t+1},m_{t+2} | m_t, \bar{e}_t). 
            \end{split}
\end{equation} 

The resulting partition $\epsilon^2_t$ is then a refinement of the partition $\epsilon_t$. To see this note that 
\begin{equation}
\label{eq:epsmargin}
   p_{M_{t+1}}(m_{t+1}|m_t,e_t) =\sum_{m_{t+2} \in \M_{t+2}} p_{M_{t+1},M_{t+2}}(m_{t+1},m_{t+2} | m_t, e_t)
\end{equation} 
such that all environments  $\hat{e}_t, \bar{e}_t \in \E_t$ that are in distinct blocks of $\epsilon_t$ i.e.\ those with
\begin{equation}
\label{eq:epsdistinct}
  p_{M_{t+1}}(m_{t+1} |  m_t,\hat{e}_t) \neq p_{M_{t+1}}(m_{t+1} | m_t, \bar{e}_t)
\end{equation} 
are also in distinct blocks of $\epsilon^2_t$ because \cref{eq:epsmargin,eq:epsdistinct} imply
\begin{equation}
  p_{M_{t+1},M_{t+2}}(m_{t+1},m_{t+2} | m_t, \hat{e}_t) \neq p_{M_{t+1},M_{t+2}}(m_{t+1},m_{t+2} | m_t, \bar{e}_t).
\end{equation} 
The more time-steps into the future we consider the finer the induced partition of the environment. Conversely, the partition of the environment obtained by considering only one next time-step is a coarsening of those obtained by considering more time-steps.

\subsection{Steps to get perception for entities}

In order to get a notion of perception for entities that is similar to the notion of perception based on influence that we used for the perception-action loop the intuition is then simply 
\begin{itemize}
  \item ``take an entity at time $t$'' (analogous to $m_t$ ),
  \item get the ``possible next time-slices of this entity'' at $t+1$ (analogous to $\M_{t+1}$ above),
  \item obtain ``the'' conditional probability distribution over these ``next time-slices'' given the current one and the environment (analogous to $p_{M_{t+1}}(.|m_t,e_t)$)
  \item classify the environments according to their influence on the transitions to these ``next time-slices''.
\end{itemize}
There are, however, multiple problems which complicate the formal definition of the required notions. Some more obvious ones are listed next. We discuss them and more subtle ones in more detail in the subsequent sections.

\begin{enumerate}
  \item An entity at time $t$ is either a spatial entity (no temporal extension) or it is a time-slice of an entity. If it is a spatial entity then it has no ``next-time slice''. If it is a time-slice we are not ``taking'' the entire entity. If we take the entire entity then it comes with possibly long past and future extension. It can also have a past and an empty next time-slice. 
  \item Related to the previous point is another problem. Assuming an entity that has future extension i.e.\ the next-time slice is not empty, then there are no other ``possible next time-slices of the entity''. As mentioned before the entity is defined in its entirety. Accordingly the next time-slice of an entity (not only if it is empty) is uniquely defined just like its entire future.
  \item Assuming we have obtained some ``next time-slices'' in a reasonable way, these may not be mutually exclusive and exhaustive unlike the values $m_{t+1} \in \M_{t+1}$. This means multiple next time-slices can occur together (if they are not mutually exclusive) or none of the next time-slices occurs (if they are not exhaustive). This makes the construction of the conditional probability distribution complicated.
\end{enumerate}

It turns out that the steps to get a generalisation of perception for entities are more clearly presented in a slightly different order. The overarching goal remains the construction of a conditional probability distribution that generalises $p_{M_{t+1}}(.|m_t,e_t): \M_{t+1} \rightarrow [0,1]$. The steps we take in the next sections are then:
\begin{itemize}
  \item Define entities with identical pasts up to $t$ as analogues of ``an entity at time $t$''. These are the \textit{co-perception entities}.
  \item Define the entire futures of the co-perception entities as the proto-analogues of ``possible next time-slices of the entity'' and only later focus on the actual next time-slices of these entities via the ``branching partition''.
  \item Devise a way to deal with the problems of non-exhaustion on the level of the entire futures of entities. To define the conditional probability distribution we need an exhaustive set of possible outcomes/futures since the sum over the possible outcomes must equal one.
  \item Restrict the environments that can be classified by perception to those that can co-occur with the entities.
  \item Deal with the problem of mutual exclusion of entity futures, which can be done by a further assumption of non-interpenetration of entities. We will see that, to define the conditional probability distribution, we need a mutually-exclusive set of possible outcomes/futures.
  \item Partition all co-perception entities into blocks (called branches) of entities with identical next time-slices since co-perception entities may differ at even later times only. This partition is called the branching-partition. The final conditional probability distribution over ``next time-slices'' will then be over the branches of his partition.
  \item The environments are then classified according to the conditional probability distribution over the branches. 
\end{itemize}

\subsection{Co-perception entities}
\label{sec:copent}
We first discuss the problem that entities have unique next time-slices and therefore the set of ``possible next time-slices'' only contains a single time-slice. This will be resolved by using the ``co-perception entities'' in order to provide a set of possible next time-slices. These do not come from the same entity but from the co-perception entities. Here we motivate and discuss these entities. 

First note that any part of an entity $x_A$ (which is a STP) can also be a part of another entity $y_B$. This means that $A \cap B \neq \emptyset$ and $x_{A\cap B}=y_{A\cap B}$. Therefore we can also have entities $x_A,y_B$ that are identical at some time $t$, 
i.e.\
\begin{equation}
  x_{A_t} = y_{B_t}.
\end{equation} 
These can in general have different next time-slices. The next time-slices of all entities that are equal to $x_A$ at time $t$ (where $x_{A_t}$ is not empty) are then a first candidate for the conditional probability distribution to range over. 

Note however that these entities can also have different pasts. Since we want to define the perception of a single entity we therefore only consider entities that are identical up to some time $t$, 
i.e.\
\begin{equation}
  x_{A_\pet} = y_{B_\pet}.
\end{equation} 
The set of entities with identical pasts up to time $t$ can be interpreted as the set of entities that are the most like $x_A$ up to $t$. These are \textit{different} entities but they only differ in the future. Their futures (including their next time-slices) are therefore a close analogue to the ``possible next time-slices of the entity''. To make sure however that the entities have a next time-slice we also require that they have \textit{non-empty} next time-slice. These requirements together define the notion of the \textit{co-perception entities of an entity $x_A$ at time $t$}. These are entities that also perceive something (maybe the same thing) at $t$ (in their trajectories) if $x_A$ perceives something at $t$.    
\begin{mydef}[Co-perception entities of an entity at $t$]
\label{def:copent}
  Let $\Xv$ be a multivariate Markov chain with $V=J\times T$ and entity set $\Ent$. Let $x_A \in \Ent$ be an entity with non-empty time-slices at $t$ and $t+1$. The set of \textit{co-perception entities $\perstp(x_A,t)$ of entity $x_A$ at $t$} is the set of entities with non-empty time-slices at $t$ and $t+1$, and that are identical up to $t$:
  \begin{equation}
    \perstp(x_A,t):=\{y_B \in \Ent: B_t,B_{t+1} \neq \emptyset, y_{B_\pet}=x_{A_\pet}\}.
  \end{equation} 
\end{mydef}
%

As mentioned before the time-slices at $t+1$ of the co-perception entities $\perstp(x_A,t)$ provide an analogue of the ``possible next time-slices of the entity $x_A$''. The next step would then be to define a conditional probability distribution over this set given the past $x_{A_\pet}$ of the entity and any environment $x_{V_t \bs A_t}$ at $t$. Instead of only looking at the next time-slices at $t+1$ we can also consider the whole future $t\prec = [t+1:n-1]$ ($n-1$ is the last time-step in $T$) of the co-perception entities. As we have indicated in \cref{sec:experception} if we define the conditional probability distribution over the whole futures we will obtain a finer classification of the environments. We can then still refocus on the next time-slice afterwards. This general viewpoint is also more suitable for the formal development of the theory.

In order to discuss the associated problems we consider the special case of a set of co-perception entities that contains only two entities quite thoroughly. So assume that there are only two co-perception entities (including $x_A$ itself) i.e.\ $\perstp(x_A,t)=\{x_A,y_B\}$. 

Apart from the conditions on co-perception entities (\cref{def:copent}) the entities in $\perstp(x_A,t)$ are arbitrary STPs since we are trying to define perception for arbitrary entity sets. 

Still we can note that since $\{x_A,y_B\}$ forms a set the two entities are not equal 
\begin{equation}
x_A \neq y_B 
\end{equation} 
and since $x_{A_\pet} = y_{B_\pet}$ (due to \cref{def:copent}) we then know that 
\begin{equation}
  x_{A_{t\prec}} \neq y_{B_{t\prec}}.
\end{equation} 
But we do not know at which time-slices they differ. For example they could be equal at $t+1$ or any other particular future time-step $t+r$ with $r \in [1,n-1-t]$ (at the last time step $n-1$ of the multivariate Markov chain there is no perception since there is no future). For the next time-slices $x_{A_{t+1}}$ and $y_{B_{t+1}}$ of $x_A$ and $y_B$ we have the extra condition that they are non-empty, i.e.\
\begin{equation}
  A_{t+1} \cap V_{t+1} \neq \emptyset
\end{equation} 
and 
\begin{equation}
  B_{t+1} \cap V_{t+1} \neq \emptyset.
\end{equation} 
Apart from this, the entities in $\perstp(x_A,t)$ are completely arbitrary.
For the following it is important to keep the possible relations between the time-slices of entities in $\perstp(x_A,t)$ in mind. We therefore take a look at these possible relations as well as their implications for the co-occurrence (i.e.\ the joint probabilities) of the time-slices. The time-slices can occupy the same random variables and have the same values (i.e.\ be identical), occupy the same random variables and have different values, occupy partly the same random variables and have the same values at the random variables in the intersection, occupy partly the same random variables and have the different values at the random variables in the intersection, and occupy only different random variables.
Formally, for the time-slices $x_{A_{t+r}}$ and $y_{B_{t+r}}$ of $x_A$ and $y_B$ we can have the following situations:
\begin{enumerate}
   \item $A_{t+r}=B_{t+r}$ and $x_{A_{t+r}} = y_{B_{t+r}}$,
 \item $A_{t+r}=B_{t+r}$ and $x_{A_{t+r}} \neq y_{B_{t+r}}$, 
  \item $A_{t+r}\neq B_{t+r}$, $A_{t+r}\cap B_{t+r} \neq \emptyset$  and $x_{A_{t+r}\cap B_{t+r}} = y_{A_{t+r}\cap B_{t+r}}$,
\item $A_{t+r}\neq B_{t+r}$, $A_{t+r}\cap B_{t+r} \neq \emptyset$  and $x_{A_{t+r}\cap B_{t+r}} \neq y_{A_{t+r}\cap B_{t+r}}$,
\item $A_{t+r}\cap B_{t+r} = \emptyset$.
\end{enumerate}
So in general we have to write the probability that both of the time-slices occur (given\footnote{This conditioning can also be removed in the following calculation. However, since we are only interested in probabilities under these conditions in this section we keep it.} an arbitrary environment $x_{V_t \bs A_t}$ and the identical past $x_{A_\pet}$) as:
\begin{align}
&Pr(X_{A_{t+r}}=x_{A_{t+r}},X_{B_{t+r}}=y_{B_{t+r}} |x_{V_t \bs A_t},x_{A_\pet}) 
\\
\begin{split}
=&Pr(X_{A_{t+r}\bs B_{t+r}}=x_{A_{t+r}\bs B_{t+r}},X_{A_{t+r}\cap B_{t+r}}=x_{A_{t+r}\cap B_{t+r}},
\\
&\phantom{Pr(} 
X_{A_{t+r}\cap B_{t+r}}=y_{A_{t+r}\cap B_{t+r}},X_{B_{t+r}\bs A_{t+r}}=y_{B_{t+r}\bs A_{t+r}}
|x_{V_t \bs A_t},x_{A_\pet})
\end{split}\\
\begin{split}
=&\delta_{x_{A_{t+r}\cap B_{t+r}}}(y_{A_{t+r}\cap B_{t+r}})
Pr(X_{A_{t+r}\bs B_{t+r}}=x_{A_{t+r}\bs B_{t+r}},
\\
&\phantom{\delta_{x_{A_{t+r}\cap B_{t+r}}}(y_{A_{t+r}\cap B_{t+r}})Pr(}
X_{A_{t+r}\cap B_{t+r}}=x_{A_{t+r}\cap B_{t+r}},
\\
&\phantom{\delta_{x_{A_{t+r}\cap B_{t+r}}}(y_{A_{t+r}\cap B_{t+r}})Pr(}
X_{B_{t+r}\bs A_{t+r}}=y_{B_{t+r}\bs A_{t+r}} |x_{V_t \bs A_t},x_{A_\pet}) 
\end{split}
\\
\begin{split}
=&\delta_{x_{A_{t+r}\cap B_{t+r}}}(y_{A_{t+r}\cap B_{t+r}}) \\
&\phantom{\delta_{x_{A_{t+r}\cap B_{t+r}}}}
p_{A_{t+r}\cup B_{t+r}}(x_{A_{t+r}\bs B_{t+r}},x_{A_{t+r}\cap B_{t+r}},y_{B_{t+r}\bs A_{t+r}}|x_{V_t \bs A_t},x_{A_\pet})
\end{split}
\end{align} 

Consequently the five situation above imply the following for the probability of co-occurrence: 
\begin{enumerate}
  \item $A_{t+r}=B_{t+r}$ and $x_{A_{t+r}} = y_{B_{t+r}}$ implies
    \begin{align}
Pr(X_{A_{t+r}}=x_{A_{t+r}},X_{B_{t+r}}=y_{B_{t+r}} |x_{V_t \bs A_t},x_{A_\pet})
=&p_{A_{t+r}}(x_{A_{t+r}}|x_{V_t \bs A_t},x_{A_\pet})\\
=&p_{B_{t+r}}(y_{B_{t+r}}|x_{V_t \bs A_t},x_{A_\pet}).
\end{align} 
  \item $A_{t+r}=B_{t+r}$ and $x_{A_{t+r}} \neq y_{B_{t+r}}$ implies: 
  \begin{equation}
Pr(X_{A_{t+r}}=x_{A_{t+r}},X_{B_{t+r}}=y_{B_{t+r}} |x_{V_t \bs A_t},x_{A_\pet}) =0
\end{equation} 
\item $A_{t+r}\neq B_{t+r}$, $A_{t+r}\cap B_{t+r} \neq \emptyset$  and $x_{A_{t+r}\cap B_{t+r}} = y_{A_{t+r}\cap B_{t+r}}$ implies
 \begin{align}
Pr(&X_{A_{t+r}}=x_{A_{t+r}},X_{B_{t+r}}=y_{B_{t+r}} |x_{V_t \bs A_t},x_{A_\pet})\\
&=p_{A_{t+r}\cup B_{t+r}}(x_{A_{t+r}\bs B_{t+r}},x_{A_{t+r}\cap B_{t+r}},y_{B_{t+r}\bs A_{t+r}}|x_{V_t \bs A_t},x_{A_\pet})
\end{align}
\item $A_{t+r}\neq B_{t+r}$, $A_{t+r}\cap B_{t+r} \neq \emptyset$  and $x_{A_{t+r}\cap B_{t+r}} \neq y_{A_{t+r}\cap B_{t+r}}$ implies
  \begin{equation}
Pr(X_{A_{t+r}}=x_{A_{t+r}},X_{B_{t+r}}=y_{B_{t+r}} |x_{V_t \bs A_t},x_{A_\pet}) =0
\end{equation}
\item $A_{t+r}\cap B_{t+r} = \emptyset$ implies
\begin{align}
Pr(&X_{A_{t+r}}=x_{A_{t+r}},X_{B_{t+r}}=y_{B_{t+r}} |x_{V_t \bs A_t},x_{A_\pet})\\
&=p_{A_{t+r},B_{t+r}}(x_{A_{t+r}},y_{B_{t+r}}|x_{V_t \bs A_t},x_{A_\pet}).
\end{align}
\end{enumerate}
%

\subsection{The problems of exhaustion and mutual-exclusion}
\label{sec:exmuex}
In order to define our notion of perception we need to define a suitable conditional probability distribution over 
the next time-slices or the futures of the co-perception entities $\perstp(x_A,t)$. Intuitively, we want to know with what probability 
which entity occurs at the next time-slice in order to classify the environments accordingly. We are then only interested in cases where one of the co-perception entities' (non-empty) futures actually occurs. Other situations should not be taken into consideration since they do not concern the co-perception entities. However, in general it is possible that none of the co-perception entities occurs. This also poses a formal problem since the probability distribution should range over a set of possible outcomes such that one of them always occurs i.e.\ the sum over the probabilities of all outcomes must be one. In other words the outcomes must be exhaustive. Another requirement is that the outcomes are mutually exclusive, i.e.\ only one (and with the previous requirement exactly one) of the outcomes occurs. This property is also not satisfied in general for co-perception entities. In the following we will take a short look at how to construct a probability distribution over a set of events that is not exhaustive. This construction will also require that the events are mutually exclusive. The construction is basically elementary and well known probability theory. We expose it here in some detail to show why we require mutual exclusion and exhaustion and how these requirements are implicit in the perception-action loop. 

%
%

Consider again the simple case where $\perstp(x_A,t)=\{x_A,y_B\}$. From the multivariate Markov chain we know the probabilities for each of them given the identical past and the environment i.e.\ we know
\begin{equation}
  q:= p_{A_{t\prec}}(x_{A_{t\prec}}|x_{V_t \bs A_t},x_{A_\pet})
\end{equation} 
and 
\begin{equation}
  r:= p_{B_{t\prec}}(y_{B_{t\prec}}|x_{V_t \bs A_t},x_{A_\pet}).
\end{equation}
It is not guaranteed that any of the two STPs will occur. It is generally guaranteed if $A_{t\prec} = B_{t\prec}$ and $X_{A_{t\prec}}$ can only take these two values i.e.\  $\X_{A_{t\prec}}=\{x_{A_{t\prec}},y_{A_{t\prec}}\}$ or at least if for all $z_{A_{t\prec}} \in \X_{A_{t\prec}}$ and environments $x_{V_t \bs A_t}$ we have
\begin{equation}
  p_{A_{t\prec}}(z_{A_{t\prec}}|x_{V_t \bs A_t},x_{A_\pet})=0.
\end{equation} 
In all other cases the random variables with indices in $A_{t\prec} \cup B_{t\prec}$ can take values that lead to neither $x_{A_{t\prec}}$ nor $y_{B_{t\prec}}$ occurring.\footnote{For example if we let $C_{t\prec}:=A_{t\prec} \cup B_{t\prec}$ and define $z_{C_{t\prec}}$ in such a way that it differs from both $x_{A_{t\prec}}$ and $y_{B_{t\prec}}$ i.e.\ there exist nodes $i \in A_{t\prec}, j \in B_{t\prec}$ (possibly with $i =j$)  such that $z_i \neq x_i$ and $z_j \neq y_j$. Then if $p_{C_{t\prec}}(z_{C_{t\prec}}|x_{V_t \bs A_t},x_{A_\pet})>0$ we have $q+r<1$.}

In order to state the property of exhaustiveness formally we first define a notation for the probability that \textit{one or more} elements of a set $\{x^k_{A^k}\}_{k \in I}$ of STPs occurs. We here stop conditioning on $x_{V_t \bs A_t},x_{A_\pet}$ here for readability. Conditioning all following probabilities does not interfere with the argument.
\begin{mydef}
   Let $\Xv$ be a multivariate Markov chain with $V=J\times T$ also let $\stpset=\{x^k_{A^k}\}_{k \in I}$ be a set of STPs. We then write
   \begin{equation}
     Pr\left(\bigcup \stpset\right):=Pr\left(\bigcup_{k \in I} \{X_{A^k}=x^k_{A^k}\}\right):=Pr\left(\bigcup_{k \in I} \T(x^k_{A^k})\right).
   \end{equation} 
   Where $\T(x^k_{A^k})$ is the set of trajectories that $x^k_{A^k}$ occurs in (\cref{def:stptraset}).
\end{mydef}

Then we define that a set of STPs $\stpset$ is exhaustive if the probability that one or more of them occur is one.
\begin{mydef}[Exhaustiveness of a set of STPs]
\label{def:exhaustive}
   Let $\Xv$ be a multivariate Markov chain with $V=J\times T$ also let $\stpset=\{x^k_{A^k}\}_{k \in I}$ be a set of STPs. We say that $\stpset$ is \textit{exhaustive} if 
   \begin{equation}
     Pr\left(\bigcup \stpset\right)=1.
   \end{equation} 
   Else, we call $\stpset$ non-exhaustive.
\end{mydef}
Remark:
\begin{itemize}
  \item It is important to note that for any subset $A \subseteq V$ of $V$ the set $\X_A$ is an exhaustive set of STPs. An example we have seen in \cref{sec:experception} before is the set $\M_{t+1}$ of possible next values of the agent-process. An example we will encounter later in \cref{sec:palooprelation} is $\M_{t\prec}$, the set of possible futures starting from time $t+1$ of the agent process in a perception action loop.  
\end{itemize}

Since there is no guarantee that the set $\perstp(x_A,t)$ of co-perception entities is exhaustive, the question is how to construct a (conditional) probability distribution over a set of non-exhaustive STPs. The standard approach is to use the definition of conditional probabilities of two events $E,F$ (with $p(F)>0$) of the form:
\begin{equation}
\label{eq:condprob}
  p(E|F):=\frac{p(E \cap F)}{p(F)}.
\end{equation} 
With this we can define the probability of any single STP $x^k_{A^k}\in \stpset$ given that one or more of the STPs in a non-exhaustive set $\stpset$ occur. At least as long as one or more of the STPs \textit{can} occur. I.e.\ if 
\begin{equation}
  \Pr\left(\bigcup \stpset\right)>0.
\end{equation} 
we can replace $F \rightarrow \bigcup \stpset$ and $E \rightarrow \{X_{A_k}=x^k_{A^k}\}$ in \cref{eq:condprob} to get
\begin{align}
  \Pr\left(X_{A_k}=x^k_{A^k}|\bigcup \stpset\right) &= \frac{\Pr\left(\{X_{A_k}=x^k_{A^k}\} \cap \bigcup \stpset\right)}{\Pr(\bigcup \stpset)}\\
  &=\frac{\Pr\left(\T(x^k_{A^k}) \cap \bigcup \stpset\right)}{\Pr(\bigcup \stpset)}\\
  &=\frac{\Pr\left(\T(x^k_{A^k}) \cap \bigcup_{l \in I} \T(x^l_{A^l})\right)}{\Pr(\bigcup \stpset)} \\
  &=\frac{\Pr\left(\T(x^k_{A^k})\right)}{\Pr(\bigcup \stpset)}\\
  &=\frac{\Pr(X_{A_k}=x^k_{A^k})}{\Pr(\bigcup \stpset)}\\
  &=\frac{p_{A_k}(x^k_{A^k})}{\Pr(\bigcup \stpset)}.
\end{align} 
While conditioning on $\stpset$ guarantees that one of the STPs $x^k_{A^k}$ occurs it still does not necessarily result in a probability distribution since the sum over all STPs in $\stpset$ may not be equal to one:
\begin{align}
  \sum_{k \in I} \Pr\left(X_{A_k}=x^k_{A^k}|\bigcup \stpset\right) &=  \frac{\sum_{k \in I} p_{A_k}(x^k_{A^k})}{\Pr\left(\bigcup \stpset\right)}\\
  &\geq 1.
\end{align} 
However, if the STPs in $\stpset$ are mutually exclusive this changes. 
\begin{mydef}[Mutual exclusion]
Let $\Xv$ be a multivariate Markov chain with $V=J\times T$ also let $\stpset=\{x^k_{A^k}\}_{k \in I}$ be a set of STPs. We say that $\stpset$ is \textit{a set of mutually exclusive STPs} if for all $k,l \in I$ we have 
\begin{equation}
  \Pr(X_{A_k}=x^k_{A^k},X_{A_l}=x^l_{A^l}) = 0.
\end{equation} 
\end{mydef}
Remark:
\begin{itemize}
  \item It is important to note that for any subset $A \subseteq V$ of $V$ the set $\X_A$ is a mutually exclusive set of STPs. Both $\M_{t+1}$ and $\M_{t\prec}$ of the agent process in a perception action loop are examples of this.  
\end{itemize}

If $\stpset$ is mutually exclusive then we get:
\begin{equation}
  \Pr\left(\bigcup \stpset\right)= \sum_{k \in I} p_{A_k}(x^k_{A^k})
\end{equation} 
and 
\begin{align}
  \sum_{k \in I} \Pr\left(X_{A_k}=x^k_{A^k}|\bigcup \stpset\right) &=  1
\end{align} 
in general. We can then also write:
\begin{align}
  \Pr\left(X_{A_k}=x^k_{A^k}|\bigcup \stpset\right) &=\frac{p_{A_k}(x^k_{A^k})}{\sum_{l \in I} p_{A_l}(x^l_{A^l})}
\end{align} 
in our usual notation. Then the above defines a probability for each element of the set $\stpset$ or equivalently for each index $k \in I$. We can then define the probability distribution  $p_\stpset:\stpset \rightarrow [0,1]$ via
\begin{align}
  p_\stpset(x^k_{A^k}):=\frac{p_{A_k}(x^k_{A^k})}{\sum_{l \in I} p_{A_l}(x^l_{A^l})}
\end{align} 
In summary, if we have a set $\stpset$ of non-exhaustive but mutually exclusive STPs we now know how to define a probability distribution over them. For reference we put this in a theorem.\footnote{This is not an original theorem of this thesis. We presented the preceding arguments since they help to understand subsequent notions.}
\begin{thm}[Probability distribution construction for non-exhaustive but mutually exclusive sets of STPs]
\label{thm:pdistconst}
  Let $\Xv$ be a multivariate Markov chain with $V=J\times T$ also let $\stpset=\{x^k_{A^k}\}_{k \in I}$ be a set of STPs. If $\stpset$ is mutually exclusive (not necessarily exhaustive) and 
  \begin{equation}
  \label{eq:nonzeronenner}
    \sum_{l \in I} p_{A_l}(x^l_{A^l}) > 0
  \end{equation} 
   we can define a probability distribution $p_\stpset:\stpset \rightarrow [0,1]$ via
\begin{align}
  p_\stpset(x^k_{A^k}):=\frac{p_{A_k}(x^k_{A^k})}{\sum_{l \in I} p_{A_l}(x^l_{A^l})}.
\end{align} 
For convenience we also write this as a probability distribution over an index set of $\perstp$:
\begin{align}
  p_\stpset(k):=p_\stpset(x^k_{A^k}).
\end{align}
\end{thm}
\begin{proof}
  Along the lines of the preceding argument. What is missing is the calculation that 
  \begin{equation}
  \Pr\left(\bigcup \stpset\right)= \sum_{k \in I} p_{A_k}(x^k_{A^k})
\end{equation} 
follows from mutual exclusion. This is straightforward but tedious. The idea is that since the probability of all intersections of the sets of trajectories $\T(x^k_{A^k})$ vanishes (due to mutual exclusion) we get the same result \textit{as if} all $\T(x^k_{A^k})$ were disjoint. The probability of a union of disjoint sets is the sum over the probabilities of the sets. 
\end{proof}

For the previous example of a simple co-perception set $\perstp(x_A,t)=\{x_A,y_B\}$ this means that if they are mutually exclusive i.e.\
\begin{equation}
  \Pr(X_{A_{t\prec}}=x_{A_{t\prec}},X_{B_{t\prec}}=y_{B_{t\prec}} |x_{V_t \bs A_t},x_{A_\pet}) =0
\end{equation} 
and at least one of their conditional probabilities is positive i.e.\ 
\begin{equation}
q+r>0
\end{equation} 
we can write $x^1_{A^1}:=x_A$ and $x^2_{A^2}=y_B$, and $b \in \{1,2\}$ to get a conditional probability distribution\footnote{Since all probabilities involved are conditioned on the same STPs the above argument and \cref{thm:pdistconst} hold equally for conditional probabilities.} 
\begin{align}
\label{eq:bimorph}
  p_{\perstp(x_A,t)}(b|x_{V_t \bs A_t},x_{A_\pet}):&= \frac{p_{A^b}(x^b_{A^b_{t\prec}}|x_{V_t \bs A_t},x_{A_\pet})}{\sum_{c \in \{1,2\}}p_{A^c}(x^c_{A^c_{t\prec}}|x_{V_t \bs A_t},x_{A_\pet})}\\
  &=\frac{p_{A^b}(x^b_{A^b_{t\prec}}|x_{V_t \bs A_t},x_{A_\pet})}{q+r}.
\end{align}


With such a conditional probability distribution we can define perception in basically the same way as for the perception-action loop.

\subsection{Co-perception environments}
\label{sec:copenv}
\Cref{eq:bimorph} is already a step towards entity perception since it is a conditional probability distribution over futures of (co-perception) entities i.e.\ over things already quite similar to ``next time-slices of the entity''.  
There are two remaining problems however. The construction of the conditional probability distribution relies on \cref{thm:pdistconst}. For this to apply we need mutual exclusion of the STPs (in this case the co-perception entities) and we need \cref{eq:nonzeronenner} to hold. In general the set of co-perception entities $\perstp(x_A,t)$ is \textit{not} mutually exclusive. This will be discussed further in \cref{sec:nonint}. In this section we discuss the second problem. Our solution may seem like it includes a strong requirement on the environments. We then show that this requirement is implicit in the perception-action loop as well.

The second problem is the condition of \cref{eq:nonzeronenner} which in the case of co-perception entities concerns the sum over the probabilities of the next time-slices of the co-perception entities $\perstp(x_A,t)$ given the identical past:
\begin{equation}
  \sum_{k \in I}p_{A^k}(x^k_{A^k_{t\prec}}|x_{V_t \bs A_t},x_{A_\pet}).
\end{equation} 
This is not necessarily greater than zero for all environments $x_{V_t \bs A_t} \in \X_{V_t \bs A_t}$. In fact if 
\begin{equation}
  p_{V_t,A_\pet}(x_{V_t \bs A_t},x_{A_\pet})=0
\end{equation} 
then 
\begin{equation}
  p_{A^k}(x^k_{A^k_{t\prec}}|x_{V_t \bs A_t},x_{A_\pet})
\end{equation} 
is not even defined. To be able to use \cref{thm:pdistconst} we must therefore require of environments $x_{V_t \bs A_t}$ that can be classified that 
\begin{equation}
  p_{V_t,A_\pet}(x_{V_t \bs A_t},x_{A_\pet})>0
\end{equation}
and that there is at least one element $x^k_{A^k_{t\prec}} \in \perstp(x_A,t)$ with
\begin{equation}
  p_{A^k_{t\prec},V_t,A_\pet}(x^k_{A^k_{t\prec}},x_{V_t \bs A_t},x_{A_\pet})>0
\end{equation} 
We can summarise these two conditions as the condition that there exists $x^k_{A^k} \in \perstp(x_A,t)$ with
\begin{equation}
\label{eq:coenvcondition}
  p_{A^k,V_t}(x^k_{A^k},x_{V_t \bs A_t})>0
\end{equation} 
where we used that $A^k_\pet=A_\pet$. We can also get rid of the need for the existence quantifier by writing this condition as:
\begin{equation}
\label{eq:copenvcondition2}
  \Pr\left(\bigcup \perstp(x_A,t) \cap \{X_{V_t \bs A_t} = x_{V_t \bs A_t}\}\right) >0.
\end{equation} 
We call the subset of such environments the \textit{co-perception environments}. 
\begin{mydef}[Co-perception environments]
\label{def:copenv}
  Let $\Xv$ be a multivariate Markov chain with $V=J\times T$ and entity set $\Ent$. Let $x_A \in \Ent$ be an entity with non-empty time-slices at $t$ and $t+1$ and $\perstp(x_A,t)$ its co-perception entities. Then define the \textit{associated co-perception environments $\X^\perstp_{V_t \bs A_t} \subseteq \X_{V_t \bs A_t}$} by 
  \begin{equation}
   \label{eq:copenv}
    \X^\perstp_{V_t \bs A_t}:= \{\bar{x}_{V_t \bs A_t} \in \X_{V_t \bs A_t} : \exists y_B \in \perstp(x_A,t), p_{B,V_t\bs A_t}(y_B,\bar{x}_{V_t \bs A_t}) > 0\}.
  \end{equation} 
\end{mydef}
Remark:
\begin{itemize}
  \item The co-perception environments of a co-perception set $\perstp(x_A,t)$ are then the spatial patterns $\X_{V_t \bs A_t}$ at $t$ that can co-occur with at least one co-perception environment. 
\end{itemize}

It may seem like a (too) strong requirement that the co-perception environments are compatible with an entire entity including its (entire) future. This also seems strange from a causal perspective. One way to interpret this is to say that the co-perception environments are just the environments that are ever going to be classified by an entity. Whenever an environment that is not in $\X^\perstp_{V_t \bs A_t}$ occurs at $t$ together with the identical past $x_{A_\pet}$ there will not be any perception since no entity with the identical past will be there at $t+1$. Whenever there is an entity with the identical past at $t+1$ the environment is in $\X^\perstp_{V_t \bs A_t}$.

At the same time this assumption is also implicit in the perception-action loop. Say (as is the case in the perception-action loop) the futures of the co-perception entities are just the possible values of a set $C$ of random variables i.e.\ if there exists $C \subseteq V_{t \prec}$ such that 
\begin{equation}
\perstp(x_A,t)_{t\prec}:=\{y_{B_{t \prec}}: y_B \in \perstp(x_A,t)\} =\X_C.
\end{equation} 
Then futures of the co-perception entities are exhaustive and mutually exclusive. In that case it turns out that it is sufficient to require that the co-perception environments can co-occur with the identical past $x_{A_\pet}$. The requirement that it can co-occur with at least one future of a co-perception entity is then automatically satisfied and vice versa. 
So if $\perstp(x_A,t)_{t\prec} = \X_C$ it is sufficient to define
 \begin{equation}
 \label{eq:copenvtot}
    \X^\perstp_{V_t \bs A_t}:= \{\bar{x}_{V_t \bs A_t} \in \X_{V_t \bs A_t} : p_{A_\pet,V_t\bs A_t}(x_{A_\pet},\bar{x}_{V_t \bs A_t}) > 0\}.
  \end{equation} 
This condition \textit{always} needs to be satisfied for probabilities $p(.|x_{A_\pet},\bar{x}_{V_t \bs A_t})$, that condition on the identical past $x_{A_\pet}$ and environment $\bar{x}_{V_t \bs A_t}$ to be defined.
To see that the two sets of \cref{eq:copenv,eq:copenvtot} are equal recall with \cref{eq:copenvcondition2} that if 
%
$\perstp(x_A,t)_{t \prec}=\X_C$ an environment $\bar{x}_{V_t \bs A_t}$ is in $\X^\perstp_{V_t \bs A_t}$ if 
\begin{equation}
  \Pr\left(\bigcup \X_C \cap \{X_{A_\pet}=x_{A_\pet}\} \cap \{X_{V_t \bs A_t} = \bar{x}_{V_t \bs A_t}\}\right) > 0.
\end{equation} 
Using the mutual exclusion and exhaustiveness of the random variable $X_C$ we note that
\begin{align}
&\Pr\left(\bigcup \X_C \cap \{X_{A_\pet}=x_{A_\pet}\} \cap \{X_{V_t \bs A_t} = \bar{x}_{V_t \bs A_t}\}\right) \\
=&\sum_{\hat{x}_C \in \X_C} \Pr\left( \{X_C =\hat{x}_C\}\cap \{X_{A_\pet}=x_{A_\pet}\} \cap \{X_{V_t \bs A_t} = \bar{x}_{V_t \bs A_t}\}\right) \\
=&\sum_{\hat{x}_C \in \X_C} p_{C,A_\pet,V_t \bs A_t}(\hat{x}_C, x_{A_\pet},\bar{x}_{V_t \bs A_t}) \\
=& p_{A_\pet,V_t \bs A_t}(x_{A_\pet},\bar{x}_{V_t \bs A_t}).
\end{align} 
Where we used mutual exclusion from the second to the third line and exhaustiveness from the fourth to the fifth. So the two sets of \cref{eq:copenv,eq:copenvtot} are identical in this case. Without mutual exclusion and exhaustiveness the two sets are not equal.

More intuitively this can be understood by noting that assuming that the entities exhaust a set of future random variables $X_C$ means that there is a future of a co-perception entity in \textit{every} trajectory compatible with $x_{A_\pet}$. Since all environments that are compatible with $x_{A_\pet}$ must occur in one of those trajectories \cref{eq:copenv} is automatically satisfied. 
Our definition of co-perception environments therefore does not contain an additional assumption compared to cases where futures of entities exhaust sets of random variables like in the case of the perception-action loop or the models of biological individuals of \citet{krakauer_information_2014}. 



%

\subsection{Non-interpenetration and mutual exclusion}
\label{sec:nonint}
In this section we define the formal assumptions of general non-interpenetration and (past specific) non-interpenetration for entity sets. This leads to mutual-exclusion of entities in ways that enable a unique definition of our notion of entity perception. Without these assumptions we can still define entity perception but there will an arbitrary choice involved which influences the perceptions (we will see this in \cref{sec:branchmorph}).

So different choices mean that the extracted perceptions are different. This is not a desirable situation since we are making this choice. The goal of this thesis is, however, that all the notions only depend on the multivariate Markov chain itself. Non-interpenetration is therefor a desirable property.

General non-interpenetration requires that any two STPs $x_A,y_B$ that partly occupy the same random variables i.e.\ 
\begin{equation}
  A \cap B \neq \emptyset
\end{equation} 
never co-occur i.e.\
\begin{equation}
  \Pr(X_A=x_A,X_B=y_B)=0.
\end{equation} 
This notion of non-interpenetration treats spatial and temporal overlap (we call $A \cap B$ the \textit{overlap}) equally. There is a philosophical debate about whether interpenetration is possible for real objects (see e.g.\ \citet{sep-location-mereology}). Here we only want to suggest that non-interpenetration may be a reasonable assumption for entities. Non-interpenetration says that the same spatiotemporal region cannot be occupied by two different entities. This is intuitively true for solid objects. It is however somewhat problematic when we think of multicellular organisms as entities that may contain cells that are also entities by themselves. In that case these cells would be parts of two different entities and violate non-interpenetration. It is also possible that the cells by themselves are actually not entities according to some other entity criterion. We also note that on the level of cells we do not have interpenetration. Two different cells never occupy the same spatiotemporal region. A cell may divide, but that is one spatiotemporal entity dividing \textit{spatially} into two at some point in time. This suggests that there may be some levels of organisation or hierarchies\footnote{Whether the disintegration hierarchies are related is beyond the scope of this thesis but an interesting avenue to pursue in the future.} of entities involved. We will not further discuss this here. Instead we only note that the formal property of general non-interpenetration 
relates to our notion of perception by providing mutual exclusion of co-perception entities. We now state the definition of general non-interpenetration for future reference.
\begin{mydef}[General non-interpenetration]
  Let $\Xv$ be a multivariate Markov chain with $V=J\times T$. An entity set $\Ent \subseteq \bigcup_{B \subseteq V} \X_B$ \textit{satisfies general non-interpenetration or is generally non-interpenetrating} if for all $y_B,z_C \in \Ent$ with $y_B \neq z_C$ we have
  \begin{equation}
    B \cap C \neq \emptyset \Rightarrow \Pr(X_B=y_B,X_C=z_C)=0.
  \end{equation} 
\end{mydef}
It turns out that general non-interpenetration is not necessary for our purposes. We require only that all co-perception entities are mutually exclusive. For this it is already sufficient that for two entities $x_A,y_B$ if there is a time $t \in T$ such that they have identical \textit{pasts} up to $t$ 
\begin{equation}
x_{A_\pet} = y_{B_\pet}
\end{equation} 
 but are then different at some time in the future 
 \begin{equation}
   x_{A_{t\prec}} \neq y_{B_{t\prec}}
 \end{equation} 
 must be mutually exclusive \textit{given that their pasts occurred}:
\begin{equation}
  \Pr(X_{A_{t\prec}}=x_{A_{t\prec}},X_{B_{t\prec}}=y_{B_{t\prec}}|x_{A_\pet})=0.
\end{equation} 
%
This means that there cannot be two different entities which are identical up to some point in time $t$ and then, in the same \textit{single} trajectory (with positive probability), at some point ``reveal'' their difference. If entities with identical pasts \textit{ever} reveal their difference they must be in different trajectories i.e.\ they must be mutually exclusive. We could call this ``past specific non-interpenetration'' but since we only need this notion outside of this section we will just refer to it as non-interpenetration.
\begin{mydef}[Non-interpenetration]
\label{def:noninterpen}
  Let $\Xv$ be a multivariate Markov chain with $V=J\times T$. An entity set $\Ent \subseteq \bigcup_{B \subseteq V} \X_B$ \textit{satisfies non-interpenetration or is non-interpenetrating} if for all $y_B,z_C \in \Ent$ we have
  \begin{equation}
    \begin{split}\exists t \in T : y_{B_\pet} &= z_{C_\pet} \text{ and }   y_{B_{t\prec}} \neq z_{C_{t\prec}} \\
&\Rightarrow   \Pr(X_{B_{t\prec}}=y_{B_{t\prec}},X_{B_{t\prec}}=z_{C_{t\prec}}|y_{B_\pet})=0.
    \end{split}
  \end{equation} 
\end{mydef}
Remark:
\begin{itemize}
  \item We note here that non-interpenetration is not necessarily satisfied by $\ci$-entities as we will see in \cref{sec:actperceptexamples}.
\end{itemize}

Non-interpenetration implies that co-perception entities are mutually exclusive:
\begin{thm}
\label{thm:nonintmuex}
  Let $\Xv$ be a multivariate Markov chain with $V=J\times T$ and entity set $\Ent$. Let $x_A \in \Ent$ be an entity with non-empty time-slices at $t$ and $t+1$ and $\perstp(x_A,t)$ its co-perception entities. If $\Ent$ satisfies non-interpenetration then $\perstp(x_A,t)$ is mutually exclusive.
\end{thm}
\begin{proof}
  Let $y_B,z_C \in \perstp(x_A,t)$ with $y_B \neq z_C$. Then they have identical pasts and so we have $y_{B_\pet} = z_{C_\pet}$. From non-interpenetration we then get 
  \begin{equation}
    \Pr(X_B=y_B,X_C=z_C)=0.
  \end{equation} 
%
\end{proof}
Remark:
\begin{itemize}
  \item Note that non-interpenetration does not imply anything about exhaustiveness. We can have non-interpenetrating co-perception entities that are not exhaustive. But since we know how to define a conditional probability distribution for non-exhaustive sets of STPs (\cref{thm:pdistconst}) this is not a problem.
\end{itemize}

This means that under non-interpenetration we can always define a conditional probability distribution over the entire futures of a set $\perstp(x_A,t)$ of co-perception entities. 
\begin{mydef}[Co-perception morph]
 Let $\Xv$ be a multivariate Markov chain with $V=J\times T$ and a non-interpenetrating entity set $\Ent$. Let $x_A \in \Ent$ be an entity with non-empty time-slices at $t$ and $t+1$ and $\perstp(x_A,t)=\{x^k_{A^k}\}_{k \in I}$ its co-perception entities. Furthermore let $\X^\perstp_{V_t \bs A_t} \subseteq \X^\perstp_{V_t \bs A_t}$ be the set of co-perception environments.
%
Then the conditional probability distribution $p_\perstp:\perstp \rightarrow [0,1]$
\begin{align}
  p_{\perstp(x_A,t)}(k|x_{V_t \bs A_t},x_{A_\pet}):&= \frac{p_{A^k}(x^k_{A^k_{t\prec}}|x_{V_t \bs A_t},x_{A_\pet})}{\sum_{l \in I}p_{A^l}(x^l_{A^l_{t\prec}}|x_{V_t \bs A_t},x_{A_\pet})}
\end{align}
is well defined and we call it the \textit{co-perception morph}.
\end{mydef}

The co-perception morph is a conditional probability distribution over the entire futures of the co-perception entities. It can be used to partition the co-perception environments by assigning environments to the same block / perception if the lead to the same co-perception morph. However, we wanted to define perception that occurs from $t$ to $t+1$. In the co-perception morph there may be entities that only start differing far in the future. If two environments have different influences only on these two entities and equal influence on all other co-perception entities the co-perception morph still distinguishes the two environments at $t$ already. This does not seem like a good definition of perception from $t$ to $t+1$. We therefore partition all co-perception entities according to their next time-slices in the next section.


%

\subsection{Branching partition}
In this section we will present a partition of the co-perception entities called the branching partition. This will put all entities into the same blocks (called \textit{branches}) that have identical time-slices at $t+1$. 
This construction is intuitive as it ignores differences between co-perception entities that only become apparent at times later than $t+1$. 
Since we are interested in the perceptions that happen in the transition from $t$ to $t+1$ such differences should be ignored. The branches (blocks) of the branching partition are therefore an even better analogue to the ``possible next time-slices'' than the set of co-perception entities directly. 

First some more detail: at each transition from time-step $t$ to $t+1$ the co-perception entities $\perstp(x_A,t)$ split up into sets of entities that are identical up to $t+1$ (we will call these sets the \textit{branches}). Only one of these sets is the set $\perstp(x_A,t+1)$. For example an entity $y_B \in \perstp(x_A,t)$ with the same past up to $t$ but with a different time-slice at $t+1$ i.e.\ $y_{B_{t+1}} \neq x_{A_{t+1}}$ is part of a different branch. In that case this branch is $\perstp(y_B,t+1)$ and we have $\perstp(y_B,t+1) \cap \perstp(x_A,t+1) = \emptyset$. In summary then the dynamics of the system split up the co-perception entities of $x_A$ up to $t$ into disjoint sets (the branches) of entities with identical pasts up to $t+1$. We can then interpret the branches at the time $t+1$ as the distinctions among the co-perception entities that are revealed at time $t+1$. Further distinctions among the co-perception entities are only revealed at later times. This also means that these are \textit{all} differences that could possibly be due to the influence of the environment at $t$ and that show their effect at $t+1$ (not later). In this way the perceptions at $t$ should also be defined with respect to these branches. We call the partition that is defined via the identification of entities in $\perstp(x_A,t)$ that are identical up to $t+1$ the \textit{branching partition}. 
\begin{mydef}[Branching partition]
\label{def:compart}
Let $\Xv$ be a multivariate Markov chain with $V=J\times T$ and entity set $\Ent$. Let $x_A \in \Ent$ be an entity with non-empty time-slices at $t$ and $t+1$ and $\perstp(x_A,t)$ its co-perception entities. 
Then define the \textit{branching partition $\eta(x_A,t)$ of $\perstp(x_A,t)$} as the partition induced by the equivalence classes of the equivalence relation
   \begin{equation}
    \begin{split}y_B \sim & z_C \\
    &\Leftrightarrow  y_{B_{t+1}}=z_{C_{t+1}},
    \end{split}
  \end{equation} 
where $y_B,z_C \in \perstp(x_A,t)$.
\end{mydef}
Remark:
\begin{itemize}
  \item The definition of the branching partition can easily be generalised to more than one time-step into the future. Instead of requiring equality at $t+1$ we can require equality for the next $r$ time-steps:
   \begin{equation}
    \begin{split}y_B \sim & z_C \\
    &\Leftrightarrow y_{B_{t+1:t+r}}=z_{C_{t+1:t+r}}
    \end{split}
  \end{equation} 
  This leads to a partition of $\perstp(x_A,t)$ which is a refinement of $\eta(x_A,t)$. The branches of $t+1$ are further partitioned according to the equality of the contained entities at $t+2$, and similarly at each subsequent time-step. This may be used to construct a kind of multi-time-step perception which is more precise than one-time-step perception. Here we focus only on the one-step-perception, nonetheless all further notions are easily adapted to the multi-step case.
\end{itemize}
The branches of the branching partition are the final analogue notion of ``the entity's next possible time-slices''. We then define a conditional probability distribution over the branches (called the branch-morph) and classify the co-perception environments accordingly.

\subsection{Branch-morph}
\label{sec:branchmorph}
Given the branching partition $\eta(x_A,t)$ for a non-interpenetrating entity set we can then define a conditional probability distribution over the branches by just summing up the probabilities of all entities in each branch (remember that they are all mutually exclusive) to get the probability of a branch i.e.\ write for every block $b \in \eta(x_A,t)$:
 \begin{equation}
   p(b|\hat{x}_{V_t \bs A_t},x_{A_{\preceq t}}):= \sum_{y_B \in b} p_{B_{t\prec},V_t\bs A_t}(y_{B_{t\prec}}|\hat{x}_{V_t \bs A_t},x_{A_{\preceq t}}).
 \end{equation} 
Since the branches are also mutually exclusive (because all co-perception entities are mutually exclusive) we can divide by the sum of probabilities of the branches to get a probability distribution. 
\begin{equation}
   p_{\eta(x_A,t)}(b|\hat{x}_{V_t\bs A_t},x_{A_{\preceq t}}):=\frac{p(b|\hat{x}_{V_t \bs A_t},x_{A_{\preceq t}})}{\sum_{c \in \eta(x_A,t)} p(c|\hat{x}_{V_t \bs A_t},x_{A_{\preceq t}})}.
 \end{equation} 

This is the idea behind \cref{def:branchmorph} below. 

However, 
%
we do not necessarily need the mutual exclusion of the \textit{entire} set of co-perception entities if we want to define \textit{some} perception. 
For this we can use a subset $\zeta(x_A,t) \subset \perstp(x_A,t)$ with $x_A \in \zeta(x_A,t)$ of mutually exclusive co-perception entities. We can then still use the branching partition on this subset and define perception in the same way as below by replacing $\perstp(x_A,t)$ with $\zeta(x_A,t)$ everywhere.

 
 However, for a set of co-perception entities that is not naturally mutually exclusive as in the case of non-interpenetration there are many possible choices of such subsets which lead to different perceptions. For example say $\perstp(x_A,t)=\{x_A,y_B,z_C\}$. Then if they are not all mutually exclusive, we can have that $x_A$ and $y_B$ are mutually exclusive and $x_A$ and $z_C$ are mutually exclusive but $y_B$ and $z_C$ are \textit{not} mutually exclusive. So then to get a proxy $\zeta(x_A,t)$ of $\perstp(x_A,t)$  (consisting only of mutually exclusive co-perception entities) we can either choose $\zeta(x_A,t) = \{x_A,y_B\}$ or $\zeta(x_A,t) =\{x_A,z_C\}$. These choices will in general lead to different perceptions such that perception is not unique if we don't have mutual exclusion of all co-perception entities. So non-interpenetration is an attractive property for entity sets if we want to use our notion of perception since it allows us to uniquely define it.
 
 We now finally state the definition of the branch-morph, the co-perception environment partition and the perceptions which are just the blocks of the co-perception environment partition.
%
%
%
%
%
%
%

\begin{mydef}[Branch-morph]
\label{def:branchmorph}
 Let $\Xv$ be a multivariate Markov chain with index set $V=J\times T$ and entity set $\Ent$. Let $x_A \in \Ent$ be an entity with non-empty time-slices at $t$ and $t+1$ and $\perstp(x_A,t)$ its co-perception entities and $\eta(x_A,t)$ the branching partition.
 Furthermore, let $\X^\perstp_{V_t \bs A_t} \subseteq \X_{V_t \bs A_t}$ be the associated co-perception environments. Also write for every block $b \in \eta(x_A,t)$:
 \begin{equation}
 \label{eq:beforebranchmorph}
   p(b|\hat{x}_{V_t \bs A_t},x_{A_{\preceq t}}):= \sum_{y_B \in b} p_{B_{t\prec},V_t\bs A_t}(y_{B_{t\prec}}|\hat{x}_{V_t \bs A_t},x_{A_{\preceq t}}).
 \end{equation} 
  Then for each $\hat{x}_{V_t \bs A_t} \in \X^\perstp_{V_t \bs A_t}$ we define the \textit{branch-morph} over $\eta(x_A,t)$ as the probability distribution $p_{\eta(x_A,t)}(.|\hat{x}_{V_t\bs A_t},x_{A_{\preceq t}}):\eta(x_A,t) \rightarrow [0,1]$ with 
 \begin{equation}
 \label{eq:branchmorph}
   p_{\eta(x_A,t)}(b|\hat{x}_{V_t\bs A_t},x_{A_{\preceq t}}):=\frac{p(b|\hat{x}_{V_t \bs A_t},x_{A_{\preceq t}})}{\sum_{c \in \eta(x_A,t)} p(c|\hat{x}_{V_t \bs A_t},x_{A_{\preceq t}})},
 \end{equation} 
 for all $b \in \eta(x_A,t)$.
\end{mydef}

With the branch-morph we can then define, as expected, the perceptions as equivalence classes of the co-perception environments with respect to the associated branch-morph. First we define a partition of the co-perception \textit{environments} called the co-perception environment partition. The perceptions are then the blocks of this partition.

\begin{mydef}
Let $\Xv$ be a multivariate Markov chain with index set $V=J\times T$ and entity set $\Ent$. Let $x_A \in \Ent$ be an entity with non-empty time-slices at $t$ and $t+1$ and $\perstp(x_A,t)$ its co-perception entities and $\eta(x_A,t)$ the branching partition.
 Furthermore, let $\X^\perstp_{V_t \bs A_t} \subseteq \X_{V_t \bs A_t}$ be the associated co-perception environments.
  Then define the \textit{co-perception environment partition $\pi^\perstp(x_A,t)$ of $\X^\perstp_{V_t \bs A_t}$} as the partition induced by the equivalence classes of the equivalence relation
  \begin{equation}
    \begin{split}\hat{x}_{V_t \bs A_t} \sim & \bar{x}_{V_t \bs A_t} \\
    &\Leftrightarrow \forall b \in \eta(x_A,t):p_{\eta(x_A,t)}(b|\hat{x}_{V_t\bs A_t},x_{A_{\preceq t}}) = p_{\eta(x_A,t)}(b|\bar{x}_{V_t\bs A_t},x_{A_{\preceq t}}).
    \end{split}
  \end{equation} 
\end{mydef}
Remarks:
\begin{itemize}
  \item This means all associated co-perception environments in the same block of $\pi^\perstp(x_A,t)$ have the same branch-morph. In other words they lead to the same branch of entity futures (i.e.\ the same future branch) with the same probabilities. Then all elements of these environment blocks have identical effects on the future branches and these branches cannot distinguish between environments within the blocks.
\end{itemize}

\begin{mydef}[Perceptions]
Let $\Xv$ be a multivariate Markov chain with index set $V=J\times T$ and entity set $\Ent$. Let $x_A \in \Ent$ be an entity with non-empty time-slices at $t$ and $t+1$ and $\perstp(x_A,t)$ its co-perception entities.
 Furthermore, let $\X^\perstp_{V_t \bs A_t} \subseteq \X_{V_t \bs A_t}$ be the associated co-perception environments and $\pi^\perstp(x_A,t)$ its co-perception environment partition.

 Then the blocks of $\pi^\perstp(x_A,t)$ are called the \textit{perceptions of $x_A$ at $t$}.

\end{mydef}

\section{Entity action and perception in the perception-action loop}
\label{sec:palooprelation}
%
%
%
%

We now show that agent-environment systems as modelled by the perception-action loop are multivariate Markov chains containing a specific choice of entity sets. 

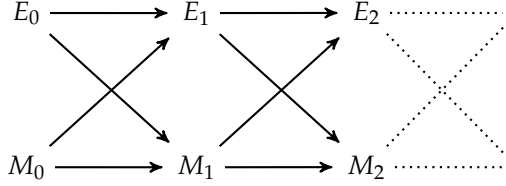
\begin{figure}
\begin{center}
  \begin{tikzpicture}

  \matrix (m) [matrix of math nodes, nodes in empty cells,row sep=1.5cm,column sep=1.5cm]
  {
     E_0 &      E_1 &      E_2 & \vphantom{E_3} \\
     M_0 &      M_1 &      M_2 & \vphantom{M_3} \\
   };
  \path[myarrow]
    (m-1-1) edge node {} (m-2-2)
            edge node {} (m-1-2)
    (m-2-1) edge node {} (m-2-2)
            edge node {} (m-1-2)
    

    (m-1-2) edge node {} (m-2-3)
            edge node {} (m-1-3)
    (m-2-2) edge node {} (m-2-3)
            edge node {} (m-1-3)
            
    (m-1-3) edge[-,dotted] node {} (m-2-4)
            edge[-,dotted] node {} (m-1-4)
    (m-2-3) edge[-,dotted] node {} (m-2-4)
            edge[-,dotted] node {} (m-1-4)

%
    ;            
\end{tikzpicture}
  \caption{First time-steps of the Bayesian network of the perception-action loop. 
The processes represent environment $\Et$, 
and agent memory $\Mt$. 
}
  \label{fig:palooprep}
\end{center}
\end{figure}

In this section we interpret the perception-action loop from the perspective of our own concept of agents. We identify an entity set and the subset of the entity set that are agents from the perception-action loop perspective. Both of these choices turn out to be nonrestrictive. The perception-action loop itself puts no strong constraints on the ``agents'' (often only referred to as \textit{systems}) it models. Its Bayesian network structure of \cref{fig:palooprep} is also compatible with two independent and identically distributed random variables. Accordingly there have been attempts to identify further restrictions or measures that quantify further distinctions among such systems. 
The most relevant in our context is the measure of \textit{autonomy} for an agent in the perception-action loop \citep{bertschinger_autonomy_2008}. We show how our notion of actions is sufficient for the property of non-heteronomy which is part of the requirement for autonomy. We also show that, as expected due to our construction, our notion of perception specialises to the notion of  perception in the perception action (\cref{sec:paloopformal}). The main point of this chapter is to show how our notion of an acting and perceiving entity can be seen as a generalisation of (partly autonomous) agents as they are modelled in perception-action loops. Our notions are more general because they are well defined for non-exhaustive and extent varying entity-sets.

%

%

Recall that an agent in a perception-action loop (\cref{def:paloop}) is a stochastic process $\Mt$ interacting with an environment process $\Et$. If we want to make the interactions explicit we can use the extended perception-action loop of \cref{def:expaloop}. In this case we also have an action process $\At$ and a sensor process $\St$. For convenience we again show the Bayesian network (of the non-extended version) in \cref{fig:palooprep}.

In the perception-action loop each trajectory $x_V$ is considered to consist of a time-evolution $m_T$ of the agent and a time-evolution of the environment $e_T$. The agent therefore occurs in every trajectory and occupies the same degree of freedom in every trajectory. According to our working definition (\cref{def:agent}) all agents are entities and according to our concept of entities (\cref{def:entset}), entities are STPs. Each of the time-evolutions $m_T$ is a STP in the perception-action loop. We can then define the entity set $\Ent$ of a perception-action loop as the set of time-evolutions of the agent process i.e.\
\begin{equation}
  \Ent^\pal:=\{m_T \in \prod_{t \in T} \M_t\}.
\end{equation}
With this definition every time-evolution of the agent process corresponds to an entity and every such entity corresponds to the time-evolution of an agent (or \textit{the} agent). Similarly, we can define entities for the environments and add them to $\Ent^\pal$. The symmetry of the perception-action loop makes no a priori difference between agents and environments. We will focus on the agent process here and do not need environment entities. 

Compared to $\ci$-entities (\cref{def:cientset}) the entity set $\Ent^\pal$ is not very restrictive. Even if each $M_t$ for $t \in T$ is a independently distributed random variable the sequences $m_T$ would still be considered time evolutions of entities. Furthermore, each of these entities would be considered an agent in this picture. In order to introduce stronger conditions efforts have been made to distinguish informationally closed \citep{bertschinger_information_2006} and autonomous \citep{bertschinger_autonomy_2008} agents. Our own concept of agents also puts stronger constraints on the notion of an entities. We require actions, perception, and goal-directedness from entities that can be counted as agents. We have not defined a notion of goal-directedness, but the notions of entity actions (\cref{sec:actions}) and entity perceptions (\cref{sec:perceptions}) can be used for the perception-action loop entity set $\Ent^\pal$. We will consider this next and see that our more general requirements of actions corresponds to a requirement by \citeauthor{bertschinger_autonomy_2008} in the case of perception-action loop entity sets.

\subsection{Entity actions in the perception-action loop}

According to our definition (\cref{def:action}) an entity $x_A$ performs an action at time $t$ in a trajectory $x_V$ if there is a co-action entity $y_B$ occurring in a co-action trajectory $y_V$ with $x_{V_t \bs A_t}=y_{V_t \bs B_t}$. In the case of the perception action loop we can write every trajectory as a pair $(m_T,e_T)$ where $m_T$ is an entity. The entity $m_T$ then performs an action at time $t$ in trajectory $(m_T,e_T)$ with $p_{M_T,E_T}(m_T,e_T)>0$ if there is an entity $\bar{m}_t$ such that
\begin{itemize}
   \item $\bar{m}_T$ occurs in $(\bar{m}_T,\bar{e}_T) \neq (m_T,e_T)$ with $p_{M_T,E_T}(\bar{m}_T,\bar{e}_T)>0$, 
   \item at $t$ the entities $m_T$ and $\bar{m}_T$ occupy the same random variables, which is the case for all entities in $\Ent^\pal$,
   \item at $t$ the environments of $m_T$ and $\bar{m}_T$ are identical: $e_t = \bar{e}_t$,
   \item at $t+1$ the entities are different: $m_{t+1} \neq \bar{m}_{t+1}$.
\end{itemize}
Since all entities occupy the same random variables we can only have value actions in the perception-action loop.

We now show that these conditions can be related to conditions for autonomous systems/agents proposed by \citet{bertschinger_autonomy_2008}. We can say that the more entities perform actions at $t$ the higher is potentially the non-heteronomy component of a measure of autonomy proposed by \citeauthor{bertschinger_autonomy_2008}.

If we assume that these conditions are fulfilled at some time $t$ for two entities $m_T,\bar{m}_T$ we can derive that the conditional entropy $\HS(M_{t+1}|E_t)$ of the next agent state given the current environment state is greater than zero:
\begin{equation}
  \HS(M_{t+1}|E_t) > 0.
\end{equation}
\begin{proof}
  From $p_{M_T,E_T}(m_T,e_T)>0$ and $p_{M_T,E_T}(\bar{m}_T,\bar{e}_T)>0$ it directly follows that $p_{M_{t+1}}(m_{t+1}|e_t)>0$, $p_{M_{t+1}}(\bar{m}_{t+1}|e_t)>0$ and $p_{E_t}(e_t)>0$. Then:
  \begin{align}
    \HS(M_{t+1}|E_t):&=-\sum_{\hat{e}_t\in \E_t} p_{E_t}(\hat{e}_t) \sum_{\hat{m}_{t+1}\in \M_{t+1}} p_{M_{t+1}}(\hat{m}_{t+1}|\hat{e}_t) \log p_{M_{t+1}}(\hat{m}_{t+1}|\hat{e}_t) \label{eq:nonhet}\\
    \begin{split}&\geq -p_{E_t}(e_t) \left(p_{M_{t+1}}(m_{t+1}|e_t) \log p_{M_{t+1}}(m_{t+1}|e_t)\right.\\
&\phantom{\geq -p_{E_t}(e_t) \left(\right.}\left.+ p_{M_{t+1}}(\bar{m}_{t+1}|e_t) \log p_{M_{t+1}}(\bar{m}_{t+1}|e_t)\right)
    \end{split}\\
    &>0.
  \end{align}
\end{proof}
We can also see from this that the more entities perform actions at $t$ the more terms in \cref{eq:nonhet} are positive. The final value of $\HS(M_{t+1}|E_t)$ depends on the actual probabilities but the maximum value for $n$ positive terms is $\log n$. So the more different co-action entities there are for a time $t$ the higher the conditional entropy $\HS(M_{t+1}|E_t)$ can get. Also note that if there are no actions at $t$ i.e.\ no co-action entity in no co-action trajectory at $t$ then $\HS(M_{t+1}|E_t)=0$. Entity actions of entities in $\Ent^\pal$ are therefore necessary and sufficient for $\HS(M_{t+1}|E_t)>0$.

The conditional entropy $\HS(M_{t+1}|E_t)$ measures the uncertainty about the next agent state when the current environment state is known. It has been proposed as a measure of non-heteronomy in \citet{bertschinger_autonomy_2008}. Non-heteronomy means that the agent is not determined by the history of the environment. We only treat here the case where the history length is just one time-step $E_t$ but generalisations to multiple time-steps $\HS(M_{t+1}|E_{t-l:t})$ are straightforward (see remark to \cref{def:action}). We have argued above that non-heteronomy at time $t$ depends on the existence of entity actions performed at $t$ and is limited by the number of different such actions at $t$. The entity actions as we have defined them are therefore like building blocks that make up the non-heteronomy of a stochastic process. In other words they are a local version of non-heteronomy in the case of value actions. Furthermore our definition applies to extent actions as well. 

The measure of autonomy proposed by \citet{bertschinger_autonomy_2008} contains another component measuring self-determination i.e.\ the degree to which the current agent state determines its next state. This is not ensured by our definition of actions. If $M_{t+1}$ is an independently and uniformly distributed random variable then there are $|\M_{t+1}|$ co-actions at $t$ and $\HS(M_{t+1}|E_t)=\HS(M_{t+1})=\log |\M_{t+1}|$. The role of ensuring self-determination in our case is delegated to the entity set. The perception-action loop entity set $\Ent^\pal$ is too nonrestrictive for this purpose. Our notion of $\ci$-entities on the other hand would not count a value $m_{t+1}$ of an independently distributed random variable as part of a larger entity 
. 

We expect that there are multiple action definitions that coincide with conditions similar to autonomy/non-heteronomy in the case of perception-action loop entities. An example would be requiring that the environments at $t+1$ differ in some way. We have only presented one definition here. Investigating and comparing further possible notions is future work.

We note that \citet{ikegami_uncertainty_1998} propose to use possible/compatible counterfactual trajectories of game players as signs of autonomy. This idea is similar to ours. We construct the capability to act from the counterfactual trajectories. Actions are arguably the basic units that realise autonomy over a longer period of time. In this sense we have in fact used the counterfactual trajectories to allow for a kind of autonomy. 

\subsection{Entity perception in the perception-action loop}

We now look at how entity perception as defined in \cref{sec:perceptions} specialises to the case of the perception-action loop. This argument in effect constitutes a proof that our \cref{def:branchmorph} of the branch-morph is a generalisation of the conditional probability distributions $p_{M_{t+1}}(.|m_t,e_t):\M_{t+1} \rightarrow [0,1]$ to non-interpenetrating, co-perception entities that (in contrast to the case of the perception-action loop) do not exhaust a set of future random variables and may exhibit counterfactual variation in extent. This result is not surprising since we set out to do just this but it is also instructive to work through the recovery of the original expression of the conditional probability distribution starting from the general branch-morph.


We pick an entity $m_T$ from the entity set $\Ent^\pal$ and consider its perceptions at an arbitrary time-step $t\in T$. In order to get the perceptions at $t$ we need 
\begin{enumerate}
  \item the co-perception entities $\perstp(m_T,t)$ of $m_T$ at $t$,
  \item the branching partition $\eta(m_T,t)$ with its branches,
  \item the co-perception environments,
  \item the branch-morphs for each environment,
  \item and the co-perception environment partition $\pi^\perstp(x_A,t)$ with its blocks, the perceptions.
\end{enumerate}
These can be identified in the following way.
\begin{enumerate} 
  \item 
The co-perception entities $\perstp(m_T,t)$ are the entities in $\Ent^\pal$ that have non-empty time-slices at $t,t+1$, and that are identical to $m_T$ up to $t$. All entities in $\Ent^\pal$ have non-empty time slices at all times. So we have:
\begin{equation}
  \perstp(m_T,t)=\{\bar{m}_T \in \Ent^\pal:\bar{m}_{\preceq t}=m_{\preceq t}\} \\
\end{equation} 
Note that as $t$ increases there are less and less co-perception entities. At $t=n-1$ (recall that $T=0:n-1$) we eventually have $\perstp(m_T,t)=\{m_T\}$. Also note that the futures of the co-perception entities exhaust the future random variables $M_{t \prec}$ i.e.\
\begin{equation}
  \perstp(m_T,t)_{t \prec} = \{\bar{m}_{t \prec}: \bar{m}_T \in \perstp(m_T,t)\} = \M_{t \prec}.
\end{equation} 

\item
First recall that the entity set $\Ent^\pal$ satisfies non-interpenetration since they all occupy the same set $\Mt$ of random variables. Therefore $\perstp(m_t,t)$ is mutually exclusive and we get unique perception via the branching partition $\eta(m_T,t)$ of the entire set $\perstp(m_t,t)$. The branching partition $\eta(m_T,t)$ is composed out of blocks (the branches) of co-perception entities that 
%
are identical up to $t+1$ i.e.\  
   \begin{equation}
    \begin{split}\hat{m}_T \sim & \bar{m}_T \\
    &\Leftrightarrow \hat{m}_{t+1}=\bar{m}_{t+1}.
    \end{split}
  \end{equation} 
We can therefore identify the blocks of $\eta(m_T,t)$ i.e.\ the future branches by the values that the entities take at $t+1$. 
Define the branch $b(\bar{m}_{t+1})$ associated to $\bar{m}_{t+1} \in \M_{t+1}$ via 
\begin{equation}
\label{eq:pabranches}
  b(\bar{m}_{t+1}):=\{\hat{m}_T \in \perstp(m_T,t) : \hat{m}_{t+1} = \bar{m}_{t+1}\}.
\end{equation}
The branching partition is then: 
\begin{equation}
  \eta(m_T,t)= \{b(\bar{m}_{t+1})\subseteq \perstp(m_T,t): \bar{m}_{t+1} \in \M_{t+1}\}. 
\end{equation} 

\item
The co-perception environments are the STPs $x_{V_t \bs A_t}$ compatible with at least one co-perception entity. For the perception-action loop and entity $m_T$ at $t$ we have $\X_{V_t \bs A_t}=\E_t$ and therefore $\X^\perstp_{V_t \bs A_t}= \E^\perstp_t$. Where $\E^\perstp_t$ is 
\begin{equation}
  \E^\perstp_t= \{e_t \in \E_t:\exists \bar{m}_T \in \perstp(m_T,t), p_{M_T,E_t}(\bar{m}_T,e_t)>0\}.
\end{equation} 
As we have noted in \cref{sec:copenv} since the co-perception entities exhaust $M_{t\prec}$ this requirement is equivalent to
\begin{equation}
  \E^\perstp_t= \{e_t \in \E_t: p_{M_t,E_t}(\bar{m}_t,e_t)>0\}.
\end{equation} 

\item
The branch-morphs are the probability distributions $p_{\eta(m_T,t)}(.|e_t,m_\pet):\eta(m_T,t)\rightarrow [0,1]$ over the branches for each co-perception environment $e_t \in \E^\perstp_t$. These are defined using \cref{eq:beforebranchmorph} which for the perception-loop becomes
\begin{equation}
   p(b(\bar{m}_{t+1}),e_t|m_\pet):= \sum_{\hat{m}_T \in b(\bar{m}_{t+1})} p_{M_{t\prec},E_t}(\hat{m}_{t\prec},e_t|m_\pet).
 \end{equation} 
We can rewrite the sum on the right hand side using \cref{eq:pabranches} for $b(\bar{m}_{t+1})$ and then $\perstp(m_T,t)_{t \prec}=\M_{t \prec}$:
\begin{align}
 p(b(\bar{m}_{t+1}),e_t|m_\pet) &= \sum_{\{\hat{m}_T \in \perstp(m_T,t) : \hat{m}_{t+1} = \bar{m}_{t+1}\}} p_{M_{t\prec},E_t}(\hat{m}_{t\prec},e_t|m_\pet)\\
&= \sum_{\{\hat{m}_{t \prec} \in \M_{t \prec} : \hat{m}_{t+1} = \bar{m}_{t+1}\}} p_{M_{t\prec},E_t}(\hat{m}_{t\prec},e_t|m_\pet)\\
  &=\sum_{\hat{m}_{t+1 \prec} \in \M_{t+1 \prec}} p_{M_{t\prec},E_t}(\bar{m}_{t+1},\hat{m}_{t+1 \prec},e_t|m_\pet)\\
  &=p_{M_{t+1},E_t}(\bar{m}_{t+1},e_t|m_\pet).
\end{align}
The definition of the branch-morph for the perception-action loop is
\begin{align}
   p_{\eta(m_T,t)}(b(\bar{m}_{t+1})|e_t,m_\pet):&=\frac{p(b(\bar{m}_{t+1}),e_t|m_\pet)}{\sum_{b \in \eta(m_T,t)} p(b,e_t|m_\pet)} \\
 \end{align}
 which we can rewrite now
\begin{align}
   p_{\eta(m_T,t)}(b(\bar{m}_{t+1})|e_t,m_\pet)&=\frac{p_{M_{t+1},E_t}(\bar{m}_{t+1},e_t|m_\pet)}{\sum_{\hat{m}_{t+1} \in \M_{t+1}} p_{M_{t+1},E_t}(\hat{m}_{t+1},e_t|m_\pet)} \\
   &=\frac{p_{M_{t+1},E_t}(\bar{m}_{t+1},e_t|m_\pet)}{p_{E_t}(e_t|m_\pet)} \\
   &=p_{M_{t+1}}(\bar{m}_{t+1}|e_t,m_\pet) \\
   &=p_{M_{t+1}}(\bar{m}_{t+1}|e_t,m_t).
 \end{align}
In the last line we used the Bayesian network of the perception-action loop. 

\item The co-perception environment partition $\pi^\perstp(m_T,t)$ of $\E^\perstp_t=$ is the partition induced by the equivalence classes of the equivalence relation
 \begin{equation}
    \begin{split}\hat{e}_t \sim & \bar{e}_t \\
    &\Leftrightarrow \forall b \in \eta(m_T,t):p_{\eta(m_T,t)}(b|\hat{e}_t,m_{\preceq t}) = p_{\eta(m_T,t)}(b|\bar{e}_t,m_{\preceq t}).
    \end{split}
  \end{equation} 
Using the branch-morph above this is equivalent to
 \begin{equation}
    \begin{split}\hat{e}_t \sim & \bar{e}_t \\
    &\Leftrightarrow \forall m_{t+1} \in \M_{t+1}:p_{M_{t+1}}(m_{t+1}|\hat{e}_t,m_t) = p_{M_{t+1}}(m_{t+1}|\bar{e}_t,m_t)
    \end{split}
  \end{equation} 
which is just the equivalence relation of \cref{eq:perceptionequiv} used to extract the sensor-values in \cref{sec:paloopformal}.

%
\end{enumerate}
So we have seen that our definitions of \cref{sec:perceptions} specialise in the case of the perception-action loop to the same concept of perception as in \cref{sec:paloopformal}. 

More interesting for future research is that the branch-morphs are generalisations of the conditional probability distribution $p_{M_{t+1}}(.|e_t,m_t)$. These conditional probability distributions play a role in various information theoretic concepts formulated for the perception-action loop.  Examples include informational closure \citep{bertschinger_information_2006}, autonomy \citep{bertschinger_autonomy_2008}, and morphological computation \citep{zahedi_quantifying_2013}. We have also related entity actions to information theoretic measures in the perception action loop. The branch-morph (and also the entity actions) therefore suggest that it is possible to generalise these measures to entity sets that are non-exhaustive and vary counterfactually in extent. As we have argued in \cref{sec:entinmvmc} such entities should be considered in a general definition of agents in multivariate Markov chains. The branch-morphs (and possibly other similar constructions) therefore provide a new tool to formulate such a general definition of agents.

\chapter{The structure of spatiotemporal patterns in small Markov chains}
\label{sec:exstp}
%
%


%
In this chapter we investigate the structure of integrated and completely locally integrated spatiotemporal patterns as it is revealed by the disintegration hierarchy. This will expose many of the formal notions introduced in \cref{ch:fb} in practice. For this we will use two very simple multivariate Markov chains. We will use the disintegration theorem (\cref{thm:disintegration}) to extract the completely locally integrated spatiotemporal patterns. We will also use the SLI symmetry theorem and its corollary (\cref{thm:symmpart,thm:symmpartcor}) to explain the structure of the disintegration hierarchies. 

In \cref{sec:cliasentities} we proposed to employ the completely locally integrated spatiotemporal patterns as a formal definition for entities. The entity set obtained in this way are called the $\ci$-entities. We will calculate the entity sets for 

We will see in \cref{sec:exampleconnection} that the three phenomena that entity definitions should not preclude (compositionality, degree of freedom traversal, counterfactual variation) are exhibited by $\ci$-entities in the example systems.

In \cref{sec:actions,sec:perceptions} we defined entity actions and entity perceptions for arbitrary entity-sets. In \cref{sec:actperceptexamples} we present examples of entity actions and entity perceptions of $\ci$-entities in the example systems.

In \cref{sec:exdiscussionoutlook} we discuss the results of this chapter, point out weaknesses, and propose further research directions. While there are some promising signs with respect to using $\ci$-entities as entity sets and our notions of entity action and entity perception there are also some problems that need to be addressed before these proposals can claim to capture agents that only lack goal-directedness. 

This chapter exclusively investigates examples of original notions that we have presented in \cref{ch:fb,ch:agents}. Therefore, almost its entire content is original. For reference the contributions of this chapter are:
\begin{itemize}
  \item Computation and presentation of disintegration and refinement-free disintegration hierarchies for two simple systems.
  \item Explanation of the occurrence of multiple disconnected components in the partially ordered disintegration levels via the SLI symmetry theorems.
  \item Computation and presentation of the completely locally integrated spatiotemporal patterns of two simple systems.
  \item Examples of $\ci$-entities that exhibit the three phenomena compositionality, degree of freedom traversal, and counterfactual variation that we argued for in \cref{sec:entinmvmc}.
  \item Examples of entity actions of $\ci$-entities. 
  \item Example of interpenetrating $\ci$-entities showing that they do not necessarily obey non-interpenetration.
  \item Example of an entity perception and a branch-morph using a proxy for a co-perception partition.
  \item Example of an entity action and entity perception of the same $\ci$-entity at the same time-step.
  \item Discussion of the results on $\ci$-entities as entity sets in the example systems.
\end{itemize}

%

As we will see in \cref{sec:platticenumbers} the computational complexity of computing the completely locally integrated spatiotemporal patterns increases rapidly. Since an important aspect of our proposal to use these patterns as entities is that they do not rely on further intuitions or prior knowledge about what entities are we must consider all possibilities exhaustively. The multivariate Markov chains we choose in this section are therefore extremely small. While this limits the relevance for the interpretation of the completely locally integrated spatiotemporal patterns as agents it allows us to discuss these patterns themselves more thoroughly.

The systems we will look at are the following:
\begin{enumerate}
  \item $\MCconst$ which consists of two constant and independent binary random variables.
  \item $\MCnoise$ which consists of two binary random variables that are mostly constant but where a noise term $\epsilon$ makes every other transition possible.
\end{enumerate}
In all cases we choose a uniform initial distribution in order to exhaust the dynamics of all trajectories of the Markov chains. 




\section{Properties of partition lattices}
\label{sec:platticenumbers}
\begin{figure}
\begin{center}
  \input{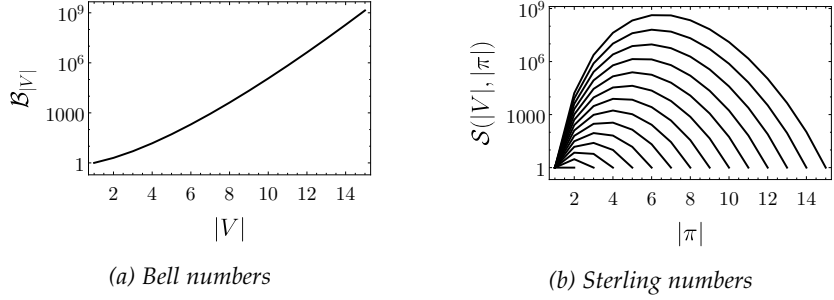}
  \caption{Bell and Sterling numbers. \subref{fig:bell} Logarithmic plot of the Bell numbers $\B_{|V|}$ for showing the number of partitions of a set $V$ with $|V|=\{1,...,15\}$. \subref{fig:sterling} Logarithmic plot of the Sterling numbers showing the number of partitions $\pi$ with $|\pi|$ blocks. The different lines correspond to different cardinalities $|V|$ of the set $V$ that the $\pi$ partition. The number at which a line ends indicates $|V|$.}
  \label{fig:bellsandsterlings}
\end{center}
\end{figure}

Before we look at examples of partition lattices in the following chapters we quickly recall some properties that hold for all such lattices. As mentioned in \cref{thm:bellno} the number of elements $|\Latt(V)|$ in a partition lattice $\Latt(V)$ of a set $V$ is given by the Bell number $\B_{|V|}$. Asymptotically, this number grows faster than exponentially with $|V|$ \citep{debruijn_asymptotic_1970} as is suggested by the more than linear growth in the logarithmic plot \cref{fig:bellsandsterlings}\subref{fig:bell}. We also mentioned that the number of partitions $\pi$ with a fixed number $|\pi|=k$ of blocks is given by the Sterling number $\S(|V|,k)$. The most partitions then turn out to exist for the number of blocks $k$ around $|V|/2$. This can be seen in \cref{fig:bellsandsterlings}\subref{fig:sterling}.

\section{Number of STP}
\label{sec:stpnumber}
The number $\patno(\Xv)$ of STPs in a Bayesian network $\Xv$ is equal to the the sum over all $k$ of the subsets of size $k$ times the number of different STP on this subset of size $k$. The number of different STP on a subset $A \subseteq V$ is $|\X_A|$ so we get:
\begin{equation}
\label{eq:stpcount}
\patno(\Xv)=\sum_{A\subseteq V} |\X_A| =
  \sum_{k = 1}^{|V|} \sum_{\{A \subseteq V: |A| = k\}} |\X_A|. 
\end{equation} 
If we assume that the state spaces of all random variables in the network are equal i.e.\ $|\X_i| =n$ for all $i \in V$ then $|\X_A|=n^{|A|}$ and we get:
\begin{equation}
  \patno(\Xv)=\sum_{k = 1}^{|V|} \binom{|V|}{k} n^{k}. 
\end{equation} 
To get the number $\slino(\Xv)$ of SLI that have to be evaluated to check every partition of each STP $x_A$ we have to further multiply the number of STPs $|\X_A|$ in \cref{eq:stpcount} by the number of partitions $\B_{|A|}$ of these STPs. So the number of SLI values to evaluate is:
\begin{equation}
\label{eq:slivaluecount}
  \slino(\Xv)=\sum_{k = 1}^{|V|} \sum_{\{A \subseteq V: |A| = k\}} |\X_A| \B_{k}. 
\end{equation} 
For equal state spaces we get:
\begin{equation}
\label{eq:slivaluecountCA}
  \slino(\Xv)=\sum_{k = 1}^{|V|} \binom{|V|}{k} n^{k} \B_{k} 
\end{equation} 
SLI evaluations. 

If we use the disintegration theorem we evaluate all partitions of the entire Bayesian network for each trajectory i.e.\
\begin{equation}
\label{eq:slinodis}
 \slino^{\dis}(\Xv)= |\X_V| \B_{|V|}
\end{equation} 
partitions which is only the last term for $k= |V|$ of the sums over $k$ in \cref{eq:slivaluecount,eq:slivaluecountCA}. For equal state spaces this becomes
\begin{equation}
\label{eq:slinodiseq}
 \slino^{\dis}(\Xv)= n^{|V|} \B_{|V|}.
\end{equation} 
We still obtain what we are most interested in which are all the completely integrated patterns within the trajectories. However we also have to obtain the refinement free disintegration hierarchy which requires us to find the finest partitions at each disintegration level and check if they have refinements at preceding levels. If we ignore this for the moment, the disintegration theorem saves us 
\begin{equation}
  \slino(\Xv)-\slino^{\dis}(\Xv)=\sum_{k = 1}^{|V|-1} \sum_{\{A \subseteq V: |A| = k\}} |\X_A| \B_{k}
\end{equation} 
evaluations. Which in the case of equal state spaces is:
\begin{equation}
  \slino(\Xv)-\slino^{\dis}(\Xv)=\sum_{k = 1}^{|V|-1} \binom{|V|}{k} n^{k} \B_{k}
\end{equation}
evaluations. However as can be seen from \cref{eq:slinodis,eq:slinodiseq} the superexponential growth of the number of evaluations with the size of the index set $V$ remains even if we use the disintegration theorem. The extra burden of finding the refinement free disintegration hierarchy has to be added to this as well.

We also note here that the above considerations do not include the computational resources needed to calculate the probabilities needed for the evaluation of specific local integrations. In order to calculate the disintegrations of a trajectory we need the global probability distribution over the entire Bayesian network. This means we need the probability (a real number between $0$ and $1$) of each trajectory. If we only have binary random variables, the number of trajectories is $2^{|V|}$ which make the straightforward computation of disintegration hierarchies unrealistic even for quite small systems. If we take a seven by seven grid of the game of life cellular automaton and want to look at three time-steps we have $|V|= 147$. If we use $32$ bit floating numbers this give us $5 \times 10^30$ petabytes of storage needed for this probability distribution. This suggests that formal proofs are more useful for the investigation of specific local integration and disintegration hierarchies than simulations. Nonetheless we here show some very simple systems to get a better understanding of the formal notions.

\section{Two constant and independent binary random variables: \texorpdfstring{$\MCconst$}{MCconst}}
\label{sec:id2}
\subsection{Definition}

\begin{figure}
\begin{center}
\begin{tikzpicture}
    [myarrow,node distance=2cm]

    \node[] (1) [] {$X_{1,1}$};
    \node[] (2) [right of=1] {$X_{1,2}$};
    \node[] (3) [right of=2] {$X_{1,3}$};
    \node[] (5) [below of=1,node distance=1.5cm] {$X_{2,1}$};
    \node[] (6) [below of=2,node distance=1.5cm] {$X_{2,2}$};
    \node[] (7) [below of=3,node distance=1.5cm] {$X_{2,3}$};


    \path[myarrow]
      (1) edge node {} (2)
      (2) edge node {} (3)

      (5) edge node {} (6)
      (6) edge node {} (7)




      ;
  \end{tikzpicture}
  \caption{Bayesian network of $\MCconst$. There is no interaction between the two processes.}
  \label{fig:2cellidbn}
\end{center}
\end{figure}
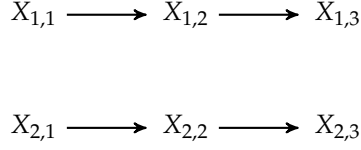

Define the time- and space-homogeneous multivariate Markov chain $\MCconst$ with Bayesian network $\{X_{j,t}\}_{j \in \{1,2\}, t \in \{0,1,2\}}$ and 
\begin{itemize}
 \item 
 \begin{equation}
\pa(j,t)=\begin{cases} 
                                     \emptyset &\text{ if } t=0, \\
                                      \{(j,t-1)\} &\text{ else,}
                                   \end{cases}
 \end{equation}
                                   
 \item \begin{equation}
 \label{eq:id2markovmatrix}
p_{j,t}(x_{j,t}|x_{j,t-1})=\delta_{x_{j,t-1}}(x_{j,t}) =\begin{cases} 
                                     1 &\text{ if } x_{j,t}=x_{j,t-1}, \\
                                     0 &\text{ else,}
                                   \end{cases}
\end{equation} 
\item \begin{equation}
\label{eq:id2initialdist}
p_{j,0}(x_{j,0}) = 1/4.                             
\end{equation} 
\end{itemize}
The Bayesian network can be seen in \cref{fig:2cellidbn}.

\subsection{Trajectories}
\begin{figure}[ht!]
\input{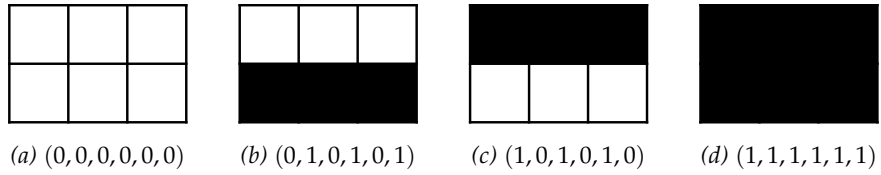}
\caption{%
        Visualisation of the four possible trajectories of $\MCconst$. In each trajectory the time index increases goes from left to right. There are two rows corresponding to the two random variables at each time step and three columns corresponding to the three time-steps we are considering here. 
     }%
   \label{fig:2cellidbntras}
\end{figure}

In order to get the disintegration hierarchy $\dis(x_V)$ we have to choose a trajectory $x_V$ and calculate the SLI of each partition $\pi \in \Latt(V)$. There are only four different trajectories possible in $\MCconst$ and they are:
\begin{equation}
  x_V=(x_{1,0},x_{2,0},x_{1,1},x_{2,1},x_{1,2},x_{2,2})=\begin{cases} (0,0,0,0,0,0) &\text{ if } x_{1,0}=0, x_{2,0}=0; \\
								      (0,1,0,1,0,1) &\text{ if } x_{1,0}=0, x_{2,0}=1; \\
								      (1,0,1,0,1,0) &\text{ if } x_{1,0}=1, x_{2,0}=0; \\
								      (1,1,1,1,1,1) &\text{ if } x_{1,0}=1, x_{2,0}=1.
                                                        \end{cases}
\end{equation} 
Each of these trajectories has probability $p_V(x_V)=1/4$ and all other trajectories have $p_V(x_V)=0$. We call the four trajectories the \textit{possible trajectories}.
We visualise the possible trajectories as a grid with each cell corresponding to one variable. The spatial indices are constant across rows and time-slices $V_t$ correspond to the columns. A white cell indicates a $0$ and a black cell indicates a $1$. This results in the grids of \cref{fig:2cellidbntras}.
\subsection{Partitions of trajectories}
The disintegration hierarchy is composed out of all partitions in the lattice of partitions $\Latt(V)$. Note that we are partitioning the entire spatially and temporally extended index set $V$ of the Bayesian network and not only the time-slices. Blocks in the partitions of $\Latt(V)$ are then, in general, spatiotemporal patterns and not only spatial patterns.

The number of partitions $|\Latt(V)|$ of a set of $|V|=6$ elements is $\B_6=203$ (see \cref{thm:bellno}). These partitions $\pi$ can be classified according to their cardinality $|\pi|$ (number of blocks in the partition). The number of partitions of a set of cardinality $|V|$ into $|\pi|$ blocks is the Sterling number $\S(|V|,|\pi|)$. For $|V|=6$ we find the Sterling numbers: 
  \begin{equation}   
    \begin{array}{|l|c|c|c|c|c|c|}
\hline
|\pi| & 1 & 2 & 3 & 4 & 5 & 6 \\
\hline
\S(|V|,|\pi|) & 1 & 31 & 90 & 65 & 15 & 1 \\
\hline
\end{array}
\end{equation}

It is important to note that the partition lattice $\Latt(V)$ is the same for all trajectories as it is composed out of partitions of $V$. On the other hand the values of SLI $\mi_\pi(x_V)$ with respect to the partitions in $\Latt(V)$ generally depend on the trajectory $x_V$.  

\subsection{SLI values of the partitions}
\begin{figure}[ht!]
\begin{center}
  \includegraphics[width=.9\linewidth]{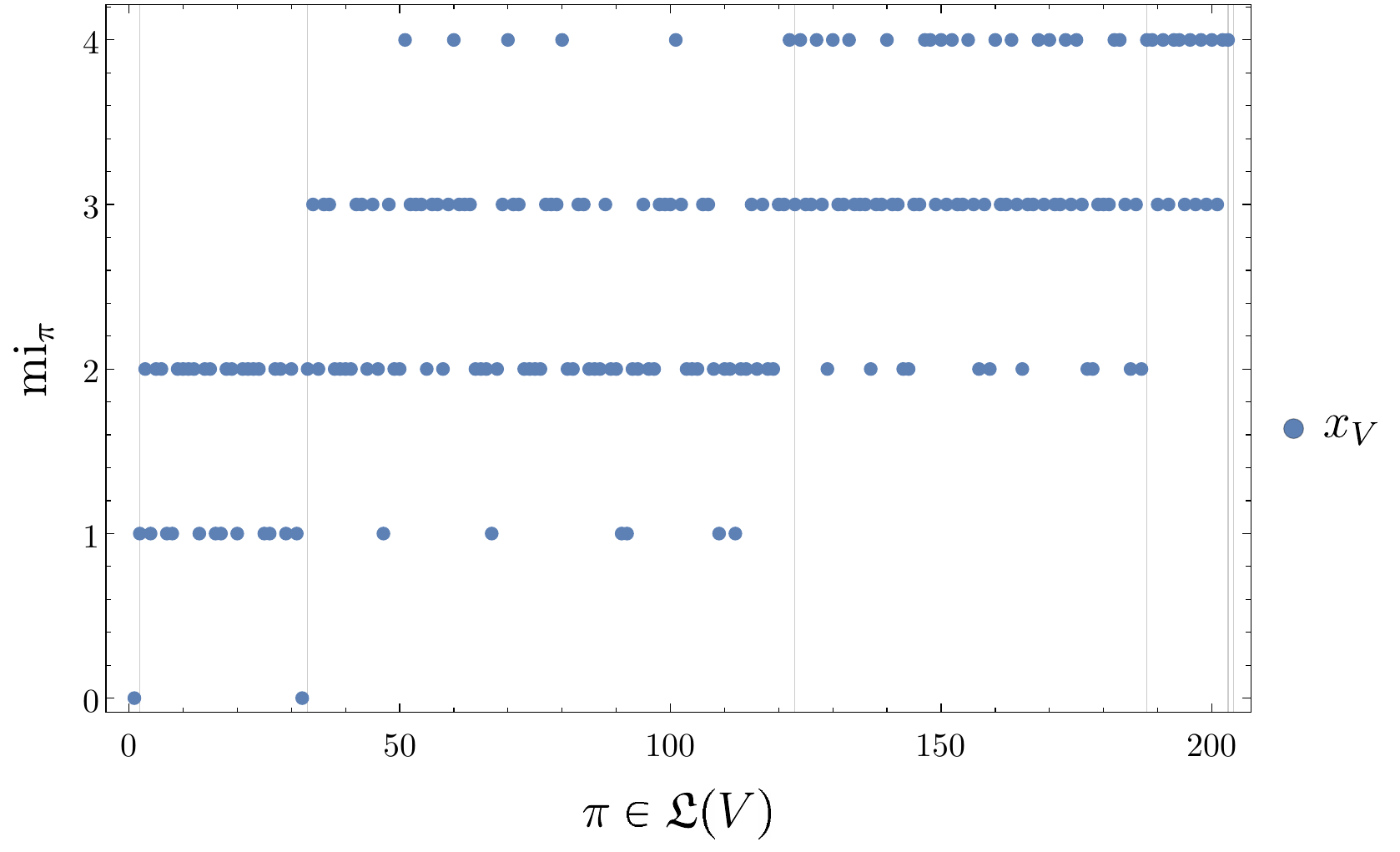}
\end{center}
\caption{%
        Specific local integrations $\mi_\pi(x_V)$ of any of the four trajectories $x_V$ seen in \cref{fig:2cellidbntras} with respect to all $\pi \in \Latt(V)$. The partitions are ordered according to an enumeration with increasing cardinality $|\pi|$ (see \citealp[chap. 4.3.3]{pemmaraju_computational_2009} for the method). We indicate with vertical lines at what partitions the cardinality $|\pi|$ increases by one.}%
   \label{fig:2cellidbnsli}
\end{figure}

We can calculate the SLI $\mi_\pi(x_V)$ of every trajectory $x_V$ with respect to each partition $\pi \in \Latt(V)$ according to \cref{def:sli}:
\begin{equation}
 \mi_\pi(x_V):= \log \frac{p_V(x_V)}{\prod_{b \in \pi} p_b(x_b)}.
\end{equation} 
In the case of $\MCconst$ the SLI values with respect to each partition do not depend on the trajectories. For an overview we plotted the values of SLI with respect to each partition $\pi \in \Latt(V)$ for any trajectory of $\MCconst$ in \cref{fig:2cellidbnsli}. 
We can see in \cref{fig:2cellidbnsli} that the cardinality does not determine the value of SLI. At the same time there seems to be a trend to higher values of SLI with increasing cardinality of the partition. We can also observe that only five different values of SLI are attained by partitions on this trajectory. 
We will collect these classes of partitions with equal SLI values in the disintegration hierarchy next. 

\subsection{Disintegration hierarchy}
\begin{figure}[ht!]
\begin{center}
  \includegraphics[width=.9\linewidth]{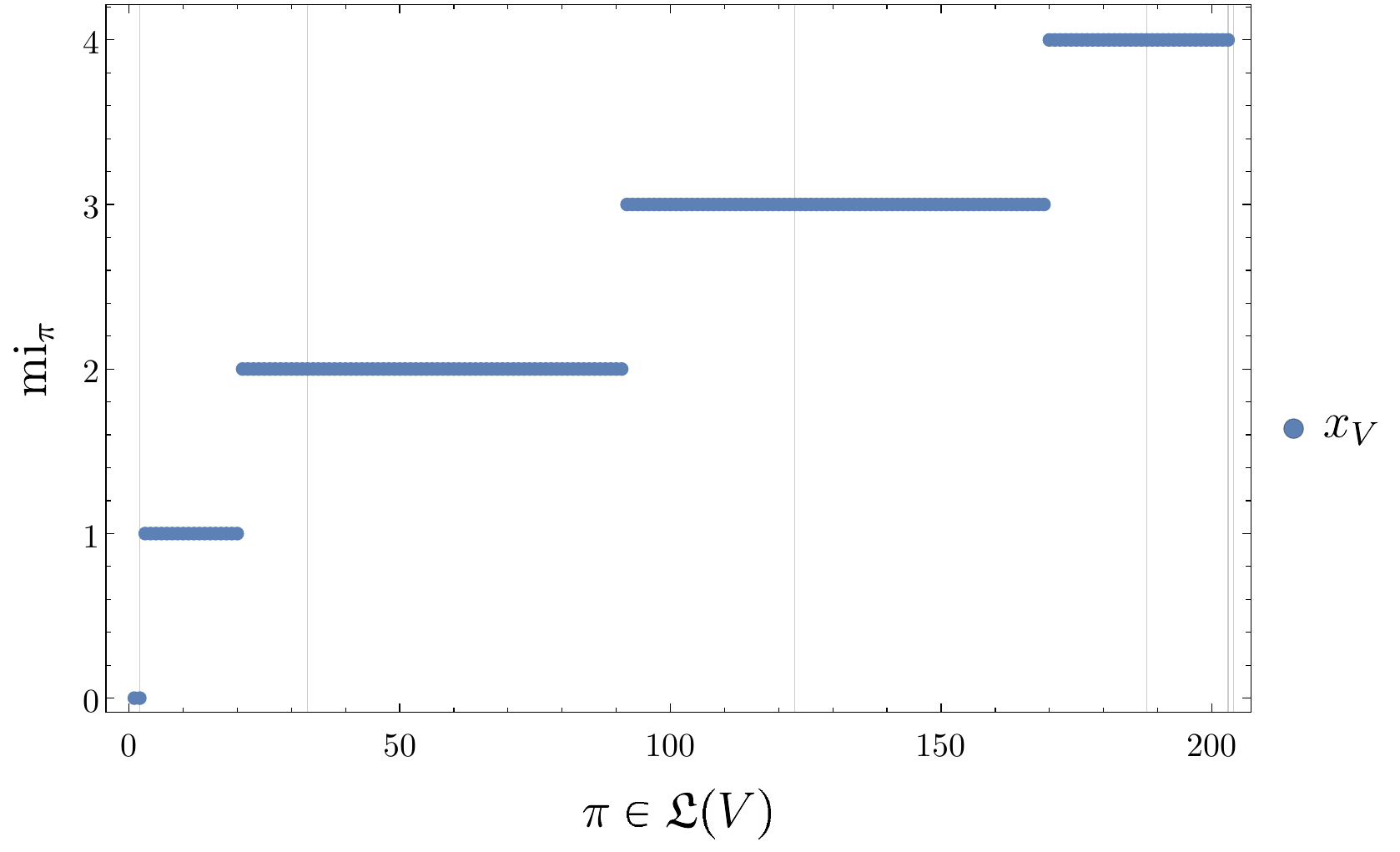}
\end{center}
\caption{%
        Same as \cref{fig:2cellidbnsli} but with the partitions sorted according to increasing SLI.
        }%
   \label{fig:2cellidbnslisorted}
\end{figure}

In order to get insight into the internal structure of the partitions of a trajectory $x_V$ we obtain the disintegration hierarchy $\dis(x_V)$ (see \cref{def:dishier}) look at the Hasse diagrams of each of the disintegration levels $\dis_i(x_V)$. If we sort the partitions of any trajectory of $\MCconst$ according to increasing SLI value we obtain \cref{fig:2cellidbnslisorted}. There we see groups of partitions attaining the SLI values $\{0,1,2,3,4\}$ these groups are the disintegration levels $\{\dis_1(x_V),\dis_2(x_V),\dis_3(x_V),\dis_4(x_V),\dis_5(x_V)\}$. The exact numbers of partitions in each of the levels are:
  \begin{equation}   
    \begin{array}{|l|c|c|c|c|c|c|}
\hline
i       & 1 & 2 & 3 & 4 & 5 \\
\hline
\mi_\pi & 0 & 1 & 2 & 3 & 4 \\
\hline
|\dis_i| & 2 & 18 & 71 & 78 & 34 \\
\hline
\end{array}
\end{equation}
\begin{figure}[ht!]
\input{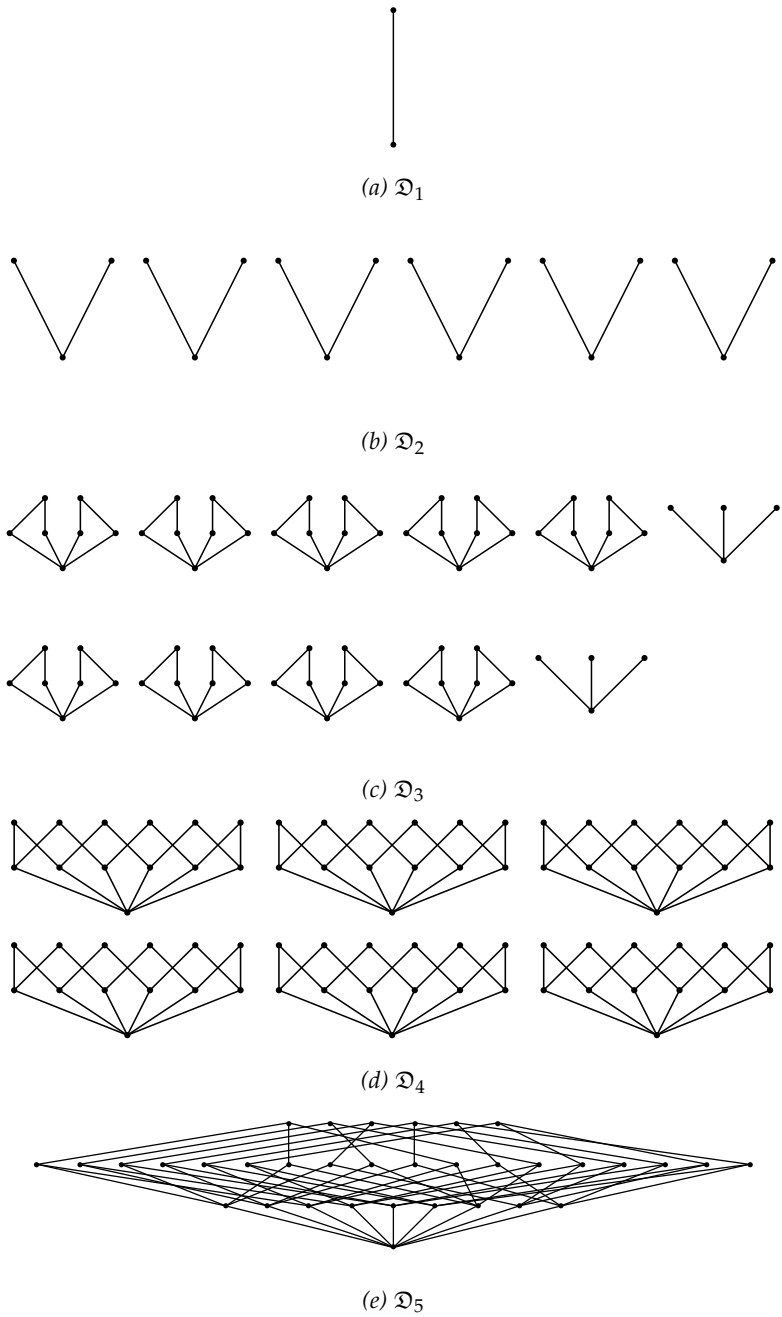}
\caption{%
        Hasse diagrams of the five disintegration levels of the trajectories of $\MCconst$. Every vertex corresponds to a partition and edges indicate that the lower partition refines the higher one.
     }%
   \label{fig:disid2}
\end{figure}
Next we look at the Hasse diagram of each of those disintegration levels. Since the disintegration levels are subsets of the partition lattice $\Latt(V)$ they are in general not lattices by themselves. The Hasse diagrams visualise the set of partitions in each disintegration level partially ordered by refinement $\lpre$ (see \cref{def:refinement}). Recall that in Hasse diagrams of such posets the partitions are arranged such that if $\pi \neq \xi$ and $\pi \lpre \xi$ then $\pi$ is drawn below $\xi$. Also, an edge is drawn from partition $\pi$ to $\xi$ if one covers the other e.g.\ if $\pi \lpre: \xi$. The Hasse diagrams are shown in \cref{fig:disid2}. We see immediately that within each disintegration level apart from the first and the last the Hasse diagrams contain multiple connected components.

\begin{figure}[ht!]
\begin{center}
  \includegraphics[width=\textwidth]{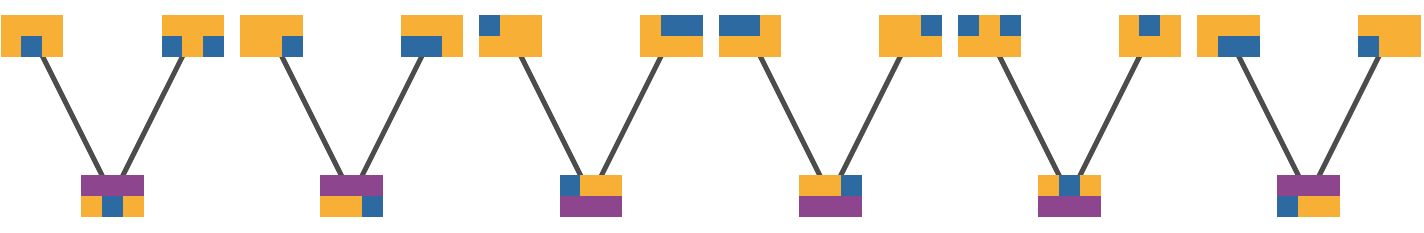}
\end{center}
\caption{%
        Hasse diagram of $\dis_2$ of $\MCconst$ trajectories. Here we visualise the partitions at each vertex. The blocks of a partition are the cells of equal colour. Note that we can obtain all six disconnected components by permuting the indices via spatial inversion $h_{\inv}(j,t)=(|J|+1-j,t)$ and ``global'' time shifts $h_\shift(j,t)=(j,(t-1) \mod 3)$. For example  acting on the partitions in the first component from the left we obtain: the second component via $h_\shift$, the third component via $h_{\inv} \circ h_\shift^{-1}$, the fourth via $h_{\inv} \circ h_\shift$, the fifth via $h_{\inv}$, and the sixth via $h_\shift^{-1}$.
     }%
   \label{fig:disid2D2part}
\end{figure}

Furthermore, within a disintegration level the connected components often have the same Hasse diagrams. For example in $\dis_2$ (\cref{fig:disid2}\subref{fig:disid2D2}) we find six connected components with three partitions each. The identical refinement structure of the connected components is related to the symmetries of the probability distribution over the trajectories. This will be discussed in \cref{sec:mcconstsymmetries}. We can visualise the partitions themselves in the Hasse diagrams as in \cref{fig:disid2D2part}. 

\begin{figure}[ht!]
\begin{center}
  \includegraphics[width=\textwidth]{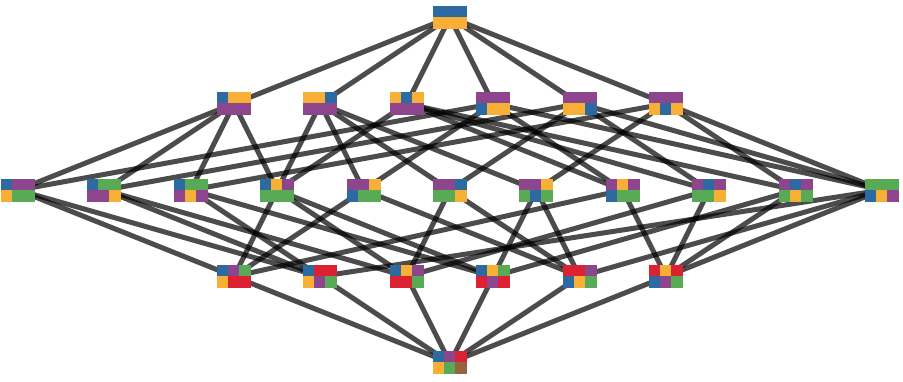}
\end{center}
\caption{%
        Hasse diagrams of the refinement-free disintegration hierarchy $\dis^\lmin$ of $\MCconst$ trajectories. Here we visualise the partitions at each vertex. The blocks of a partition are the cells of equal colour. It turns out that partitions that are on the same \textit{horizontal} level in this diagram correspond exactly to a level in the refinement-free disintegration hierarchy $\dis^\lmin$. The $i$-th horizontal level starting from the top corresponds to $\dis^\lmin_i$. Take for example the second horizontal level from the top. The partitions on this level are just the minimal elements of the poset $\dis_2$ which was visualised in \cref{fig:disid2D2part}. We have shown the posets of the other disintegration levels only without their partitions in \cref{fig:disid2} but their minimal elements are all present at the according horizontal level in this diagram.   
%
     }%
   \label{fig:rfdisid2}
\end{figure}

Recall that due to the disintegration theorem (\cref{thm:disintegration}) we are interested especially in partitions that do not have refinements at their own or any preceding (i.e.\ lower indexed) disintegration level. These partitions consist of blocks that are completely integrated i.e.\ all possible partitions of each of the blocks results in a positive SLI value or is a single node of the Bayesian network. The refinement-free disintegration hierarchy $\dis^\lmin(x_V)$ contains only these partitions and is shown in a Hasse diagram in \cref{fig:rfdisid2}.

\subsection{Symmetries}
\label{sec:mcconstsymmetries}
 As shown in \cref{thm:ontrapartsymm1} the symmetries of the trajectory $x_V$ that are also symmetries of $p_V$ generate partitions with respect to which the SLI has the same value. More formally, if for all elements $h$ of a group of permutations $\gr{H}$ we have both $h p_V = p_V$ and $h x_V = x_V$ (for the particular trajectory $x_V$, not necessarily for all trajectories) then $\mi_{h \pi}(x_V)= \mi_\pi(x_V)$. This means if we start with one partition $\pi$ with a particular SLI value (i.e.\ on a particular disintegration level) then we can generate the orbit of partitions $\gr{H} \pi := \{h \pi : h \in \gr{H}\}$ under $\gr{H}$ which contains only partitions with equal SLI value. Since any permutation preserves the refinement relation between and cardinality of partitions (see \cref{thm:refinementpreserve,thm:cardinalitypreserve} respectively) we can also take a set of partitions partially ordered by refinement and generate identical posets (with identical Hasse diagrams) of partitions of respectively equal cardinality. In the following we will find the symmetries of both $p_V$ and of the four possible trajectories. 
 
  We will now establish symmetries of $p_V$. It is not difficult to just look at the probability distribution $p_V$ which attributes the probability $1/4$ for each of the four possible trajectories and infer symmetries by visual inspection. 
We only need to find permutations of the index set $V$ that transform all trajectories into trajectories with the same probability. Since all four possible trajectories in \cref{fig:2cellidbntras} have the same probability every permutation that maps these trajectories onto each other is a symmetry of $p_V$. Note that permutations which map one of those trajectories into a trajectory with probability zero cannot be symmetries of $p_V$. To get an intuition for this take one of the less symmetric possible trajectories of $\MCconst$ from \cref{fig:2cellidbntras} e.g.\ the second one
\begin{center}	    
            \includegraphics[width=0.2\textwidth]{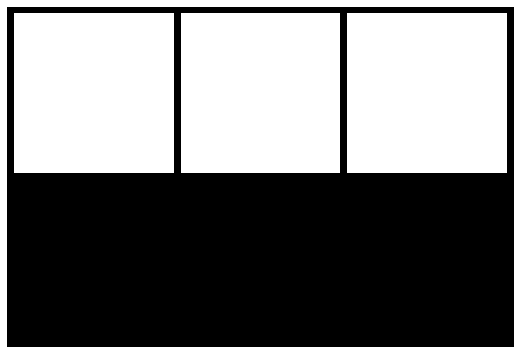}
\end{center}
Any permutation of the cells in this grid is a permutation of $V$. Since the other possible trajectories are either more symmetric (like the first and the fourth) or share the same symmetry (like the third) any permutation which maps this trajectory to one of the four is a symmetry of $p_V$. Such symmetries are ``spatial inversion'' (flipping the top and bottom row in \cref{fig:2cellidbntras}), ``global time-slice permutations'' (reordering the columns in \cref{fig:2cellidbntras} in any way)
, and ``individual time-slice permutation'' (reordering the cells within a single row in \cref{fig:2cellidbntras} in any way). Spatial inversion maps the second trajectory above to the third and vice versa, the first and second trajectory are left invariant. The time-slice permutations both leave all possible trajectories invariant. Note that the global time-slice permutations can actually be constructed from the individual time-slice permutations. An example of a permutation that is not a symmetry is to invert only a single time-slice and not change the others (flipping the top and bottom cell in the second column). Next, we will formally derive theses intuitions from the definition of $\MCconst$.

First let us look at ``individual time-slice permutations'' more precisely this means permutations of the form $h(j,t_1) = (j,t_2)$ where for different spatial indices the time-shift may be different i.e.\ for $j_1 \neq j_2$ we may have $h(j_1, t_1)=(j_1,t_2)$, $h(j_2,t_1)=(j_2,t_3)$, and $t_2 \neq t_3$. We can separate such permutations into two sets, those that permute the time indices only for the first spatial index and those that only permute them for the second. Let $\gr{T}_j = \symg_{(j,T)} \subset \symg_V$ be the subgroup of permutations only affecting the indices in $(j,T) = \{(j,t):t \in T\}$ ( we will in the following loosely refer to $(1,T),(2,T)$ as the first and second, or top and bottom \textit{row} since this is visually intuitive given our representation of trajectories and patterns as grids). Note that by combining elements $h_1 \in \gr{T}_1$ and $h_2 \in \gr{T}_2$ to get $h_1 \circ h_2$ we can affect both $(1,T)$ and $(2,T)$ so that we can also permute whole time-slices at once if both $h_1$ and $h_2$ permute the respective sets in the same way. This gives the ``global time-slice permutations''. 


Now note that for the present case both $\gr{T}_j$, $j \in \{1,2\}$ are actually symmetry groups of all possible trajectories $x_V$. This means they are necessarily symmetry groups of $p_V$ since for any arbitrary $p_V$ if $x^g_V=x_V$ then of course $g p_V(x_V) = p_V(x^g_V)=p_V(x_V)$. Formally, for any $j \in \{1,2\}$ and $h_j \in \gr{T}_j$ we have
\begin{align}
  h_j x_V = h_j x_{J,T} &= (x_{i,t})^{h_j^{-1}}_{i \in J, t \in T} \\
  &= \{X_{i,t}=x_{i,t}\}^{h_j^{-1}}_{i \in J, t \in T} \\
  &= \{X_{i,t}=x_{h_j^{-1}(i,t)}\}_{i \in J, t \in T} \\
  &= \{X_{i,t}=x_{i,t}\}_{i \neq j, t \in T} \cup \{X_{i,t}=x_{h_j^{-1}(j,t)}\}_{j, t \in T} 
\end{align}
If we now recall that for all possible trajectories $x_V$ and $t,s \in T$ we have $x_{j,t}=x_{k,s}$ so that $x_{h_j^{-1}(j,t)}=x_{j,t}$ we see that  
\begin{align}
h_j x_V = x_V
\end{align} 
for all possible trajectories. Since all permutations are bijective, the impossible trajectories must also be mapped to impossible trajectories such that all $h_j \in \gr{T}_j$ with $j \in \{1,2\}$ are symmetries of all trajectories and therefore also symmetries of $p_V$. 

Next we will look at spatial inversion. Note first that spatial inversion does not leave all the possible trajectories invariant since it transforms the second into the third and vice versa. So spatial inversion is only a symmetry of $p_V$.

It would not be difficult to derive that spatial inversion is a symmetry of $p_V$ directly from looking at its effect on $p_V$. However, we here want to exhibit how \cref{thm:spatialsymm} can be used in establishing symmetries. 


 \Cref{thm:spatialsymm} tells us that a group of spatial symmetries of both the Markov matrix and the initial distribution is also a symmetry group of $p_V$. Define the spatial inversion via $h_{\inv}(j,t)=(|J|+1-j,t)$. Then $h_{\inv} \circ h_{\inv} = \id$, $h_{\inv}^{-1}=h_{\inv}$ which means $\{h_{\inv},\id\}$ form a subgroup of the spatial permutations $\symg_J \times \{\id\}$. We now show that this is a symmetry group of $p_V$. Note that $V_0 =(J,0)= \{(1,0),(2,0)\}$, and recall that we chose the uniform distribution as initial distribution such that for any $x_{V_0},\bar{x}_{V_0} \in \X_{V_0}$ we have $p_{V_0}(x_{V_0})=p_{V_0}(\bar{x}_{V_0})=1/4$. Since by construction $x^{h_{\inv}}_{V_0} \in \X_{V_0}$ 
 we have for any $x_{V_0} \in \X_{V_0} =\{(0,0),(0,1),(1,0),(1,1)\}$:  
 \begin{align}
   h_{\inv} p_{V_0}(x_{V_0}) &= p_{V_0}(x^{h_{\inv}}_{V_0}) 
   \\
   &=\Pr(\{X_{1,0}=x_{1,0},X_{2,0}=x_{2,0}\}^{h_{\inv}}) \\
   &=\Pr(X_{1,0}=x_{h_{\inv}(1,0)},X_{2,0}=x_{h_{\inv}(2,0)}) \\
   &=\Pr(X_{1,0}=x_{2,0},X_{2,0}=x_{1,0}) \\  
   &= p_{V_0}(x_{V_0}).
 \end{align}
Or in short $h_{\inv} p_{V_0} =p_{V_0}$. We also have $h_{\inv} (P p_{V_t}) = P (h_{\inv} p_{V_t})$. To see this recall the definition of the dynamics of each of the two random variables of $\MCconst$ in \cref{eq:id2markovmatrix} which was:
\begin{equation*}
\tag{\ref{eq:id2markovmatrix} revisited}
p_{j,t}(x_{j,t}|x_{j,t-1})=\delta_{x_{j,t-1}}(x_{j,t}) =\begin{cases} 
                                     1 &\text{ if } x_{j,t}=x_{j,t-1}, \\
                                     0 &\text{ else.}
                                   \end{cases}
\end{equation*} 
From this we get the Markov matrix via
\begin{align}
  p_{V_{t+1}}(x_{V_{t+1}}|x_{V_t}) &= \prod_{i \in V_{t+1}} p_i(x_i|x_{\pa(i)}) \\
  &= p_{1,t}(x_{1,t+1}|x_{1,t})p_{2,t}(x_{2,t+1}|x_{2,t}) \\
&= \delta_{x_{1,t}}(x_{1,t+1}) \delta_{x_{2,t}}(x_{2,t+1}).  
\end{align} 
So that (being extra verbose as it is the only such calculation in this thesis)
\begin{align}
  p_{V_{t+1}}(x^{h_{\inv}}_{V_{t+1}}|x^{h_{\inv}}_{V_t}) &=\Pr(\{X_{j,t+1}=x_{j,t+1}\}^{h_{\inv}}_{j \in J} |\{X_{j,t}=x_{j,t}\}^{h_{\inv}}_{j \in J}) \\
  \begin{split}&= \Pr((X_{1,t+1}=x_{1,t+1})^{h_{\inv}}|(X_{1,t}=x_{1,t})^{h_{\inv}}) \\
    &\phantom{\Pr((X_{1,t+1}=}\times \Pr((X_{2,t+1}=x_{2,t+1})^{h_{\inv}}|(X_{2,t}=x_{2,t})^{h_{\inv}}) 
  \end{split}\\
  \begin{split}&= \Pr(X_{1,t+1}=x_{h_{\inv}(1,t+1)}|X_{1,t}=x_{h_{\inv}(1,t)}) \\
    &\phantom{\Pr((X_{1,t+1}=}\times \Pr(X_{2,t+1}=x_{h_{\inv}(2,t+1)}|X_{2,t}=x_{h_{\inv}(2,t)}) 
  \end{split}\\  
  \begin{split}&= \Pr(X_{1,t+1}=x_{2,t+1}|X_{1,t}=x_{2,t}) \\
    &\phantom{\Pr((X_{1,t+1}=}\times \Pr(X_{2,t+1}=x_{1,t+1}|X_{2,t}=x_{1,t}) 
  \end{split}\\ 
  &= p_{1,t+1}(x_{2,t+1}|x_{2,t}) p_{2,t+1}(x_{1,t+1}|x_{1,t})\\
  &= \delta_{x_{2,t}}(x_{2,t+1}) \delta_{x_{1,t}}(x_{1,t+1}) \\
  &= \delta_{x_{1,t}}(x_{1,t+1}) \delta_{x_{2,t}}(x_{2,t+1}) \\
  &= p_{V_{t+1}}(x_{V_{t+1}}|x_{V_t}).
\end{align}
which implies 
\begin{equation}
  p_V(x_{J,t+1}|x_{J,t}) = p_V(x^{h_{\inv}}_{J,t+1}|x^{h_{\inv}}_{J,t}).
\end{equation} 
By \cref{thm:symmP2} this implies $h_{\inv} (P p_{V_t}) = P (h_{\inv} p_{V_t})$ and together with $h_{\inv} p_{V_0} =p_{V_0}$ \cref{thm:spatialsymm} then implies $h_{\inv} p_V = p_V$ which shows that $h_{\inv}$ is a symmetry of $p_V$. Clearly, $\{\id,h_{\inv}\}$ is then a symmetry group of $p_V$.

We have now established that $\gr{T}_1,\gr{T}_2$, and $\{\id,h_\inv\}$ are symmetry groups of $p_V$. Since they are all subgroups of $\symg_V$ we can combine their elements via function composition (e.g.\ $h_j \circ h_\inv$) to get (possibly further) elements of $\symg_V$. The set of elements that can be formed in this way is a subgroup of $\symg_V$ called the subgroup generated by $\gr{T}_1,\gr{T}_2$, and $\{\id,h_\inv\}$. Since all elements of this subgroup are also symmetries of $p_V$ we call the subgroup generated by $\gr{T}_1,\gr{T}_2$, and $\{\id,h_\inv\}$ the symmetry group of $\MCconst$ and denote it by $\gr{G}_{\MCconst}$. 

Let us now come back to the symmetries of SLI. If we look at the first trajectory
\begin{center}	    
            \includegraphics[width=0.2\textwidth]{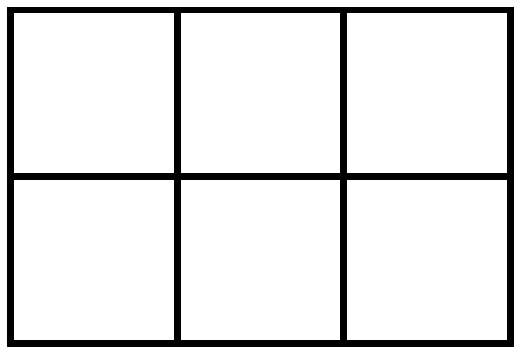}
\end{center}
 we see that it is symmetric with respect to any permutation of $V$. So every symmetry of $p_V$ is also a symmetry of $x_V$ and we can use the entire group $\gr{G}_{\MCconst}$ of symmetries of $p_V$ to generate equally disintegrating partitions. Formally, for each $g \in \gr{G}_{\MCconst}$ we have $g x_V = x_V$ so that $\mi_{g \pi}(x_V)=\mi_\pi(x_V)$. As we have seen in \cref{fig:disid2D2part} combining $h_\inv \in \gr{G}_{\MCconst}$ and $h_\shift \in \gr{G}_{\MCconst}$ (a time shift to the right of all indices) generates all the disconnected components of the poset of $\dis_2$. 
 
\begin{figure}[ht!]
\input{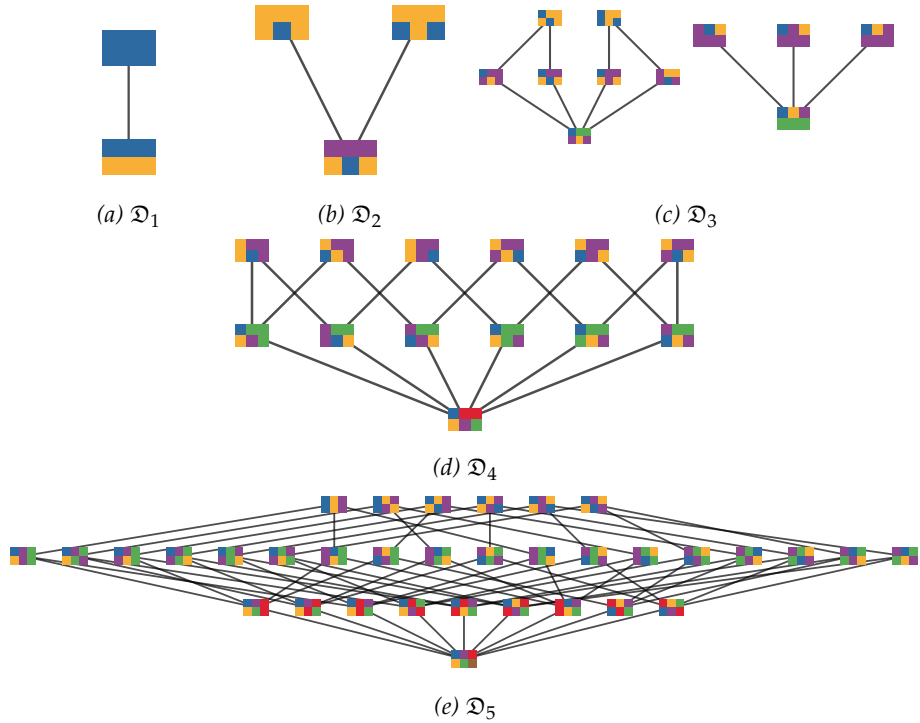}
\caption{%
        Representatives of the distinct Hasse diagrams within each disintegration level of $\MCconst$. Again we visualise the partitions at each vertex with the blocks of a partition of equal colour. Note that at level $\dis_3$ (in \subref{fig:d3id2singlespart}) there are two distinct Hasse diagrams whereas on the other levels there is only one per level. 
     }%
   \label{fig:id2singles}
\end{figure}

 If we look at the other disintegration levels we can see that $\gr{G}_{\MCconst}$ explains the occurrence of multiple disconnected components also in $\dis_4$. It is compatible also with $\dis_1$ in the sense that for the two partitions in $\dis_1$ (see \cref{fig:id2singles}\subref{fig:d1id2singlespart}) we have $g \pi = \pi$ for all $g \in \gr{G}_{\MCconst}$. Similarly we only expect a single component in $\dis_5$ since the minimal partition there, the zero of $\Latt(V)$, also has $g \lzero =\lzero$ for all $g \in \gr{G}_{\MCconst}$. 
 
\begin{figure}[ht!]
\begin{center}
  \includegraphics[width=\textwidth]{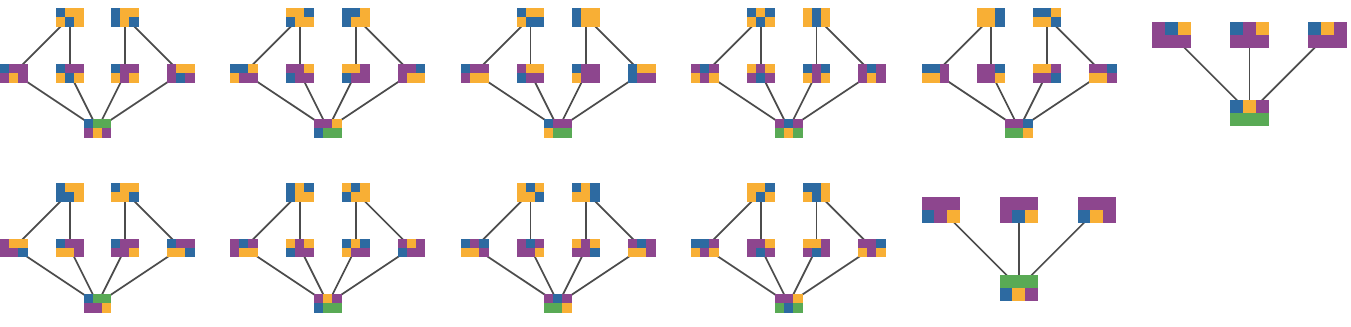}
\end{center}
\caption{%
        Hasse diagram of $\dis_3$ of $\MCconst$ trajectories with visualised partitions. The blocks of a partition are the cells of equal colour. Note that we can obtain all nine disconnected components with three horizontal levels by permuting the indices via spatial inversion $h_{\inv}(j,t)=(|J|+1-j,t)$ and ``individual time slice permutations'' permuting indices in one row independent of the other. However, we cannot obtain the partitions of the two components (at the right edge of both rows) with two horizontal levels from the partitions of the components with three horizontal level components. The equality of SLI 
     }%
   \label{fig:disid2D3part}
\end{figure}

 The only outlier in this hierarchy is $\dis_3$. We show the poset of partitions of $\dis_3$ in \cref{fig:disid2D3part}. It shows two different kinds of disconnected component. One consisting of seven partitions which occurs nine times and one consisting of four partitions occurring two times. According to \cref{thm:refinementpreserve,thm:cardinalitypreserve} we should not be able to use symmetries to map partitions in the first kind of component to partitions in the second kind. 

  This can be seen by noting for example that in most cases the cardinalities of the blocks of the partitions in the components of the first kind are different from the cardinalities of the blocks in the components of the second kind. In case of the first kind the cardinalities are: top layer: $\{4,2\}$ and $\{3,3\}$, second layer: $\{4,1,1\},\{3,2,1\},\{2,2,2\}$, and bottom layer $\{2,2,1,1\}$. In the the case of the second kind: top layer: $\{4,1,1\}$, bottom layer: $\{3,1,1,1\}$. Since all permutations in $\symg_V$ maintain the cardinalities of all blocks only the partitions with block cardinalities $\{4,1,1\}$ occur in either kind and could be transformed into each other by an elements of $\gr{G}_\MCconst$. However, we can observe that this is impossible as we only have $h_\inv$ which exchanges the entire rows and cannot exchange individual cells in the top row with those in the bottom row. 

  So the symmetries of $p_V$ due to permutations of $V$ do not explain why the components of the first and second kind occur on the same disintegration level i.e.\ why the partitions they contain have the same SLI values. This is then a kind of degeneracy of the disintegration level $\dis_3$.  
  
  In order to get an overview of only the different components of the posets in each disintegration level we have drawn them including visualised partitions in \cref{fig:id2singles}.


Finally let us look at the second trajectory again
\begin{center}	    
            \includegraphics[width=0.2\textwidth]{./images/2cellidbn2}
\end{center}
and apply $h_\inv$ we get
\begin{center}	    
            \includegraphics[width=0.2\textwidth]{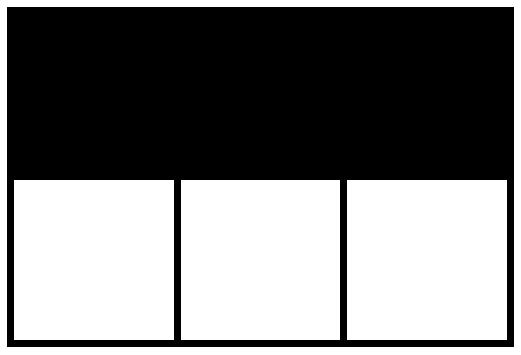}
\end{center}
which is the third trajectory. So $h_\inv$ is not a symmetry of the second trajectory and we don't have $h_\inv x_V=x_V$. This means that the conditions for \cref{thm:ontrapartsymm1} are not satisfied. Nonetheless, as we mentioned before, the disintegration hierarchies of all four possible trajectories are identical. The reason for this is \cref{thm:ontrapartsymm2} which asserts that even if $g x_V \neq x_V$ we still have $\mi_{g \pi}(x_V)=\mi_\pi(x_V)$ if for all blocks $b \in \pi$ we have $p_b(x^g_b) = p_b(x_b)$. This is in fact the case here for $g \in \gr{G}_\MCconst$ and all partitions on the second and third trajectories.

To see this note first that for $x_V$ either the second or third possible trajectory any $b \subseteq V$, $x^g_b= (x_{g(i)})_{i \in b}$ is still a pattern that can occur on at least one of the four possible trajectories. In other words, it is impossible that $x^g_b$ is a pattern that cannot occur. Also keep in mind that $x^g_b$ concerns exactly the same random variables in the Bayesian network as $x_b$ even if the values it fixes may differ. Then distinguish two situations. First, let $b \subset (j,T)$ i.e.\ $b$ is part of a single row indicated by $j$. Then $x^g_b$ fixes some of the values in the one of the rows and $x_b$ also fixes some of the values (possibly different) in one of the rows (possibly the other one). Recall that only four trajectories are possible and fixing the value (independent of what value) of any random variable in one of the rows selects two possible trajectories from the four. Then $p_b(x^g_b)$ is a sum of the probabilities of these two trajectories and since their probabilities are all $1/4$ this sum is $1/2$ for both $p_b(x^g_b)$ and $p_b(x_b)$. Second, let $b \cap (1,T) \neq \emptyset$ and $b \cap (2,T) \neq \emptyset$ then $x^g_b$ and also $x_b$ fixes the values in both rows so there is only a single possible trajectory selected whose probability is $1/4$ in every case. So again $p_b(x^g_b) = p_b(x_b)$.  

\begin{figure}[ht!]
\begin{center}
\
  \includegraphics[width=\textwidth]{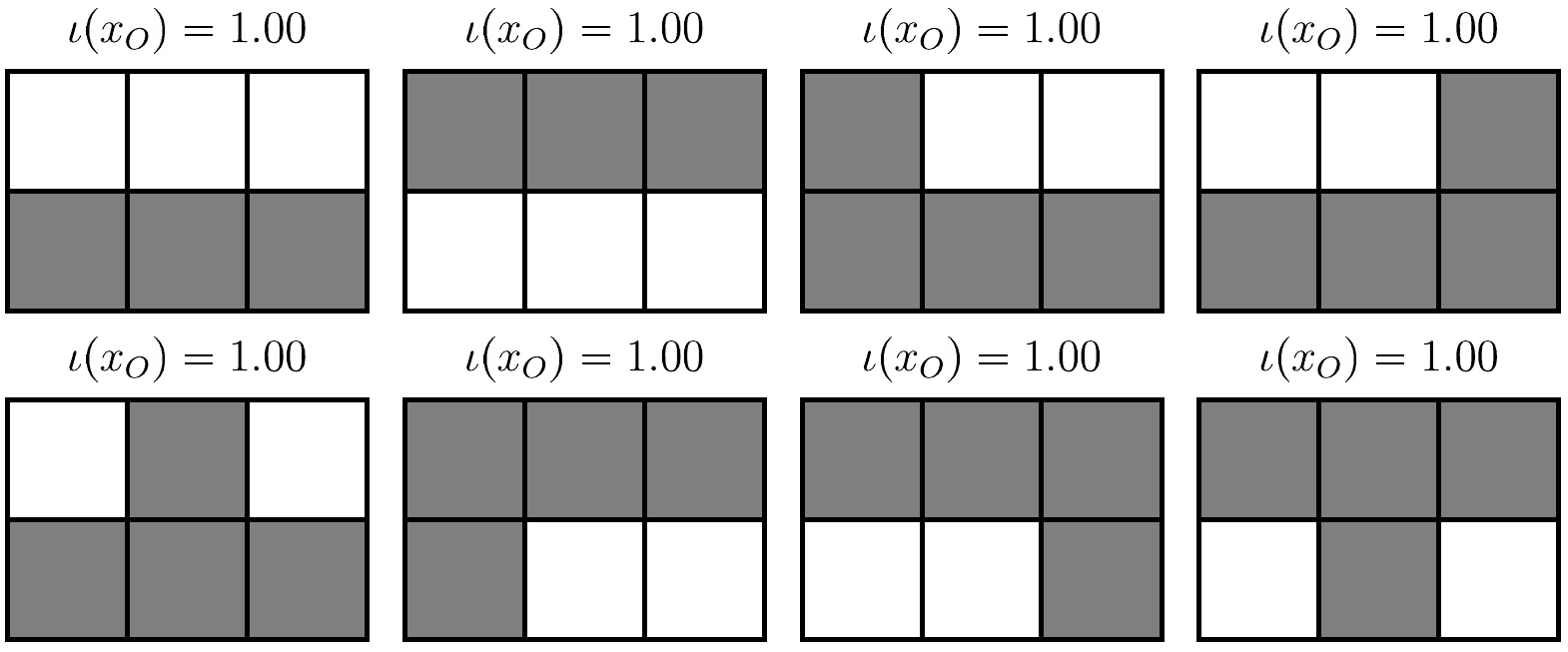}
\end{center}
\caption{%
All distinct completely integrated STPs on the first possible trajectory of $\MCconst$. The value of complete local integration is indicated above each STP. We display STPs by colouring the cells corresponding to random variables that are not fixed to any value by the STP in grey. Cells corresponding to random variables that are fixed by the STP are coloured according to the value i.e.\ white for $0$ and black for $1$. 
     }%
   \label{fig:id2cipat}
\end{figure}

\begin{figure}[ht!]
\begin{center}
\
  \includegraphics[width=\textwidth]{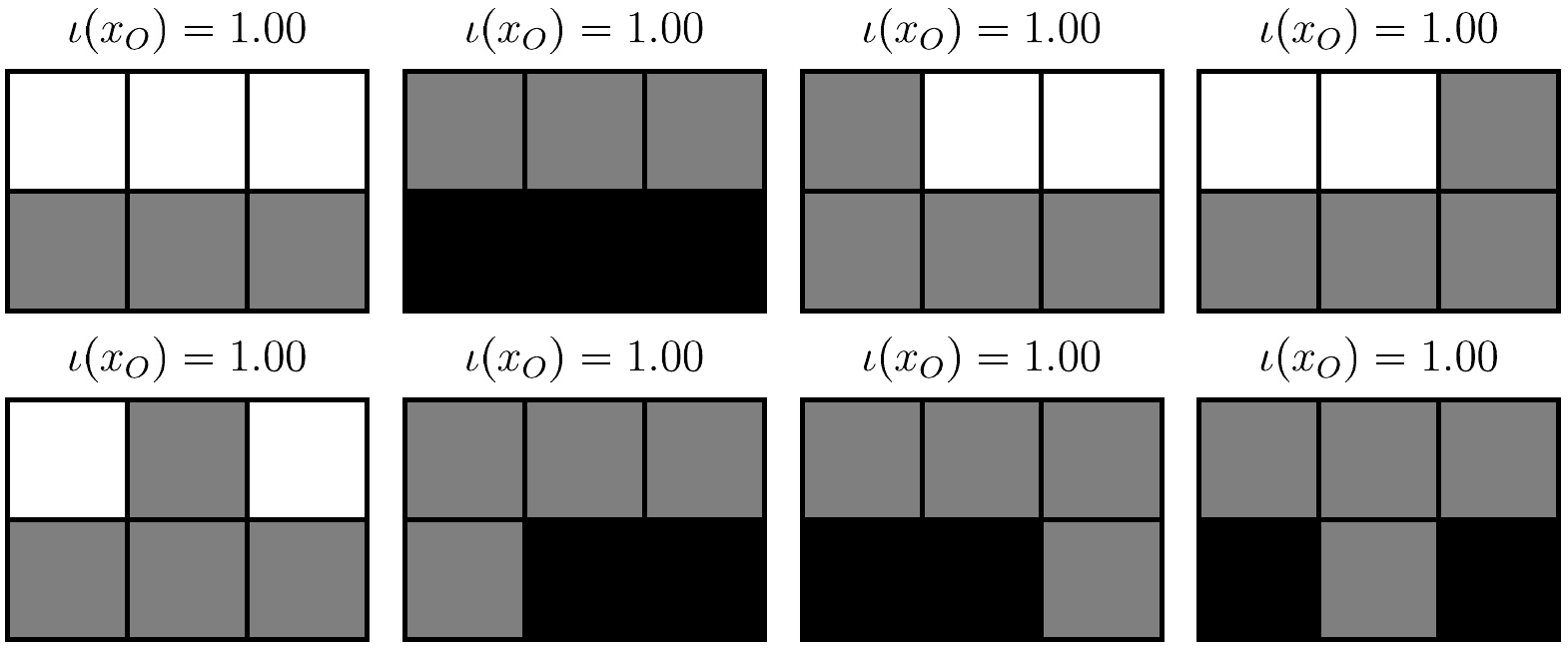}
\end{center}
\caption{%
All distinct completely integrated STPs on the second possible trajectory of $\MCconst$. The value of complete local integration is indicated above each STP.
     }%
   \label{fig:id2cipat2}
\end{figure} 

\subsection{Completely integrated STPs}

After having looked at the disintegration hierarchy extensively we now make use of it by extracting the completely (locally\footnote{When it is clear from context that we are talking about complete local integration we drop ``local'' for the sake of readability.}) integrated STPs of the four trajectories of $\MCconst$. Recall that due to the disintegration theorem (\cref{thm:disintegration}) we know that all blocks in partitions that occur in the refinement-free disintegration hierarchy are either singletons or correspond to completely integrated STPs. If we look at the refinement-free disintegration hierarchy in \cref{fig:rfdisid2} we see that many blocks occur in multiple partitions and across disintegration levels. We also see that there are multiple blocks that are singletons. If we ignore singletons since they are trivially integrated as they cannot be partitioned we end up with eight different blocks. Since the disintegration hierarchy is the same for all possible trajectories these blocks are also the same for each of them. However, the STPs that result are different due to the different values within the blocks. We show the eight completely integrated STPs and their complete local integration (\cref{def:ci}) on the first trajectory in \cref{fig:id2cipat} and on the second trajectory in \cref{fig:id2cipat2}.

Since the disintegration hierarchies are the same for the four possible trajectories of $\MCconst$ we get the same refinement-free partitions and therefore the same blocks containing the completely integrated STPs. This is apparent when comparing \cref{fig:id2cipat,fig:id2cipat2} and noting that each STP occurring on the first trajectory has a corresponding STP on the second trajectory that differs (if at all) only in the values of the cells it fixes and not in what values it fixes. More visually speaking, for each STP in \cref{fig:id2cipat} there is a corresponding STP in \cref{fig:id2cipat} leaving the same cells grey. 

\begin{figure}[ht!]
\begin{center}
\
  \includegraphics[width=\textwidth]{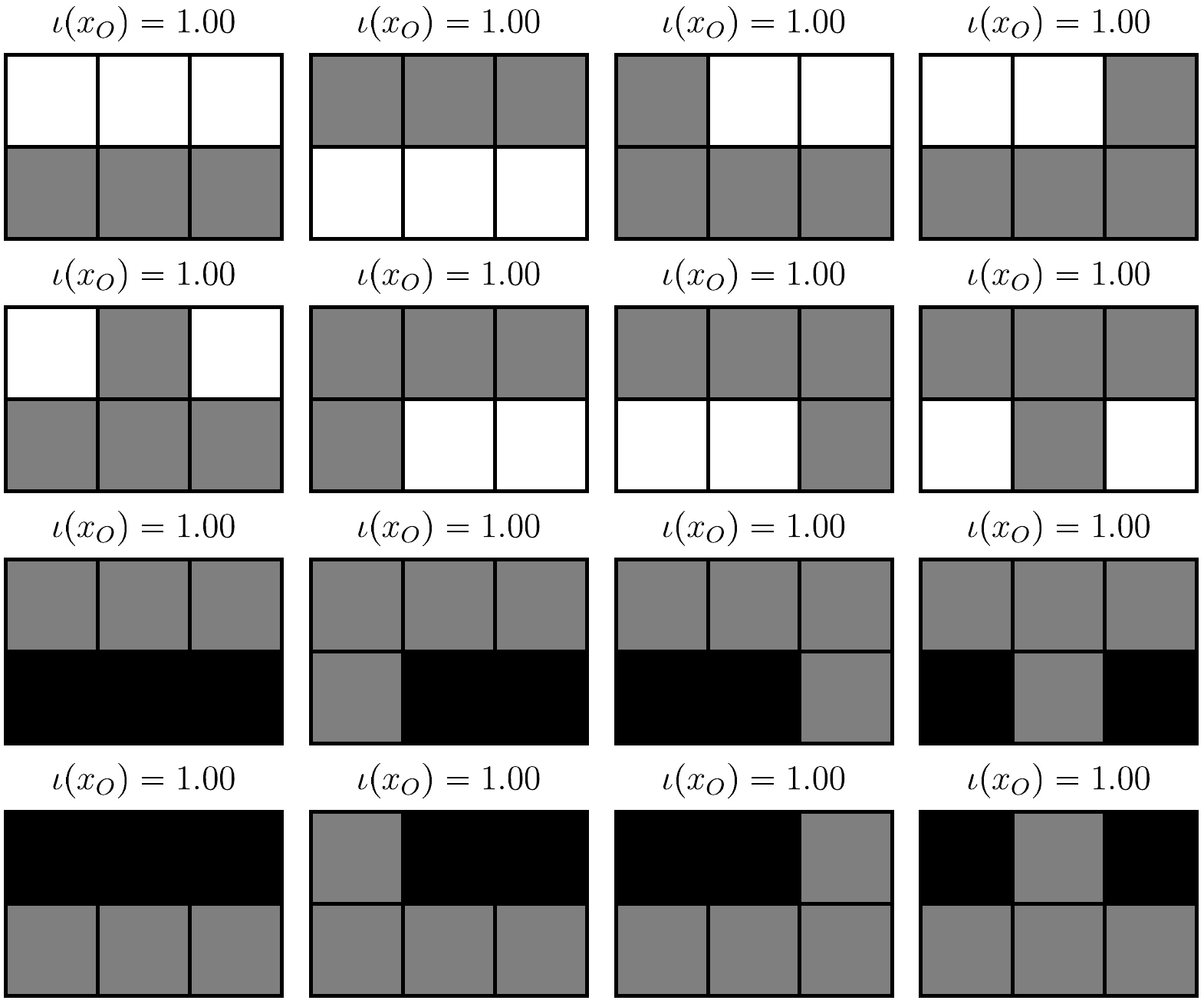}
\end{center}
\caption{%
All distinct completely integrated STPs on all four possible trajectories of $\MCconst$. The value of complete local integration is indicated above each STP.
     }%
   \label{fig:id2alltracipat}
\end{figure}

If we are not interested in a particular trajectory we can also look at all different completely integrated STP on any trajectory. For $\MCconst$ these are shown in \cref{fig:id2alltracipat}
We see that all completely integrated STPs $x_O$ have the same value of complete local integration $\ci(x_O)=1$. This can be explained using the deterministic expression for the SLI of \cref{eq:detmipi} and noting that for $\MCconst$ if any of the values $x_{j,t}$ is fixed by a STP then $(x_{j,s})_{s \in T}=x_{j,T}$ are determined since they must be the same value. This means that the number of trajectories $N(x_{j,S})$ in which any pattern $x_{j,S}$ with $S \subseteq T$ occurs is either $N(x_{j,S}) =0$, if the pattern is impossible, or $N(x_{j,S}) =2$ since there are two trajectories compatible with it. Note that all blocks $x_b$ in any of the completely integrated STP and all STP $x_O$ themselves are of the form $x_{j,S}$ with $S \subseteq T$. Let $N(x_{j,S})=:N$ and plug this into \cref{eq:detmipi} for an arbitrary partition $\pi$:
\begin{align}
    \mi_\pi(x_O)&= (|\pi|-1) \log |\X_{V_0}| - \log \frac{\prod_{b \in \pi} N(x_b)}{N(x_O)} \\
    &= (|\pi|-1) \log |\X_{V_0}| - \log \frac{N^{|\pi|}}{N} \\
    &= (|\pi|-1) \log \frac{|\X_{V_0}|}{N}.
\end{align}
To get the complete local integration value we have to minimise this with respect to $\pi$ where $|\pi|\geq 2$. So for $|\X_{V_0}|=4$ and $N=2$ we get $\ci(x_O)=1$.

Another observation is that the completely integrated STPs are all limited to one of the two rows. This shows on a simple example that, as we would expect, completely integrated patterns cannot extend from one independent process to another.

\section{Two random variables with small interactions}
\label{sec:noise}
In this section we look at a system almost identical to that of \cref{sec:id2} but with a kind of noise introduced. This allows all trajectories to occur and is designed to test whether the spatiotemporal patterns maintain integration in the face of noise.    
\subsection{Definition}
We define the time- and space-homogeneous multivariate Markov chain $\MCnoise$ via the Markov matrix $P$ with entries 
\begin{equation}
P_{f(x_{1,t+1},x_{2,t+1}),f(x_{1,t},x_{2,t})}=p_{J,t+1}(x_{1,t+1},x_{2,t+1}|x_{1,t},x_{2,t})
\end{equation} 
where we define the function $f:\{0,1\}^2\rightarrow [1:4]$ via 
\begin{equation}
f(0,0)=1,f(0,1)=2,f(1,0)=3,f(1,1)=4. 
\end{equation} 
With this convention $P$ is
\begin{equation}
P =\left(
\begin{array}{cccc}
 1-3 \epsilon  & \epsilon  & \epsilon  & \epsilon  \\
 \epsilon  & 1-3 \epsilon  & \epsilon  & \epsilon  \\
 \epsilon  & \epsilon  & 1-3 \epsilon  & \epsilon  \\
 \epsilon  & \epsilon  & \epsilon  & 1-3 \epsilon  \\
\end{array}
\right)
\end{equation} 
%
%
%
The initial distribution is again the uniform distribution
\begin{equation}
p_{j,0}(x_{j,0}) = 1/4.                             
\end{equation} 
Writing this multivariate Markov chain as a Bayesian network is possible but the conversion is tedious. The Bayesian network one obtains can be seen in \cref{fig:2cellnoisebn}. The state of \textit{both} random variables remains the same with probability $1-3 \epsilon$ and transitions into each other possible combination with probability $\epsilon$. In the following we set $\epsilon = 1/100$.

\begin{figure}
\begin{center}
\begin{tikzpicture}
    [myarrow,node distance=2cm]

    \node[] (1) [] {$X_{1,1}$};
    \node[] (2) [right of=1] {$X_{1,2}$};
    \node[] (3) [right of=2] {$X_{1,3}$};
    \node[] (5) [below of=1,node distance=1.5cm] {$X_{2,1}$};
    \node[] (6) [below of=2,node distance=1.5cm] {$X_{2,2}$};
    \node[] (7) [below of=3,node distance=1.5cm] {$X_{2,3}$};


    \path[myarrow]
      (1) edge node {} (2)
      (1) edge node {} (6)      
      (2) edge node {} (3)
      (2) edge node {} (6)
      (2) edge node {} (7)
      (3) edge node {} (7)

      (5) edge node {} (2)
      (5) edge node {} (6)
      (6) edge node {} (3)
      (6) edge node {} (7)
      

%



      ;
  \end{tikzpicture}
  \caption{Bayesian network of $\MCnoise$.}
  \label{fig:2cellnoisebn}
\end{center}
\end{figure}

\subsection{Trajectories}
\begin{figure}[ht!]
\input{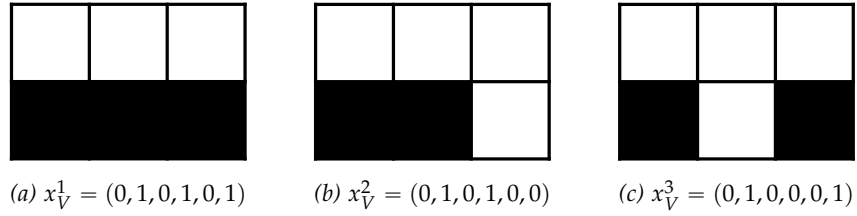}
\caption{%
        Visualisation of three trajectories of $\MCnoise$. In each trajectory the time index increases from left to right. There are two rows corresponding to the two random variables at each time step and three columns corresponding to the three time-steps we are considering here. We can see that the first trajectory (in \subref{fig:noisetra1}) makes no $\epsilon$-transitions, the second (in \subref{fig:noisetra2}) makes one from $t=2$ to $t=1$, and the third (in \subref{fig:noisetra3}) makes two.
     }%
   \label{fig:2cellnoisebntras}
\end{figure}

In this system all trajectories are possible trajectories. This means there are $2^6=64$ possible trajectories, since every one of the six random variables can be in any of its two states. There are three classes of trajectories with equal probability of occurring. The first class with the highest probability of occurring are the four possible trajectories of $\MCconst$. Then there are $24$ trajectories that make a single $\epsilon$-transition (i.e.\ a transition where the next pair is not the same as the current one $(x_{1,t+1},x_{2,t+1}) \neq (x_{1,t},x_{2,t})$, these transitions occur with probability $\epsilon$), and $36$ trajectories with two $\epsilon$-transitions. We pick only one trajectory from each class. The representative trajectories are shown in \cref{fig:2cellnoisebntras} and will be denoted $x^1_V,x^2_V$, and $x^3_V$ respectively. The probabilities are $p_V(x^1_V)=0.235225,p_V(x^2_V)=0.0024250,p_V(x^3_V)=0.000025$.
\begin{figure}[ht!]
\begin{center}
\
  \includegraphics[width=\textwidth]{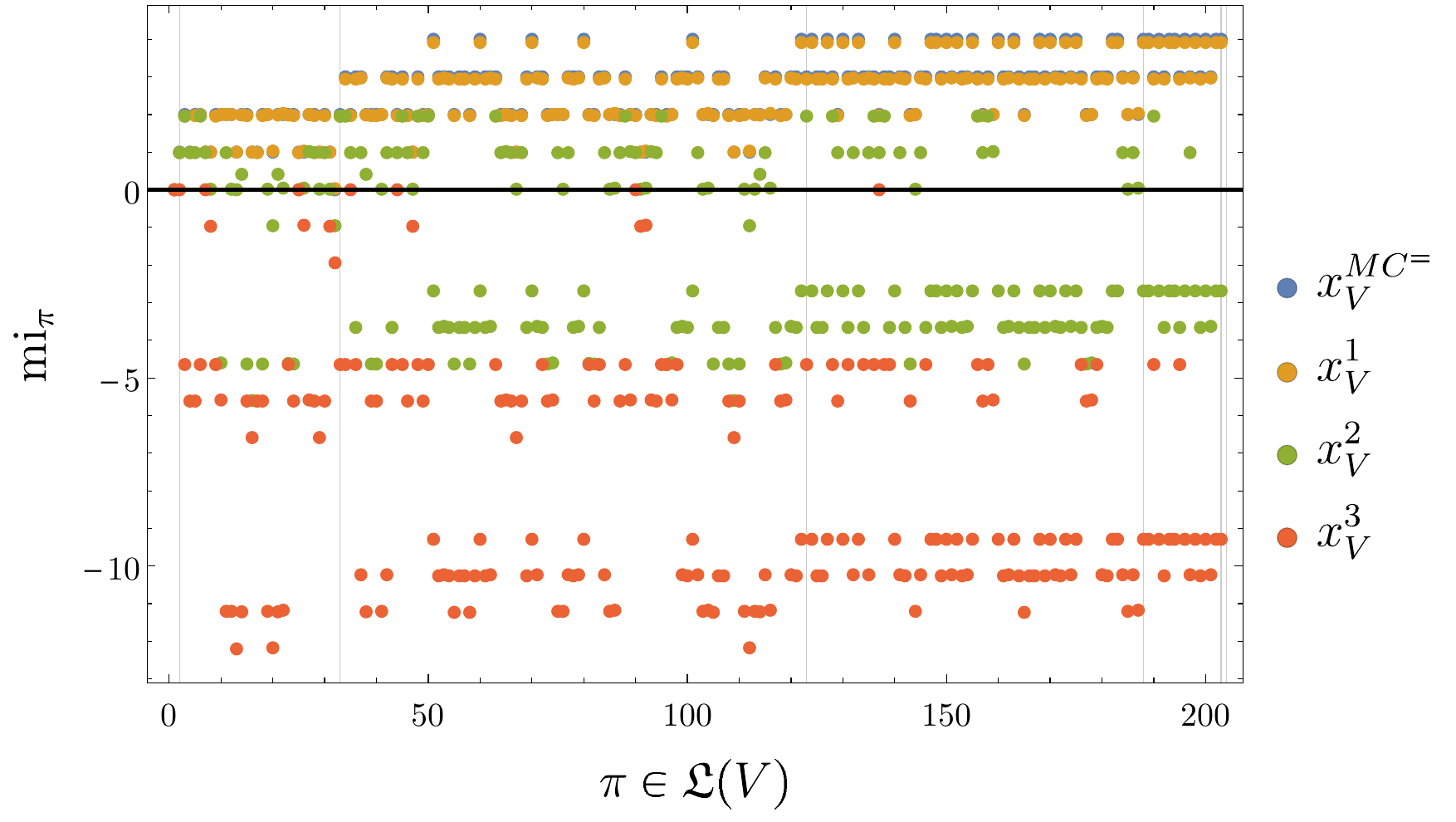}
\end{center}
\caption{%
        Specific local integrations $\mi_\pi(x_V)$ of one of the four trajectories of $\MCconst$ (measured w.r.t. the probability distribution of $\MCconst$), here denoted $x^\MCconst_V$, and the three representative trajectories $x^k_V, x \in \{1,2,3\}$ of $\MCnoise$ (measured w.r.t. the probability distribution of $\MCnoise$) seen in \cref{fig:2cellnoisebntras} with respect to all $\pi \in \Latt(V)$. The partitions are ordered as in \cref{fig:2cellidbnsli} with increasing cardinality $|\pi|$. Vertical lines indicate partitions where the cardinality $|\pi|$ increases by one. Note that the values of $x^\MCconst_V$ are almost completely hidden from view by those of $x^1_V$. }%
   \label{fig:2cellnoisebnsli}
\end{figure}

\subsection{SLI values of the partitions}
Again we calculate the SLI $\mi_\pi(x_V)$ of every trajectory $x_V$ with respect to each partition $\pi \in \Latt(V)$. 
In contrast to $\MCconst$ the SLI values with respect to each partition of $\MCnoise$ do depend on the trajectories. We plot the values of SLI with respect to each partition $\pi \in \Latt(V)$ for the three representative trajectories in  \cref{fig:2cellnoisebnsli}. 

It turns out that the SLI values of $x^1_V$ are almost the same as those of $\MCconst$ in \cref{fig:2cellidbnsli} with small deviations due to the noise. This should be expected as $x^1_V$ is also a possible trajectory. Also note that trajectories $x^2_V,x^3_V$ exhibit negative SLI with respect to some partitions. In particular, $x^3_V$ has non-positive SLI values with respect to any partition. This is due to the low probability of this trajectory compared to its parts. The blocks of any partition have so much higher probability than the entire trajectory that the product of their probabilities is still greater or equal to the trajectory probability.

\begin{figure}[ht!]
\begin{center}
\
  \includegraphics[width=\textwidth]{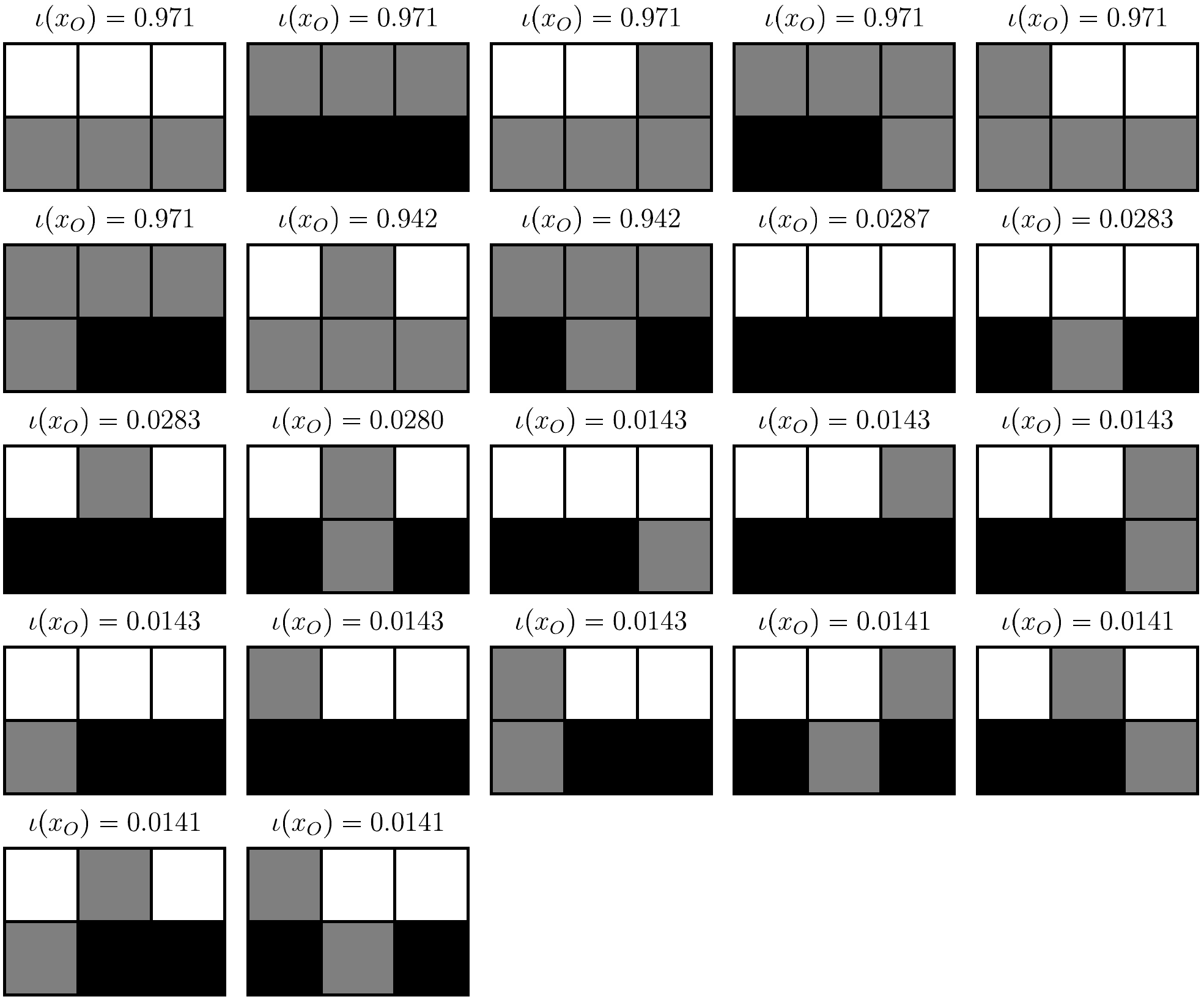}
\end{center}
\caption{%
All distinct completely integrated STPs on the first trajectory $x^1_V$ of $\MCnoise$. The value of complete local integration is indicated above each STP. See \cref{fig:id2cipat} for colouring conventions.
     }%
   \label{fig:noise2cipat1}
\end{figure} 

\subsection{Completely integrated STPs}
\begin{figure}[ht!]
\begin{center}
\
  \includegraphics[width=\textwidth]{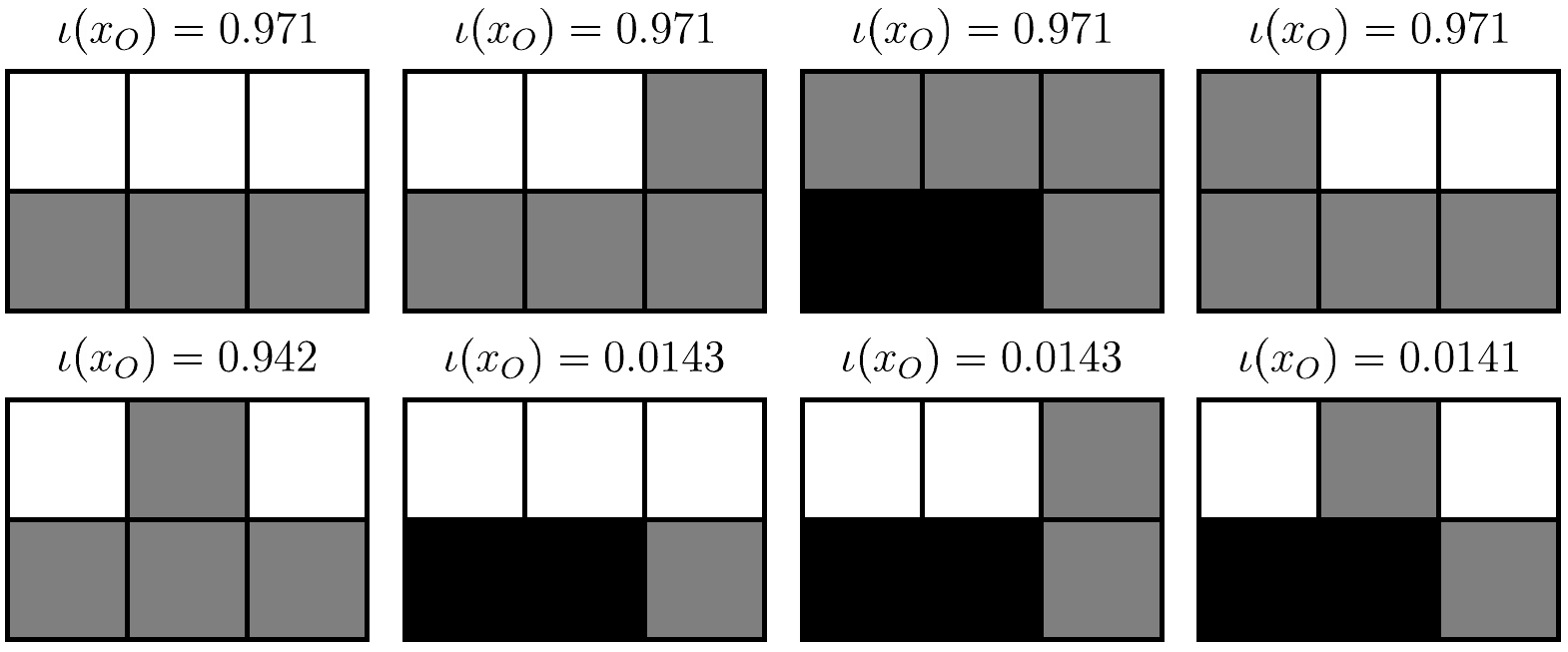}
\end{center}
\caption{%
All distinct completely integrated STPs on the second trajectory $x^2_V$ of $\MCnoise$. The value of complete local integration is indicated above each STP.
     }%
   \label{fig:noise2cipat2}
\end{figure}

\begin{figure}[ht!]
\begin{center}
\
  \includegraphics[width=\textwidth]{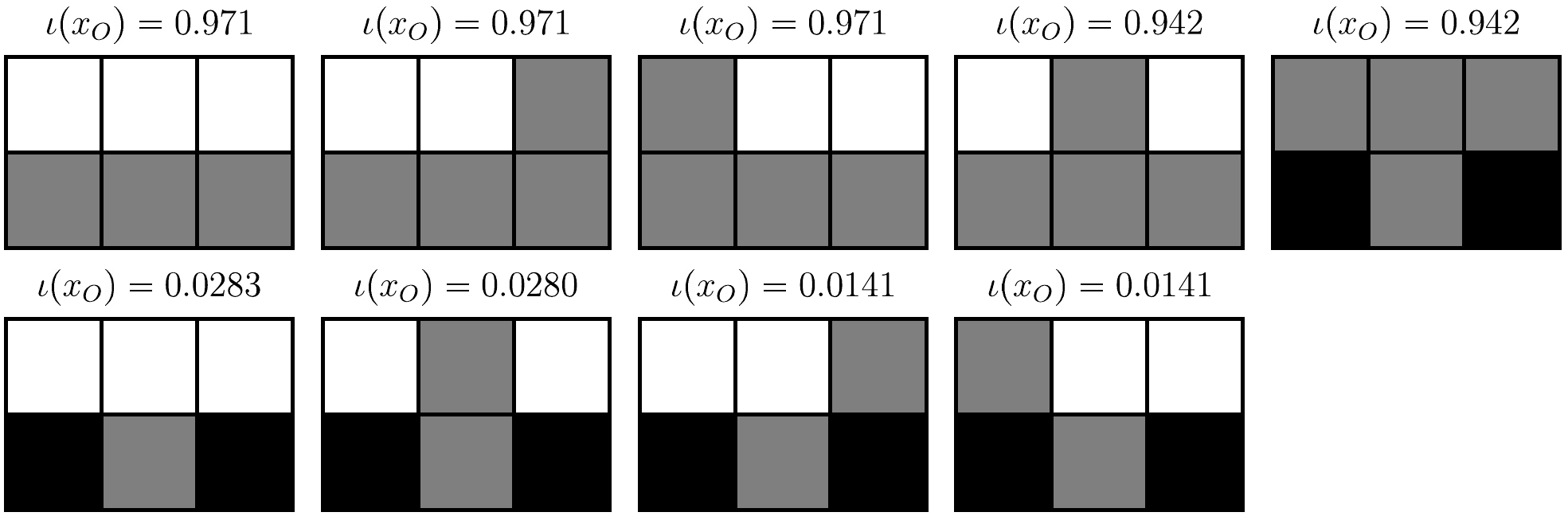}
\end{center}
\caption{%
All distinct completely integrated STPs on the third trajectory $x^3_V$ of $\MCnoise$. The value of complete local integration is indicated above each STP.
     }%
   \label{fig:noise2cipat3}
\end{figure}

In this section we look at the completely integrated STPs for each of the three representative trajectories $x^k_V, k \in \{1,2,3\}$. They are visualised together with their complete local integration values in \cref{fig:noise2cipat1,fig:noise2cipat2,fig:noise2cipat3}. In contrast to the situation of $\MCconst$ we now have completely integrated STPs with varying values of complete local integration. 

On the first trajectory $x^1_V$ we find all the eight STPs that are completely locally integrated in $\MCconst$ (see \cref{fig:id2cipat2}). These are also more than an order of magnitude more integrated than the rest of the completely integrated STPs.


\section{Completely locally integrated spatiotemporal patterns as entities}
\label{sec:exampleconnection}
In \cref{sec:identity} we have proposed to use the set of completely integrated spatiotemporal patterns to solve the identity problem. This means using the completely integrated spatiotemporal patterns as an entity set. There we called this entity set the $\ci$-entities. In this section we look at the completely integrated spatiotemporal patterns found in \cref{sec:id2,sec:noise} with regard to the phenomena and properties of entities described in \cref{sec:entinmvmc}.

In \cref{sec:entinmvmc} we have described three phenomena that should not be precluded by a formal notion of entities. These are 
\begin{enumerate}
  \item compositionality,
  \item degree of freedom traversal, and
  \item counterfactual variation.
\end{enumerate}
We show by example that $\ci$-entities can exhibit these three phenomena.

Regarding compositionality we can see in \cref{fig:id2alltracipat} that $\MCconst$ contains $\ci$-entities that are composed of multiple temporal parts. More precisely $\MCconst$ contains $\ci$-entities that occupy random variables in multiple time-slices. For example the $\ci$-entity:
\begin{center}
  \includegraphics[width=.28\linewidth]{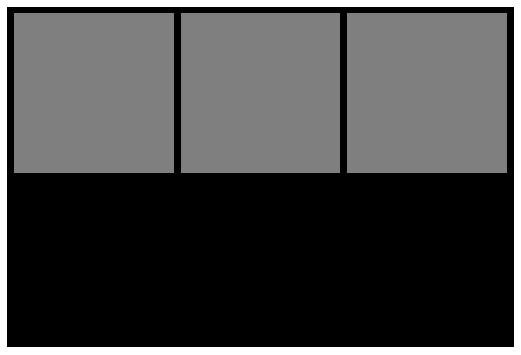}
\end{center}

While $\MCconst$ contains no $\ci$-entities that are composed of multiple spatial parts such $\ci$-entities exist in $\MCnoise$ on all three kinds of trajectories as we can see in \cref{fig:noise2cipat1,fig:noise2cipat2,fig:noise2cipat3}. An example would be the $\ci$-entity:
\begin{center}
  \includegraphics[width=.28\linewidth]{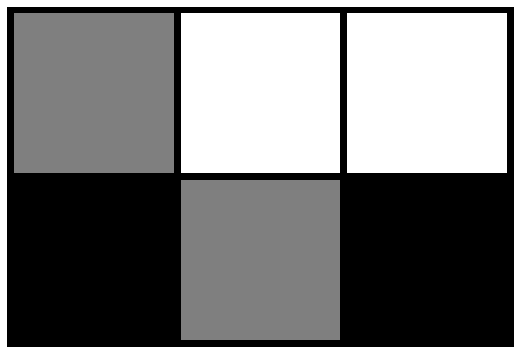}
\end{center}
This entity is also temporally composite.
These entities still have much lower $\ci$ values than for the entities that occupy only one of the random variables in multiple time-slices. However, this still shows that there are spatially, temporally, and spatiotemporally composite $\ci$-entities.

Regarding degree of freedom traversal, we can see that, for example, the $\ci$-entity 
\begin{center}
  \includegraphics[width=.28\linewidth]{./images/tra1doftraentity.pdf}
\end{center}
changes the degrees of freedom that it occupies from the bottom one to the top one and then to both in the two time-steps. So there are degree of freedom traversing $\ci$-entities.\footnote{Since there are also entities that do not traverse degrees of freedom in the first time-step we can also not just rename the indices at the second time-step to get rid of all degree of freedom traversing entities.} 

Regarding counterfactual variation, the $\ci$-entities are counterfactual \textit{in value only} for $\MCconst$. For example the following two $\ci$-entities from the first and second possible trajectory differ in value on the same set of of occupied random variables:
\begin{center}
  \includegraphics[width=.6\linewidth]{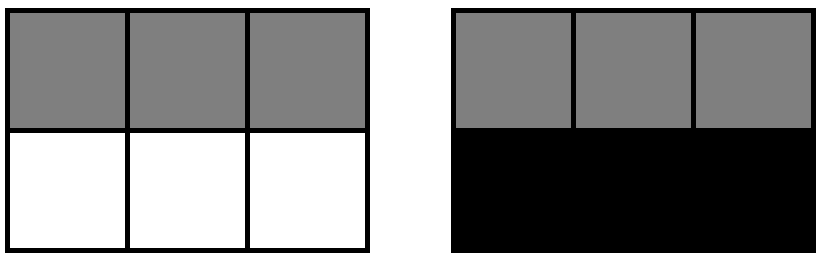}
\end{center}
But the sets of random variables occupied by the $\ci$-entities of $\MCconst$ are the same in all trajectories. 

In $\MCnoise$ on the other hand we find that the sets of random variables occupied by the $\ci$-entities differ from one trajectory to another. For example the $\ci$-entities
\begin{center}
  \includegraphics[width=.6\linewidth]{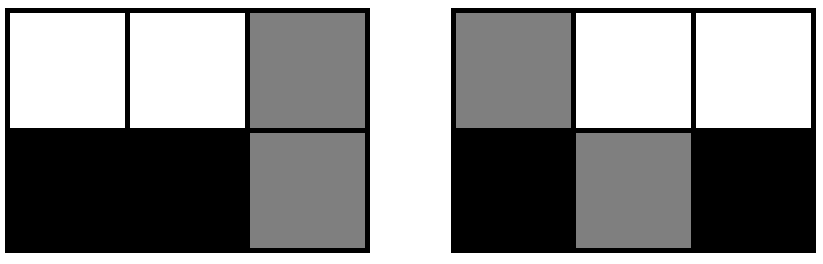}
\end{center}
which occur in $x^2_V$ and $x^3_V$ respectively occupy sets of random variables that no entity in the other trajectory occupies. So there the entity set of $\ci$-entities for $\MCnoise$ is counterfactual in extent.

We only note briefly that there are some counter intuitive $\ci$-entities that skip the second times-step for both $\MCconst$ and $\MCnoise$ \cref{fig:noise2cipat1,fig:noise2cipat2,fig:noise2cipat3}. Whether these are due to the small scale of the system or do occur more generally will be further investigated in the future.

\section{Action and perception of \texorpdfstring{$\ci$}{j}-entities}
\label{sec:actperceptexamples}
In this section we briefly present examples of $\ci$-entities that exhibit actions and perceptions according to our definitions. We prove by example that that $\ci$-entities can perform value and extent actions. We also show that non-interpenetration is not necessarily satisfied by $\ci$-entities as it is not satisfied in our example systems $\MCconst$ and $\MCnoise$. This implies that the co-perception entities are not necessarily mutually exclusive and in fact we find example co-perception entities that are not mutually exclusive. We can still use a subset of mutually exclusive entities with common past to extract some perceptions. These are not uniquely defined however and we also choose a different subset and show it has a slightly different branch-morph as well. We also show that the same $\ci$-entity can perform an action and perceive something in the same transition from one step to the next. 

\subsection{Actions of \texorpdfstring{$\ci$}{j}-entities in \texorpdfstring{$\MCconst$}{MCconst} and \texorpdfstring{$\MCnoise$}{MCeps}}
According to our definition of entity action (\cref{def:action}) there are actions performed by $\ci$-entities in $\MCconst$ and $\MCnoise$. For example the $\ci$-entities
\begin{center}
  \includegraphics[width=.6\linewidth]{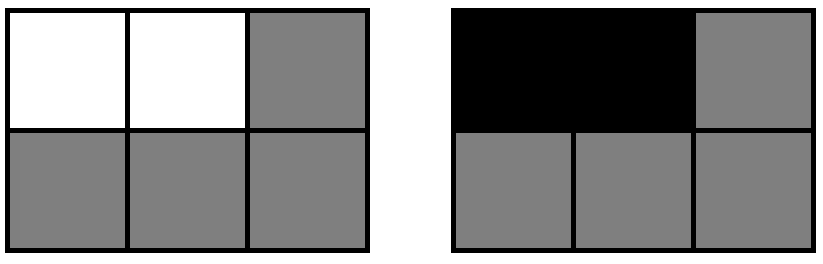}
\end{center}
are co-action entities at the first time-step and their second time-step time-slices are co-actions. 
\begin{itemize}
  \item They occur in the different and possible co-action trajectories:
\begin{center}
  \includegraphics[width=.6\linewidth]{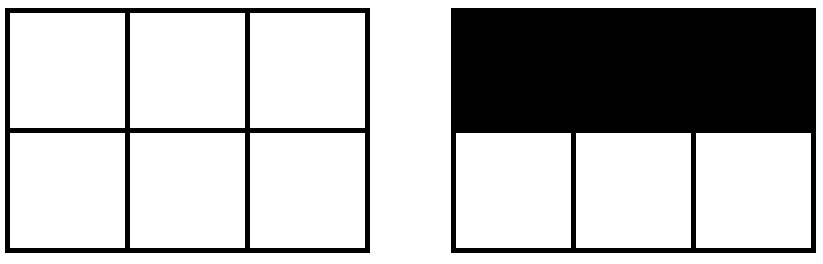}
\end{center}
\item They occupy the same random variable at the first time-step.
\item The environments at the first time-step in the two trajectories are identical.
\item And at the second time-step the two $\ci$-entities differ.
\end{itemize}
For the environment it is therefore impossible to be sure about the next configuration of the entity that co-occurs at the first time-step. 

Since the above spatiotemporal patterns are $\ci$-entities in the shown trajectories in both $\MCconst$ and $\MCnoise$ they are co-action entities in both chains. 

Note that the above is a value action. The time-slices at the second time-step differ only in the value they assign to the top random variable. In $\MCconst$ there are no extent actions, but in $\MCnoise$ there are. The following two entities are co-action entities at the first time-step in co-action trajectories $x^2_V$ and $x^3_V$ of $\MCnoise$:

\begin{center}
  \includegraphics[width=.6\linewidth]{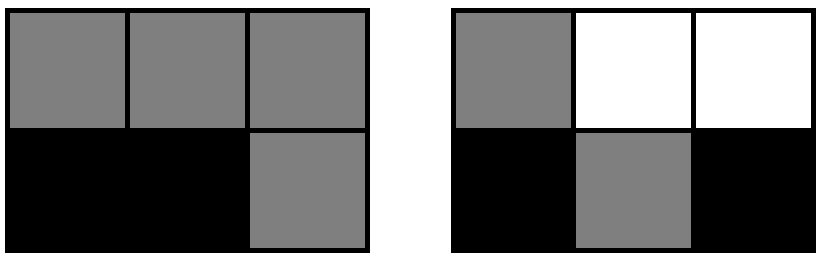}
\end{center}

These entities occur in the different and possible co-action trajectories $x^2_V$ and $x^3_V$:
\begin{center}
  \includegraphics[width=.6\linewidth]{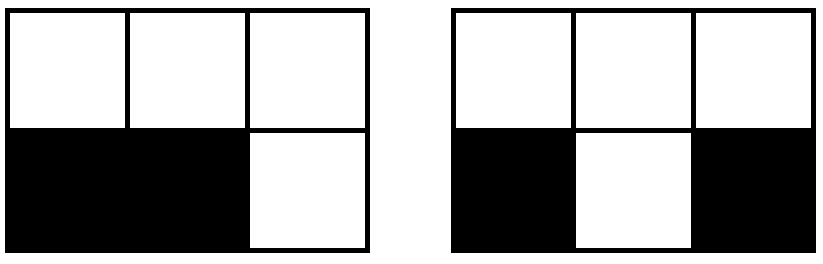}
\end{center}
and also fulfil the other conditions for actions mentioned above. The action is an extent action since the occupied variable is the bottom variable for the left entity and the top variable for the right entity. This shows that there are $\ci$-entities that perform value actions and those that perform extent actions. 

The extent action above also reveals a challenge to $\ci$-entities. We argued in \cref{sec:actions} that the differences at $t+1$ of two entities with the same environments at $t$ should be due to the entity or random. In the above case the entities (and their environments in the according trajectories) are equal at $t$ so the differences between the two at $t+1$ are due to the noise. We suggested that it is the task of the entity-set to exclude such random parts from entities. The $\ci$-entities might therefore need further adaptations or the notion of actions might need to be changed. A third possibility is that due to the ``global'' construction of entities there are some non-intuitive effects. Another such effect can be seen are the entities that skip an entire times-step that we mentioned in \cref{sec:exampleconnection}.

Finally, we note that there are also more intuitive co-action entities that perform extent actions. For example the co-action entities
\begin{center}
  \includegraphics[width=.6\linewidth]{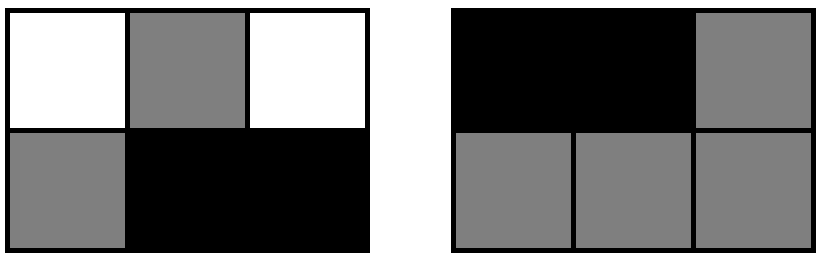}
\end{center}
which occur in the co-action trajectories
\begin{center}
  \includegraphics[width=.6\linewidth]{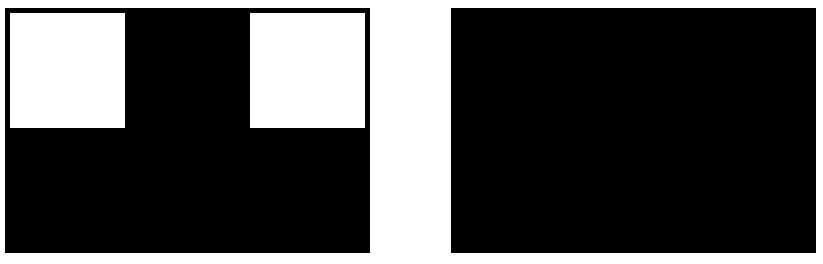}
\end{center}
have different ``internal'' values at the first time-step which can be interpreted as the reason for the different extent at the second time-step. In the light of the previous extent action however this interpretation is questionable and needs further investigation.

\subsection{Perceptions of \texorpdfstring{$\ci$}{j}-entities in \texorpdfstring{$\MCconst$}{MCconst} and \texorpdfstring{$\MCnoise$}{MCeps}}
Regarding perception we first note that the example chains $\MCconst$ and $\MCnoise$ allow interpenetration of $\ci$-entities. We can see that there are interpenetrating $\ci$-entities in all trajectories of the systems considered here. For example the $\ci$-entities 
\begin{center}
  \includegraphics[width=.6\linewidth]{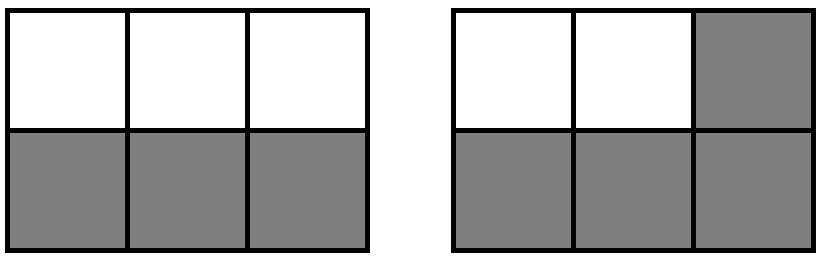}
\end{center}
both occur in each of the three trajectories of $\MCnoise$. They are not identical patterns and still both fix partly the same random variables. Yet, as they occur in the same trajectories the probability that they co-occur is non-zero. This violates non-interpenetration as defined in \cref{def:noninterpen}. Therefore the dynamics of the systems considered here do not support the definition of perception proposed in \cref{sec:perceptions}. This implies that by itself the choice of the $\ci$-entities as an entity set does not necessarily lead to non-interpenetration. As mentioned before it is not clear at what level non-interpenetration should be required. 

Next, note that in $\MCconst$ there are no co-perception entities for any of its $\ci$-entities (see \cref{fig:id2alltracipat}). For any given entity $x_A$ with non-empty time-slices at $t$ and $t+1$ there is no other entity $y_B$ that has identical past $y_{B_\pet}=x_{A_\pet}$ and differs at $t+1$. This is due to the fact that the future of any entity is completely determined by its current state regardless of the rest of the system / environment. 

In $\MCnoise$ we do find co-perception entities. Due to the interpenetration of entities however we cannot use the full set of co-perception entities.
As mentioned in \cref{sec:branchmorph} we can use a mutually exclusive subset $\zeta(x_A,t)$ of the co-perception entities as a proxy for the co-perception partition. The simplest case is to use only two entities, the original entity $x_A$ and a co-perception entity $y_B$ so that $\zeta(x_A,t) = \{x_A,y_B\}$. The co-perception entity $y_B$ must be chosen such that
\begin{itemize}
\item $\Pr(X_A=x_A,X_B=y_B)=0$,
\item $x_{A_{t+1}} \neq y_{B_{t+1}}$.
\end{itemize}
The first condition so that it is mutually exclusive the second so that it is part of a different branch in the branching partition. This gives us a branching partition $\eta(x_A,t)=\{\{x_a\},\{y_B\}\}$.
An example of such a co-perception pair are the following $\ci$-entities of $\MCnoise$:
\begin{center}
  \includegraphics[width=.6\linewidth]{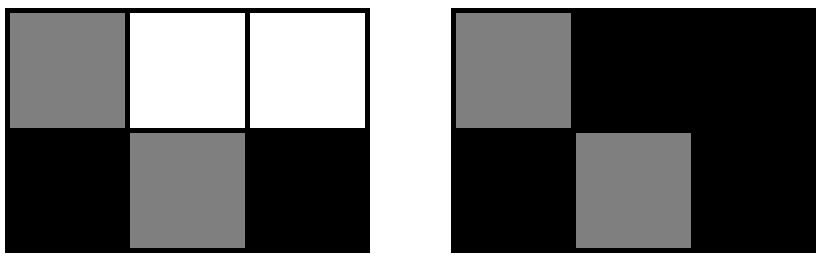}
\end{center}
So here $A=\{(2,0),(1,1),(1,2),(2,2)\}$ with $x_A =(1,0,0,1)$ and $B=A$ with $y_B=(1,1,1,1)$. Since $x_{A_\pet}=y_{B_\pet}$ and $A_0,A_1 \neq \emptyset$ we indeed have two co-perception entities $y_B \in \perstp(x_A,0)$. The two co-perception entities perceive the difference between two co-perception environments at the first time-step $t=0$. Since every trajectory is possible in $\MCnoise$ every environment at the first time-step is a co-perception environment so that $\X^\perstp_{V_t \bs A_t} = \X_{V_t \bs A_t}$. The environments that will be classified are then the possible values $\{0,1\}$ of the random variable $X_{V_t \bs A_t}=X_{1,0}$ that is not fixed by the two co-perception entities. 

Each environment $x_{1,0} \in \X_{1,0}$ has an associated branch-morph $p(.|x_{1,0},x_{A_\pet}):\eta(x_A,t)\rightarrow [0,1]$ over the two entities. The two morphs turn out to be:

  \begin{equation}   
    \begin{array}{|c|c|c|}
\hline
  & x_A & y_B \\
\hline
p(.|0,x_{A_\pet})      & 4705/4754 & 49/4754  \\
\hline
p(.|1,x_{A_\pet}) & 49/4754 & 4705/4754  \\
\hline
\end{array}
\end{equation}
We see that they are indeed different such that the branch-morph partitions $\pi^\perstp$ consists of two blocks containing one co-perception environment each $\pi^\perstp=\{\{X_{1,0}=0\},\{X_{1,0}=1\}\}$. The elements of this partition are then the perceptions.

Finally, we can see here that the choice of a co-perception entity is not unique. We can also choose an entity $z_C\neq y_B$ to get another co-perception pair. Another such possibility with the same $x_A$ is the following pair:
\begin{center}
  \includegraphics[width=.6\linewidth]{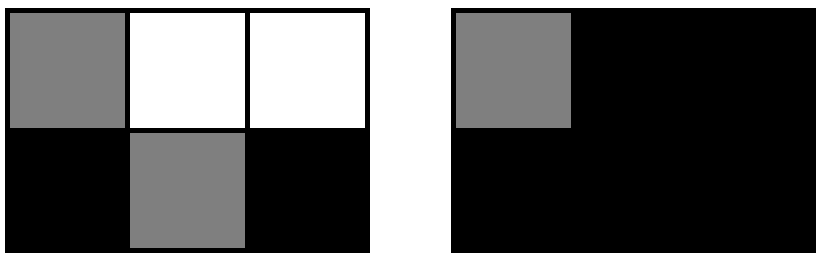}
\end{center}
Here $C=\{(2,0),(1,1),(2,1),(1,2),(2,2)\}$ and $z_C=(1,1,1,1,1)$. Note that $z_C$ and $y_B$ differ at the second time-slice but are not mutually exclusive:
\begin{equation}
  Pr(X_B=y_B,X_C=z_C)=4753/20000>0.
\end{equation} 
Therefore, they cannot be put into the same branch (because they differ at $t+1$) but they also cannot be put into different branches (because they can occur together). This is possible because they interpenetrate and makes it difficult to find a unique co-perception partition. 

The perceptions of the co-perception pair $x_A,z_C$ are the same as for $x_A,y_B$ but this is due to the limits of our example. The branch morphs are different:
  \begin{equation}   
    \begin{array}{|c|c|c|}
\hline
  & x_A & z_C \\
\hline
p(.|0,x_{A_\pet})      & 9410/9507 & 97/9507  \\
\hline
p(.|1,x_{A_\pet}) & 98/9507 & 9409/9507  \\
\hline
\end{array}
\end{equation}
This indicates that in general the perceptions of different choices of $\zeta(x_A,t)$ co-perception entities are also different.

\subsection{Action and perception of the same \texorpdfstring{$\ci$}{j}-entity at the same time}
In the previous sections we have seen co-action entities and co-perception entities. We only want to emphasise here that an entity can have a co-perception pair and a (different) co-action pair at the same time $t$. This means that action and perception do not preclude each other. An example is the entity $x_A$ from our perception example. It is the entity on the left in
\begin{center}
  \includegraphics[width=.6\linewidth]{./images/coperpair.pdf}
\end{center}
shown together with its co-perception entity for the first time-step. However, $x_A$ also occurs as a co-action at the first time-step of another entity. These co-actions (with $x_A$ on the right) are 
\begin{center}
  \includegraphics[width=.6\linewidth]{./images/coextendactions.pdf}
\end{center}
So the same $\ci$-entity can perform actions and have perceptions the same time-step.

\section{Discussion}
In this chapter we presented two very simple multivariate Markov chains $\MCconst$ and $\MCnoise$. In \cref{sec:id2} We calculated the disintegration hierarchy and the refinement free disintegration hierarchy of $\MCconst$. We explained the occurrence of multiple identical disconnected components in the partially ordered disintegration levels. These are due to the invariance of the specific local integration as revealed by \cref{thm:symmpart,thm:symmpartcor}. We then presented the set of completely locally integrated spatiotemporal patterns of $\MCconst$. This corresponds to the $\ci$-entities of $\MCconst$. These exhibit compositionality in time but not in space and counterfactual variation in value but not in extend. They do not exhibit degree of freedom traversal as should be expected due to the independent dynamics. All the entities of $\MCconst$ have the same $\ci$ value of $1$ bit. This is due to the four possible trajectories having identical probabilities and the deterministic dynamics. 

In \cref{sec:noise} we extracted the $\ci$-entities of $\MCnoise$. These exhibit compositionality in time and space, counterfactual variation in value and extend, and degree of freedom traversal. The $\ci$-entities of $\MCnoise$ also attain various $\ci$ values ranging from $0.014$ bit to $0.971$ bit. 

In \cref{sec:actperceptexamples} we turned our attention to entity action and entity perception of the $\ci$-entities of $\MCconst$ and $\MCnoise$. There are entity actions in $\MCconst$ but only in value. For our weak definition of actions this is not surprising since the entities in $\MCconst$ are self-determining and independent of the rest of the system.

In $\MCnoise$ we also find entity actions in extend. Here we also find an example of an extent action that suggests that either the entity action requirements are too weak or the $\ci$-entities do not provide the internal connection between their parts that we expected. We will discuss this further below.

Concerning perception, we find no perceptions in $\MCconst$ which should be expected for parts of an independent process. In our formalisation of perception this leads to an empty set of co-perception entities. For $\MCnoise$ there are co-perception entities. However, we cannot use the unique construction of a branching partition due to interpenetration of $\ci$-entities and non-exclusion of co-perception entities. We therefore used the approach that only relies on a set $\zeta$ of mutually exclusive co-perception entities to define the branch-morph. This resulted in perception being defined for entities in $\MCnoise$. We also saw that another choice of a co-perception entity can lead to a (quantitatively) different perception. This further confirmed that without non-interpenetration perception is not necessarily uniquely defined.

Finally, we showed that the same $\ci$-entity can perform entity actions and entity perception at the same time-step $t$. This is achieved simply by finding a co-action entity and a (usually different) co-perception entity for the same original entity at the same time-step. The story of perceptions and actions of an entity is then defined via the co-action and co-perception entities along its time-evolution.

The combination of $\ci$-entities with entity action and entity perception then fulfils quite a few expectations and requirements that we have discussed before. However, there are also some things that are not easily interpreted. One thing briefly mentioned before are entities that skip a time-step. Such entities occur in both systems $\MCconst$ and $\MCnoise$. A second thing is the extent action that seemed to be random even though it was part of an entity. This could mean at least four different things: 
\begin{enumerate}
  \item The entity action definition is too weak and should require explicitly that the time-slice $x_{A_t}$ of the entity at $t$ determines to some degree the time-slice $x_{A_{t+1}}$ at $t+1$. This seems to speak against the idea of starting from entities in the first place. If actions explicitly require a connection between them from one step to the next why have an additional notion of entity? Can we not only define actions (and perceptions) in this case and either have the entity emerge or ignored it completely? This is a valid approach and is sometimes discussed \footnote{Personal communication with Nathaniel Virgo.} but not much in keeping with the ``entity-first'' point of view in this thesis.
  \item The notion of $\ci$-entity is not strong enough to ensure that there are internal relations between all parts of the entities. This is possible but would be surprising. The requirement that all possible partitions have lower probability than the whole seems like a quite strong condition. 
  \item The situation is not really a problem but is just not intuitive. We have also mentioned the time-step skipping $\ci$-entities which are not intuitive in a Markov chain. It may be that the global way that $\ci$-entities are defined connects the parts of entities in subtle ways. Maybe the action that seemed random above is not that random after all if seen from a global perspective (here ``global'' includes the entire Bayesian network, all time-steps and degrees of freedom). This ,however, still questions our motivation for our action definition. There we argued that if the environments are equal and the next time-slices different then the difference at the next time-step must originate from the agent (which it does not in the above example) or is random (which must then be wrong). So there either must be a third possibility due to some global effects or we have to accept that randomness is a kind of (proto-)action. A third possibility seems strange to us but on the other hand there are sometimes strange effects in probability and information theory. 
  \item Goal-directedness saves the situation. We expect that goal-directedness induces a connection between the entity perceptions and the entity actions. The entity actions have to somehow be adapted to the entity perceptions in order to get goal-directed behaviour. Then the example of an action above could stay an ``action'' or proto-action but the according entity could never be an agent since it cannot be goal-directed. This view still questions our motivation for the entity action definition. Again it seems we would have to accept random events as actions.   
\end{enumerate}
None of these possibilities is unproblematic. So more research is needed.

A third thing concerns non-interpenetration. We have seen that it provides uniquely defined branch-morphs and perceptions. However the $\ci$-entities do interpenetrate. This leads to multiple coexisting but different perceptions for the same entity. There might be ways to define perception in a way similar to ours that deals with interpenetrating entities in a unique way and there might be completely different definitions of perception where the problem does not occur. However, the solution closest to the work in this thesis might be to get rid of non-interpenetrating entities. There could be well motivated ways to only use a (non-interpenetrating) subset of all $\ci$-entities as an entity set. The partitions in the disintegration hierarchy all partition the system into non-interpenetrating blocks. We have in this thesis considered all blocks of all partitions together as the $\ci$-entities. It is maybe not too far fetched to select blocks from only a few partitions or even a single partition as the entity-set. For example only blocks of partitions that are in some way on the same ``scale'' as others e.g.\ same cardinality, same block size, etc. This is related to the question of whether only entities on the same ``scale'' cannot interpenetrate but those on different ``scales'' can. We could also try to settle for a single partition, possibly the one in the refinement-free disintegration hierarchy with the absolute minimum of SLI. This would lead to non-interpenetration. This is future research.

There are also further requirements for entities that may be derived from phenomena associated to living organisms. Examples of such phenomena are birth, death, growth, replication, etc. The implications of those will also be investigated in the future.

On the more technical side we would like to extend our definition of the branch-morph and try to generalise information theoretic measures for it. 

Finally, there is the lack of a definition of \textit{entity goal-directedness}. Even in the event that the definitions presented in this thesis turn out to be correct, entity goal-directedness has to weave the entity actions and entity perceptions together. So that they make some kind of sense as a sequence. This is essential if we want to give a full account of agents in lawful systems.

\chapter{Conclusion}
\label{ch:conclusion}
\label{sec:exdiscussionoutlook}
The overarching goal of this thesis was to further improve our understanding of how the notion of agents can be compatible with a lawful universe. For this we attempted further formalisations of the notions identified in the literature as constitutive of agents. These are entities, perception, action, and goal-directedness. We made no contribution to the formalisation of goal-directedness. 

With respect to entities we contributed a novel measure of integration called complete local integration (CLI). We have proposed this also as possible basis for a definition of entities. However, we first analysed this notion in the more general context of Bayesian networks. CLI is based on the specific local integration (SLI). We proved upper bounds constructively and constructed a candidate for a lower bound for SLI. We defined the disintegration-hierarchy and its refinement free version. Using these we revealed a relation between the finest partitions of global trajectories achieving a certain SLI value and the CLI of the blocks of these partitions. All blocks of such partitions are completely locally integrated. Conversely, all completely locally integrated spatiotemporal patterns are a block in at least one such partition. We expect that this result can be used to further investigate CLI and SLI formally. We presented an interpretation of this result from coding theory. This still needs rigorous analysis but seems promising (\cref{sec:interpret}).

We also established the transformation of SLI under permutations of nodes in Bayesian networks and showed how they can be applied for the explanation of the disintegration hierarchy in an example \cref{sec:exstp}. The transformation properties are also important for future theoretical work.

In \cref{ch:agents} we presented three criteria for formal definitions of entities. These are compositionality, degree of freedom traversal, and counterfactual variation. Counterfactual variation has two realisations: in value and in extent. Based on these criteria we concluded that sets of random variables are not suitable for general agent definitions. We therefore defined general entity-sets as subsets of the set of all spatiotemporal patterns of a multivariate Markov chain. This definition turned out to be of considerable conceptual value. The main reason for this is that the perception-action loop also has an associated entity-set. By using this entity-set the new notions of entity action and entity perception defined for arbitrary entity-sets naturally transform into notions known from the perception-action loop literature. The perception-action loop entity set is exhaustive,  does not traverse degrees of freedom and only varies counterfactually in value. The general entity sets can vary counterfactually in extent, traverse degrees of freedom freely, and need not be exhaustive. 

Our entity action definition was shown to imply non-heteronomy, an information theoretically defined notion related to autonomy due to \citet{bertschinger_autonomy_2008}. Entity perception was shown to correspond to a standard concept of perception in the perception-action loop. Entity perception is based on our construction of the branch-morph. This is a generalisation of the probability of the next agent state given current agent and environment states. This conditional probability distribution underlies the standard construction of perception in the perception-action loop. However, it also occurs in other concepts defined for the perception-action loop. The generalisation of such notions should now be feasible. Constructions similar to the branch-morph for other conditional probability distributions are also thinkable. We hope that this will lead to a generalised theory of perception and action for entity-sets. 

During the construction of the branch-morph it became clear that it is uniquely defined if the entity-set satisfies non-interpenetration. This notion therefore seems worth investigating with respect to an axiom for entity sets. We were also able to show that the assumptions we need to make on the co-perception environments (which are the environments that can be perceived or distinguished by an entity) only appear stronger than the assumptions inherent in the perception-action loop and are actually equivalent. This is further support for our method of generalisation. 

We noted that both entity actions and entity perception have multi-time-step analogues. Similar constructions are also used in perception-action loops. We also noted in the discussion of related work that our notion of perception can be seen as a formal generalisation to stochastic settings of the macroperturbations of the cognitive domain as defined for the glider in \citet{beer_cognitive_2014}.

Concerning simulation results we have calculated the disintegration hierarchies, visualised them and explained their structure using the SLI symmetry theorems. We also calculated the $\ci$-entities (the completely locally integrated spatiotemporal patterns) for both example systems. For the example system with noise term we selected three trajectories. These results support our formal results.

With respect to the usefulness of the $\ci$-entities as an entity-set the results are insufficient for a final assessment. 
 As expected we found $\ci$-entities that exhibit compositionality in time and space, counterfactual variation in value and extend, and degree of freedom traversal. However, we also found counter intuitive examples of $\ci$-entities that skip a time-step completely. In other words they disappear for a time-step and reappear again. This could be due to the small system size but this is not certain and needs more investigation.

Concerning entity action and entity perception of the $\ci$-entities. We found entity actions in value and extent. However we also find an example of an extent action that is counter intuitive and questions either our notion of entity actions or the $\ci$-entities. A decisive result is still lacking however.

Concerning perception, we found perceptions only where we expected it i.e.\ not for an independent process. However we could not define perceptions uniquely. A unique construction is guaranteed for non-interpenetrating entity-sets. However the $\ci$-entities turned out to allow interpenetration. As expected we then also found non-exclusive co-perception entities. In order to still get perceptions we used the approach that only relies on a set $\zeta$ of mutually exclusive co-perception entities to define the branch-morph. We obtained perceptions but we also saw that another choice of a co-perception entity can lead to a (quantitatively) different perception. This further confirmed that without non-interpenetration perception is not necessarily uniquely defined.

Finally, we showed that the same $\ci$-entity can perform entity actions and entity perception at the same time-step $t$. 

All together the $\ci$-entities need further investigation. One interesting next step is to use only the partition with the lowest SLI value for each trajectory as a source for entities. Such an entity-set would be non-interpenetrating. The other big question concerns the strange action we found. This also must be investigated further.

\bibliographystyle{apalike}
\bibliography{bibliography}

\end{document}